\documentclass[twoside]{article}

\usepackage{amsmath}
\usepackage{amsthm}
\usepackage{amssymb}
\usepackage{mathtools}
\usepackage[cal=euler,scr=boondoxupr,bb=bboldxLight]{mathalpha}
\usepackage{upgreek}
\usepackage{mleftright}
\usepackage[pdftex]{hyperref}
\usepackage[sf,bf]{titlesec}
\usepackage{bm}
\usepackage{euscript}
\usepackage{url}            
\usepackage{booktabs}       
\usepackage{caption,makecell}
\usepackage{threeparttable}
\usepackage{amsfonts}       
\usepackage{nicefrac}       
\usepackage{microtype}      
\usepackage{color}
\usepackage{tabu}
\usepackage{multirow}
\usepackage{float}
\usepackage{algorithm}
\usepackage{algorithmic}
\usepackage{comment}
\usepackage{graphicx}
\usepackage{epstopdf}
\usepackage{subcaption}
\usepackage{framed}
\usepackage{enumitem}
\usepackage{textcomp}
\usepackage{setspace}
\usepackage{cleveref}
\usepackage{latexsym}
\usepackage{tcolorbox}
\usepackage[numbers,sort&compress]{natbib}
\usepackage{alltt}
\usepackage{stmaryrd}
\usepackage{xspace}
\usepackage{xfrac}    
\usepackage{tikz}
\usepackage{tikz-qtree}
\usepackage{todonotes}
\usepackage{inconsolata} 
\usepackage[normalem]{ulem}

\usepackage[margin=1in]{geometry}
\usepackage{fancyhdr}
\fancypagestyle{plain}{\fancyhf{} 
	
	\fancyfoot[C]{\textsf{\thepage}}
}

\definecolor{mydarkblue}{rgb}{0,0.08,0.45}
\hypersetup{ %
	colorlinks=true,
	linkcolor=mydarkblue,
	citecolor=mydarkblue,
	filecolor=mydarkblue,
	urlcolor=mydarkblue}

\newcommand{\calB}{\mathcal{B}}
\newcommand{\calC}{\mathcal{C}}
\newcommand{\calD}{\mathcal{D}}

\newcommand{\calU}{\mathcal{U}}

\newcommand{\calZ}{\mathcal{Z}}

\newcommand{\scrM}{\mathscr{M}}

\newcommand{\scrO}{\mathscr{O}}

\newcommand{\Ex}{\mathbb{E}}

\newcommand{\one}{\mathbb{1}}
\newcommand{\zero}{\bm{0}}
\newcommand{\One}{\bm{1}}

\newcommand{\RR}{\mathbb{R}}

\newcommand{\Rp}{\RR_+}
\newcommand{\Rpp}{\RR_{++}}
\newcommand{\bbS}{\mathbb{S}}

\newcommand{\NN}{\mathbb{N}}

\newcommand{\diff}{\mathrm{d}}
\DeclareMathOperator{\prox}{prox}
\DeclareMathOperator{\rprox}{rprox}

\DeclareMathOperator{\soft}{soft}

\newcommand{\Id}{\mathrm{Id}}
\newcommand{\e}{\mathrm{e}}
\newcommand{\bA}{\bm{A}}

\newcommand{\bD}{\bm{D}}

\newcommand{\bH}{\bm{H}}
\newcommand{\bI}{\bm{I}}

\newcommand{\bM}{\bm{M}}

\newcommand{\bQ}{\bm{Q}}

\newcommand{\bx}{\bm{x}}
\newcommand{\by}{\bm{y}}
\newcommand{\bz}{\bm{z}}
\newcommand{\bbeta}{\bm{\beta}}

\DeclareMathOperator*{\argmin}{argmin}

\DeclareMathOperator*{\dom}{dom}
\DeclareMathOperator*{\interior}{int}

\newcommand{\bp}{\bm{p}}
\newcommand{\bmu}{\bm{\mu}}

\newcommand{\bxi}{\bm{\xi}}

\newcommand{\bw}{\bm{w}}

\newcommand{\sumK}{\sum_{k=1}^K}

\newcommand{\sumd}{\sum_{i=1}^d}

\newcommand{\bu}{\bm{u}}

\newcommand{\bomega}{\bm{\omega}}
\newcommand{\blambda}{\bm{\lambda}}

\newcommand{\bSigma}{\bm{\Sigma}}

\newcommand{\tr}{\mathrm{tr}}

\newcommand{\sfN}{\mathsf{N}}

\newcommand{\sfU}{\mathsf{U}}

\newcommand{\sfW}{\mathsf{W}}

\newcommand{\sfZ}{\mathsf{Z}}

\newcommand{\oRR}{\overline{\RR}}

\newcommand{\tbx}{\widetilde{\bx}}

\newcommand{\tbu}{\widetilde{\bu}}

\newcommand{\iiddist}{\stackrel{\text{i.i.d.}}{\sim}}
\renewcommand{\mid}{\,|\,}
\newcommand{\midd}{\,|\kern-0.25ex|\,}
\newcommand{\setn}{\llbracket n\rrbracket}

\newcommand{\setK}{\llbracket K\rrbracket}

\newcommand{\set}[1]{\llbracket #1\rrbracket}
\newcommand{\KL}{\mathrm{KL}}
\newcommand{\TV}{\mathrm{TV}}
\newcommand{\dotp}[2]{\left\langle #1, #2\right\rangle}

\newcommand{\env}{\mathrm{env}}
\newcommand{\lenv}{\overleftarrow{\env}}
\newcommand{\renv}{\overrightarrow{\env}}
\newcommand{\lprox}[2]{\overleftarrow{\operatorname{P}}_{\negthinspace\negthinspace #1}^{#2}}
\renewcommand{\rprox}[2]{\overrightarrow{\operatorname{P}}_{\negthinspace\negthinspace #1}^{#2}}

\newcommand{\lU}[2]{\overleftarrow{U}_{\negthinspace\negthinspace #1}^{#2}}
\newcommand{\rU}[2]{\overrightarrow{U}_{\negthinspace\negthinspace #1}^{#2}}

\DeclareMathOperator*{\arsinh}{arsinh}

\DeclareMathAlphabet\rsfscr{U}{rsfso}{m}{n}

\theoremstyle{plain}
\newtheorem{theorem}{Theorem}[section]

\theoremstyle{definition}
\newtheorem{definition}[theorem]{Definition}

\theoremstyle{remark}

\crefname{assumption}{Assumption}{Assumptions}
\Crefname{assumption}{Assumption}{Assumptions}
\crefname{problem}{Problem}{Problems}
\Crefname{problem}{Problem}{Problems}
\crefname{example}{Example}{Examples}
\Crefname{example}{Example}{Examples}

\let\le\leqslant
\let\ge\geqslant

\let\hat\widehat
\let\tilde\widetilde

\makeatletter
\DeclareFontFamily{OMX}{MnSymbolE}{}
\DeclareSymbolFont{MnLargeSymbols}{OMX}{MnSymbolE}{m}{n}
\SetSymbolFont{MnLargeSymbols}{bold}{OMX}{MnSymbolE}{b}{n}
\DeclareFontShape{OMX}{MnSymbolE}{m}{n}{
	<-6>  MnSymbolE5
	<6-7>  MnSymbolE6
	<7-8>  MnSymbolE7
	<8-9>  MnSymbolE8
	<9-10> MnSymbolE9
	<10-12> MnSymbolE10
	<12->   MnSymbolE12
}{}
\DeclareFontShape{OMX}{MnSymbolE}{b}{n}{
	<-6>  MnSymbolE-Bold5
	<6-7>  MnSymbolE-Bold6
	<7-8>  MnSymbolE-Bold7
	<8-9>  MnSymbolE-Bold8
	<9-10> MnSymbolE-Bold9
	<10-12> MnSymbolE-Bold10
	<12->   MnSymbolE-Bold12
}{}

\let\llangle\@undefined
\let\rrangle\@undefined
\DeclareMathDelimiter{\llangle}{\mathopen}%
{MnLargeSymbols}{'164}{MnLargeSymbols}{'164}
\DeclareMathDelimiter{\rrangle}{\mathclose}%
{MnLargeSymbols}{'171}{MnLargeSymbols}{'171}
\makeatother

\newcommand{\norm}[1]{\left\lVert#1\right\rVert}
\newcommand{\euclidnorm}[1]{\left\lVert#1\right\rVert_2}
\newcommand{\vecnorm}[2]{\left\| #1 \right\|_{{#2}}}
\newcommand{\matsnorm}[2]{|\kern-0.25ex|\kern-0.25ex| #1 |\kern-0.25ex|\kern-0.25ex|_{{#2}}}

\newcommand{\onenorm}[1]{\vecnorm{#1}{1}}

\newcommand{\specnorm}[1]{\matsnorm{#1}{\mathrm{S}}}

\newcommand{\iid}{i.i.d.\xspace~}

\renewcommand{\left}{\mleft}
\renewcommand{\right}{\mright}

\begin{document}
	\title{\sffamily Non-Log-Concave and Nonsmooth Sampling \\via Langevin Monte Carlo Algorithms}
	\author{Tim Tsz-Kit Lau%
        \thanks{Department of Statistics and Data Science, Northwestern University, Evanston, IL 60208, USA; 
        Email: \href{mailto:timlautk@u.northwestern.edu}{\texttt{timlautk@u.northwestern.edu}}.}
        \thanks{The University of Chicago Booth School of Business, Chicago, IL 60637, USA; Email: \href{mailto:timtsz-kit.lau@chicagobooth.edu}{\texttt{timtsz-kit.lau@chicagobooth.edu}}.}
        \and 
        Han Liu%
        \thanks{Department of Computer Science, Northwestern University, Evanston, IL 60208, USA; Email: \href{mailto:hanliu@northwestern.edu}{\texttt{hanliu@northwestern.edu}}.} \footnotemark[1]
        \and 
        Thomas Pock%
        \thanks{Institute of Computer Graphics and Vision, Graz University of Technology, 8010 Graz, Austria; Email: 
        \href{mailto:pock@icg.tugraz.at}{\texttt{pock@icg.tugraz.at}}.}
	}
	
	\maketitle
	
	\numberwithin{equation}{section}

	\begin{abstract}
		We study the problem of approximate sampling from non-log-concave distributions, e.g., Gaussian mixtures, which is often challenging even in low dimensions due to their multimodality. We focus on performing this task via Markov chain Monte Carlo (MCMC) methods derived from discretizations of the overdamped Langevin diffusions, which are commonly known as Langevin Monte Carlo algorithms. Furthermore, we are also interested in two nonsmooth cases for which a large class of proximal MCMC methods have been developed: (i) a nonsmooth prior is considered with a Gaussian mixture likelihood; (ii) a Laplacian mixture distribution. Such nonsmooth and non-log-concave sampling tasks arise from a wide range of applications to Bayesian inference and imaging inverse problems such as image deconvolution. We perform numerical simulations to compare the performance of most commonly used Langevin Monte Carlo algorithms. 
	\end{abstract}

    \section{Introduction} 
    \label{sec:intro}
        Sampling efficiently from a high-dimensional target distribution $\pi\propto\e^{-U}$ on $\RR^d$ has been a long-standing problem in various scientific and engineering disciplines, including statistics, applied probability, physics, machine learning and signal processing.     
        The task of drawing samples efficiently from high-dimensional complex probability distributions enables us to perform inference using complex statistical models from large amounts of data, where uncertainty quantification is of paramount importance to understand the intrinsic risk associated with every decision made with models learned from data. The ability to quantify uncertainty when comparing a theoretical or computational model to observations is critical to conducting a sound scientific investigation, particularly in machine-learned models and in the physical sciences like physics \cite{gal2022bayesian}. More specifically, Bayesian inference \cite{van2021bayesian,gelman2013bayesian} is a prominent method for linking models and observations and estimating uncertainties, in which sampling techniques are widely adopted, which also finds applications to various areas such as imaging processing and inverse problems (see e.g., \cite{durmus2022proximal}), and Bayesian neural networks and deep learning \cite{mackay1992practical}, etc. 
            
        While Markov chain Monte Carlo (MCMC) methods \cite{robert2004monte} have been the major workhorse of such sampling tasks, most traditional MCMC algorithms were regarded as unscalable to high dimensions. In particular, in modern large-scale applications such as Bayesian deep learning in the overparameterized regime in which we want to make posterior inference on the neural network weights, traditional MCMC algorithms become computationally prohibitive in such high dimensions and alternative approaches such as variational inference (VI; see e.g., \cite{blei2017variational}) have been widely adopted. 
        
        The Langevin Monte Carlo (LMC) algorithm (possibly with Metropolis--Hastings adjustment), which is derived based on the overdamped Langevin diffusion, has become a popular MCMC method for high-dimensional continuously differentiable distributions since it only requires access to a gradient oracle of the potential of the distribution, which can be computed easily using automatic differentiation softwares such as \textsf{PyTorch} \cite{paszke2019pytorch}, \textsf{TensorFlow} \cite{tensorflow2015-whitepaper} and \textsf{JAX} \cite{jax2018github}. Furthermore, in Bayesian inference tasks where large datasets are used, the gradient oracle required can be provided with stochastic approximation \cite{robbins1951}, leading to stochastic gradient Langevin dynamics (SGLD) \cite{welling2011bayesian}. 
        
        Theoretical convergence guarantees of the LMC algorithm have been rather well understood in the literature when the target potential is (strongly) convex and Lipschitz smooth (see e.g., \cite{dalalyan2017theoretical,durmus2017nonasymptotic,cheng2018convergence,durmus2019high,erdogdu2022convergence,freund2022convergence} for unadjusted Langevin algorithm and \cite{roberts1996exponential,roberts1998optimal,bou2013nonasymptotic,dwivedi2019log,chen2020fast,lee2020logsmooth,mangoubi2019nonconvex,chewi2021optimal,wu2022minimax,altschuler2023faster} for Metropolis-adjusted Langevin algorithm). However, the understanding theoretical properties of the LMC algorithm for sampling from \emph{non-log-concave} and \emph{non-(log-)smooth} target distributions is very limited in the literature. To facilitate our understanding under such settings, in this work, we are interested in sampling from more general distributions, modeled with mixture distributions. 
        For instance, mixtures of Gaussians are so expressive that they can be treated as a universal approximation of arbitrary probability measures (see e.g., \cite{delon2020wasserstein}). However, sampling from mixture distributions remains notoriously difficult with standard MCMC methods due to their multimodality \cite{celeux2000computational,jasra2005markov,chopin2012free}.  
        
        In this work, we provide an empirical study on MCMC algorithms based on the overdamped Langevin diffusion for general \emph{non-log-concave} and \emph{non-(log-)smooth} distributions, which are exemplified by mixtures of Gaussians possibly with nonsmooth Laplacian priors and mixtures of Laplacians. Note that this is beyond usual theoretical assumptions on the target distributions and hence beyond theoretical convergence guarantees for such LMC algorithms in most existing works in the literature.

        \subsection{Related Work}
        The multimodal Gaussian mixture distribution has long been used as a benchmarking problem for sampling algorithms, see e.g., \cite{lambert2022variational,yan2023learning}, which are based on alternative methods such as variational inference and gradient flows. Gradient-based MCMC methods have become crucial in large-scale applications such as Bayesian deep learning, while they also aroused much theoretical interest, notably due to the intriguing connection between optimization and sampling (see e.g., \cite{ma2019sampling,dalalyan2017further}). 
        
        In the rest of this section, we give a brief overview of related work on gradient-based MCMC methods in the recent literature. While we attempt to mention most of them, due to the gigantic amount of work in this field, the cited literatures here are by no means exhaustive.

        \subsubsection{Gradient-Based MCMC Algorithms}
        Since the seminal work \cite{parisi1981correlation} by Nobel laureate in Physics Giorgio Parisi, the \emph{overdamped Langevin diffusion} has motivated a large class of Langevin Monte Carlo algorithms through its discretization, which consist of the unadjusted Langevin algorithm. The Metropolis--Hastings correction step, as a standard technique used to ensure a correct distribution, together with a Langevin proposal, has led to the \emph{Metropolis-adjusted Langevin algorithm} (MALA) \cite{roberts1996exponential}. Lots of theoretical convergence results have been established in the literature, see e.g., \cite{durmus2017nonasymptotic,durmus2019high,dalalyan2017theoretical,altschuler2022concentration,altschuler2023resolving,durmus2021asymptotic,cheng2018convergence} for ULA and \cite{dwivedi2019log,chewi2021optimal,lee2021lower,altschuler2023faster,durmus2022geometric,bou2013nonasymptotic,wu2022minimax,roberts1998optimal,roberts2002langevin,boisvert2022mala} for MALA. Other works also relax the assumption of strong-log-concavity of the target distribution to log-concavity \cite{nguyen2022unadjusted}, or that the target distribution satisfies some functional inequalities \cite{mousavi2023towards}, e.g., Poincar\'{e} and log-Sobolev inequalities. 
        
        Other than the overdamped Langevin dynamics, gradient-based MCMC algorithms based on higher-order dynamics such as the underdamped Langevin or kinetic Langevin diffusion and Hamiltonian dynamics have also attracted much attention, which respectively lead to the \emph{underdamped} or \emph{kinetic Langevin Monte Carlo} (ULMC or KLMC) algorithms \cite{zhang2023improved,cheng2018underdamped,durmus2021uniform,dalalyan2020sampling} and the \emph{Hamiltonian Monte Carlo} (HMC) algorithm \cite{neal1993bayesian,neal2011mcmc,bou2023mixing,chen2022optimal,chen2020fast,bou2020coupling,li2023riemannian,mangoubi2019mixing,mangoubi2021mixing}. 
        
        Taking ideas from both Riemannian and information geometry into consideration, \cite{girolami2011riemann} introduced the \emph{manifold Metropolis-adjusted Langevin algorithm} and \emph{Riemannian manifold Hamiltonian Monte Carlo algorithm}. \cite{livingstone2014information} derived the Langevin diffusion on a Riemannian manifold and provided a more in-depth study on the subject. Geometric-informed Langevin and Hamiltonian Monte Carlo algorithms have been heavily studied since then; see e.g.,  \cite{gatmiry2022convergence,cheng2022efficient,wang2020fast,lee2018convergence,kook2022sampling,kook2023condition,zhang2022geometry,livingstone2014information,xifara2014langevin}. 
    
        While most existing theoretical results on the LMC algorithms have focused on strongly log-concave and log-Lipschitz-smooth target distributions, recent works have investigated the theoretical properties of   
        non-log-concave sampling with Langevin Monte Carlo algorithms, see e.g.,  \cite{balasubramanian2022towards,chewi2022analysis,cheng2018sharp,mou2019sampling,ge2018beyond,mou2022improved,holzmuller2023convergence,mangoubi2019nonconvex,vempala2019rapid}. 
           
        In addition to the widely used Euler--Maruyama discretization scheme, more sophisticated discretization schemes for stochastic differential equations (SDEs) are also used to reduce the discretization bias, e.g., the \emph{explicit stabilized SK-ROCK scheme} \cite{pereyra2020accelerating} and \emph{preconditioned Crank--Nicolson Langevin} (pCNL) samplers \cite{cotter2013mcmc,titsias2018auxiliary}.    
        
        We refer readers to a recent review on gradient-based MCMC algorithms \cite{vorstrup2021gradient} for more discussion on related methods, in which the class of \emph{piecewise deterministic Monte Carlo} (PDMC) algorithms derived from piecewise deterministic Markov processes (PDMPs) have been mentioned in detail, e.g., the \emph{zig-zag sampler} \cite{bierkens2019zig} and the \emph{bouncy particle sampler} \cite{bouchard2018bouncy}.

        \subsubsection{Proximal MCMC Algorithms}
        In practice, we might need to generate samples from distributions with nonsmooth potentials, e.g., shrinkage priors in Bayesian regression problems, whereas the convergence guarantees of LMC algorithms often require the Lipschitz smoothness of the target potential. 
        Motivated by proximal optimization algorithms which are widely applied to nonsmooth optimization problems, a large class of proximal MCMC algorithms have been proposed in the literature; see e.g., \cite{pereyra2016proximal,bernton2018langevin,wibisono2019proximal,salim2019stochastic,salim2020primal,brosse2017sampling,bubeck2018sampling,luu2021sampling,schreck2015shrinkage,mou2022efficient}. In particular, the central idea of \cite{pereyra2016proximal,durmus2018efficient} is to sample from a smooth surrogate distribution whose potential is the smooth approximation of the nonsmooth potential of the target distribution. Such a smooth approximation is called the \emph{Moreau envelope} or \emph{Moreau--Yosida regularization} from convex analysis \cite{bauschke2017}, whose gradient involves the proximity operator. With the Metropolis--Hastings correction step, proximal MALA was also proposed and studied theoretically \cite{crucinio2023optimal,pillai2022optimal}. In addition to Bayesian imaging applications (e.g., \cite{durmus2018efficient,vidal2020maximum1,de2020maximum2,cai2022proximal}), such proximal MCMC algorithms are also recently applied to Bayesian inference problems in statistics \cite{zhou2022proximal,heng2023bayesian}. 
        
        Alternatively, other than the Moreau envelope and proximity operator, LMC algorithms have also been extended with other tools for nonsmooth sampling, e.g., Gaussian smoothing \cite{chatterji2020langevin} and subgradient \cite{lehec2023langevin}. More recently, another type of proximal samplers is proposed based on a novel Gibbs sampling approach, which aim to correct the asymptotic bias induced by the smooth approximation of the nonsmooth part of the potential; see   \cite{gopi2023algorithmic,lee2021structured,chen2022improved,liang2022proximal,liang2023proximal} for details. 
        It is however worth noting that proximal MCMC methods can also be applied for smooth sampling.

        \subsubsection{Stochastic Gradient MCMC Algorithms}
        Similar to stochastic gradient methods in the optimization literature, replacing the full gradient of the potential by its stochastic approximation (namely, stochastic gradient or batch gradient) in gradient-based MCMC algorithms has led to a large class of \emph{stochastic gradient Markov Chain Monte Carlo} (SG-MCMC) methods \cite{ma2015complete}, which have been the major workhorse of sampling high-dimensional distributions, e.g., Bayesian deep learning. In addition to \emph{stochastic gradient Langevin dynamics} (SGLD) \cite{welling2011bayesian,ding2014bayesian,li2016preconditioned,brosse2018promises,dalalyan2019user}, stochastic gradient MCMC algorithms derived from high-order dynamics \cite{li2016high} such as \emph{stochastic gradient kinetic Langevin dynamics} (SGKLD) and \emph{stochastic gradient Hamiltonian Monte Carlo} (SGHMC) \cite{chen2014stochastic,zou2019stochastic,zou2021convergence}. For the projected and proximal variants of SGLD, see e.g.,  \cite{lamperski2021projected,durmus2019analysis}. 
        
        Moreover, when computing the stochastic gradients, data samples might have more structures other than simply being independently and identically distributed: 
        \cite{barkhagen2021stochastic,chau2021stochastic} studied SGLD for the case of dependent data streams. We also refer readers to \cite{nemeth2021stochastic} for a recent review of SG-MCMC methods.

        \subsection{Notation} 
        	\label{subsec:notation}
        	We denote by $\bI_d\in\RR^{d\times d}$ the $d\times d$ identity matrix. 
        	We also define $\oRR \coloneqq \RR\cup\{+\infty\}$, $\Rp \coloneqq \left[0, +\infty\right[$, $\Rpp \coloneqq \left]0, +\infty\right[$. $\NN$ denotes the set of nonnegative integers and $\NN^* \coloneqq \NN\setminus\{0\}$ denotes the set of positive integers. Then we write $\setn \coloneqq \{1, \ldots, n\}$ for $n\in\NN^*$. 
        	Let $\bbS^d_{++}$ denote the set of symmetric positive definite matrices in $\RR^{d\times d}$. 
            For every $\bQ\in\bbS^d_{++}$, let $\norm{\cdot}_{\bQ}\coloneqq\dotp{\cdot}{\bQ\cdot}^{\sfrac12}$. 
        	The domain of a function $f\colon\RR^d\to\oRR$ is $\dom f \coloneqq \{\bx\in\RR^d : f(\bx)<+\infty\}$. 
        	The set $\Gamma_0(\RR^d)$ denotes the class of lower-semicontinuous convex functions from $\RR^d$ to $\oRR$ with a nonempty domain (i.e., proper). 
        	The convex indicator function $\iota_\calC(x)$ of a closed convex set $\calC\ne\varnothing$ at $x$ equals $0$ if $x\in\calC$ and $+\infty$ otherwise.  
            The $0$-$1$ indicator function $\one_{\calC}(x)$ of a set $\calC$ at $x$ equals $1$ if $x\in\calC$ and $0$ otherwise.  
        	The $(d-1)$-dimensional probability simplex is denoted by $\Delta^d \coloneqq \{\bp\in[0, 1]^d : \dotp{\One_d}{\bp} = 1\}$, where $\One_d$ is the $d$-dimensional all-one vector. 
        	We denote by $\calB(\RR^d)$ the Borel $\sigma$-field of $\RR^d$. 
        	For two probability measures $\mu$ and $\nu$ on $\calB(\RR^d)$, the relative entropy or the Kullback--Leibler (KL) divergence from $\mu$ to $\nu$ is $D_{\KL}(\mu\midd\nu) \coloneqq \int_{\RR^d} \log(\diff\mu/\diff\nu)  \,\diff\mu$ if $\mu$ is absolutely continuous w.r.t.~$\nu$ (denoted by $\mu\ll\nu$) with the Radon--Nikodym derivative $\diff\mu/\diff\nu$ and $+\infty$ otherwise. 
         	The total variation (TV) distance between $\mu$ and $\nu$ is defined by $\|\mu - \nu\|_{\TV} = \sup_{A\in\calB(\RR^d)}|\mu(A) - \nu(A)|$. 
             For $p\in\left[1,+\infty\right[$, the $p$-Wasserstein distance between $\mu$ and $\nu$ is defined by 
             	\begin{equation*}
             		\sfW_p(\mu, \nu) \coloneqq \left(\inf_{\pi\in\Pi(\mu, \nu)}\int_{\Omega\times\Omega} \norm{\bx - \by}^p\,\diff\pi(\bx, \by)\right)^{\negthickspace\sfrac1p},
             	\end{equation*}
             	where $\norm{\cdot}$ is the Euclidean norm on $\RR^d$, and $\Pi(\mu, \nu)$ denotes the set of joint distributions on $\RR^d\times\RR^d$ with $\mu$ and $\nu$ as marginals. 
            We also write $a\wedge b$ for $\min\{a, b\}$ for $a,b\in\RR$. 
        	In this work, unless otherwise specified, $(\bxi_n)_{n\in\NN}$ denotes a sequence of $d$-dimensional \iid standard Gaussian random variables $\sfN(\zero_d, \bI_d)$. 
    
        \section{Preliminaries}
        In this section, we introduce the formulation of our problem of interest and necessary technical tools to facilitate the discussion of various LMC algorithms. 
    
    	\subsection{Problem Formulation}
    	We are concerned with the problem of generating samples from a probability distribution $\pi$ on $(\RR^d, \calB(\RR^d))$ which admits a density, with slight abuse of notation, also denoted by $\pi$, with respect to the Lebesgue measure 
    	\begin{equation}\label{eqn:Leb}
    		(\forall \bx\in\RR^d) \quad \pi(\bx) = \left.\e^{-U(\bx)} \,\middle/\, \int_{\RR^d} \e^{-U(\by)}\,\diff \by \right. , 
    	\end{equation}
    	where the \emph{potential} $U \colon \RR^d\to\oRR$ is measurable and we assume that $0<\int_{\calU} \e^{-U(\by)}\,\diff \by<+\infty$ for $\calU\coloneqq\dom U$. We also write $\pi\propto \e^{-U}$ for \eqref{eqn:Leb}. Usually, the number of dimensions $d \gg 1$. 
     
     Specifically, we consider mixture distributions, e.g., Gaussian mixtures, which are weighted averages of distributions of the same class but with possibly different parameters. For examples, Gaussian mixtures involve Gaussians with possibly different means and covariance matrices. More specifically, for $\bomega=(\omega_k)_{1\le k\le K}\in\Delta^K$, the density of a $\bomega$-mixture is given by 
        \begin{equation*}
            (\forall \bx\in\RR^d)\quad p(\bx) \coloneqq \sumK\omega_k p_k(\bx), 
        \end{equation*}
        where, for a $\bomega$-Gaussian mixture, we have, for each $k\in\setK$, 
    	\begin{equation*}
    		(\forall \bx\in\RR^d)\quad p_k(\bx) \coloneqq \frac{1}{\sqrt{\det(2\uppi\bSigma_k)}} \exp\left\{ -\frac12 \dotp{\bx - \bmu_k}{\bSigma_k^{-1}(\bx-\bmu_k)}\right\},
    	\end{equation*}
        which is the density of a $d$-dimensional Gaussian distribution with mean $\bmu_k\in\RR^d$ and covariance matrix $\bSigma_k\in\bbS_{++}^d$. 
    	The corresponding \emph{potential} is defined by $U(\bx) \coloneqq -\log p(\bx)$, which is nonconvex. 
    
        In various Bayesian imaging inverse problems and high-dimensional Bayesian sparse regression problems, in order to exploit prior knowledge available, prior distributions are often introduced. Here we consider priors $\varrho\propto\e^{-g}$ with $g\in\Gamma_0(\RR^d)$ and possibly nonsmooth, i.e., priors which are log-concave yet not necessarily log-smooth. More notable examples include the $\ell_1$-norm (i.e., Laplacian prior), and the isotropic/1D (e.g., for trend filtering \cite{kim2009ell} and fused lasso \cite{tibshirani2005sparsity,tibshirani2014adaptive}) and anisotropic/2D (for images) total-variation (TV) pseudonorms \cite{chambolle2004algorithm}: 
        \begin{equation}\label{eqn:nonsmooth}
        (\forall \bx\in\RR^d)\quad g_1(\bx) = \tau\onenorm{\bD\bx} \quad \text{or} \quad g_2(\bx) = \tau\norm{\bD \bx}_{1-2}, 
        \end{equation}
        where $\tau>0$ is a regularization parameter, $\norm{\cdot}_{1-2}$ is the composite $\ell_1$-$\ell_2$ norm and $\bD\colon\RR^d\to\RR^{d'}$ is a linear operator. For instance, $\bD = \bI_d$ in $g_1$ corresponds to the $\ell_1$-norm, and $\bD$ being the \emph{first-order differencing matrix} in $g_1$ and $g_2$ corresponds to the 1D-TV and 2D-TV pseudonorms respectively. For Bayesian posterior inference, we then want to sample from the posterior distribution $\pi(\bx) \propto p(\bx)\times\varrho(\bx)$. Notice that, in various Bayesian imaging inverse problems such as denoising, deconvolution and reconstruction, the likelihood term $p(\bx)$ is usually a less complicated Gaussian likelihood. We consider Gaussian mixtures instead for generality. 
        
        On the other hand, in numerical simulations, we are also interested in generating samples from a mixture of non-log-smooth distributions. In particular, we consider a $\blambda$-(isotropic) Laplacian mixture, where, for $k\in\setK$, each of the $K$ Laplacian distributions has a density: 
        \begin{equation}\label{eqn:laplacian}
        	(\forall \bx\in\RR^d)\quad p_k(\bx) \coloneqq \frac{\alpha_k^d}{2^d} \exp\left\{ -\alpha_k \onenorm{\bx - \bmu_k}\right\},
        \end{equation}
        where $\bmu_k\in\RR^d$ is a location parameter and $\alpha_k>0$ is a scale parameter.

        \subsection{Convex Analysis}   
        We give definitions of important notions from convex analysis \cite{rockafellar1970,rockafellar1998,tutorialscombettes_pesquet,bauschke2017}, which will be revisited in \Cref{sec:prox_lmc}.    
    
            \begin{definition}[Proximity operator]
                For $\lambda>0$, the \emph{proximity operator} of a function $h\in\Gamma_0(\RR^d)$ is defined by 
                \begin{equation*}
                    (\forall\bx\in\RR^d)\quad \prox_{\lambda h}(\bx) \coloneqq \argmin_{\by\in\RR^d}\ \left\{ h(\by) + \frac{1}{2\lambda}\euclidnorm{\by-\bx}^2\right\}. 
                \end{equation*}
            \end{definition}
            
            \begin{definition}[Moreau envelope]
                For $\lambda>0$, the \emph{Moreau envelope}, also known as the \emph{Moreau--Yosida regularization} of a function $h\in\Gamma_0(\RR^d)$, is defined by 
                \begin{equation*}
                 (\forall\bx\in\RR^d)\quad h_\lambda(\bx) \coloneqq \min_{\by\in\RR^d}\ \left\{h(\by) + \frac{1}{2\lambda}\euclidnorm{\by-\bx}^2\right\}. 
                \end{equation*}
                Since $h$ is convex, by \cite[Theorem 2.26]{rockafellar1998}, the Moreau envelope is well-defined, convex and continuously differentiable with its gradient given by 
                \[(\forall\bx\in\RR^d)\quad \nabla h_\lambda(\bx) = \lambda^{-1} \left(\bx - \prox_{\lambda h}(\bx)\right). \]
            \end{definition}
            The notions of Moreau envelope and proximity operator can be extended with Bregman divergences \cite{bregman1967}. We will need to introduce the notions of Legendre functions and Fenchel conjugate before formally defining Bregman divergences. 
        	\begin{definition}[Legendre functions
        		]
                Let $\varphi\in\Gamma_0(\RR^d)$. 
        		A function $\varphi$ is called (i) \emph{essentially smooth}, if it is differentiable on $\interior\dom\varphi\ne\varnothing$ and $\|\nabla\varphi(\bx_n)\|\to+\infty$ whenever $\bx_n\to \bx \in\operatorname{bdry}\dom\varphi$; (ii) \emph{essentially strictly convex}, if it is strictly convex on $\interior\dom\varphi$; (iii) \emph{Legendre}, if it is both essentially smooth and essentially strictly convex. 
        	\end{definition}
        	
        	\begin{definition}[Fenchel conjugate]\label{def:conjugate}
        		The \emph{Fenchel conjugate} of a proper function $h$ is defined by 
        		$h^*(\bx) \coloneqq \sup_{\by\in\RR^d} \,\left\lbrace \dotp{\by}{\bx} - h(\by) \right\rbrace$. 
        		For a Legendre function $\varphi$, it is well known that $\nabla\varphi\colon\interior\dom\varphi\to\interior\dom\varphi^*$ with $(\nabla\varphi)^{-1} = \nabla\varphi^*$. 
        	\end{definition}

        	\begin{definition}[Bregman divergence]
        		The \emph{Bregman divergence} between $\bx$ and $\by$ associated with a Legendre function $\varphi$ is defined through 
        		\begin{equation*}
        			D_{\varphi}\colon \RR^d\times\RR^d \to[0, +\infty] \colon
        			(\bx, \by) \mapsto 
        			\begin{cases*}
        				\varphi(\bx) - \varphi(\by) - \dotp{\nabla \varphi(\by)}{\bx - \by}, & if $\by\in\interior\dom\varphi$, \\
        				+\infty, & otherwise. 
        			\end{cases*}
        		\end{equation*}
        	\end{definition}
            Since the Bregman divergence is not symmetric in its arguments, we can define two Bregman proximity operators and Bregman--Moreau envelopes \cite{chen2012moreau,kan2012moreau,bauschke2006joint,bauschke2018regularizing}, as follows. 
            \begin{definition}[Bregman proximity operators]
            \label{def:bregman_prox}
            For $\lambda >0$, the \emph{left} and \emph{right Bregman proximity operators} of $h\in\Gamma_0(\RR^d)$ associated with a Legendre function $\varphi$ are respectively defined by 
            \begin{equation*}
            (\forall\bx\in\interior\dom\varphi)\quad
                \begin{aligned}
                    \lprox{\lambda, h}{\varphi}(\bx) &\coloneqq \argmin_{\by\in\RR^d}  \, \left\lbrace h(\by) + \frac{1}{\lambda}D_\varphi(\by, \bx)\right\rbrace, \\
                    \rprox{\lambda, h}{\varphi}(\bx) &\coloneqq \argmin_{\by\in\RR^d} \, \left\lbrace h(\by) + \frac{1}{\lambda}D_\varphi(\bx, \by) \right\rbrace. 
                \end{aligned}
            \end{equation*}
            \end{definition}
    
        	\begin{definition}[Bregman--Moreau envelopes]
        		\label{def:bregman_env}
        		For $\lambda >0$, the \emph{left} and \emph{right Bregman--Moreau envelopes} of $h\in\Gamma_0(\RR^d)$ associated with a Legendre function $\varphi$ are respectively defined by
        		\begin{equation*}
                (\forall\bx\in\RR^d)\quad
                \begin{aligned}
                \lenv_{\lambda, h}^\varphi(\bx) &\coloneqq \inf_{\by\in\RR^d} \, \left\lbrace h(\by) + \frac{1}{\lambda}D_\varphi(\by, \bx) \right\rbrace, \\
                \renv_{\lambda, h}^\varphi(\bx) &\coloneqq \inf_{\by\in\RR^d} \, \left\lbrace h(\by) + \frac{1}{\lambda}D_\varphi(\bx, \by) \right\rbrace. 
                \end{aligned}
                \end{equation*}
        	\end{definition}
            When $\varphi = \frac12\euclidnorm{\cdot}^2$, we recover the classical \emph{Moreau envelope} and the \emph{(Moreau) proximity operator} \cite{moreau1962fonctions,moreau1965proximite}. 
        	Under some rather technical conditions \cite{bauschke2018regularizing}, the left and right Bregman--Moreau envelopes are differentiable on $\interior\dom\varphi$ and their gradients are given respectively by 
           \begin{equation}\label{eqn:grad_Bregman_Moreau}
           (\forall\bx\in\interior\dom\varphi)\quad
           \begin{aligned}
           \nabla\lenv_{\lambda, h}^\varphi(\bx) &= \frac{1}{\lambda}\nabla^2\varphi(\bx)\left(\bx - \lprox{\lambda, h}\varphi(\bx) \right), \\
           \nabla\renv_{\lambda, h}^\varphi(\bx) &= \frac{1}{\lambda}\left(\nabla\varphi(\bx) - \nabla\varphi\left( \rprox{\lambda, h}\varphi(\bx)\right)  \right). 
           \end{aligned}        
           \end{equation}

    	\section{Langevin Monte Carlo Algorithms}
        \label{sec:lmc}
    	The Langevin Monte Carlo (LMC) algorithm is based on the Euler--Maruyama (EM) discretization of the overdamped Langevin diffusion. Lots of variants of the LMC algorithm have also been proposed in the literature to improve sampling efficiency and handle structural properties of target distributions. In this section, we give a brief review on LMC algorithms, which are used as an umbrella term to refer to as the class of MCMC algorithms motivated by the overdamped Langevin diffusion.

        \subsection{Unadjusted Langevin Algorithm}
            Let us recall that we are interested in sampling from the target distribution $\pi$ on $\RR^d$, whose density is only known up to a constant. With slight abuse of notation, $\pi\propto\e^{-U}$ also denotes the density of the target distribution with respect to the Lebesgue measure, where $U\colon\RR^d\to\oRR$ is called the \emph{potential} of the distribution $\pi$. We also assume that $U$ is Lipschitz smooth, i.e., there exists a Lipschitz constant $L>0$ such that $\euclidnorm{\nabla U(\bx) - \nabla U(\by)} \le L \euclidnorm{\bx-\by}$ for any $(\bx,\by)\in\RR^d\times\RR^d$, and that $U\in\Gamma_0(\RR^d)$.         
            The \emph{overdamped Langevin diffusion} is a stochastic differential equation given by, for $t\ge0$,         
            \begin{equation}\label{eqn:Langevin_diff}
            \diff X_t = -\nabla U(X_t)\,\diff t + \sqrt2 \,\diff W_t, 
            \end{equation}
            where $(W_t)_{t\ge0}$ is a $d$-dimensional Brownian motion. It is well-known that under mild appropriate conditions on $U$, this equation has a unique strong solution. Furthermore, if $\e^{-U}$ is integrable, then $\pi\propto\e^{-U}$ is the unique invariant distribution of the semigroup associated with \eqref{eqn:Langevin_diff}. 
            
            Using the first-order Euler--Maruyama (EM) discretization of the overdamped Langevin diffusion, we obtain the \emph{Langevin Monte Carlo} (LMC) algorithm (see e.g., \cite{parisi1981correlation,neal1993bayesian}), which is arguably the most widely-studied gradient-based MCMC algorithm: 
        	\begin{equation}\label{eqn:ULA}
        		(\forall n\in\NN) \quad \bx_{n+1} = \bx_n - \gamma\nabla U(\bx_n) + \sqrt{2\gamma}\,\bxi_n, 
        	\end{equation}
        	where $(\bxi_n)_{n\in\NN}$ is a sequence of \iid $d$-dimensional standard Gaussian random variables and $\gamma>0$ is a given step size. Possibly replaced with a varying step size $(\gamma_n)_{n\in\NN}$, the LMC algorithm is also referred to as the \emph{unadjusted Langevin algorithm} (ULA) \cite{durmus2017nonasymptotic,roberts1996exponential}. Notice that we use constant step sizes $\gamma>0$ when describing the LMC algorithms but nonincreasing step size sequences $(\gamma_n)_{n\in\NN}$ can be used in practice.

    	\subsection{Metropolis-Adjusted Langevin Algorithm}
        Due to the discretization, ULA gives biased samples depending on the step size. To adjust for the bias and ensure that the resulting algorithm has exactly the correct invariant distribution, the Metropolis--Hastings (MH) correction can be used. Applying a Metropolis--Hastings (MH) correction step at each iteration (as a proposal) of \eqref{eqn:ULA} yields the \emph{Metropolis-adjusted Langevin algorithm} (MALA) \cite{roberts1996exponential,robert2004monte,dwivedi2019log}. 
    	
        To be specific, the MH correction step operates as follows. 
    	Given a proposal $\tbx_{n+1}$ and the current sample $\bx_n$, the proposal is accepted with the acceptance probability 
        \begin{equation}\label{eqn:MH_prob}
        \alpha = \alpha(\tbx_{n+1}, \bx_n) \coloneqq \frac{p(\tbx_{n+1})q_\gamma(\bx_n \mid \tbx_{n+1})}{p(\bx_n)q_\gamma(\tbx_{n+1} \mid \bx_n)} \wedge 1, 
        \end{equation}
    	with 
    	\[q_\gamma(\tbx\mid \bx) = \phi_d(\tbx; \bx - \gamma\nabla U(\bx), 2\gamma\bI_d) \propto \exp\left\{-\frac{1}{4\gamma}\euclidnorm{\tbx - (\bx - \gamma\nabla U(\bx))}^2\right\}, \]
    	where $\phi_d(\cdot; \bmu, \bSigma)$ denotes the $d$-dimensional Gaussian density with mean $\bmu\in\RR^d$ and covariance matrix $\bSigma\in\bbS^d_{++}$. In general, the MH correction step can be applied to any proposal.     
        Now, defining $b_{n+1} \coloneqq \one_{\Rp}\left(\alpha(\tbx_{n+1}, \bx_n) - u_n\right)$ with a sequence of $\sfU(0,1)$ random variables $(u_n)_{n\in\NN}$, MALA then iterates 
    	\begin{equation}\label{eqn:MALA}
            (\forall n\in\NN) \quad 
            \begin{aligned}
            \tbx_{n+1} &= \bx_n - \gamma\nabla U(\bx_n) + \sqrt{2\gamma}\,\bxi_n, \\
            \bx_{n+1} &= b_{n+1}\tbx_{n+1} + (1-b_{n+1})\bx_n. 
            \end{aligned}
        \end{equation}

    	\subsection{Preconditioned Unadjusted Langevin Algorithm}    
        In convex optimization algorithms, to accelerate convergence, usually a preconditioning matrix can be used in order to exploit the local geometry of the objective function (cf.~Newton's method). In a similar vein, we can also apply a preconditioning matrix $\bM\in\bbS_{++}^d$ in ULA. This algorithm is called the \emph{preconditioned unadjusted Langevin algorithm} (PULA), which iterates  
        \begin{equation}\label{eqn:p_ula}
            (\forall n\in\NN) \quad \bx_{n+1} = \bx_n - \gamma \bM\nabla U(\bx_n) + \sqrt{2\gamma} \bM^{\sfrac12}\bxi_n. 
        \end{equation}
        Indeed, the preconditioning matrix can be either state-dependent or state-independent. Usually, state-dependent preconditioning matrices can help exploit the local geometry of the current sample, but would incur more computations. A common choice of the preconditioning matrix is $\bM(\bx) = [\nabla^2 U(\bx)]^{-1}$ for all $\bx\in\RR^d$, the inverse Hessian of the potential (which is indeed the \emph{Fisher metric}; see e.g., \cite{girolami2011riemann,livingstone2014information}). Under this context, PULA can be interpreted as ULA directly on the natural Riemannian manifold where the parameters live. There are also alternative strategies to choose the metric $\bM(\bx)$, see e.g., the SoftAbs metric proposed in \cite{betancourt2013general} and a Majorization-Minimization (MM) strategy proposed in \cite{marnissi2020majorize} with lower computational complexity.

    	\subsection{Mirror-Langevin Algorithm}
        The mirror-Langevin algorithm is the sampling counterpart of mirror descent \cite{beck2003mirror}. Introduced in \cite{zhang2020wasserstein}, under certain assumptions on $U$, the mirror-Langevin diffusion (MLD) takes the form: for $t\ge0$, 
    	\begin{equation*}
    		\begin{cases}
    			X_t = \nabla \varphi^*(Y_t), \\
    			\diff Y_t = -\nabla U(X_t)\,\diff t + \sqrt{2} \left[ \nabla^2 \varphi(X_t) \right]^{\sfrac{1}{2}}\,\diff W_t, 
    		\end{cases}
    	\end{equation*}
    	where $\varphi$ is a Legendre function and $\varphi^*$ is the Fenchel conjugate of $\varphi$. 
    	Its EM discretization yields a \emph{Hessian Riemannian LMC} (HRLMC) algorithm: 
    	\begin{equation}\label{eqn:HRLMC}
    		(\forall n \in\NN) \quad\bx_{n+1} = \nabla\varphi^*\left(\nabla\varphi(\bx_n) - \gamma\nabla U(\bx_n)
    		 + \sqrt{2\gamma}\left[ \nabla^2 \varphi(\bx_n) \right]^{\sfrac{1}{2}}\bxi_n \right). 
    	\end{equation}
    	This is the main discretization scheme considered in \cite{zhang2020wasserstein} and an earlier draft of \cite{hsieh2018mirrored}, and further studied in \cite{li2022mirror}, which is a specific instance of the Riemannian LMC reminiscent of the mirror descent algorithm. \cite{ahn2021efficient} consider an alternative discretization scheme motivated by the mirrorless mirror descent \cite{gunasekar2021mirrorless}, called the \emph{mirror-Langevin algorithm} (MLA): 
    	\begin{equation}\label{eqn:MLA}
    		(\forall n\in\NN)\quad
    		\begin{aligned}
    			\bx_{n+\sfrac{1}{2}} &= \nabla\varphi^* \left(\nabla\varphi(\bx_n) -\gamma\nabla U(\bx_n) \right), \\
    			\bx_{n+1} &= \nabla\varphi^*(Y_{\gamma_n}), 			
    		\end{aligned}
    	\end{equation}	
    	where 
    	\begin{equation}
        \label{eqn:diffusion}
    		\begin{cases}
    			\diff Y_t = \sqrt{2} \left[ \nabla^2 \varphi^*(Y_t) \right]^{-\sfrac{1}{2}}\,\diff W_t, \\
    			Y_0 = \nabla\varphi\left(\bx_{n+\sfrac{1}{2}}\right) = \nabla\varphi(\bx_n) -\gamma\nabla U(\bx_n). 
    		\end{cases} 
    	\end{equation}
    	MLA \eqref{eqn:MLA} is harder to implement than \eqref{eqn:HRLMC} in practice since a continuous diffusion \eqref{eqn:diffusion} has to be solved. 
        See also \cite{li2022mirror,jiang2021mirror} for related theoretical results of MLA. With slight ambiguity, we will also refer \eqref{eqn:HRLMC} as MLA in simulations as it is easier to solve. Note that PULA is a particular instance of MLA when $\varphi(\bx) = \frac12\norm{\bx}_{\bM^{-1}}^2$.

        \subsection{Comparison of Langevin Monte Carlo algorithms}
        \label{subsec:comp_lmc}
        We provide a summary of the iteration complexities of various unadjusted LMC algorithms in \Cref{tab:lmc} and the mixing times of Metropolis-adjusted Langevin algorithm (MALA) in total variation distance in \Cref{tab:malmc} from the existing literature. More specifically, in \Cref{tab:lmc}, we consider the composite potential $U=f+g$, where $f$ is the negative log-likelihood $f$ and $g$ is the log-prior. We consider two cases: (i) $f$ is $\hat{m}$-strongly convex and $M$-Lipschitz smooth and $g$ is convex and $\hat{G}$-Lipschitz continuous; (ii) $U$ is $\hat{m}$-strongly convex and $\hat{L}$-Lipschitz smooth. Note that the iteration complexities obtained in \cite{freund2022convergence} are dimension-independent. Note that MLA has biased convergence with a mixing time bound of $\tilde\scrO\left(d/\varepsilon^2\right)$ with more involved assumptions on $U$ and $\varphi$; see \cite{li2022mirror} for more details. The convergence behavior of PULA can also be understood from that of MLA, as it is a particular instance of MLA. 
        
        In \Cref{tab:malmc}, the reported mixing times are based on the total variation distance, and we define the condition number $\kappa\coloneqq\hat{L}/\hat{m}$. We hide logarithmic factors in problem parameters and also omit other logarithmic factors (e.g., precision target $\varepsilon$) for simplicity. 
        
        We refer the readers to \cite{freund2022convergence,chen2023does} for more detailed discussion on the comparison of the iteration complexities and mixing times of these (Metropolis-adjusted) LMC algorithms.

        \begin{table}[t]
            \caption{Comparison of unadjusted Langevin algorithm (ULA); cf.~\cite{freund2022convergence}}
            \label{tab:lmc}
            \centering
            \begin{threeparttable}
            \begin{tabular}{p{1.25cm}p{2cm}p{4.5cm}p{3.5cm}}
                \toprule
                Work &  Criterion & Assumption & Iteration Complexity\tnote{*} \\
                \midrule
                \cite{durmus2019analysis} & $\sfW_2^2$ & (i) & $\tilde\Omega\left(\frac{dM+\hat{G}^2}{\hat{m}^2\varepsilon^2} \right) $  \\[2mm]
                \cite{chatterji2020langevin} & $\sfW_2^2$ & (i) with Gaussian smoothing & $\tilde\Omega\left(\frac{d(M+\hat{G}^2)}{\hat{m}^2\varepsilon} \right)$ \\[2mm]
                \cite{freund2022convergence} & $\sfW_2^2$  & (i) & $\Omega\left(\frac{\hat{G}^2}{\hat{m}^2\varepsilon}\right)$ \\[2mm]
                \cite{cheng2018convergence} & KL & (ii) & $\tilde\Omega\left(\frac{d\hat{L}^2}{\hat{m}^2\varepsilon}\right)$ \\[2mm]
                \cite{freund2022convergence} & KL & (ii) & $\Omega\left(\frac{\hat{L}\tr(\hat{H})}{\hat{m}^2\varepsilon} + \frac{U(0)}{\varepsilon}\right)$ \\[2mm]
                \bottomrule
            \end{tabular}
            \begin{tablenotes}\small
                \item[*] $d$ is the parameter dimension, $\hat{m}$ is the strong convexity parameter of $f$, $M$ is the smoothness parameter of $f$, $\hat{G}$ is the Lipschitz parameter of $g$, $\hat{L}$ is the smoothness parameter of $U$, the matrix $\hat{H}$ is an upper bound of the Hessian of $U$ 
            \end{tablenotes}
            \end{threeparttable}
        \end{table}
        
        \begin{table}[t]
                \caption{Comparison of Metropolis-adjusted Langevin algorithm (MALA); cf.~\cite{chen2023does}}
                \label{tab:malmc}
                \centering
                \begin{threeparttable}
                \begin{tabular}{p{1.25cm}p{3.9cm}p{3.35cm}p{2.75cm}}
                    \toprule
                    Work & Extra Assumption & Initialization\tnote{*} & Mixing Time\tnote{$\dagger$} \\
                    \midrule
                    \cite{roberts1998optimal} & product distribution & warm & $\tilde\scrO(d^{\sfrac{1}3})$\tnote{$\ddagger$} \\
                    \cite{dwivedi2019log,chen2020fast} & N/A & warm or $\sfN(\bx^\star, \hat{L}^{-1}\bI_d)$ & $\tilde\scrO(\max\{d^{\sfrac{1}{2}}\kappa^{\sfrac{3}{2}}, d\kappa\})$ \\
                    \cite{lee2020logsmooth} & N/A &  $\sfN(\bx^\star, \hat{L}^{-1}\bI_d)$ & $\tilde\scrO(d\kappa)$ \\
                    \cite{chewi2021optimal} & N/A & warm & $\tilde\scrO(d^{\sfrac{1}{2}}\kappa^{\sfrac{3}{2}})$ \\
                    \cite{wu2022minimax} & N/A & warm & $\tilde\scrO(d^{\sfrac{1}{2}}\kappa)$\\
                    \cite{altschuler2023faster} & N/A &  $\sfN(\bx^\star, \hat{L}^{-1}\bI_d)$ & $\tilde\scrO(d^{\sfrac{1}{2}}\kappa)$ \\
                    \cite{chen2023does} & strongly Hessian Lipschitz & warm & $\tilde\scrO(d^{\sfrac{3}{7}}\kappa)$ \\
                    \bottomrule
                \end{tabular}
                \begin{tablenotes}\small
                \item[*] $\bx^\star$ denotes the unique mode of the target density 
                \item[$\dagger$] Logarithmic factors in $d$ and $\varepsilon^{-1}$ are hidden 
                \item[$\ddagger$] The dependence on $\kappa\coloneqq\hat{L}/\hat{m}$ is unknown and higher-order derivatives are assumed in \cite{roberts1998optimal} 
                \end{tablenotes}
                \end{threeparttable}
            \end{table}

        \section{Proximal Langevin Monte Carlo Algorithms}
        \label{sec:prox_lmc}
        We now move to the case where the potential $U$ takes a composite form, i.e., $U = f + g$, where $f$ is Lipschitz smooth and $g$ is possibly nonsmooth.     
        Since the LMC algorithms require the smoothness of the potential, one important idea proposed in \cite{pereyra2016proximal,durmus2018efficient} is to approximate the nonsmooth part $g$ of the potential $U$ using its Moreau envelope $g_\lambda$ with a smoothing parameter $\lambda>0$, followed by sampling from the smooth surrogate density $\pi_\lambda \propto \e^{-U_\lambda}$, where $U_\lambda \coloneqq f + g_\lambda$ (with slight abuse of notation of $U_\lambda$ for not representing its Moreau envelope), which converges weakly to the target density $\pi$ as $\lambda\to0$. 
        This idea is reminiscent of its convex optimization counterpart: when minimizing composite functions in the form of $U = f+g$, a large class of proximal splitting algorithms \cite{parikh2014proximal,tutorialscombettes_pesquet,chambolle2016introduction,condat2023proximal,combettes2021fixed} were proposed and have been used widely.  
        In the following, we introduce such proximal LMC algorithms and discuss their similarities to proximal optimization algorithms in more detail. 
        Note that we use constant smoothing parameters $\lambda>0$ when describing the proximal LMC algorithms but nonincreasing smoothing parameter sequences $(\lambda_n)_{n\in\NN}$ can be used in practice.

        \subsection{Moreau--Yosida Unadjusted Langevin Algorithm}
        Proposed in \cite{pereyra2016proximal,durmus2018efficient}, approximating the nonsmooth part $g$ by its Moreau envelope (Moreau--Yosida regularization), the \emph{Moreau--Yosida unadjusted Langevin algorithm} (MYULA) iterates
        \begin{equation}\label{eqn:myula}
        (\forall n\in\NN)\quad \bx_{n+1} = \left( 1 - \frac{\gamma}{\lambda} \right) \bx_n + \frac{\gamma}{\lambda}\prox_{\lambda g}(\bx_n) - \gamma\nabla f(\bx_n) + \sqrt{2\gamma}\,\bxi_n.
        \end{equation}
        A special case is when the step size is equal to the smoothing parameter, i.e., $\lambda = \gamma$, so that we have 
        \[(\forall n\in\NN)\quad \bx_{n+1} = \prox_{\gamma g}(\bx_n) - \gamma\nabla f(\bx_n) + \sqrt{2\gamma}\,\bxi_n. \]
        Furthermore, if $f=0$, then we recover the \emph{proximal unadjusted Langevin algorithm} (P-ULA), which can be viewed as the sampling analogue of the \emph{proximal point algorithm} \cite{rockafellar1976monotone} in convex optimization. 
            
        \subsection{Proximal Gradient Langevin Dynamics}
        Another different yet similar proximal LMC algorithm proposed in \cite{durmus2019analysis} (with stochastic gradient estimates) is the \emph{proximal gradient Langevin dynamics} (PGLD), which iterates
        \begin{equation}\label{eqn:pgld}
            (\forall n\in\NN) \quad \bx_{n+1} = \prox_{\lambda g}(\bx_n) - \gamma\nabla f(\prox_{\lambda g}(\bx_n)) + \sqrt{2\gamma}\,\bxi_n. 
        \end{equation}
        Note that, if we write $\by_n =  \prox_{\lambda g}(\bx_n)$ for all $n\in\NN$, then we get an equivalent update:
        \begin{equation*}
            (\forall n\in\NN) \quad \by_{n+1} = \prox_{\lambda g}\left( \by_n - \gamma\nabla f(\by_n) + \sqrt{2\gamma}\,\bxi_n\right), 
        \end{equation*}
        which can be viewed as the sampling analogue (in $\by_n$) of the \emph{proximal gradient algorithm} \cite{tutorialscombettes_pesquet} (a.k.a.~the \emph{forward-backward algorithm}) in convex optimization when $\lambda=\gamma$. Furthermore, when $f=0$ and $\lambda=\gamma$, PGLD is equivalent to P-ULA.

        \subsection{Proximal MALA and Moreau--Yosida Regularized MALA}
        With the MH correction step, \emph{Moreau--Yosida regularized MALA} (MYMALA) iterates    
        \begin{equation}\label{eqn:mymala}
        (\forall n\in\NN)\; 
        \begin{aligned}
        \tbx_{n+1} &= \left( 1 - \frac{\gamma}{\lambda} \right) \bx_n + \frac{\gamma}{\lambda}\prox_{\lambda g}(\bx_n) - \gamma\nabla f(\bx_n) + \sqrt{2\gamma}\,\bxi_n, \\
        \bx_{n+1} &= b_{n+1}\tbx_{n+1} + (1-b_{n+1})\bx_n,\, b_{n+1} \coloneqq \one_{\Rp}\left(\alpha(\tbx_{n+1}, \bx_n) - u_n\right),  
        \end{aligned}    
        \end{equation}
        where $\alpha$ is defined in \eqref{eqn:MH_prob} and $u_n\iiddist \sfU(0,1)$ for all $n\in\NN$. Again, letting $\lambda=\gamma$ and $f=0$, we recover \emph{proximal MALA} (P-MALA) proposed and studied in \cite{pereyra2016proximal,pillai2022optimal}. 
        
        The full MYMALA was mentioned briefly in \cite{durmus2018efficient}, but was not theoretically studied in detail with convergence guarantees in the literature until recently for the special case of $f=0$: \cite{pillai2022optimal} studied the case where $\lambda=\gamma$ and \cite{crucinio2023optimal} for the general case where $\lambda\ne\gamma$. Notice that, as mentioned in \cite{crucinio2023optimal}, the proximal MALA can also be used for sampling smooth distributions. In this context, the proximity operators associated with log-concave densities can be found in \cite{chaux2007variational}.

        \subsection{Preconditioned Proximal Unadjusted Langevin Algorithm}
        Incorporating the geometric idea with proximal LMC algorithms, an accelerated preconditioned version of P-ULA (PP-ULA) was proposed in \cite{corbineau2019preconditioned}. This involves a preconditioning matrix $\bQ^{-1}\in\bbS_{++}^d$ in the Moreau envelope and the proximity operator, or equivalently, a Bregman--Moreau envelope and a Bregman proximity operator with a Legendre function $\frac12\norm{\cdot}_{\bQ^{-1}}^2$: for $\lambda>0$, the Moreau envelope and proximity operator of $h\in\Gamma_0(\RR^d)$ with respect to the norm induced by $\bQ^{-1}\in\bbS_{++}^d$ are respectively defined as 
        \begin{equation*}
        (\forall\bx\in\RR^d)\quad 
        \begin{aligned}
        h_\lambda^{\bQ}(\bx) &\coloneqq \min_{\by\in\RR^d}\ \left\{ h(\by) + \frac{1}{2\lambda}\norm{\by-\bx}_{\bQ^{-1}}^2\right\}, \\
        \prox_{\lambda h}^{\bQ}(\bx) &\coloneqq \argmin_{\by\in\RR^d}\ \left\{ h(\by) + \frac{1}{2\lambda}\norm{\by-\bx}_{\bQ^{-1}}^2\right\}.
        \end{aligned}
        \end{equation*}
        Note that, similar to the Moreau envelope $h_\lambda$, the preconditioned Moreau envelope $h_\lambda^{\bQ}$ is also differentiable and its gradient (cf.~\eqref{eqn:grad_Bregman_Moreau}) is given by 
        \[(\forall\bx\in\RR^d)\quad \nabla h_\lambda^{\bQ}(\bx) = \frac1\lambda\bQ^{-1}\left(\bx - \prox_{\lambda h}^{\bQ}(\bx)\right). \]
        Consequently, with a preconditioning matrix $\bM\in\bbS_{++}^d$ for the Langevin diffusion, PP-ULA iterates 
        \begin{equation}\label{eqn:pp-ula}
        (\forall n\in\NN) \quad \bx_{n+1} = \bx_n - \frac{\gamma}{\lambda}\bM \bQ^{-1}\left(\bx_n - \prox_{\lambda U}^{\bQ}(\bx_n)\right)  + \sqrt{2\gamma}\,\bM^{\sfrac12}\bxi_n, 
        \end{equation}
        where $\bQ\in\bbS_{++}^d$. Again, for simplicity, letting $\lambda=\gamma$ and $\bM=\bQ$ yields
        \[(\forall n\in\NN) \quad \bx_{n+1} = \prox_{\lambda U}^{\bQ}(\bx_n)  + \sqrt{2\gamma}\,\bQ^{\sfrac12}\bxi_n. \]
        Let us recall that $U = f + g$ where $f$ is Lipschitz differentiable on $\RR^d$. While the proximity operator of the sum of two functions is generally intractable \cite{pustelnik2017proximity}, but when $\lambda$ is small, $\prox_{\lambda g}^{\bQ}(\bx-\lambda\bQ\nabla f(\bx))$ is a good approximation of $\prox_{\lambda U}^{\bQ}(\bx)$ \cite{corbineau2019preconditioned}. Thus, PP-ULA indeed iterates
        \begin{equation*}
        (\forall n\in\NN) \quad \bx_{n+1} = \prox_{\lambda g}^{\bQ}\left( \bx_n - \lambda\bQ\nabla f(\bx_n)\right)   + \sqrt{2\gamma}\,\bQ^{\sfrac12}\bxi_n, 
        \end{equation*}
        which can be viewed as the sampling analogue of the \emph{variable metric forward-backward algorithm} \cite{chouzenoux2014variable}. 
        In addition, if $\prox_{\lambda g}$ is easy to compute, then $\prox_{\lambda g}^{\bQ}$ can be obtained using the dual forward-backward (DFB) algorithm \cite{combettes2011proximity}; see also \cite{corbineau2019preconditioned} for details.

        \subsection{Forward-Backward Unadjusted Langevin Algorithm}
        A potential drawback of using the Moreau envelope in MYULA as a smooth approximation of only the possibly nonsmooth part $g$ is that it often does not maintain the maximum-a-posteriori (MAP) estimator of the original target distribution. Concurrent works \cite{eftekhari2023forward,ghaderi2024smoothing} pointed out such a limitation and proposed the use of the forward-backward envelope instead to address this issue, which leads to the \emph{forward-backward unadjusted Langevin algorithm} (FBULA) and  the \emph{smoothing unadjusted Langevin algorithm} (SULA) .       
        For $\lambda>0$, the forward-backward envelope of $U=f+g\in\Gamma_0(\RR^d)$ \cite{stella2017forward,themelis2018forward} is defined by 
        \[(\forall\bx\in\RR^d)\quad U^{f, g}_\lambda(\bx) \coloneqq \min_{\by\in\RR^d}\ \left\{f(\bx) + \dotp{\nabla f(\bx)}{\by-\bx} + g(\by) + \frac{1}{2\lambda}\euclidnorm{\by-\bx}^2\right\}, \]
        which has an explicit expression involving the gradient of $f$ and the Moreau envelope of $g$: 
        \[(\forall\bx\in\RR^d)\quad U^{f, g}_\lambda(\bx) = f(\bx) - \frac{\lambda}{2}\euclidnorm{\nabla f(\bx)}^2 + g_\lambda(\bx - \lambda\nabla f(\bx)). \]
        For $\lambda\in(0, 1/L)$, where $L>0$ is the Lipschitz constant of $\nabla f$, the forward-backward envelope is continuously differentiable and its gradient is given by 
        \[(\forall\bx\in\RR^d)\quad \nabla U^{f, g}_\lambda(\bx) = \lambda^{-1}\left( \bI_d - \lambda\nabla^2f(\bx)\right) \left(\bx - \prox_{\lambda g}( \bx-\lambda\nabla f(\bx)) \right). \]
        Here we assume the Hessian $\nabla^2f$ is continuous. Then, FBULA iterates
        \begin{equation}\label{eqn:fbula}
            (\forall n\in\NN) \quad \bx_{n+1} = \bx_n - \gamma\nabla U^{f, g}_\lambda(\bx_n)  + \sqrt{2\gamma}\,\bxi_n. 
        \end{equation}
        While motivated by a tool from convex optimization, we are not aware of a corresponding optimization algorithm in the literature except for the case of $\lambda=\gamma$ and $f=0$ in which FBULA is equivalent to P-ULA.

        \subsection{Bregman--Moreau Unadjusted Mirror-Langevin Algorithm}
        Extending the Moreau envelope using the Bregman--Moreau envelope along with the mirror-Langevin algorithm, \cite{lau2022bregman} proposed the \emph{Bregman--Moreau unadjusted mirror-Langevin algorithm} (BMUMLA).     
        Using two possibly different Legendre functions $\varphi$ and $\psi$, BMUMLA iterates             
        \begin{equation}\label{eqn:bmumla}
        	(\forall n\in\NN)\quad\bx_{n+1} = \nabla\varphi^*\left(\nabla\varphi(\bx_n) - \gamma\nabla U_\lambda^\psi(\bx_n)
        		+ \sqrt{2\gamma}\left[ \nabla^2 \varphi(\bx_n) \right]^{\sfrac{1}{2}}\bxi_n \right),  
        \end{equation}
        where $U_\lambda^\psi$ represents either of 
        \begin{equation}\label{eqn:env_potentials}
        	\lU{\lambda}{\psi} \coloneqq f + \lenv_{\lambda, g}^\psi \quad\text{and}\quad \rU{\lambda}{\psi} \coloneqq f + \renv_{\lambda, g}^\psi. 
        \end{equation}
        In particular, when choosing $\lambda=\gamma$, $\varphi=\psi$ and the right Bregman--Moreau envelope, we have 
        \[(\forall n\in\NN)\quad\bx_{n+1} = \nabla\varphi^*\left(\nabla\varphi\left( \rprox{\gamma, g}{\varphi}(\bx_n) \right)  - \gamma\nabla f(\bx_n) + \sqrt{2\gamma}\left[ \nabla^2 \varphi(\bx_n) \right]^{\sfrac{1}{2}}\bxi_n \right),  \]
        which can be viewed as a sampling analogue of the \emph{Bregman proximal gradient algorithm} \cite{van2017forward,bauschke2017descent,bolte2018first,bui2021bregman,chizat2022convergence}.

        \subsection{Primal-Dual Langevin Algorithms}
        Motivated by Chambolle--Pock's \emph{Primal-Dual Hybrid Gradient} (PDHG) algorithm \cite{chambolle2011first}, a first-order primal-dual splitting algorithm, very recently \cite{narnhofer2022posterior} proposed a primal-dual Langevin algorithm, called the \emph{unadjusted Langevin primal-dual algorithm} (ULPDA). In particular, the nonsmooth part of the target potential is now a composition of a nonsmooth function $g$ and a linear operator $\bD\colon\RR^d\to\RR^{d'}$, i.e., $U = f + g \circ\bD$, which widely appears in TV pseudonorms \eqref{eqn:nonsmooth}. Alternatively, the target potential $U$ can be written as     
        \[(\forall\bx\in\RR^d)\quad U(\bx) = f(\bx) + \sup_{\bu\in\RR^{d'}}\,\left\{\dotp{\bD\bx}{\bu} - g^*(\bu)\right\}. \]
        It can be shown that $\bu\in\partial g(\bD\bx)$ satisfies the supremum, where $\partial g$ is the subdifferential of $g$ defined later in \eqref{eqn:subdiff}.     
        Then, choosing $\bx_0\in\RR^d$ and $\bu_0=\tbu_0\in\RR^{d'}$, ULPDA iterates 
        \begin{equation}\label{eqn:ulpda}
        (\forall n\in\NN)\quad
        \begin{aligned}    
        \bx_{n+1} &= \prox_{\gamma f}(\bx_n - \gamma\bD^* \tbu_n) + \sqrt{2\gamma}\,\bxi_n, \\
        \bu_{n+1} &= \prox_{\lambda g^*}(\bu_n + \lambda\bD(2\bx_{n+1}-\bx_n)), \\
        \tbu_{n+1} &= \bu_n + \tau(\bu_{n+1} - \bu_n), 
        \end{aligned}
        \end{equation}
        where the step sizes $(\gamma, \lambda, \tau)$ satisfy $\gamma\lambda\le\specnorm{\bD}^{-2}$ and $\tau\in[0,1]$,  $\specnorm{\bD}\coloneqq\sup_{\bx\in\RR^{d}, \bx\ne\zero}\{\euclidnorm{\bD\bx}/\euclidnorm{\bx}\}$ is the spectral norm of $\bD$, and $\bD^*$ is the adjoint of $\bD$. Note that theoretical convergence properties of ULPDA remain open. 
        
        Another line of recent works regarding proximal LMC algorithms for target potentials with a linear operator in their nonsmooth part is \cite{artigas2022credibility,abry2022temporal,abry2023credibility,fort2023covid19}. It is worth mentioning that primal-dual Langevin algorithms reminiscent of other primal-dual splitting algorithms such as the Loris--Verhoeven \cite{loris2011generalization,drori2015simple,chen2013primal}, Condat--V\~{u} \cite{condat2013primal,vu2013splitting} and Combettes--Pesquet \cite{combettes2012primal} algorithms have yet been explored in the literature.

        \subsection{Comparison of Proximal Langevin Monte Carlo Algorithms}
        In \Cref{tab:prox_lmc}, we provide a summary of the iteration complexities of various (stochastic) proximal LMC algorithms from the existing literature (see also \Cref{sec:sglmc}), where logarithmic factors in the parameter dimension $d$ and the inverse of the precision target $\varepsilon^{-1}$ are hidden. Note that convergence results of proximal LMC algorithms remain underdeveloped compared to the LMC algorithms in \Cref{sec:lmc}. For instance, optimal scaling results for P-MALA are only recently developed in \cite{pillai2022optimal,crucinio2023optimal}. We also consider two cases: (i) $f$ is $\hat{m}$-strongly convex and $M$-Lipschitz smooth and $g$ is convex and $\hat{G}$-Lipschitz continuous; (ii) $U$ is convex and $\hat{M}$-Lipschitz.

        \begin{table}[!t]
                \caption{Comparison of (stochastic) proximal Langevin Monte Carlo algorithms}
                \label{tab:prox_lmc}
                \centering
                \begin{threeparttable}
                \begin{tabular}{p{2.2cm}p{1.5cm}p{4.25cm}p{3.3cm}}
                    \toprule
                    Algorithm &  Criterion & Assumption\tnote{*} & Iteration Complexity\tnote{$\dagger$} \\
                    \midrule
                    MYULA \cite{durmus2018efficient} & TV & (i) & $\tilde\scrO\left( \frac{d}{\varepsilon^2}\right) $\tnote{$\ddagger$} \\[2mm]
                    SPGLD \cite{durmus2019analysis} & KL & (i) with warm start, $\hat{m}\ge0$ & $\tilde\scrO\left(\frac{d+\hat{G}+D^2}{\varepsilon^2}\right)$ \\[2mm]
                    SPGLD \cite{durmus2019analysis} & TV & (i) with warm start, $\hat{m}\ge0$ & $\tilde\scrO\left(\frac{d+\hat{G}+D^2}{\varepsilon^4}\right)$ \\[2mm]
                    SPGLD \cite{durmus2019analysis} & $\sfW_2^2$ & (i) with warm start, $\hat{m}>0$ & $\tilde\scrO\left(\frac{d+\hat{G}+D^2}{\hat{m}\varepsilon^2}\right)$ \\[2mm]
                    SSGLD \cite{durmus2019analysis} & KL & (ii) with warm start & $\tilde\scrO\left(\frac{\hat{M}^2+D^2}{\varepsilon^2}\right)$ \\[2mm]
                    SSGLD \cite{durmus2019analysis} & TV & (ii) with warm start & $\tilde\scrO\left(\frac{\hat{M}^2+D^2}{\varepsilon^4}\right)$ \\[2mm]
                    \bottomrule
                \end{tabular}
                \begin{tablenotes}\small
                    \item[*] Warm start is based on $2$-Wasserstein distance
                    \item[$\dagger$] $d$ is the parameter dimension, $\hat{m}$ is the strong convexity parameter of $f$, $\hat{G}$ is the Lipschitz parameter of $g$, $D$ is an upper bound on the variance on the stochastic subgradient
                    \item[$\ddagger$] Dependence on $\hat{m}$, $M$ and $\hat{G}$ is hidden
                \end{tablenotes}
                \end{threeparttable}
            \end{table}
       
       \section{Stochastic Gradient Langevin Monte Carlo Algorithms}
       \label{sec:sglmc}
       In large-scale applications involving very large datasets such as Bayesian deep learning, the exact gradient of the smooth part of the potential would usually be very computationally expensive. In a similar vein to stochastic gradient methods in optimization, the gradient  
       $\nabla f$ of the Lipschitz smooth part $f\in\Gamma_0(\RR^d)$ is usually replaced by its unbiased estimate $\hat{\nabla}f$ using stochastic approximation \cite{robbins1951}. Suppose that there exists a measurable space $(\sfZ, \calZ)$ and a probability measure $\calD$ on $(\sfZ, \calZ)$, where $\sfZ\subseteq\RR^m$. A stochastic gradient $\hat\nabla f \colon \RR^d\times\sfZ \to \RR^d$ of $f$ is a measurable function which satisfies
       \[(\forall\bx\in\RR^d)\quad \Ex_{\bz\sim\calD}\left[\hat{\nabla}f(\bx, \bz)\right] = \nabla f(\bx). \]   
       ULA with stochastic gradients is known as \emph{stochastic gradient Langevin dynamics} (SGLD) \cite{welling2011bayesian}, which is also used in the context of optimization for nonconvex optimization \cite{raginsky2017nonconvex}. Developing extensions of SGLD for sampling multimodal distributions or nonconvex optimization is an active area of research, see e.g., \emph{cyclical SGLD} \cite{zhang2020cyclical} and \emph{contour SGLD} \cite{deng2020contour,deng2022adaptively,deng2022interacting}. 
        
        For nonsmooth composite target potentials, \cite{durmus2019analysis} proposed the use of stochastic gradients in PGLD, leading to \emph{stochastic proximal gradient Langevin dynamics} (SPGLD), which iterates            
    	\begin{equation}\label{eqn:spgld}
            (\forall n\in\NN) \quad \bx_{n+1} = \prox_{\lambda g}(\bx_n) - \gamma\hat\nabla f(\prox_{\lambda g}(\bx_n), \bz_n) + \sqrt{2\gamma}\,\bxi_n, 
        \end{equation}
        where $(\bz_n)_{n\in\NN}$ is a sequence of \iid random variables distributed according to $\calD$. 
        
        Other than SPGLD, \cite{durmus2019analysis} also proposed to use a subgradient of the nonsmooth part $g$ instead of its proximity operator. In particular, if $g\in\Gamma_0(\RR^d)$, its \emph{subdifferential} $\partial g$ is defined by 
        \begin{equation}\label{eqn:subdiff}
        (\forall\bx\in\RR^d)\quad \partial g(\bx) \coloneqq \left\lbrace \bu\in\RR^d : (\forall \by\in\RR^d)\;\; g(\bx) + \dotp{\bu}{\by-\bx} \le g(\by)  \right\rbrace, 
        \end{equation}
        which is nonempty. For $\bx\in\RR^d$, any element of $\partial g(\bx)$ is referred to as a \emph{subgradient} of $g$ at $\bx$. Then, the \emph{stochastic subgradient Langevin dynamics} (SSGLD) iterates
        \begin{equation}\label{eqn:ssgld}
            (\forall n\in\NN)\quad \bx_{n+1} = \bx_n - \gamma\left( \hat\nabla f(\bx_n, \bz_n) + \partial g(\bx_n)\right) + \sqrt{2\gamma}\,\bxi_n. 
        \end{equation}
        We also refer readers to \cite{salim2019stochastic,salim2020primal,vidal2020maximum1,de2020maximum2} for further works on stochastic gradient proximal Langevin algorithms.

    	\section{Numerical Simulations}
        \label{sec:sim}
        Numerical simulations are implemented with \textsf{Python}. For LMC algorithms with deterministic full gradients, the libraries \textsf{numpy} \cite{numpy2020array} and \textsf{scipy} \cite{scipy2020} are used. 
        Computation of proximity operators can be performed with the libraries \textsf{proxop} \cite{chierchia2022prox} and \textsf{PyProximal} \cite{ravasi2023pyproximal}. 
        For image processing, we use \textsf{scikit-image} \cite{scikit-image} and \textsf{PyLops} \cite{ravasi2020pylops}. 
        We also use the libraries 
        \textsf{matplotlib} \cite{matplotlib2007}
        and \textsf{SciencePlots} \cite{SciencePlots} for visualization purpose. 
        The library \textsf{POT: Python Optimal Transport} \cite{flamary2021pot} is used to compute Wasserstein distances between empirical distributions. 
        All simulations were performed with a Linux Ubuntu 18.04 workstation with Intel Xeon Silver 4214 @ 48$\times$3.2GHz, 755GB RAM and four NVIDIA GeForce RTX 2080Ti GPUs.
        The code repository for the experiments is publicly available at \url{https://github.com/timlautk/lmc-atomi}.

    	\subsection{Mixtures of Gaussians}
        \label{subsec:gaussian_mixture}
    	For easier inspection, we focus on the case of $d = 2$ in simulations, with different number of Gaussians in the mixtures $K\in\set{5}$. Note that the case of $K=1$ corresponds to the case of log-concave sampling. 
        For simplicity, we assign equal weights to the mixtures, i.e., $\bomega = \One_K/K$. 
        The mixtures of Gaussians have different locations and shapes, with mean vectors $\bmu_1 = (0,0)^\top$, $\bmu_2 = (-2,3)^\top$, $\bmu_3 = (2,-3)^\top$, $\bmu_4 = (3, 3)^\top$, $\bmu_5 = (-2, -2)^\top$, and covariance matrices  
        \begin{align*}
        \bSigma_1 &= \begin{pmatrix}
           		1 & -0.5 \\ -0.5 & 1
           	\end{pmatrix}, \quad 
           	\bSigma_2 = \begin{pmatrix}
           		0.5 & 0.2 \\ 0.2 & 0.7 
           	\end{pmatrix}, \quad
           	\bSigma_3 = \begin{pmatrix}
           		0.5 & 0.1 \\ 0.1 & 0.9
           	\end{pmatrix}, \\
           	\bSigma_4 &= \begin{pmatrix}
           		0.8 & 0.02 \\ 0.02 & 0.3
           	\end{pmatrix}, 
           	\bSigma_5 = \begin{pmatrix}
           		1.2 & 0.05 \\ 0.05 & 0.8 
           	\end{pmatrix}.
        \end{align*}
        We use the last four mean-covariance pairs for $K=4$. 
        We choose multiple constant step sizes $\gamma_n = \gamma\in\{0.01, 0.05, 0.1\}$ for all $n\in\NN$ and  generate 10000 samples for each of the following five algorithms: ULA \eqref{eqn:ULA}, MALA \eqref{eqn:MALA}, PULA \eqref{eqn:p_ula} with a constant preconditioning matrix, inverse Hessian PULA (IHPULA) and MLA \eqref{eqn:HRLMC}.     
        For preconditioned ULA, we first use a constant preconditioning matrix 
        \begin{equation}\label{eqn:precond_mat}
        \bM = \begin{pmatrix}
            1 & 0.1 \\ 0.1 & 0.5
            \end{pmatrix} \in \bbS_{++}^2.
        \end{equation}
        While we also want to use a varying conditioning matrix such as the inverse Hessian of $U$, since $U$ is nonconvex, its Hessian fails to be positive definite for all $\bx\in\RR^d$ and hence not invertible. Thus, the following preconditioning matrix is used: 
        $\bM(\bx) = [\nabla^2 U(\bx) + (|\lambda_{\min}(\nabla^2 U(\bx))|+\varepsilon)\bI_d]^{-1}$, where $\lambda_{\min}(\bA)$ represents the smallest eigenvalue of the matrix $\bA$ and $\varepsilon=0.05$ is a small positive constant for numerical stability. For MLA, we use the $\bbeta$-hyperbolic entropy as the Legendre function, defined through 
        \[(\forall\bx\in\RR^d)\quad\varphi_{\bbeta}(\bx) \coloneqq \sumd \left[x_i\arsinh(x_i/\beta_i) - \sqrt{x_i^2 + \beta_i^2} \right] , \]
        where $\bbeta\in\Rpp^d$ and 
        \begin{equation*}
        (\forall\bx\in\RR^d)\quad
        \begin{aligned}
            \nabla\varphi_{\bbeta}(\bx) &= \left( \arsinh(x_i/\beta_i)\right)_{1\le i\le d} , \\
            \nabla\varphi_{\bbeta}^*(\bx) &= \left( \beta_i\sinh(x_i)\right)_{1\le i\le d}.
        \end{aligned}    
        \end{equation*}
        For simplicity, we choose $\bbeta=(0.7, 0.3)^\top$.         
        We visualize the samples by plotting their kernel density estimates (KDEs) along with the true density in \Cref{fig:gaussians}. Additional results can be found in Appendix 2. We observe that sampling becomes much harder when the number of Gaussians increases as the potentials become more nonconvex.

        \begin{figure}[htbp]
            \centering
            \begin{subfigure}[h]{\textwidth}
                \centering
                \includegraphics[height=.3\textheight]{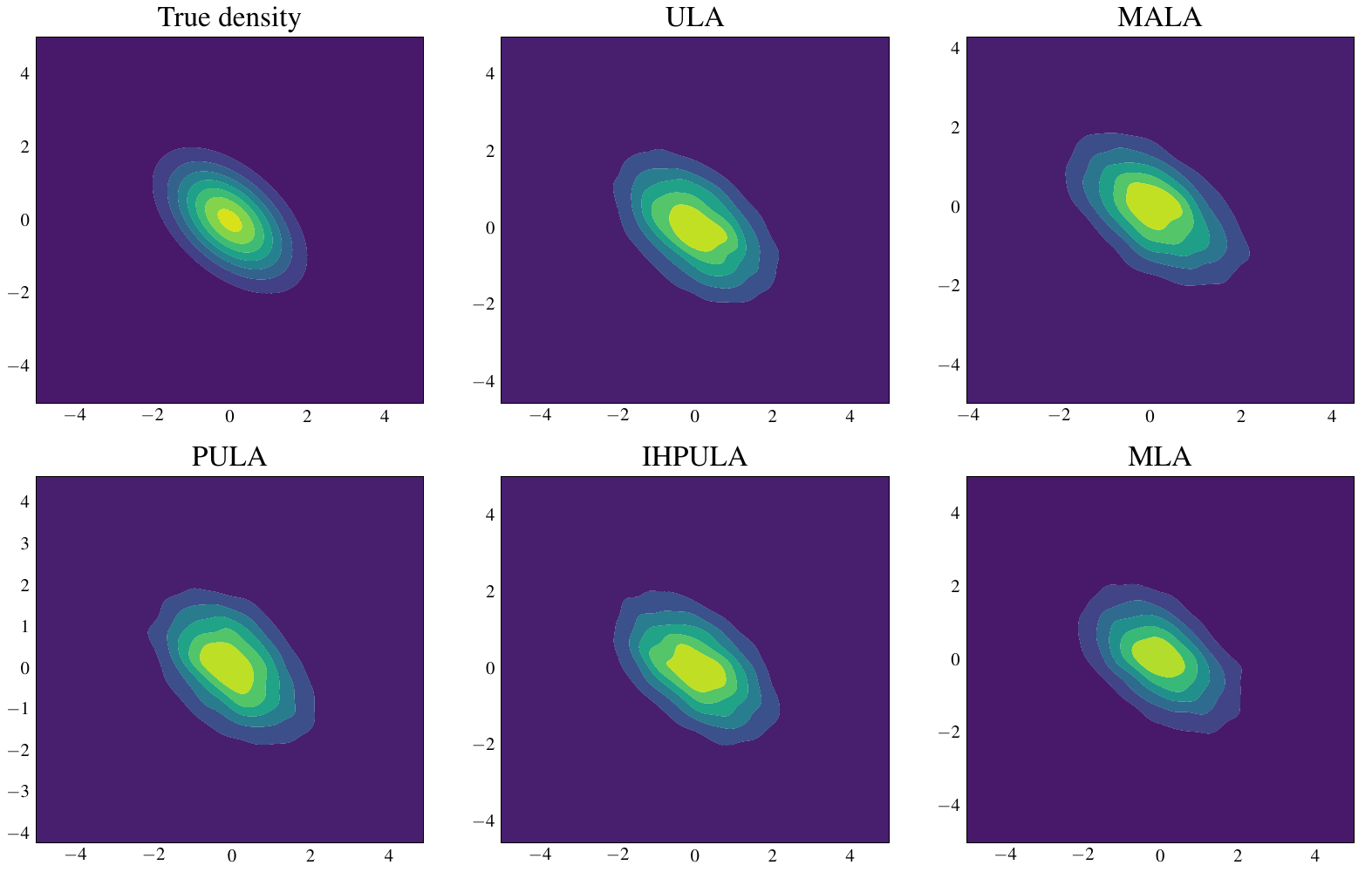}
                \caption{$K=1$}
            \end{subfigure}
            \par\vspace{2mm}
            \begin{subfigure}[h]{\textwidth}
                \centering
                \includegraphics[height=.3\textheight]{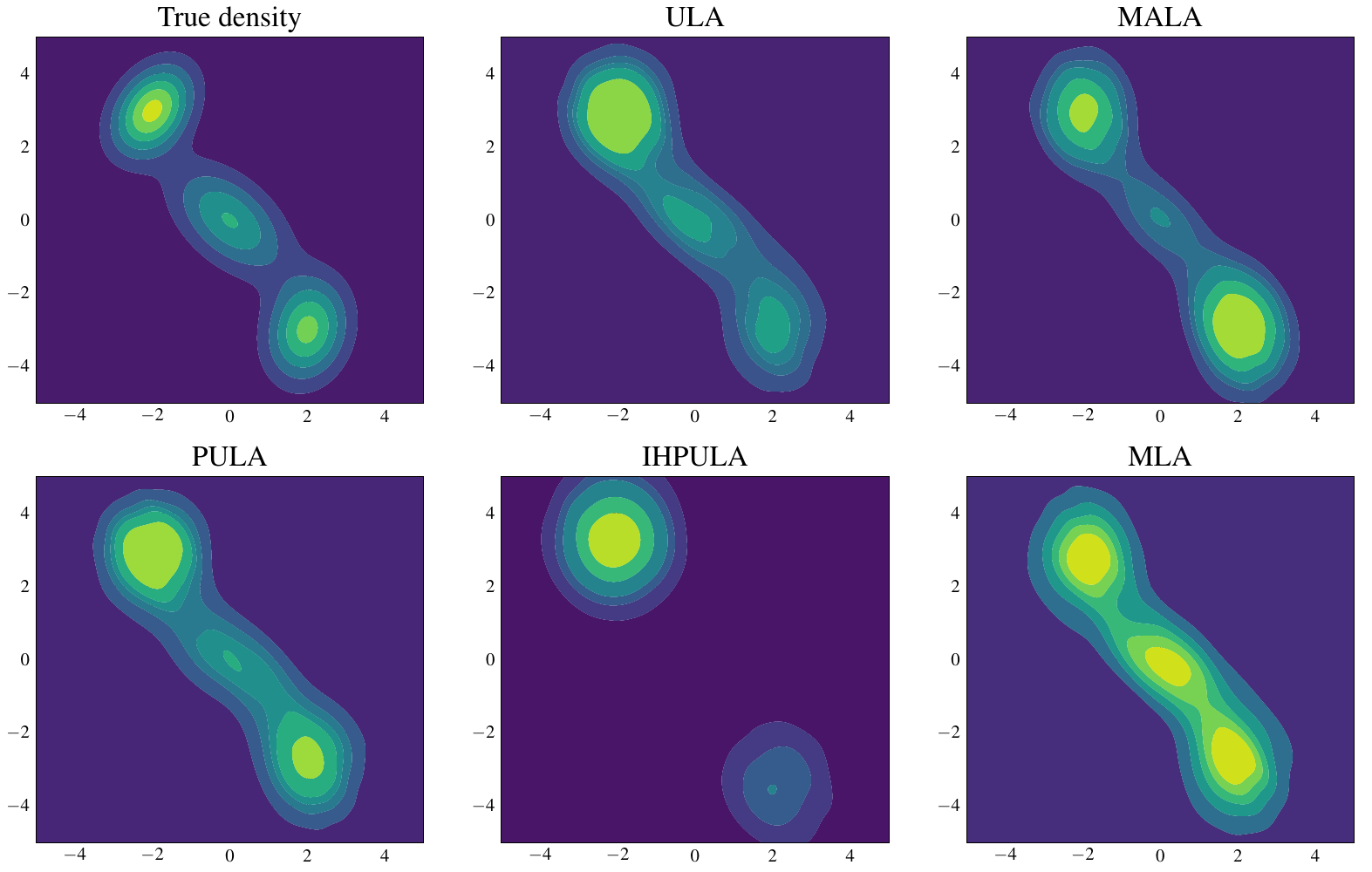}
                \caption{$K=3$}
            \end{subfigure}
            \par\vspace{2mm}
            \begin{subfigure}[h]{\textwidth}
                \centering
                \includegraphics[height=.3\textheight]{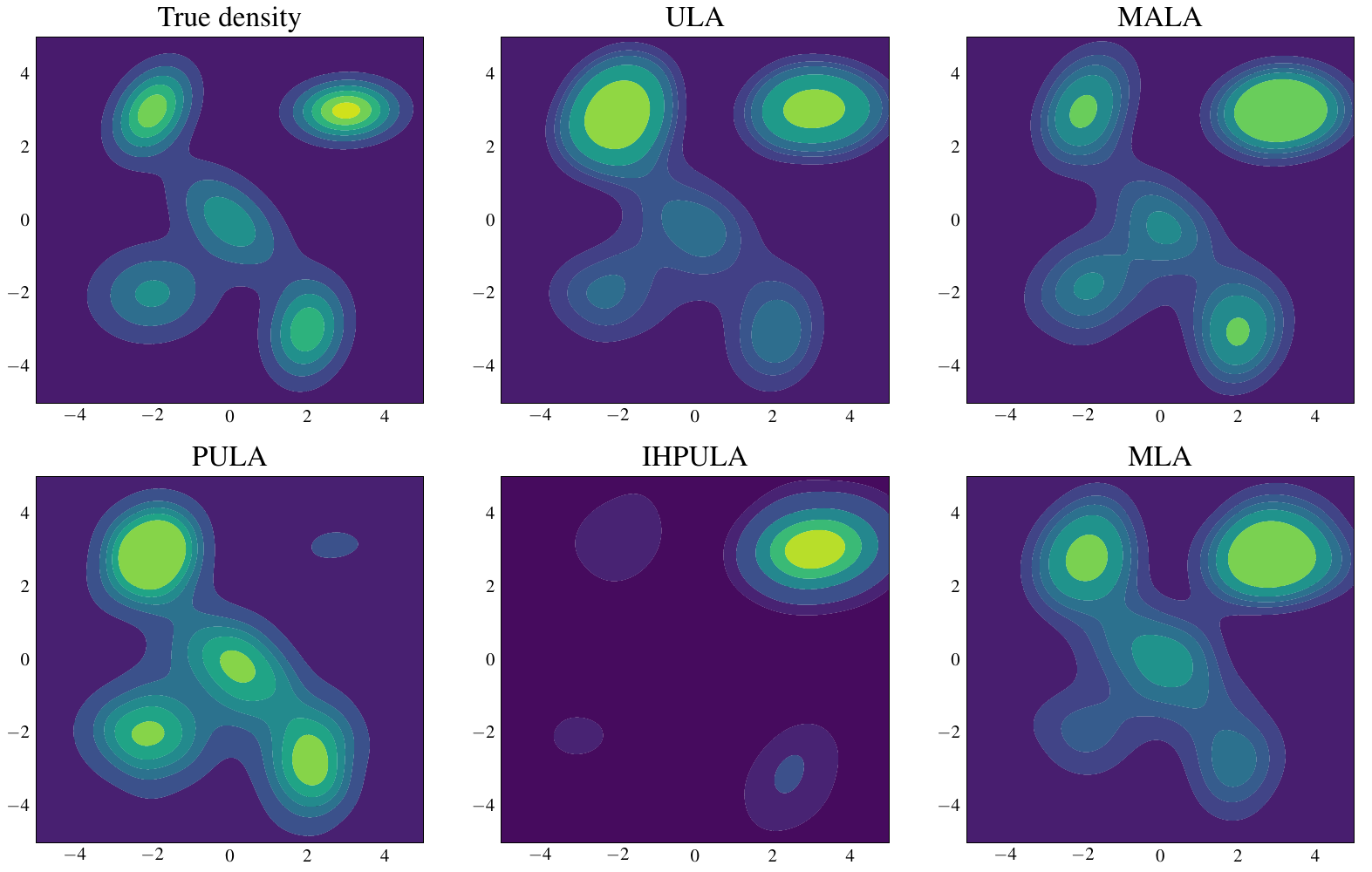}
                \caption{$K=5$}
            \end{subfigure}
            \par\vspace{-2mm}
            \caption{Mixture of $K$ Gaussians with step size $\gamma = 0.1$}
            \label{fig:gaussians}
        \end{figure}
        
\begin{figure}[htbp]
            \centering
            \begin{subfigure}[h]{0.45\textwidth}
                \centering
                \includegraphics[width=\textwidth]{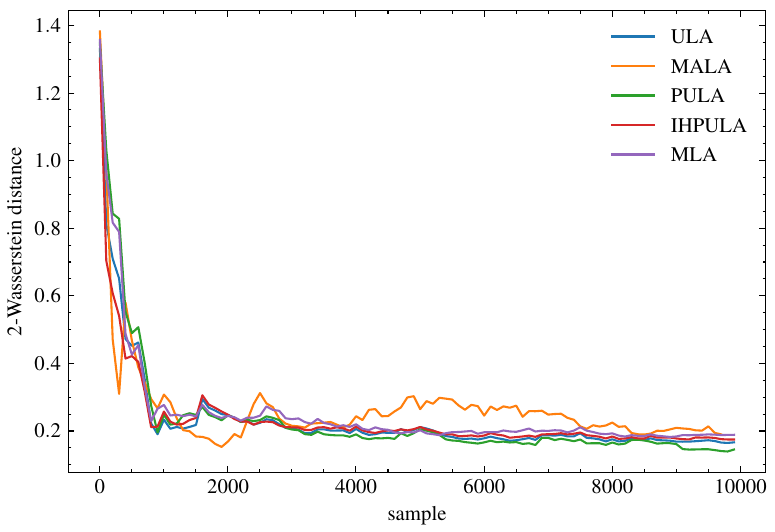}
                \caption{$K=1$}
            \end{subfigure}
            \hfill
            \begin{subfigure}[h]{0.45\textwidth}
                \centering
                \includegraphics[width=\textwidth]{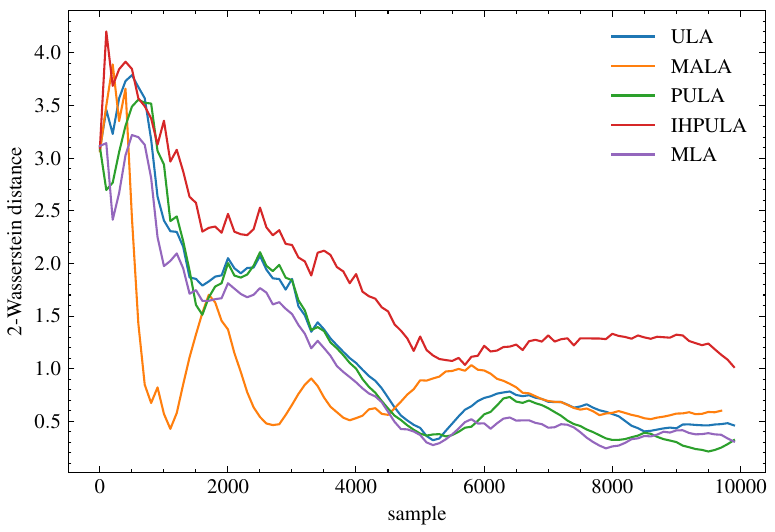}
                \caption{$K=3$}
            \end{subfigure}
            \vfill
            \begin{subfigure}[h]{0.45\textwidth}
                \centering
                \includegraphics[width=\textwidth]{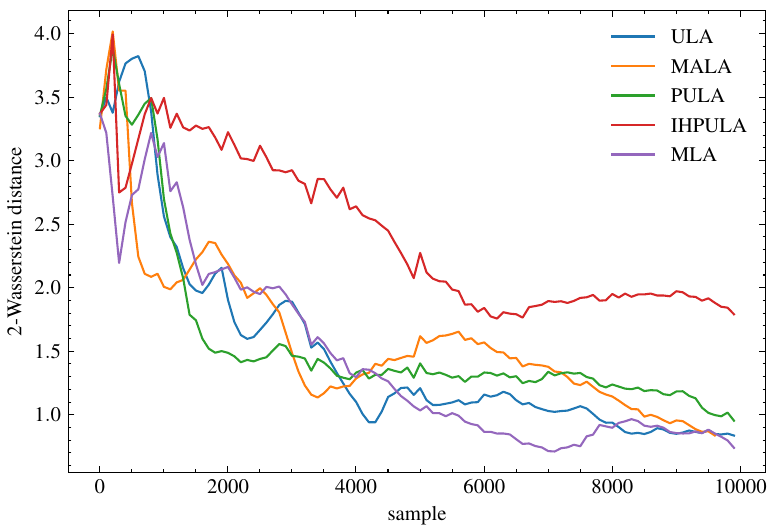}
                \caption{$K=5$}
            \end{subfigure}
            \caption{$2$-Wasserstein distances between generated samples by LMC algorithms and true samples of mixture of $K$ Gaussians with step size $\gamma = 0.1$}
            \label{fig:gaussians_wass}
        \end{figure}

        We also plot the $2$-Wasserstein distances between the 10000 generated samples and 10000 true samples generated from the Gaussian mixtures in \Cref{fig:gaussians_wass,fig:gaussians_wass_1,fig:gaussians_wass_2,fig:gaussians_wass_3}. The true samples are generated by first uniformly sampling one of the Gaussians in the mixtures and then sampling from this corresponding Gaussian. The Wasserstein distances are computed whenever an additional 100 samples are generated. We observe that the convergence behaviors of the LMC algorithms are close for the log-concave case (i.e., $K=1$), while the gaps for the non-log-concave cases (i.e., $K>1$) are large among different algorithms, hinting the importance of carefully choosing step sizes and preconditioning matrices in (preconditioned) LMC algorithms under such settings. 
        
        \subsection{Mixtures of Laplacians}
        \label{subsec:laplacian_mixture}
        In addition to nonconvexity, we further consider nonsmoothness of the target potential. 
        A natural choice extending Gaussian mixtures is mixtures of Laplacians. Note that proximal LMC algorithms such as MYULA cannot be directly applied due to the mixture structure of the target potential. However, following a similar strategy in MYULA, we propose to approximate the $\ell_1$-norms in the component Laplacian densities \eqref{eqn:laplacian} by their Moreau envelopes. i.e., letting $g_k(\bx) \coloneqq \alpha_k\onenorm{\bx - \bmu_k}$ for all $\bx\in\RR^d$, we consider the surrogate density for each component
        \[(\forall \bx\in\RR^d)\quad p_k^\lambda(\bx) \coloneqq \frac{\alpha_k^d}{2^d} \exp\left\{-g_k^\lambda(\bx)\right\},\]
        where $g_k^\lambda$ is the Moreau envelope of $g_k$ with a smoothing parameter $\lambda>0$. Then we instead sample from the surrogate density for the mixture distribution $p^\lambda = \sumK\omega_k p_k^\lambda$, with the potential $U^\lambda = - \log p^\lambda$. 
        
        We fix $\bmu_k$ to be the same as the previous section, $\alpha_k=\alpha=0.5$ for each $k\in\{1,\ldots,K\}$, and choose multiple constant step sizes $\gamma_n = \gamma\in\{0.05, 0.1, 0.15\}$ and smoothing parameters $\lambda_n=\lambda\in\{0.1,0.5,1\}$ for all $n\in\NN$. Due to the nonsmoothness, more samples are required for convergence so we generate 50000 samples using each of the following five algorithms: ULA, MALA, PULA \eqref{eqn:p_ula} with a constant preconditioning matrix \eqref{eqn:precond_mat}, and MLA \eqref{eqn:HRLMC}. We do not use the inverse Hessian PULA since we want to avoid the computation of the gradient of a proximity operator. We use the same preconditioning matrix $\bM$ in PULA and the same $\bbeta$ in MLA. 
        
        We visualize the samples by plotting their kernel density estimates (KDEs) along with the true density and the smoothed density in \Cref{fig:laplacians}. Additional results can be found in Appendix 2. We observe that the choice of step sizes matters more than the choice of smoothing parameters of Moreau envelopes.

        \begin{figure}[htbp]
            \centering
        	\begin{subfigure}[h]{\textwidth}
                \centering
                \includegraphics[height=.3\textheight]{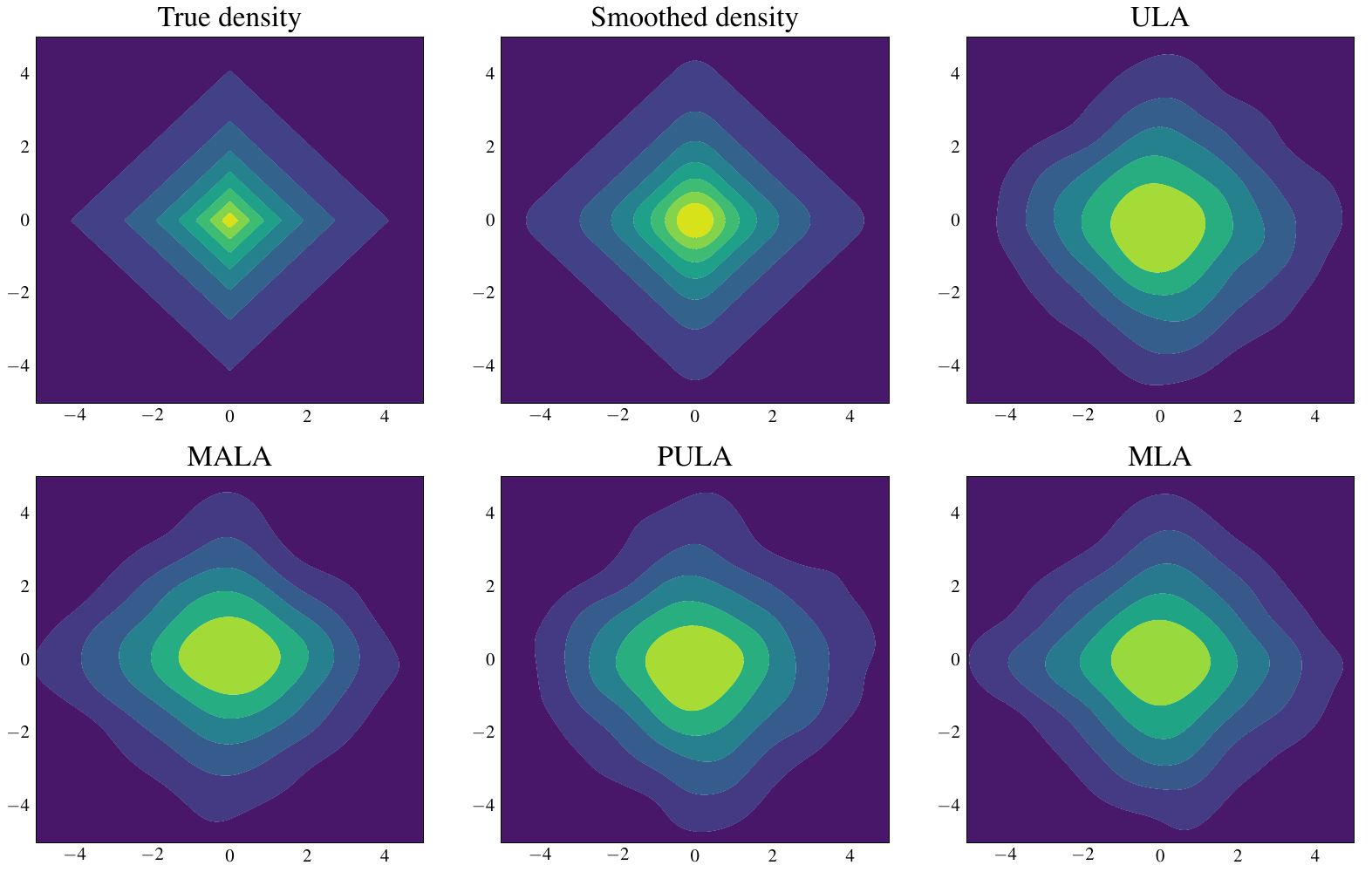}
                \caption{$K=1$}
            \end{subfigure}      
            \par\vspace{2mm}
            \begin{subfigure}[h]{\textwidth}
                \centering
                \includegraphics[height=.3\textheight]{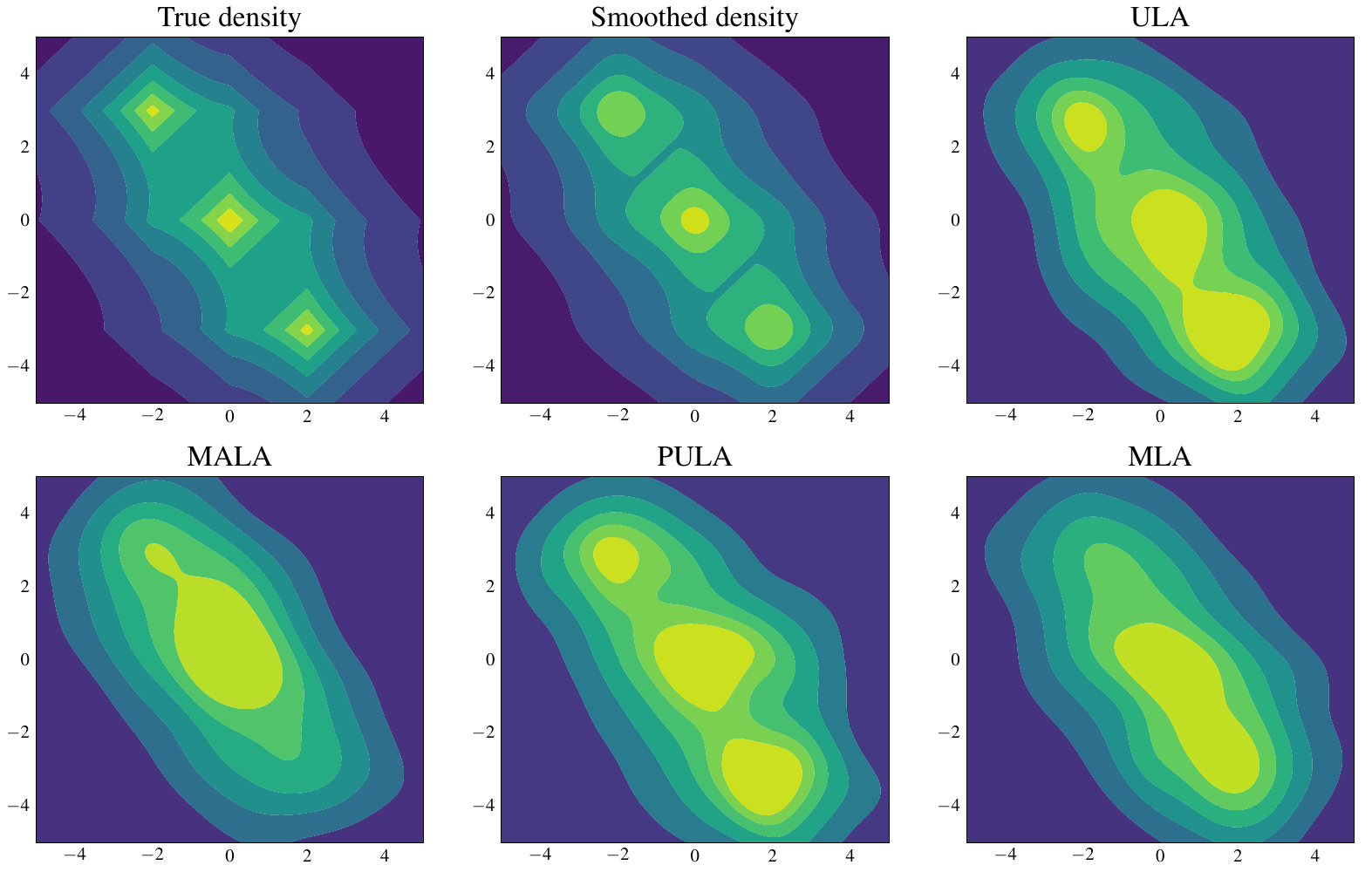}
                \caption{$K=3$}
            \end{subfigure}
            \par\vspace{2mm}
            \begin{subfigure}[h]{\textwidth}
                \centering
                \includegraphics[height=.3\textheight]{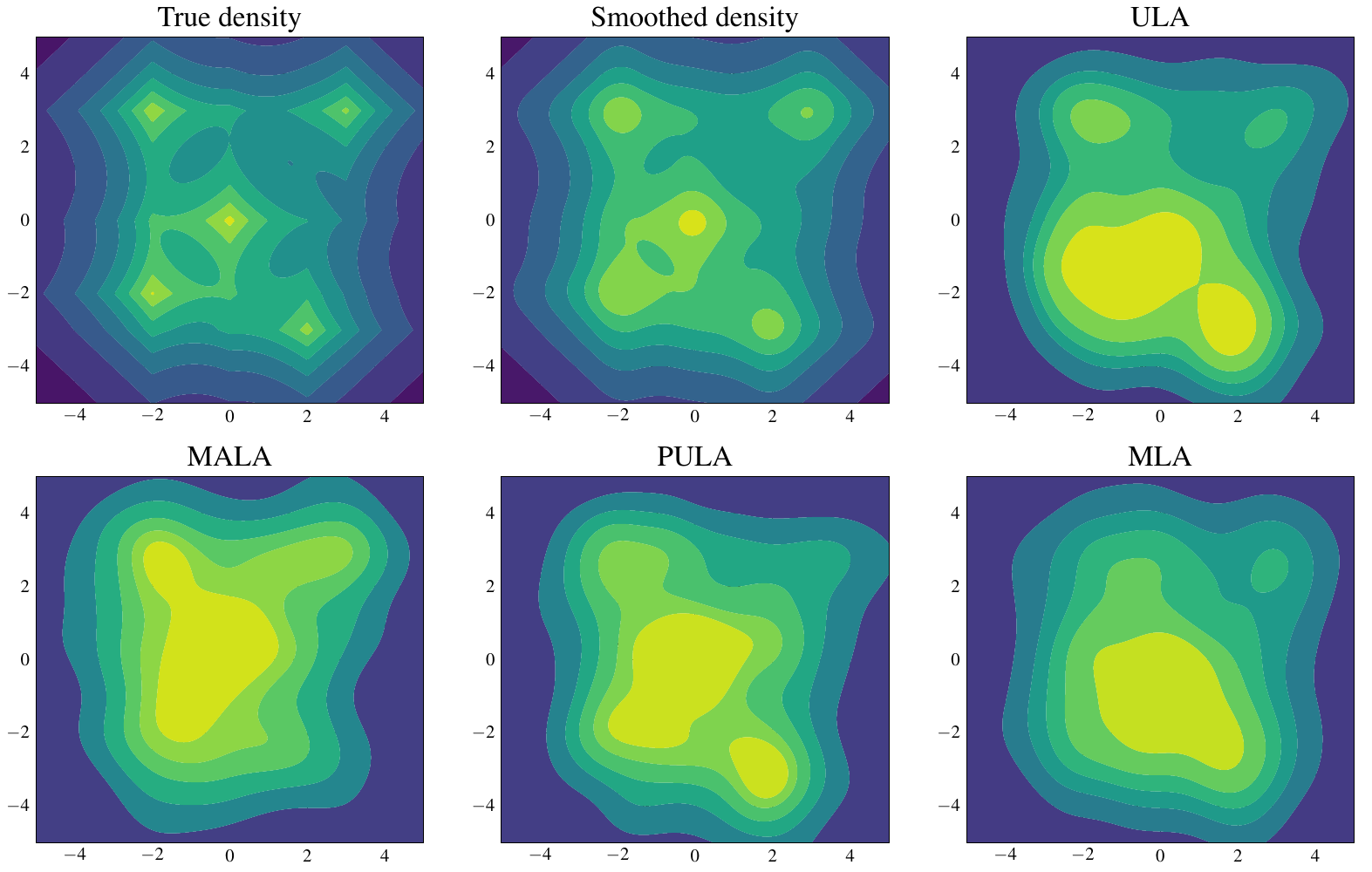}
                \caption{$K=5$}
            \end{subfigure}
            \par\vspace{-2mm}
            \caption{Mixture of $K$ Laplacians with step size and smoothing parameter pair $(\gamma, \lambda)=(0.1, 1)$}
            \label{fig:laplacians}
        \end{figure}
        
\begin{figure}[htbp]
            \centering
            \begin{subfigure}[h]{0.45\textwidth}
                \centering
                \includegraphics[width=\textwidth]{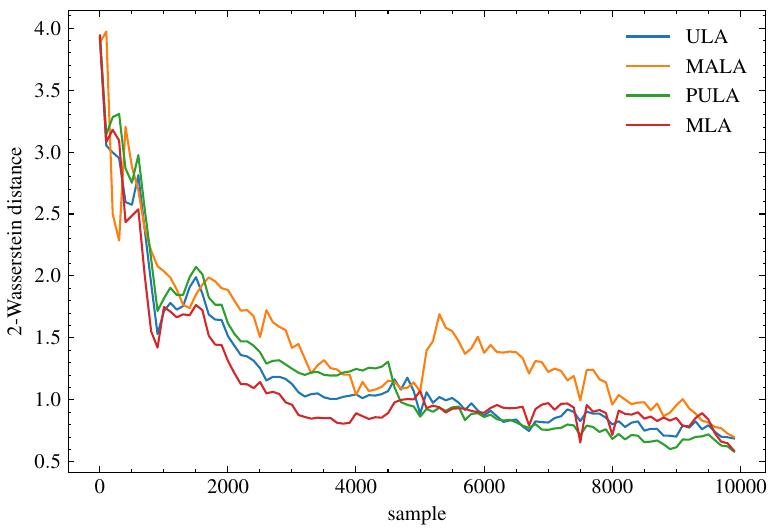}
                \caption{$K=1$}
            \end{subfigure}
            \hfill
            \begin{subfigure}[h]{0.45\textwidth}
                \centering
                \includegraphics[width=\textwidth]{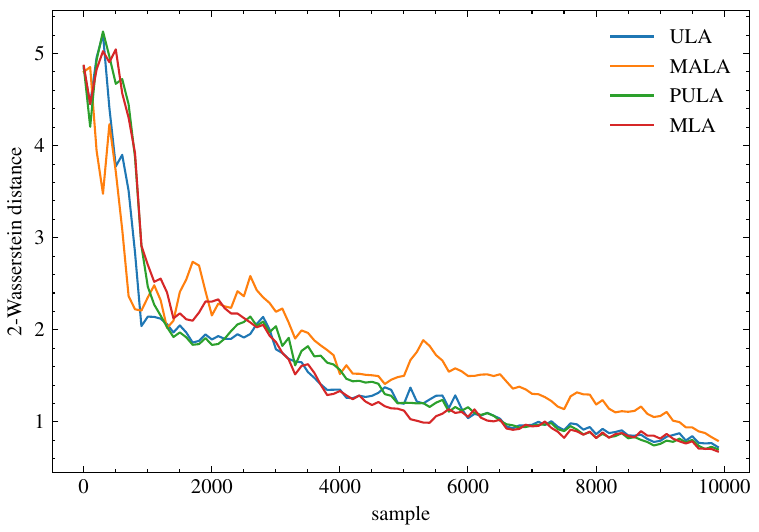}
                \caption{$K=3$}
            \end{subfigure}
            \vfill
            \begin{subfigure}[h]{0.45\textwidth}
                \centering
                \includegraphics[width=\textwidth]{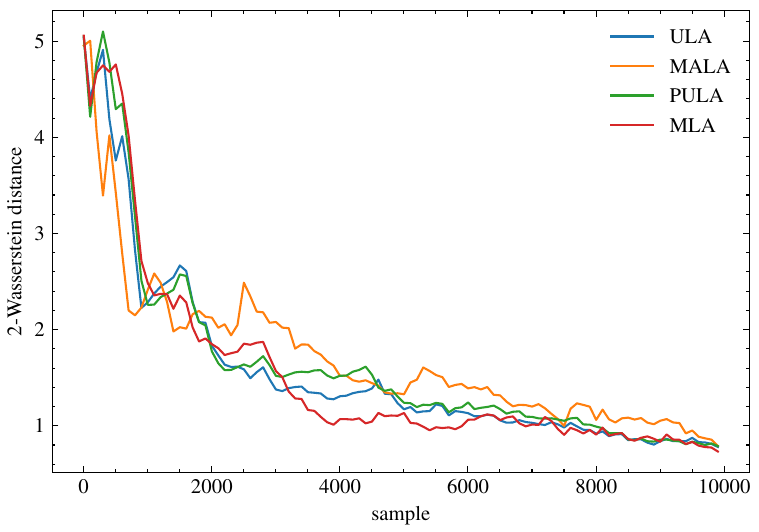}
                \caption{$K=5$}
            \end{subfigure}
            \caption{$2$-Wasserstein distances between generated samples by LMC algorithms and true samples of mixture of $K$ Laplacians with step size and smoothing parameter pair $(\gamma, \lambda)=(0.1, 1)$}
            \label{fig:laplacians_wass}
        \end{figure}

        We also plot the $2$-Wasserstein distances between the first 10000 generated samples and 10000 true samples generated from the Laplacian mixtures in \Cref{fig:laplacians_wass,fig:laplacians_wass_1,fig:laplacians_wass_2,fig:laplacians_wass_3,fig:laplacians_wass_4,fig:laplacians_wass_5,fig:laplacians_wass_6,fig:laplacians_wass_7,fig:laplacians_wass_8,fig:laplacians_wass_9}. The true samples are generated by first uniformly sampling one of the Laplacians in the mixtures and then sampling from this corresponding Laplacian. The Wasserstein distances are computed whenever an additional 100 samples are generated. We observe that the gaps in the convergence of the samples generated by different LMC algorithms are small, hinting that these algorithms might produce samples of similar quality under such settings. 
        
        \subsection{Mixtures of Gaussians with Laplacian Priors}
        In this case, the target potential is in a nonsmooth composite form, so we can apply the proximal LMC algorithms, namely PGLD \eqref{eqn:pgld}, MYULA \eqref{eqn:myula}, MYMALA \eqref{eqn:mymala}, PP-ULA \eqref{eqn:pp-ula}, FBULA \eqref{eqn:fbula}, left BMUMLA \eqref{eqn:bmumla}, each with 50000 samples. 
        We use the same Gaussian mixture likelihood as in \Cref{subsec:gaussian_mixture}, 
        and the Laplacian prior has a location parameter $(0,0)^\top$ and a scale parameter $\alpha=0.15$. In PP-ULA, we use 
        \[\bQ = \begin{pmatrix}
        1 & 0.1 \\ 0.1 & 1.5
        \end{pmatrix},\]
        and the same preconditiong matrix $\bM$ \eqref{eqn:precond_mat}. 
        We also use the hyperbolic entropy for the Bregman--Moreau envelope in BMUMLA, with $\bbeta' = (0.8,0.2)^\top$. We choose multiple constant step sizes $\gamma_n = \gamma\in\{0.05, 0.15, 0.25\}$ and smoothing parameters $\lambda_n=\lambda\in\{0.25,0.5,1\}$ for all $n\in\NN$. 
        
        We visualize the samples by plotting their kernel density estimates (KDEs) along with the true density and the smoothed density in \Cref{fig:gaussians_laplacians}. Additional results can be found in Appendix 2. We observe that the introduction of nonsmooth priors has deteriorated the convergence of the proximal LMC algorithms. Note that FBULA is not numerically stable for $K\ge2$ since $f$ is nonconvex and thus its Hessian $\nabla^2 f$ is not positive semidefinite, which appears in the gradient of the forward-backward envelope. We also observe that too much smoothing of the Laplacian priors might lead to worse sampling performance. 

        \begin{figure}[htbp]
                \centering
            \begin{subfigure}[h]{\textwidth}
                \centering
                \includegraphics[height=.3\textheight]{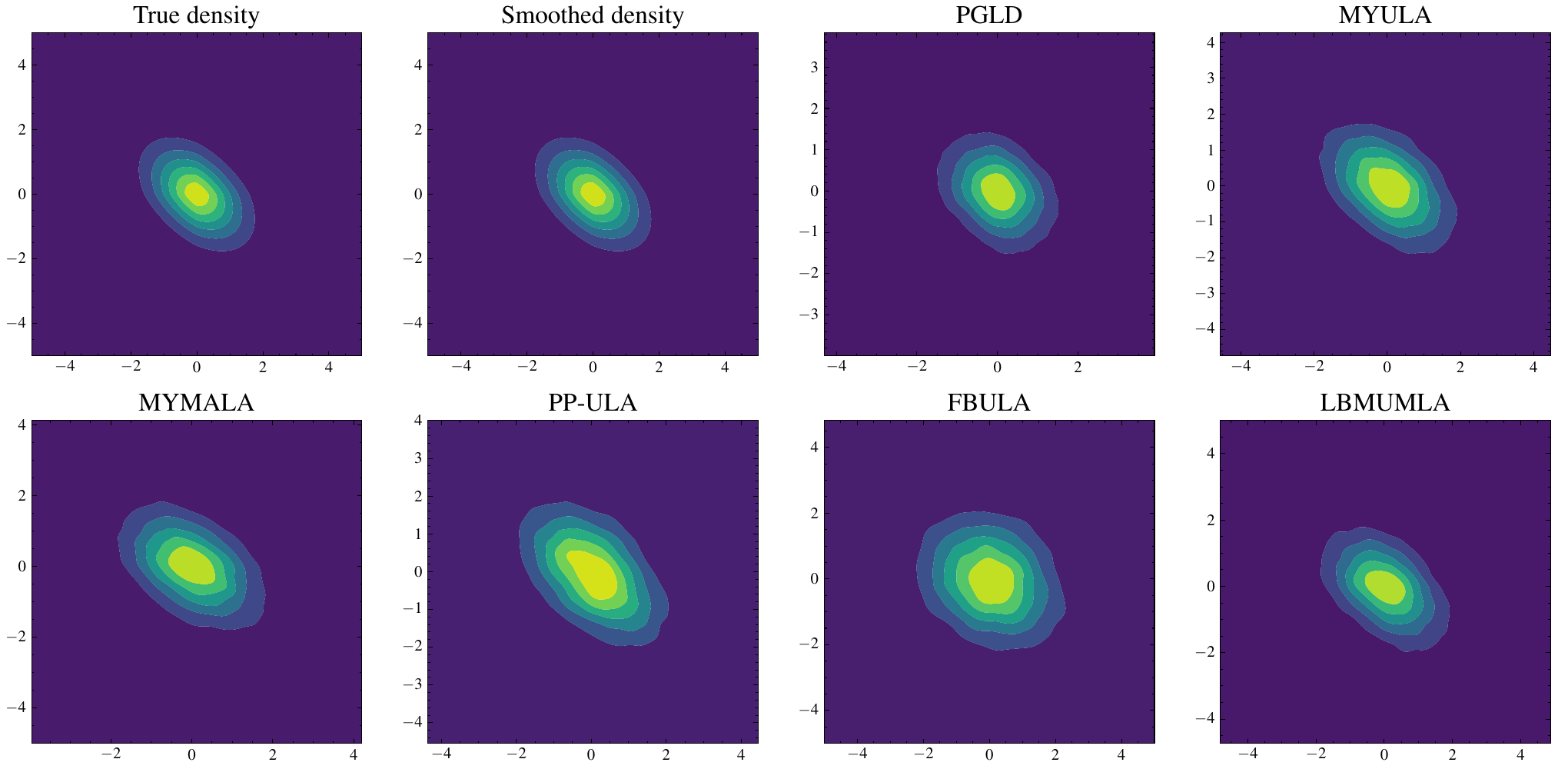}
                \caption{$K=1$}
            \end{subfigure}
            \par\vspace{2mm}
            \begin{subfigure}[h]{\textwidth}
                \centering
                \includegraphics[height=.3\textheight]{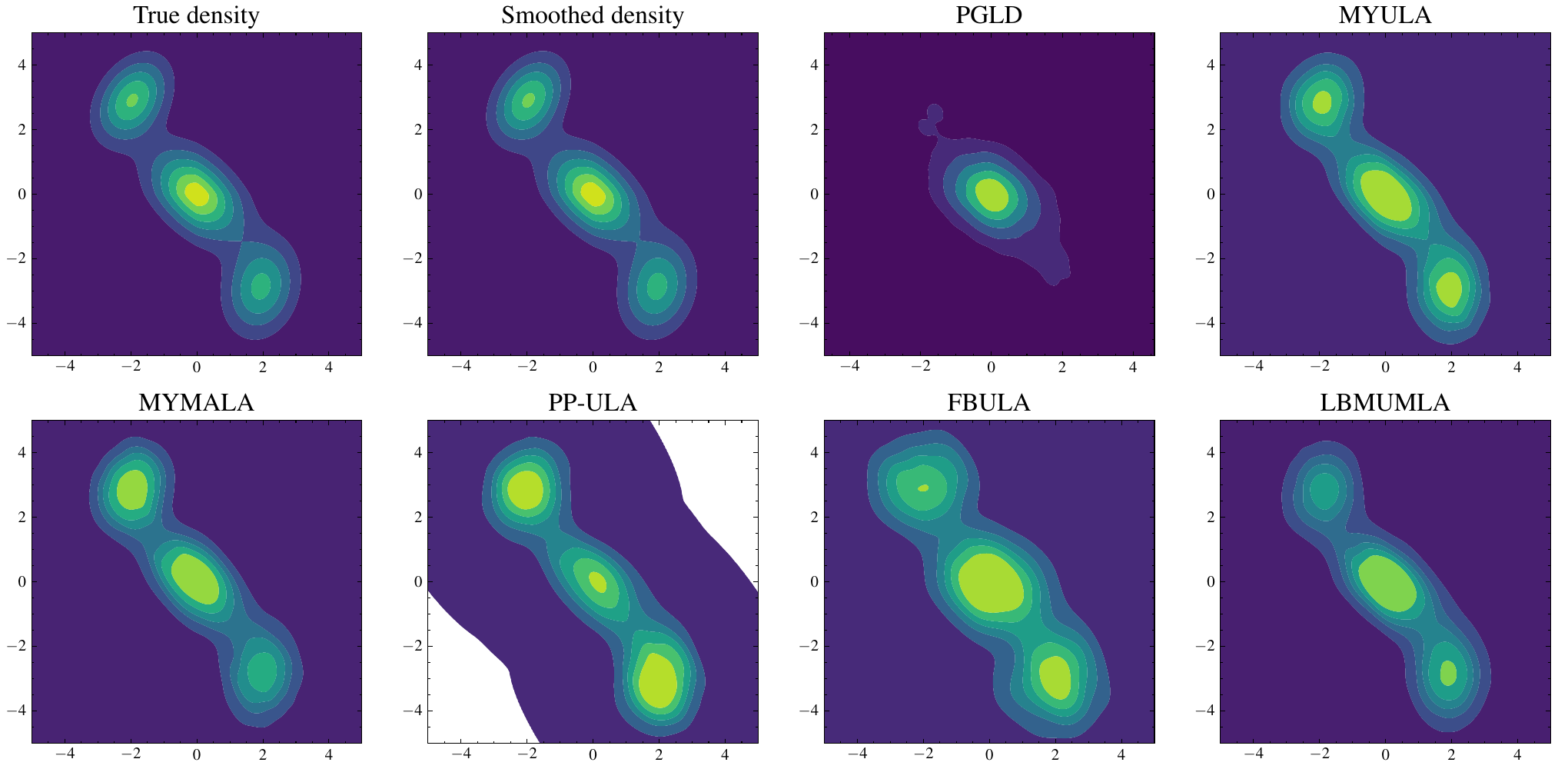}
                \caption{$K=3$}
            \end{subfigure}
            \par\vspace{2mm}
            \begin{subfigure}[h]{\textwidth}
                \centering
                \includegraphics[height=.3\textheight]{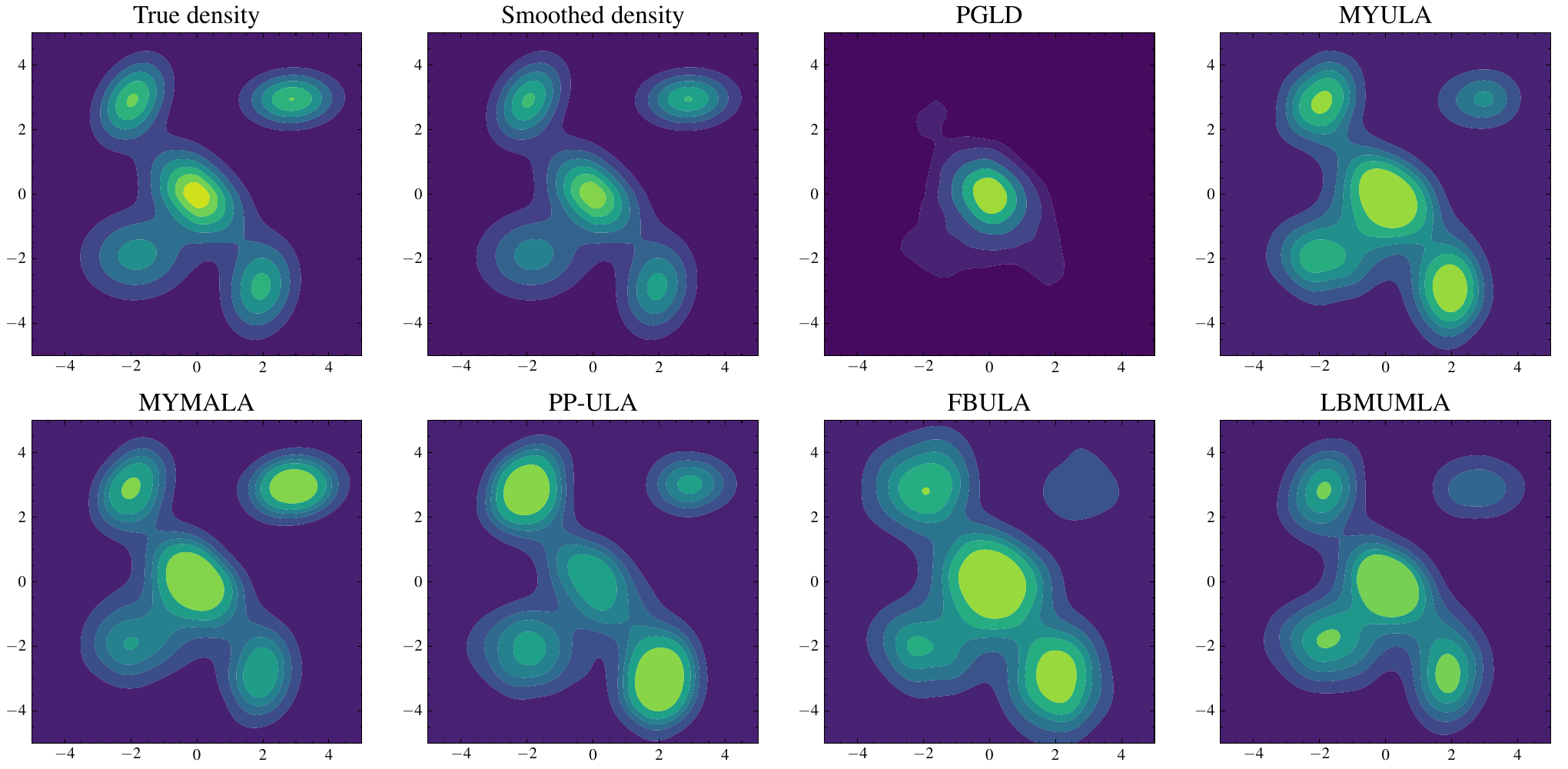}
                \caption{$K=5$}
            \end{subfigure}    
            \par\vspace{-2mm}
    	\caption{Mixture of $K$ Gaussians and a Laplacian prior with step size and smoothing parameter pair $(\gamma, \lambda)=(0.05, 0.25)$}    
        \label{fig:gaussians_laplacians}
        \end{figure}

        \subsection{Bayesian Image Deconvolution}
        \label{subsec:bayesian_image_deconv}
        In addition to the above simulations, we also want to demonstrate the LMC algorithms with real data for high-dimensional Bayesian imaging inverse problems which have non-log-concave posterior densities. We consider an image deconvolution task with a non-log-concave TV prior induced by the nonconvex TV regularizer \cite{selesnick2014convex,selesnick2020non,lanza2023convex}. The (isotropic) TV pseudonorm involves a linear operator (the first-order differencing matrix), hence allowing us to invoke a primal-dual Langevin algorithm. 
        
        We now briefly describe the problem setting, following \S4.1.2 of \cite{durmus2018efficient}. In image deconvolution, the goal is to recover a high-resolution image $\bx\in\RR^d$ from a blurred and noisy observation $\by = \bH\bx+\bw$, where $\bH\in\RR^{d\times d}$ is a circulant blurring matrix and $\bw\sim\sfN(\zero_d, \sigma^2\bI_d)$. The posterior distribution is given by    
        \[(\forall\bx\in\RR^d)\quad\pi(\bx) = p(\bx\mid \by)\propto \exp\left\{ -\frac1{2\sigma^2}\euclidnorm{\by - \bH\bx}^2 - \tau \TV(\bx) \right\}, \]
        where $\TV$ denotes the total-variation pseudonorm or its nonconvex variant such as the MC-TV \cite{selesnick2014convex,selesnick2020non,du2018minmax} and ME-TV \cite{selesnick2017total} penalties, $\sigma^2>0$ is the noise variance and $\tau>0$ is a regularization parameter. Both $\sigma^2$ and $\tau$ are assumed to be fixed in this context. We note that the MC-TV and ME-TV priors are motivated by the MC-TV and ME-TV penalties, which might lead to better recovered signals than the original convex TV penalty. Now we let $g(\bx)\coloneqq\norm{\bD \bx}_{1-2}$ for any $\bx\in\RR^d$ be the isotropic TV pseudonorm, where  $\bD\coloneqq(\bD_h^\top, \bD_v^\top)^\top\in\RR^{2d\times d}$ is the two-dimensional discrete gradient operator with $\bD_h, \bD_v\in\RR^{d\times d}$ the first-order horizontal and vertical difference matrices. We also denote $\tilde{g}\coloneqq\onenorm{\cdot}$. We consider the isotropic MC-TV (minimax-concave) and the ME-TV (Moreau envelope) priors \cite{lanza2023convex} with their corresponding potentials 
        \[(\forall\bx\in\RR^d)\quad
        \begin{aligned}
        \TV_\gamma^{\mathrm{MC}}(\bx) &\coloneqq g(\bx) - \tilde{g}_\gamma(\euclidnorm{\bD\bx}), \\
        \TV_\gamma^{\mathrm{ME}}(\bx) &\coloneqq g(\bx) - g_\gamma(\bx),
        \end{aligned}
        \]
        where $\tilde{g}_\gamma$ and $g_\gamma$ denote the Moreau envelopes of $\tilde{g}$ and $g$ with a smoothing parameter $\gamma>0$ respectively. 
        Note that we have the (in general) nonconvex posterior potential $U = f+\tau g$ with $f^{\mathrm{MC}}(\bx)\coloneqq\frac1{2\sigma^2}\euclidnorm{\by - \bH\bx}^2 - \tau \tilde{g}_{\gamma}(\euclidnorm{\bD\bx})$ and $f^{\mathrm{ME}}(\bx)\coloneqq\frac1{2\sigma^2}\euclidnorm{\by - \bH\bx}^2 - \tau g_{\gamma}(\bx)$. However, if $\gamma$ is well chosen according to the values of $\sigma^2$ and $\tau$, then $U$ could be convex \cite[Theorems 1 and 2]{lanza2023convex}. 
        
        We consider the following two LMC algorithms, MYULA \eqref{eqn:myula} and ULPDA \eqref{eqn:ulpda}. In MYULA, as we replace the nonsmooth part $\tau g$ by its Moreau envelope $\tau g_{\tau\lambda}$ with a smoothing parameter $\lambda>0$, the (smooth) surrogate posterior potential is thus 
        \[(\forall\bx\in\RR^d)\quad 
        \begin{aligned}
        U_\lambda^{\mathrm{MC}}(\bx) &= \frac1{2\sigma^2}\euclidnorm{\by - \bH\bx}^2 + \tau [g_{\tau\lambda}(\bx) - \tilde{g}_\gamma(\euclidnorm{\bD\bx})], \\
        U_\lambda^{\mathrm{ME}}(\bx) &= \frac1{2\sigma^2}\euclidnorm{\by - \bH\bx}^2 + \tau [g_{\tau\lambda}(\bx) - g_\gamma(\bx)], 
        \end{aligned}
        \]
        in which we have a difference-of-Moreau-envelopes smoothing \cite{sun2022algorithms}. The difference of Moreau envelopes are both differentiable. In particular, the gradient of $\tilde{g}_\gamma(\euclidnorm{\bD\cdot})$ is given by (see \cite[Proposition 4]{lanza2023convex} for details)
        \begin{equation}\label{eqn:grad_moreau}
        (\forall\bx\in\RR^d)\quad\nabla\tilde{g}_\gamma(\euclidnorm{\bD\bx})
        =  \bD^\top\min\left\{\frac1\gamma, \frac{1}{\euclidnorm{\bD\bx}}\right\}\bD\bx. 
        \end{equation}

        In ULPDA, we need to compute $\prox_{\lambda f}$ for some $\lambda>0$. Letting $\varphi(\bx) \coloneqq \frac1{2\sigma^2}\euclidnorm{\by - \bH\bx}^2$, then we have $f^{\mathrm{MC}}=\varphi -\tau \tilde{g}_\gamma(\euclidnorm{\bD\cdot})$ and $f^{\mathrm{ME}}(\bx)=\varphi - \tau g_{\gamma}$. Although the computation of the proximity operator of the sum of two functions is generally intractable \cite{pustelnik2017proximity}, we use a first-order Taylor expansion for the latter term to approximate the proximity operator of $f$ and introduce a forward step as follows: 
        \[(\forall\bx\in\RR^d)\quad
        \begin{aligned}
            \prox_{\lambda f^{\mathrm{MC}}}(\bx) &\approx \prox_{\lambda \varphi}\left(\bx + \lambda\tau\nabla \tilde{g}_\gamma(\euclidnorm{\bD\bx}) \right), \\
            \prox_{\lambda f^{\mathrm{ME}}}(\bx) &\approx \prox_{\lambda \varphi}\left(\left(1+\frac{\lambda\tau}{\gamma} \right) \bx - \frac{\lambda\tau}{\gamma}\prox_{\gamma g}(\bx) \right).  
        \end{aligned}
        \]
        Here $\nabla \tilde{g}_\gamma(\euclidnorm{\bD\bx})$ is given in \eqref{eqn:grad_moreau}, $\prox_{\gamma g} = \soft_\gamma$ is the proximity operator of the isotropic TV pseudonorm, and 
        \[(\forall\bx\in\RR^d)\quad \prox_{\lambda \varphi}(\bx) = \left(\Id + \frac{\lambda}{2\sigma^2}\bH^*\bH\right)^{-1}\left(\bx +\frac{\lambda}{2\sigma^2}\bH^*\by\right), \]
        where $\Id$ is the identity operator and $\bH^*$ is the adjoint of $\bH$. 
        
        In our experiments, we fix $\sigma = 0.75$ and $\tau=0.3$. We produce the blurred and noisy image $\by$ using  the $5\times5$ uniform blur operator $\bH = \bH_1$. In the absence of ground truth and information about the blur operator, we also consider the slightly misspecified $6\times6$ uniform blur operator $\bH_2$ and the strongly misspecified $7\times7$ uniform blur operator $\bH_3$. Together with the three priors, we evaluate a total of $J=9$ models, i.e.,  
        $\scrM_1$: ($\bH_1$, TV), 
        $\scrM_2$: ($\bH_1$, MC-TV), 
        $\scrM_3$: ($\bH_1$, ME-TV), 
        $\scrM_4$: ($\bH_2$, TV), 
        $\scrM_5$: ($\bH_2$, MC-TV), 
        $\scrM_6$: ($\bH_3$, ME-TV), 
        $\scrM_7$: ($\bH_3$, TV), 
        $\scrM_8$: ($\bH_3$, MC-TV), and 
        $\scrM_9$: ($\bH_3$, ME-TV).      
        We use two test images, \texttt{camera} and \texttt{einstein}, to demonstrate the effectiveness of the nonconvex TV priors. 
        
\begin{figure}[htbp]
                \centering
                \includegraphics[height=0.98\textheight]{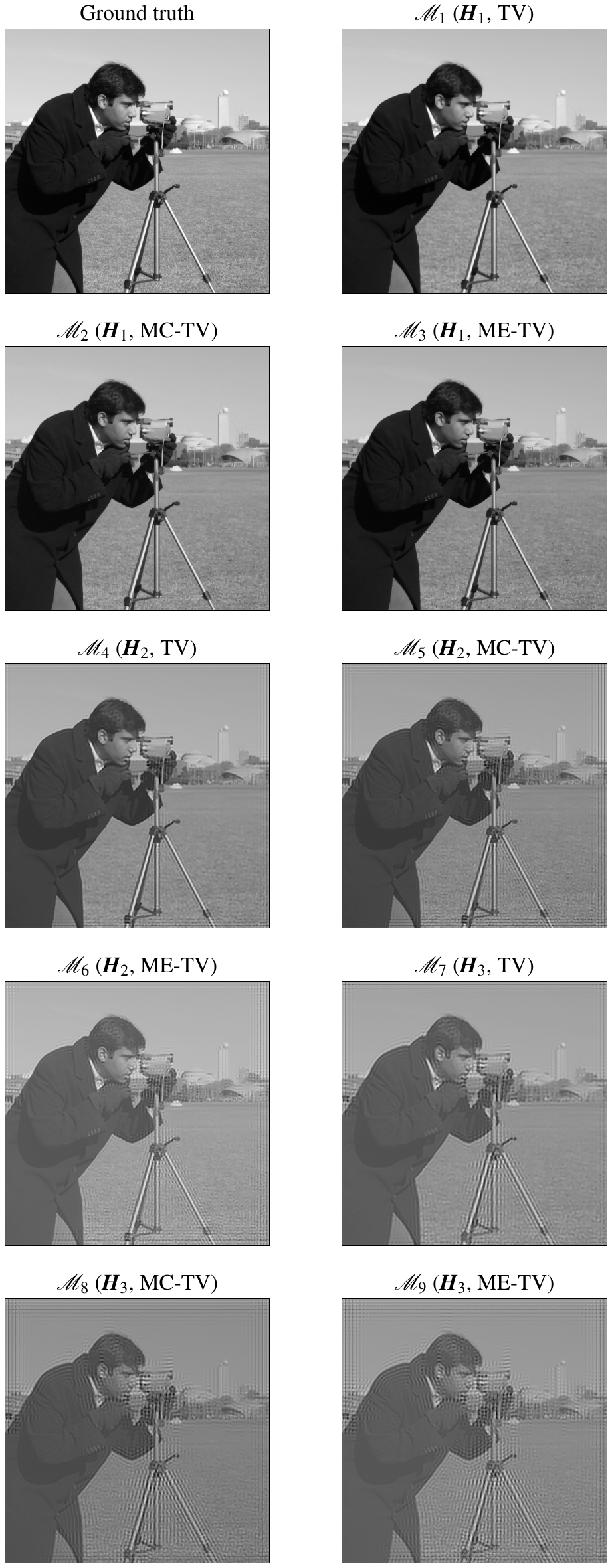}
                \caption{The \texttt{camera} test image and the MAP estimators of $\scrM_j$, $j\in\set{9}$}    
                \label{fig:camera}
        \end{figure}

        \Cref{fig:camera} presents the \texttt{camera} test image of size $d=512\times 512$ pixels and its MAP estimators for all 9 models computed using the adaptive PDHG algorithm (AdaPDHG) \cite{goldstein2013adaptive} with 1000 iterations. We take the step sizes $\gamma_{\mathrm{AdaPDHG}} = 0.95\sigma^2$ for the primal subgradient and $\lambda_{\mathrm{AdaPDHG}} = 1$ for the dual subgradient. 
        
        Then we generate 1000 samples using both MYULA and ULPDA. 
        We take the smoothing parameters of the Moreau envelopes of the MC-TV and ME-TV priors $\gamma_{\mathrm{MC}} = \gamma_{\mathrm{ME}}=15$, the step sizes of MYULA and ULPDA $\gamma_{\mathrm{MYULA}}=0.2\sigma^2$, $\gamma_{\mathrm{ULPDA}}=0.95\sigma^2$ and $\lambda_{\mathrm{ULPDA}}=1$, and the smoothing parameter of the Moreau envelope in MYULA $\lambda_{\mathrm{MYULA}}=\sigma^2$. For MYULA, the proximity operator of the TV pseudonorm is solved with the algorithm in \cite{beck2009fasttv} with 10 iterations. We present experiments for another test image, \texttt{einstein}, of size $d=667\times 877$ pixels in Appendix 3. 
        
\begin{figure}[htbp]
                \centering
                \includegraphics[height=0.98\textheight]{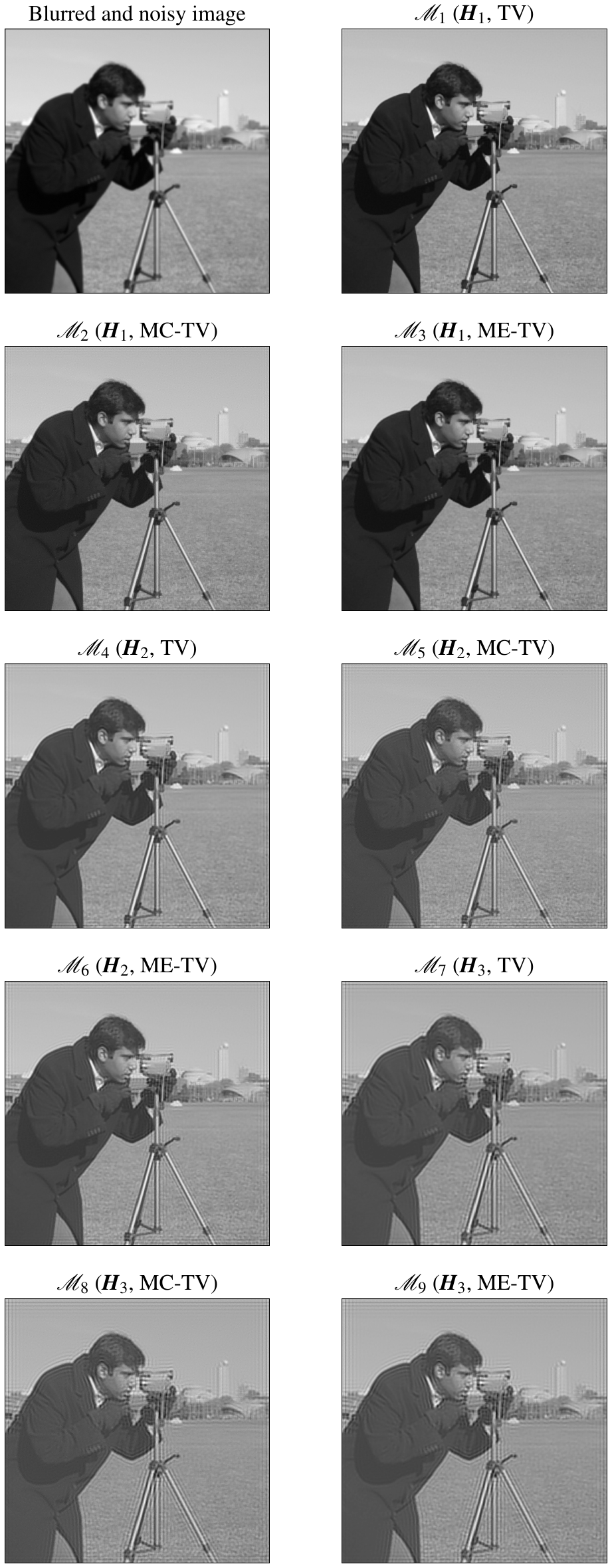}
                \caption{The blurred and noisy version of the \texttt{camera} image and the posterior means of samples of $\scrM_j$, $j\in\set{9}$, generated using MYULA}    
                \label{fig:camera_posterior_means_MYULA}
         \end{figure}

\begin{figure}[htbp]
                \centering
                \includegraphics[height=0.98\textheight]{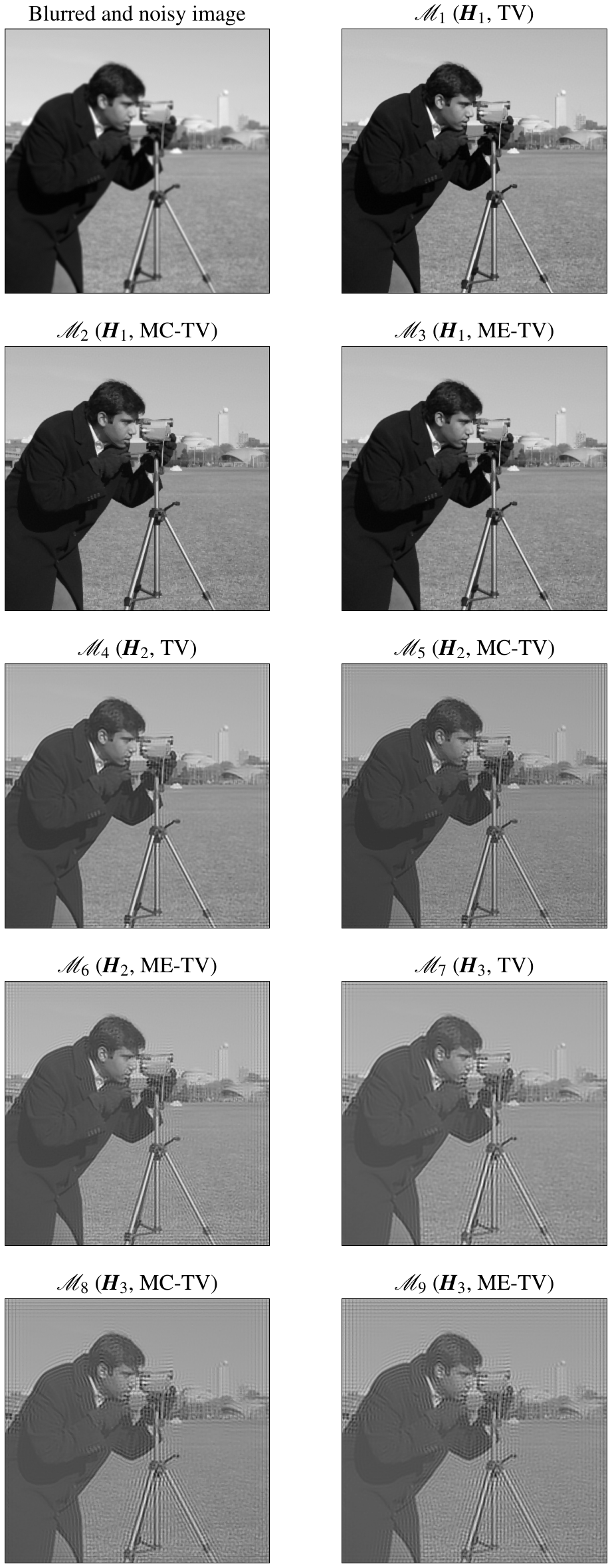}
                \caption{The blurred and noisy version of the \texttt{camera} image and the posterior means of samples of $\scrM_j$, $j\in\set{9}$, generated using ULPDA}    
                \label{fig:camera_posterior_means_ULPDA}
         \end{figure}

\begin{figure}[htbp]
                \centering
                 \begin{subfigure}[h]{0.48\textwidth}
                    \centering
                    \includegraphics[height=0.46\textheight]{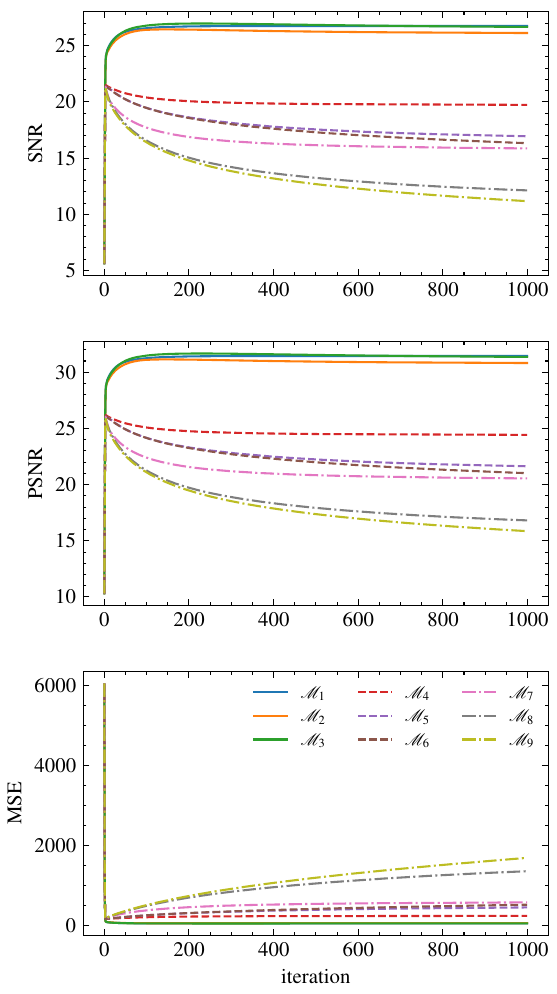}     
                    \caption{AdaPDHG}
                 \end{subfigure}
                 \begin{subfigure}[h]{0.48\textwidth}
                     \centering
                     \includegraphics[height=0.46\textheight]{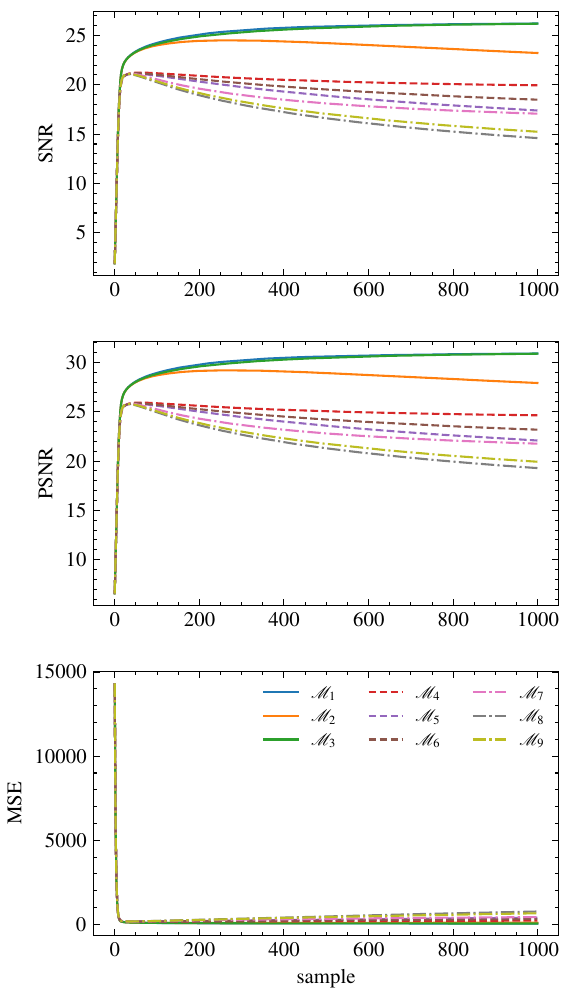}
                     \caption{MYULA}
                 \end{subfigure}
                \begin{subfigure}[h]{0.48\textwidth}
                     \centering
                     \includegraphics[height=0.46\textheight]{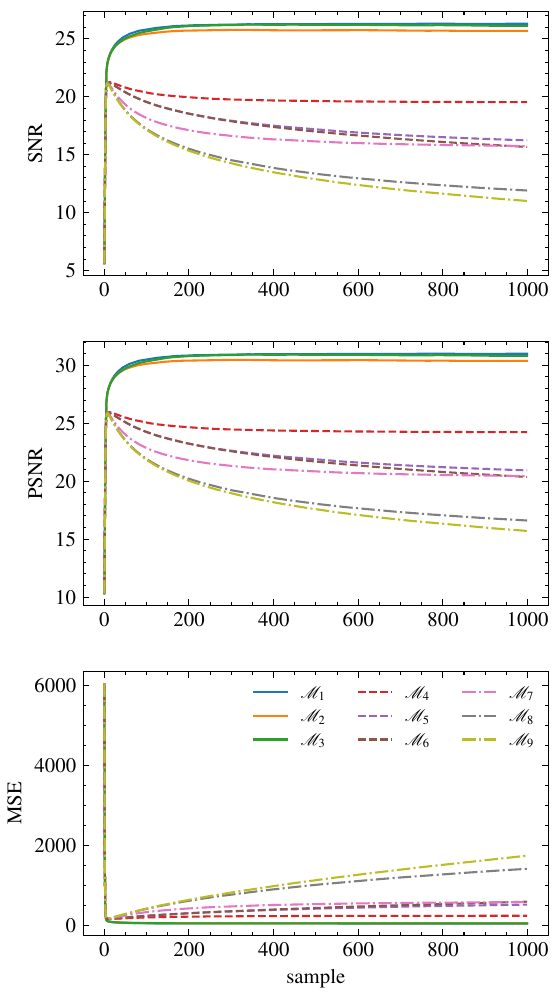}
                     \caption{ULPDA}
                 \end{subfigure}
                 \caption{SNRs, PSNRs and MSEs of iterates or samples of the \texttt{camera} image generated by different algorithms based on $\scrM_j$, $j\in\set{9}$}    
                 \label{fig:camera_snr_psnr_mse}
          \end{figure}

        To evaluate the quality of the samples, \Cref{tab:camera} presents the signal-to-noise ratios (SNR), peak signal-to-noise ratios (PSNR) and mean-squared errors (MSE) of the posterior means. \Cref{fig:camera_posterior_means_ULPDA,fig:camera_posterior_means_MYULA} present the blurred and noisy version of the test image, and the posterior means of the samples generated by ULPDA and MYULA. \Cref{fig:camera_snr_psnr_mse} plots the SNRs, PSNRs and MSEs against iterations (or samples for sampling methods). 
            
        \begin{table}[htbp]
            \begin{center}
            \caption{Signal-to-noise ratios (SNR), peak signal-to-noise ratios (PSNR) and mean-squared errors (MSE) of MAPs and posterior means of the samples based on $\scrM_j$, $j\in\set{9}$ for the \texttt{camera} image}
            \label{tab:camera}
            \begin{tabular}{@{\extracolsep{\fill}}lrrrrrrrrrrrr@{\extracolsep{\fill}}}
            \toprule%
            && \multicolumn{3}{@{}c@{}}{MAP} && \multicolumn{3}{@{}c@{}}{MYULA} && \multicolumn{3}{@{}c@{}}{ULPDA} \\
            \cmidrule{3-5}\cmidrule{7-9}\cmidrule{11-13}%
             && SNR & PSNR & MSE & & SNR & PSNR & MSE & & SNR & PSNR & MSE\\
            \midrule
            $\scrM_1$ ($\bH_1$, TV) && 26.75 & 31.44 & 46.64 &  & 26.08 & 30.77 & 54.40 &  & 26.85 & 31.54 & 45.57 \\
            $\scrM_2$ ($\bH_1$, MC-TV) && 26.12 & 30.81 & 53.90 &  & 25.09 & 29.78 & 68.47 &  & 26.91 & 31.29 & 44.97 \\
            $\scrM_3$ ($\bH_1$, ME-TV) && 26.67 & 31.36 & 47.53 &  & 26.12 & 30.81 & 53.94 &  & \textbf{27.16} & \textbf{31.85} & \textbf{42.43} \\
            $\scrM_4$ ($\bH_2$, TV) && 19.73 & 24.42 & 234.75 &  & 20.91 & 25.60 & 179.18 &  & 20.12 & 24.81 & 214.70 \\
            $\scrM_5$ ($\bH_2$, MC-TV) && 16.94 & 21.63 & 446.62 &  & 19.63 & 24.32 & 240.26 &  & 17.95 & 22.64 & 353.68 \\
            $\scrM_6$ ($\bH_2$, ME-TV) && 16.33 & 21.02 & 514.57 &  & 20.15 & 24.83 & 213.35 &  & 17.74 & 24.84 & 371.92 \\
            $\scrM_7$ ($\bH_3$, TV) && 15.86 & 20.55 & 572.61 &  & 18.90 & 23.59 & 284.39 &  & 16.83 & 23.59 & 458.29 \\
            $\scrM_8$ ($\bH_3$, MC-TV) && 12.12 & 16.81 & 1354.39 &  & 17.44 & 22.13 & 397.80 &  & 14.20 & 18.89 & 839.67 \\
            $\scrM_9$ ($\bH_3$, ME-TV) && 11.17& 15.86 & 1687.95 &  & 17.83 & 22.52 & 363.58 &  & 13.70 & 18.39 & 941.11 \\
            \bottomrule
            \end{tabular}
            \end{center}
        \end{table}
        
        From \Cref{tab:camera} and \Cref{fig:camera,fig:camera_posterior_means_ULPDA,fig:camera_posterior_means_MYULA,fig:camera_snr_psnr_mse}, we make the following observations: 
        \begin{enumerate}
        \item For the correctly specified models $\scrM_1$--$\scrM_3$, the posterior means of samples generated with ULPDA based on the MC-TV ($\scrM_2$) and ME-TV ($\scrM_3$) priors yield higher SNRs and PSNRs, as well as lower MSEs than the one based on the TV prior. More specifically, $\scrM_3$ has the best performance out of these three. 
        
        \item Compared with the MAP estimates (cf.~the optimization formulation), the posterior means of samples generated with ULPDA also have better performance in terms of the above three metrics. 
        
        \item For the misspecified models $\scrM_4$--$\scrM_9$, as opposed to the correctly specified models $\scrM_1$--$\scrM_3$, the posterior means of samples generated with both MYULA and ULPDA based on the MC-TV ($\scrM_5$ and $\scrM_8$) and ME-TV priors yield lower SNRs and PSNRs, as well as higher MSEs than the one based on the TV prior. 
        
        \item ULPDA produce better samples than MYULA for the correctly specified models  $\scrM_1$--$\scrM_3$ but worse samples for the misspecified models $\scrM_4$--$\scrM_9$. 
        
        \item The above observations also align with inspecting the recovered images in \Cref{fig:camera,fig:camera_posterior_means_ULPDA,fig:camera_posterior_means_MYULA}.
        
        \item In \Cref{fig:camera_snr_psnr_mse}, we observe that model misspecification ($\scrM_4$--$\scrM_9$) leads to degraded iterates or samples in terms of SNR, PSNR and MSE, especially for the nonconvex priors ($\scrM_5$, $\scrM_6$, $\scrM_8$ and $\scrM_9$). The correctly specified model with ME-TV prior $\scrM_3$ also has slightly better performance than the one with TV prior $\scrM_1$. Note that there is discrepancy between \Cref{tab:camera} and \Cref{fig:camera_snr_psnr_mse}, since \Cref{tab:camera} reports the metrics based on MAPs or posterior means while the plots in \Cref{fig:camera_snr_psnr_mse} are based on individual iterates or samples. 
        \end{enumerate}
        
        Note that all of the above evaluation metrics require the knowledge of the ground truth which is unrealistic in practical scenarios. While other evaluations of Bayesian posterior inference procedures such as Bayesian model selection in \cite{durmus2018efficient,cai2022proximal} can be performed in the absence of ground truth, performing them efficiently with primal-dual Langevin algorithms with nonconvex potentials for problems of high dimensions are still not well understood and we leave them as interesting future research directions. 
    
       	\section{Conclusion} 
        In this chapter, we focus on understanding the convergence behavior of MCMC algorithms motivated by the overdamped Langevin diffusion for non-log-concave and non-log-smooth sampling through numerical simulations. Similar analysis can also be performed using higher-order algorithms such as the underdamped Langevin Monte Carlo (ULMC) algorithm and the Hamiltonian Monte Carlo (HMC) algorithm, and their variants. 
            
        Other than the considered formulation in which smoothing and regularization parameters are usually pre-specified, recently the proximal MCMC methods are studied through a fully Bayesian framework with the use of epigraph priors \cite{zhou2022proximal}, avoiding the need of tuning regularization parameters. An empirical Bayesian approach is also studied in \cite{vidal2020maximum1,de2020maximum2}. On the other hand, there is also recent interest in other MCMC algorithms based on splitting methods, e.g., ADMM-type splitting and splitting in higher-order algorithms \cite{vono2022efficient,monmarche2021high,bertazzi2023splitting,shahbaba2014split,casas2022split}.         
        Note that the success of LMC algorithms hinges heavily on efficient sampling from a high-dimensional Gaussian distribution; see \cite{vono2022high} for a recent review on the topic and its connection with the proximal point algorithm. 
        Recent works also concern the development of LMC algorithms with the use of Plug \& Play (PnP) priors in the Bayesian statistical framework \cite{laumont2022bayesian,laumont2023maximum}, which are implicit priors based on highly nonconvex neural networks with more flexibility than explicit priors and have gained popularity in Bayesian imaging problems. 

        While LMC algorithms are designed mainly for continuous and unconstrained domains, recent works also attempt to adapt them for sampling from discrete distributions \cite{zanella2020informed,grathwohl2021oops,rhodes2022enhanced,zhang2022langevin,zhang2023nonasymptotic,patterson2013stochastic}.          
        We also note the intriguing connection of LMC algorithms to nonconvex optimization algorithms, in a sense that SGLD and its variants are used as global nonconvex optimization algorithms \cite{raginsky2017nonconvex,erdogdu2018global,xu2018global}.

        \section*{Acknowledgements}
        \addcontentsline{toc}{section}{Acknowledgements}
        Tim Tsz-Kit Lau would like to thank Prof.~Jean-Christophe Pesquet from CentraleSup\'{e}lec, Universit\'{e} Paris-Saclay and the organizers for their invitation and the travel support for his participation in the workshop on Advanced Techniques in Optimization for Machine learning and Imaging (ATOMI, Rome, Italy, 20--24 June 2022), during which the present work was initiated. He also thanks Luca Calatroni from CNRS, INS2I, Laboratoire I3S for his discussion on the topic of this work during the workshop. 
        Han Liu’s research is partly supported by NIH R01LM1372201, NSF CAREER 1841569 and NSF TRIPODS 1740735.

        \newpage    
        \bibliographystyle{plainnat}
        \bibliography{ref}
        	
       	\newpage
       	\appendix

       	\begin{center}
                {\LARGE \textsc{Appendix}}
        \end{center}	
        
        \tableofcontents
        
        \newpage
        \section*{Appendix 1}
        \addcontentsline{toc}{section}{Appendix 1}
        We give the expressions of the gradient and the Hessian of the potential $U$ for the Gaussian mixture density $p$:  
        \begin{equation*}
        (\forall \bx\in\RR^d)\quad 
         \nabla U(\bx) = -\dfrac{\nabla p(\bx)}{p(\bx)}, \quad
         \nabla^2 U(\bx) = \dfrac{\nabla p(\bx) \nabla p(\bx)^\top}{p(\bx)^2} - \dfrac{\nabla^2 p(\bx)}{p(\bx)}, 
        \end{equation*}
        where, for any $ \bx\in\RR^d $, the gradient and the Hessian of the density $p$ of the Gaussian mixture are 
        \begin{align*}
        \nabla p(\bx) &= \sumK \omega_k \nabla p_k(\bx)\\
        &= \sumK \frac{\omega_k}{\sqrt{\det(2\uppi\bSigma_k)}} \exp\left\{ -\frac12 \dotp{\bx - \bmu_k}{\bSigma_k^{-1}(\bx-\bmu_k)}\right\}\bSigma_k^{-1}(\bmu_k-\bx),      		
        \end{align*}
        and 
        \begin{align*}
        \nabla^2 p(\bx) &= \sumK \omega_k \nabla^2 p_k(\bx)\\
        &= \sumK \frac{\omega_k}{\sqrt{\det(2\uppi\bSigma_k)}} \exp\left\{ -\frac12 \dotp{\bx - \bmu_k}{\bSigma_k^{-1}(\bx-\bmu_k)}\right\}\left( \bSigma_k^{-1}(\bx-\bmu_k)(\bx-\bmu_k)^\top\bSigma_k^{-1}  - \bSigma_k^{-1}\right). 
        \end{align*}     
        On the other hand, the gradient of the surrogate potential $U^\lambda$ for the Laplacian mixture surrogate density $p^\lambda$ take the expressions:
        \begin{equation*}
            (\forall \bx\in\RR^d)\quad 
             \nabla U^\lambda(\bx) = -\nabla p^\lambda(\bx)/p^\lambda(\bx),  
        \end{equation*} 
        where, for any $ \bx\in\RR^d $, 
        \begin{align*}
        \nabla p^\lambda(\bx) &= \sumK \omega_k \nabla p_k^\lambda(\bx)= -\sumK \frac{\omega_k\alpha_k^d}{2^d} \exp\left\{-g_k^\lambda(\bx)\right\}\nabla g_k^\lambda(\bx)\\
        &= \sumK \frac{\omega_k\alpha_k^d}{2^d \lambda} \exp\left\{-g_k^\lambda(\bx)\right\}\left( \prox_{\lambda g_k}(\bx) - \bx\right).      		
        \end{align*}

        \newpage
        \section*{Appendix 2}
        \addcontentsline{toc}{section}{Appendix 2}
        We give additional simulation results for \Cref{sec:sim} with different number of Gaussians or Laplacians, step sizes and smoothing parameters.

        \begin{figure}[htbp]
            \centering
            \begin{subfigure}[t]{.48\textwidth}
                \centering
                \includegraphics[height=.18\textheight]{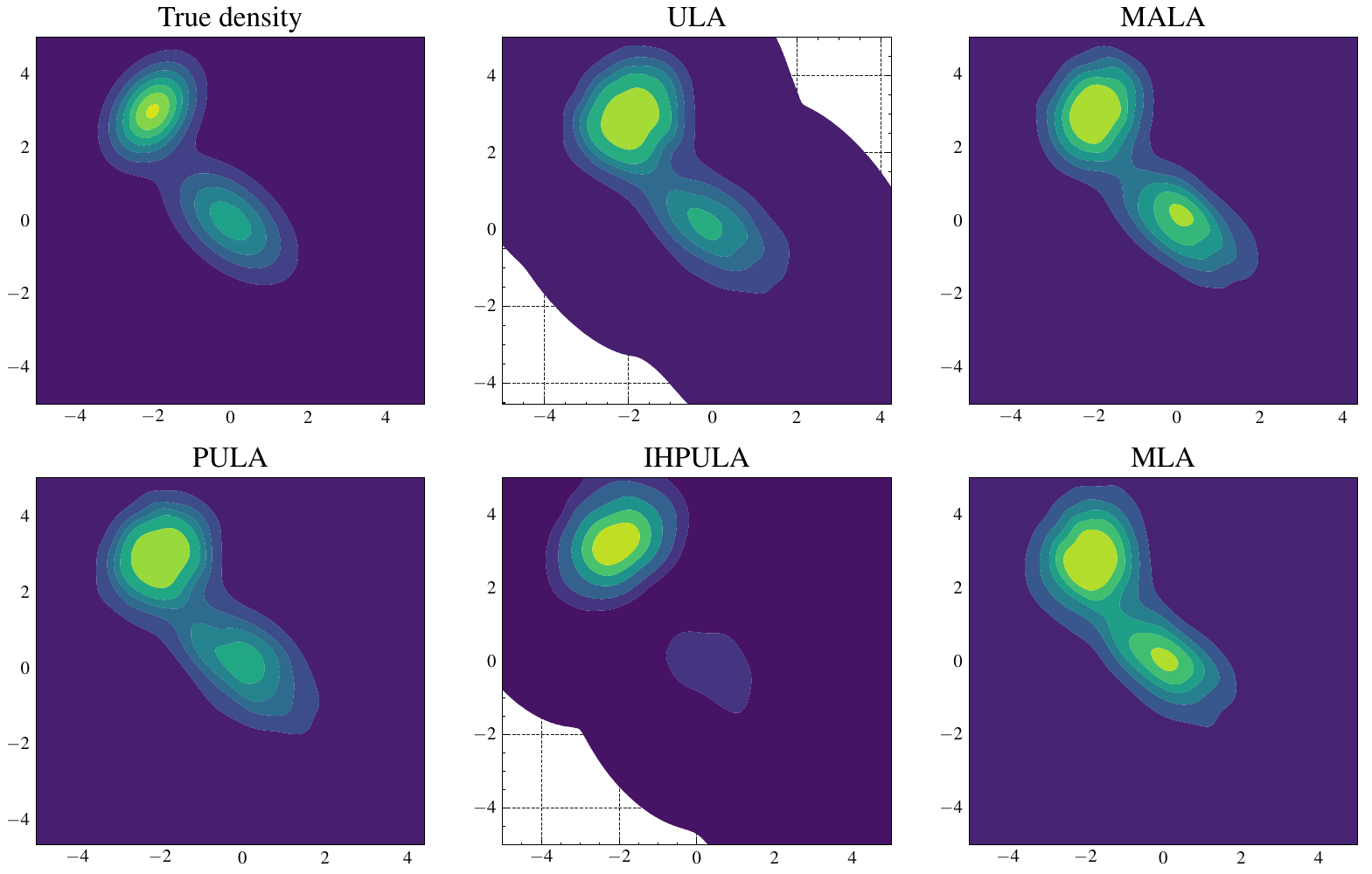}
                \caption{$K=2$}
                \vspace*{4mm}
            \end{subfigure}
            \hfill
            \begin{subfigure}[t]{.48\textwidth}
                \centering
                \includegraphics[height=.18\textheight]{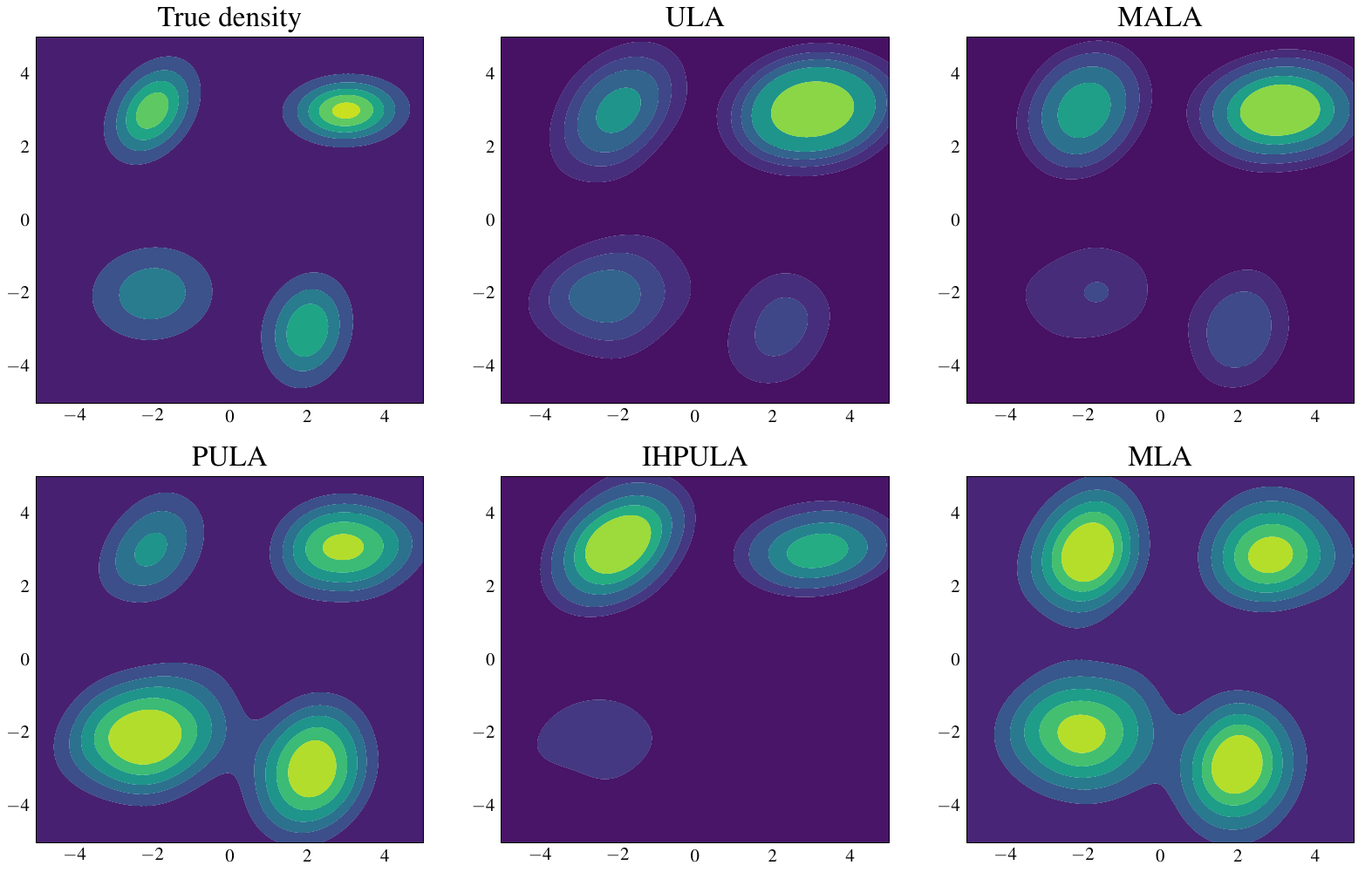}
                \caption{$K=4$}
            \end{subfigure}
            \caption{Mixture of $K$ Gaussians with $\gamma = 0.1$}
        \end{figure}

        \begin{figure}[htbp]
             \centering
            \begin{subfigure}[t]{.48\textwidth}
                \centering
                \includegraphics[height=.18\textheight]{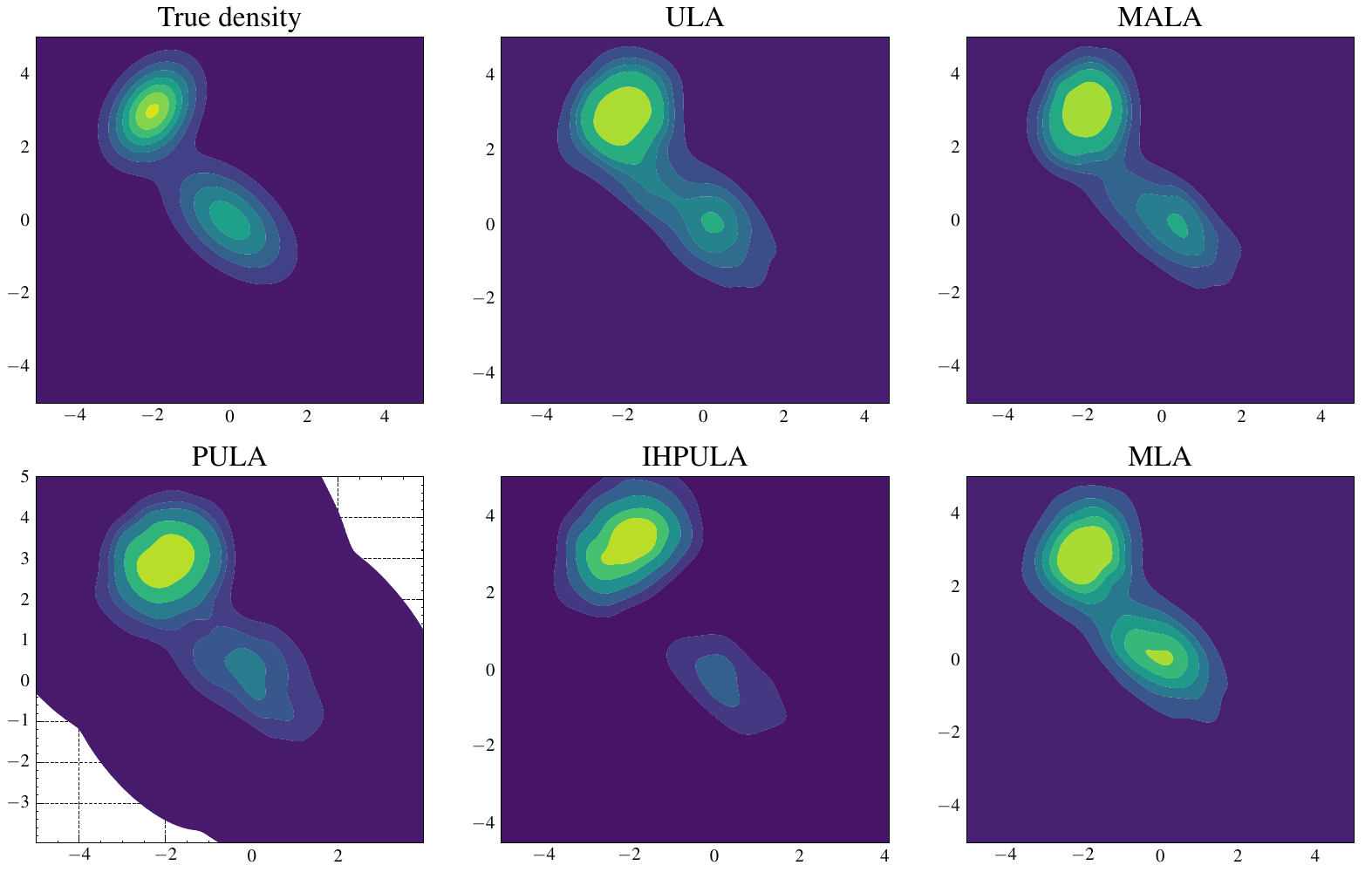}
                \caption{$K=2$}
                \vspace*{1mm}
            \end{subfigure}    
            \hfill
            \begin{subfigure}[t]{.48\textwidth}
                \centering
                \includegraphics[height=.18\textheight]{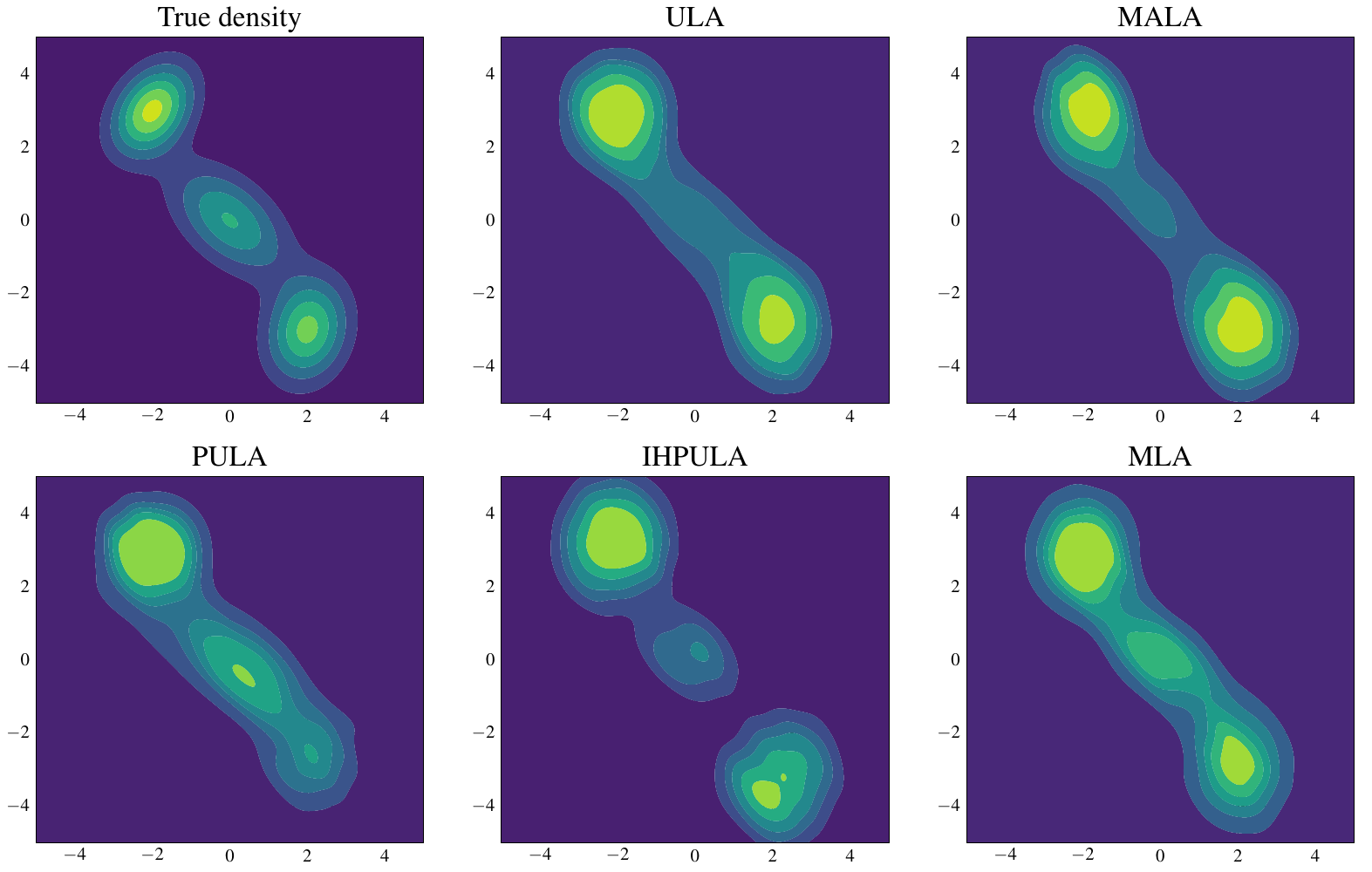}
                \caption{$K=3$}
                \vspace*{1mm}
            \end{subfigure}     
            \par\vspace{2mm}
            \begin{subfigure}[t]{.48\textwidth}
                \centering
                \includegraphics[height=.18\textheight]{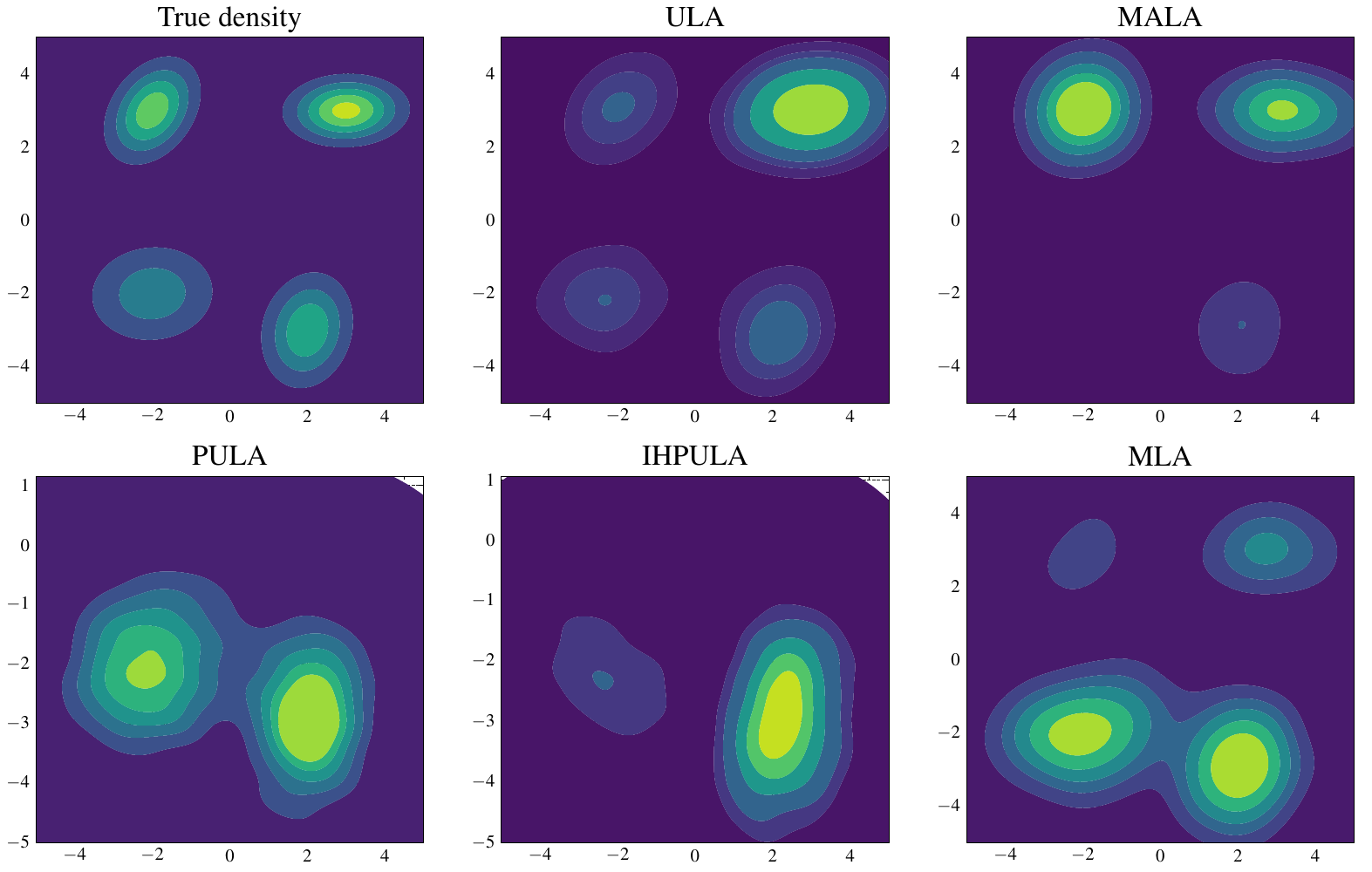}
                \caption{$K=4$}
            \end{subfigure} 
            \hfill
            \begin{subfigure}[t]{.48\textwidth}
                \centering
                \includegraphics[height=.18\textheight]{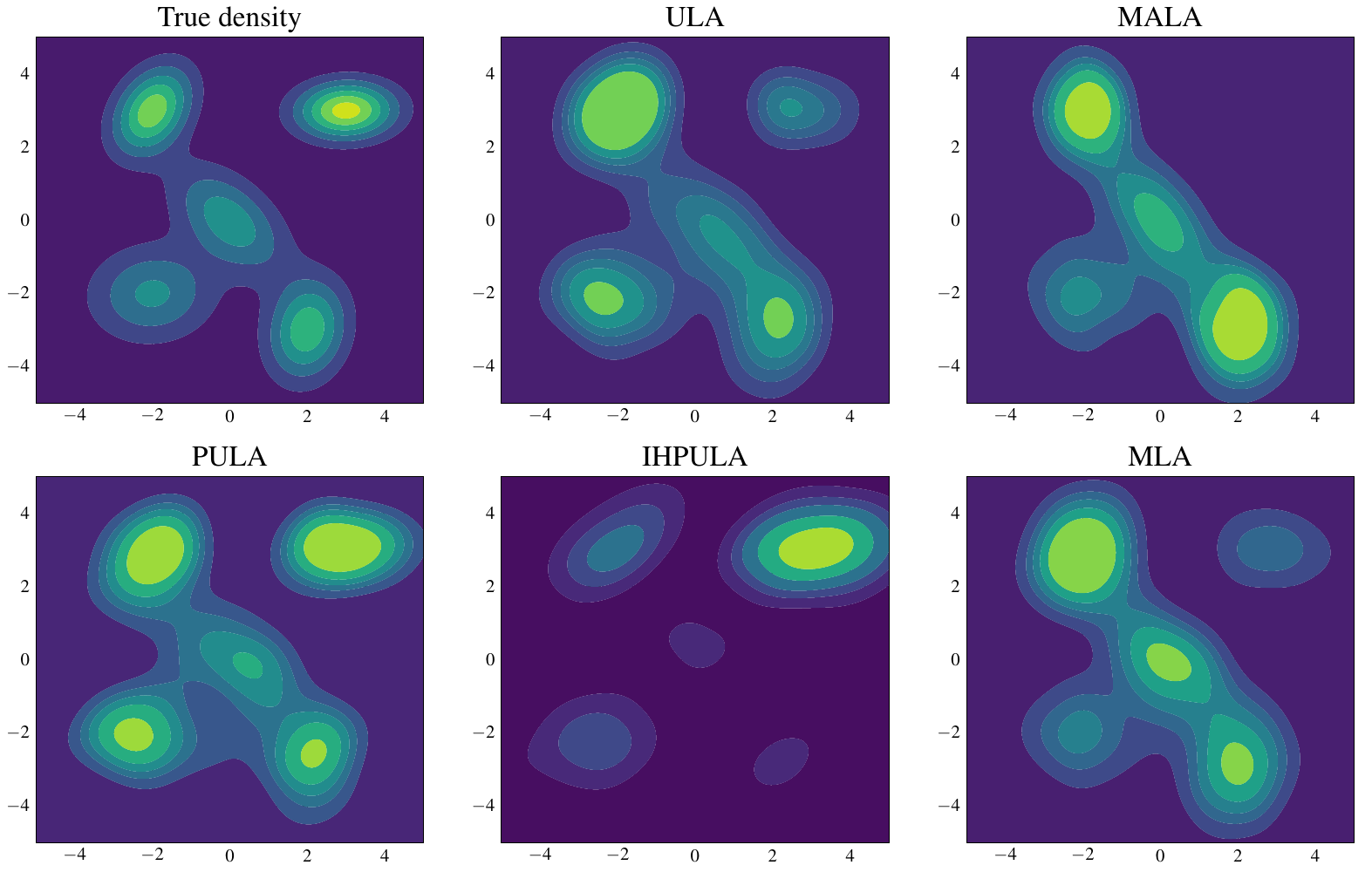}
                \caption{$K=5$}
            \end{subfigure}
            \caption{Mixture of $K$ Gaussians with $\gamma = 0.05$}
        \end{figure} 
        
        \begin{figure}[htbp]
            \begin{subfigure}[t]{.48\textwidth}
                \centering
                \includegraphics[width=\textwidth]{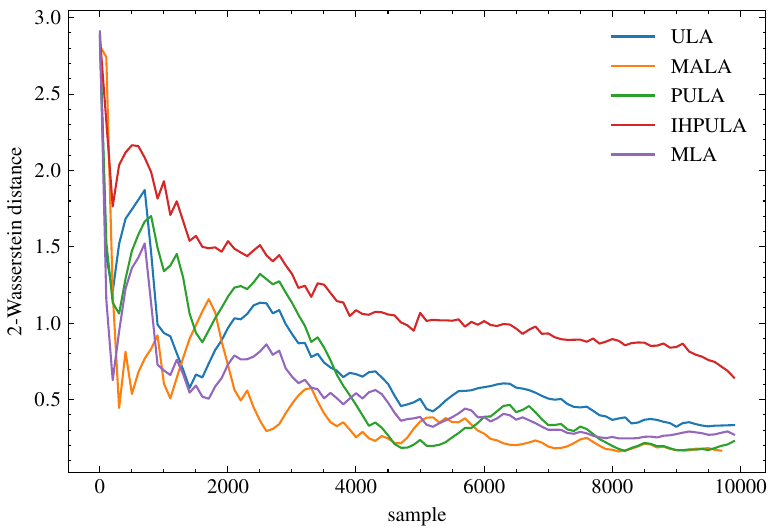}
                \caption{$K=2$}
                \vspace*{1mm}
            \end{subfigure}    
            \hfill
            \begin{subfigure}[t]{.48\textwidth}
                \centering
                \includegraphics[width=\textwidth]{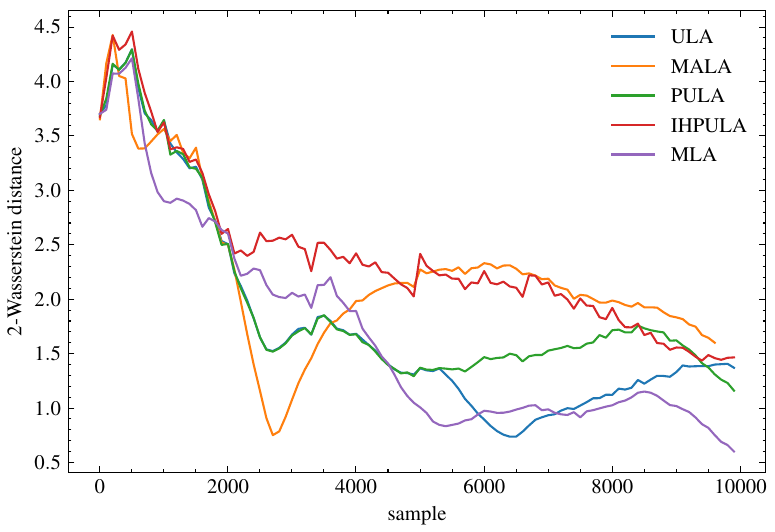}
                \caption{$K=4$}
                \vspace*{1mm}
            \end{subfigure}     
            \caption{$2$-Wasserstein distances between generated samples by LMC algorithms and true samples of mixture of $K$ Gaussians with step size $\gamma = 0.1$}
            \label{fig:gaussians_wass_1} 
        \end{figure}
        
        \begin{figure}[htbp]
            \begin{subfigure}[t]{.48\textwidth}
                \centering
                \includegraphics[width=\textwidth]{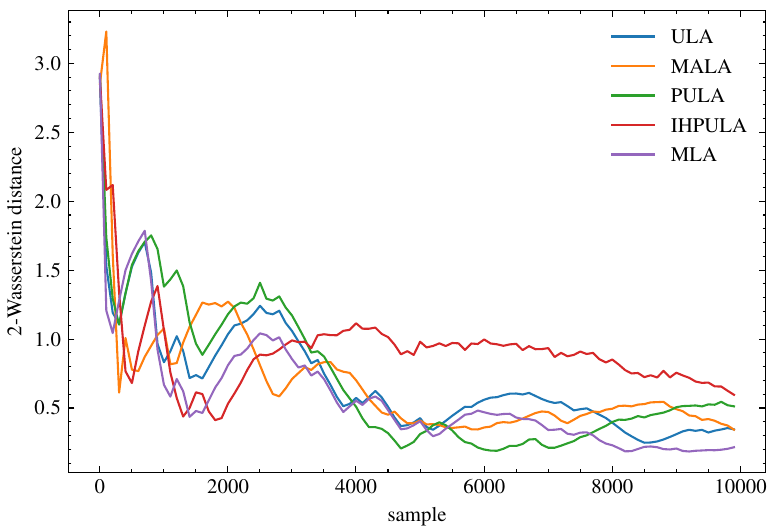}
                \caption{$K=2$}
                \vspace*{1mm}
            \end{subfigure}    
            \hfill
            \begin{subfigure}[t]{.48\textwidth}
                \centering
                \includegraphics[width=\textwidth]{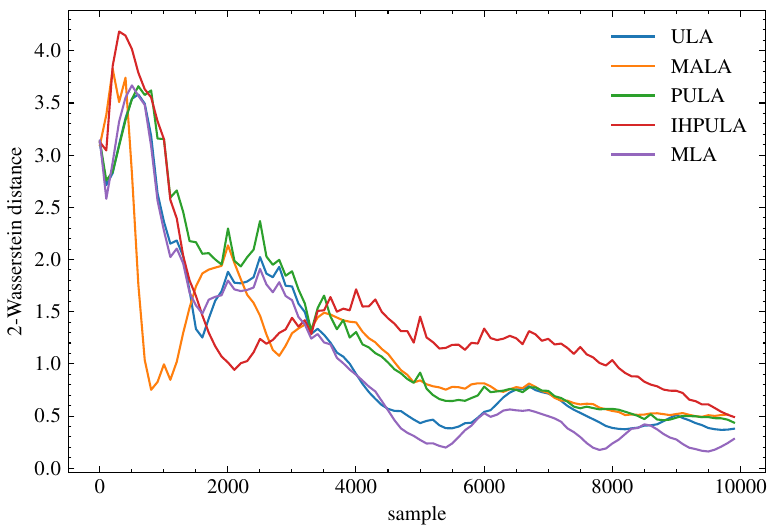}
                \caption{$K=3$}
                \vspace*{1mm}
            \end{subfigure}     
            \par\vspace{2mm}
            \begin{subfigure}[t]{.48\textwidth}
                \centering
                \includegraphics[width=\textwidth]{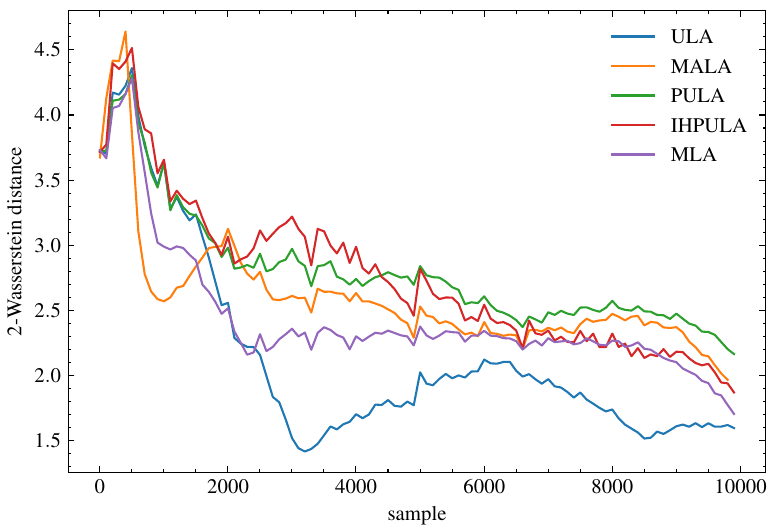}
                \caption{$K=4$}
            \end{subfigure} 
            \hfill
            \begin{subfigure}[t]{.48\textwidth}
                \centering
                \includegraphics[width=\textwidth]{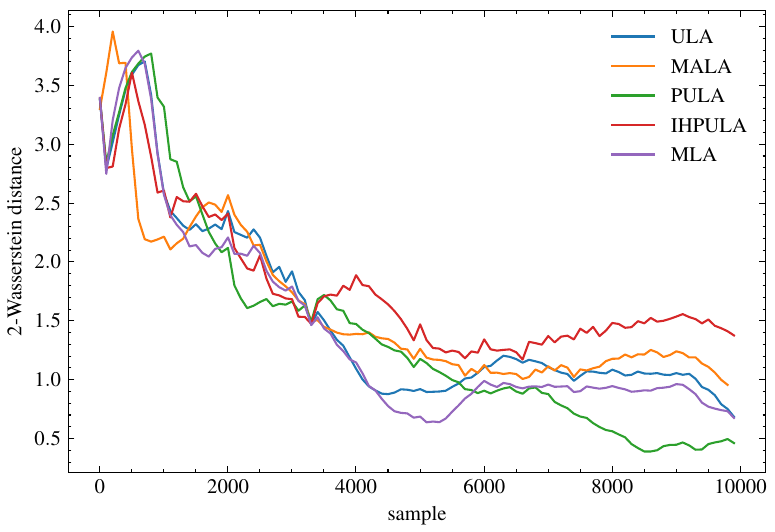}
                \caption{$K=5$}
            \end{subfigure}
            \caption{$2$-Wasserstein distances between generated samples by LMC algorithms and true samples of mixture of $K$ Gaussians with step size $\gamma = 0.05$}
            \label{fig:gaussians_wass_2}
        \end{figure}

        \begin{figure}[htbp]
             \centering
            \begin{subfigure}[h]{.48\textwidth}
                \centering
                \includegraphics[height=.18\textheight]{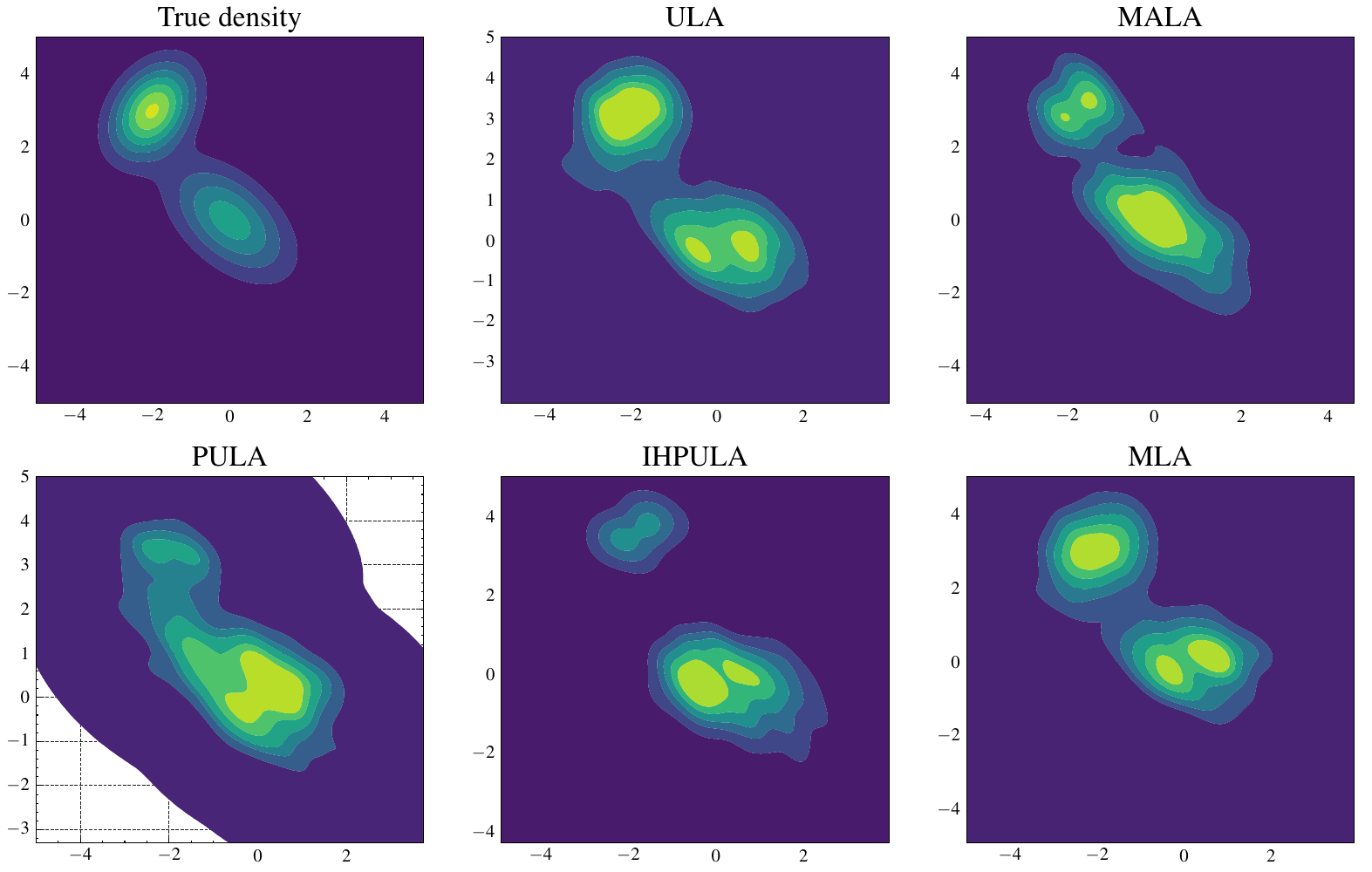}
                \caption{$K=2$}
                \vspace*{1mm}
            \end{subfigure} 
            \hfill
            \begin{subfigure}[h]{.48\textwidth}
                \centering
                \includegraphics[height=.18\textheight]{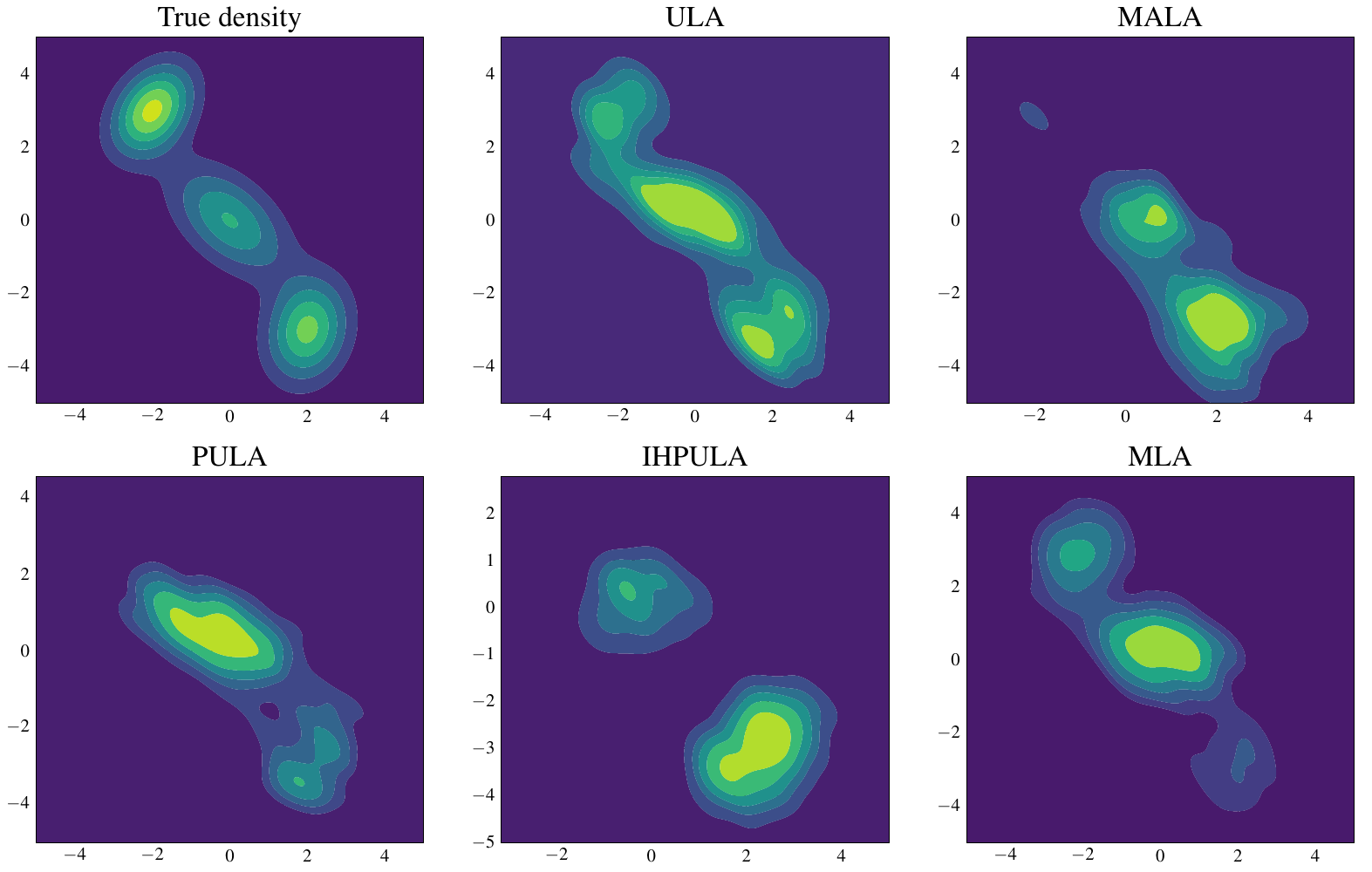}
                \caption{$K=3$}
                \vspace*{1mm}
            \end{subfigure} 
            \par\vspace{2mm}
            \begin{subfigure}[h]{.48\textwidth}
                \centering
                \includegraphics[height=.18\textheight]{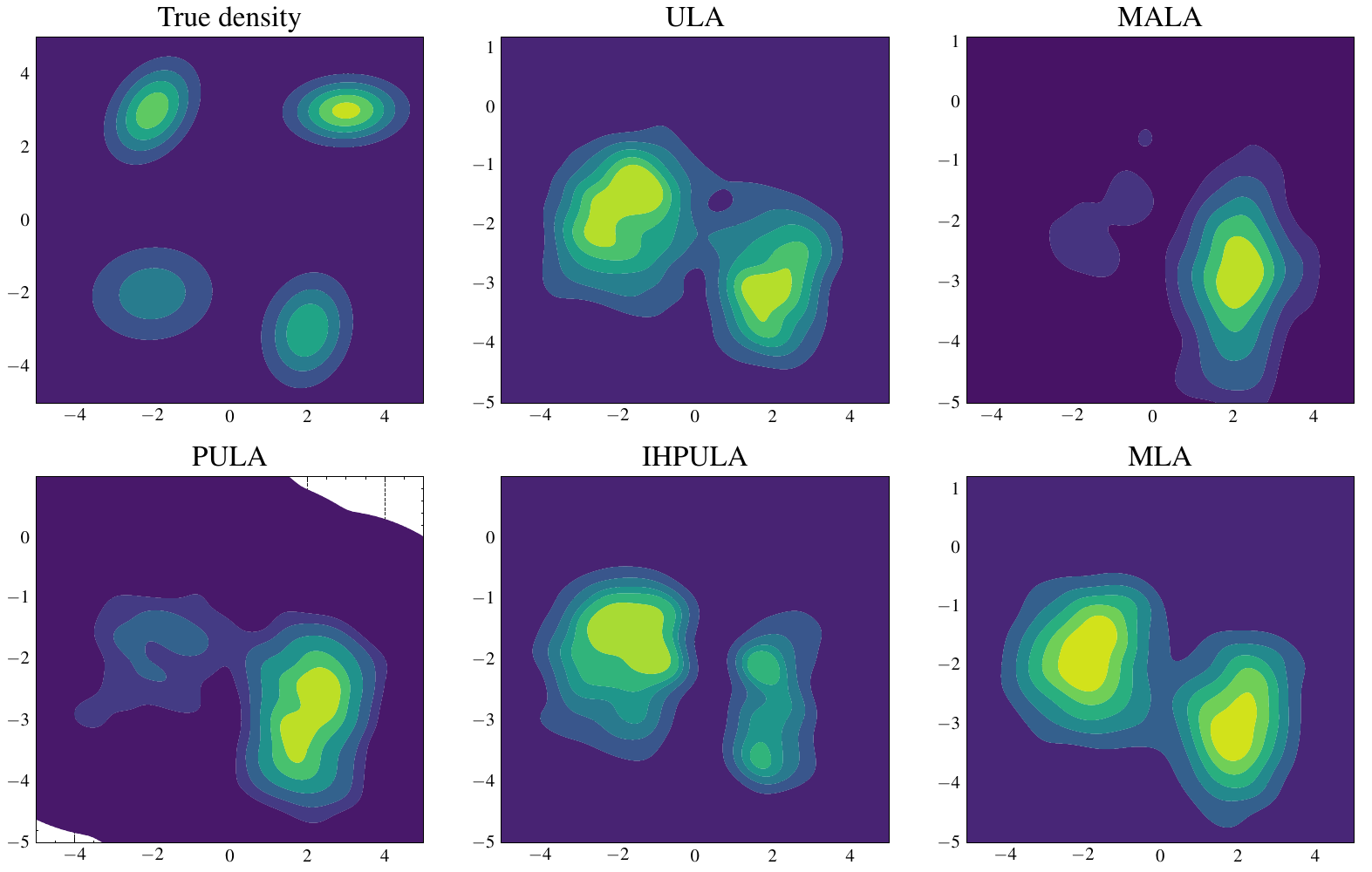}
                \caption{$K=4$}
            \end{subfigure} 
            \hfill
            \begin{subfigure}[h]{.48\textwidth}
                \centering
                \includegraphics[height=.18\textheight]{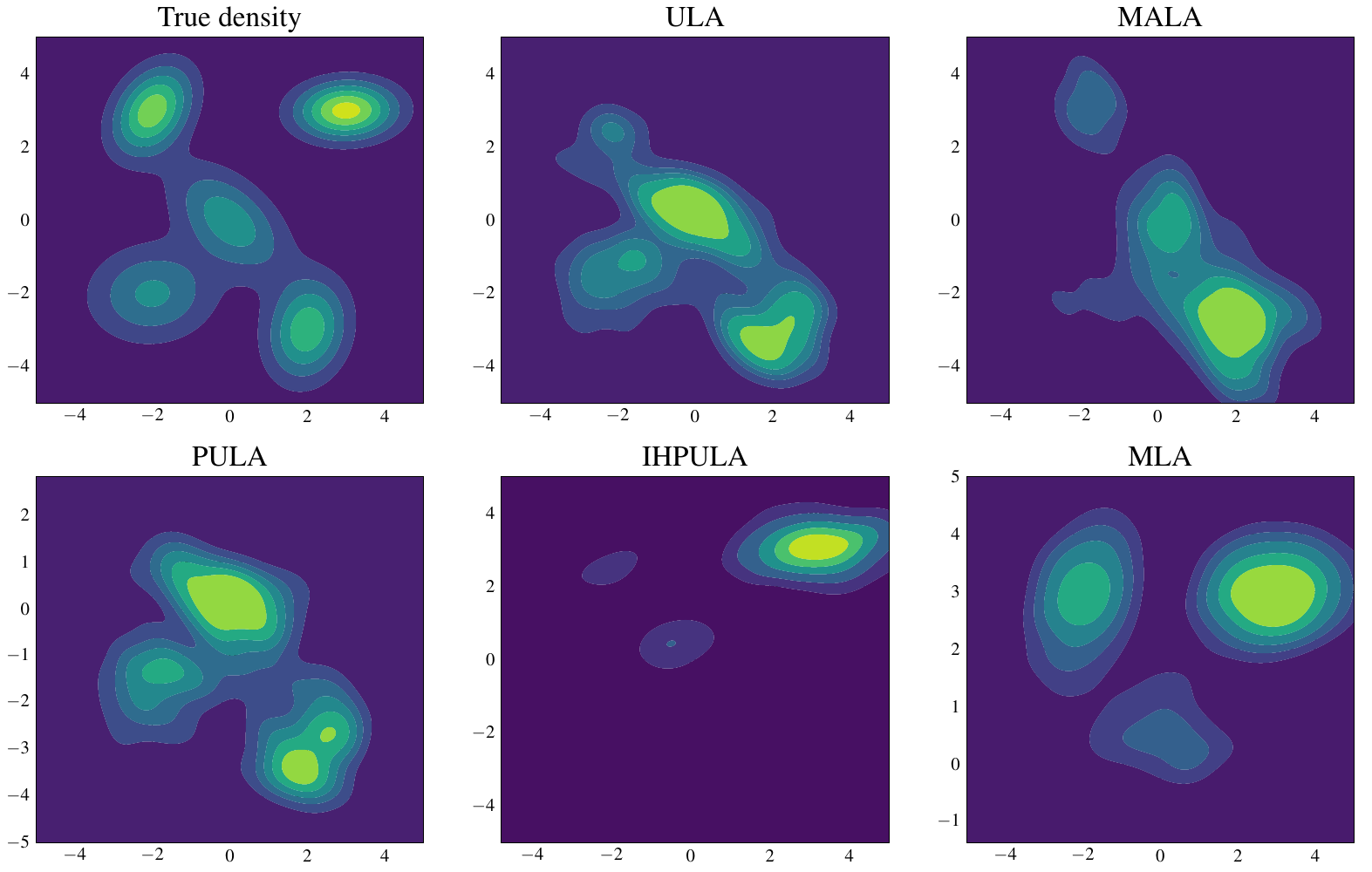}
                \caption{$K=5$}
            \end{subfigure} 
            \caption{Mixture of $K$ Gaussians with $\gamma = 0.01$}
        \end{figure}

        \begin{figure}[htbp]
            \centering
            \begin{subfigure}[h]{\textwidth}
                \centering
                \includegraphics[height=.18\textheight]{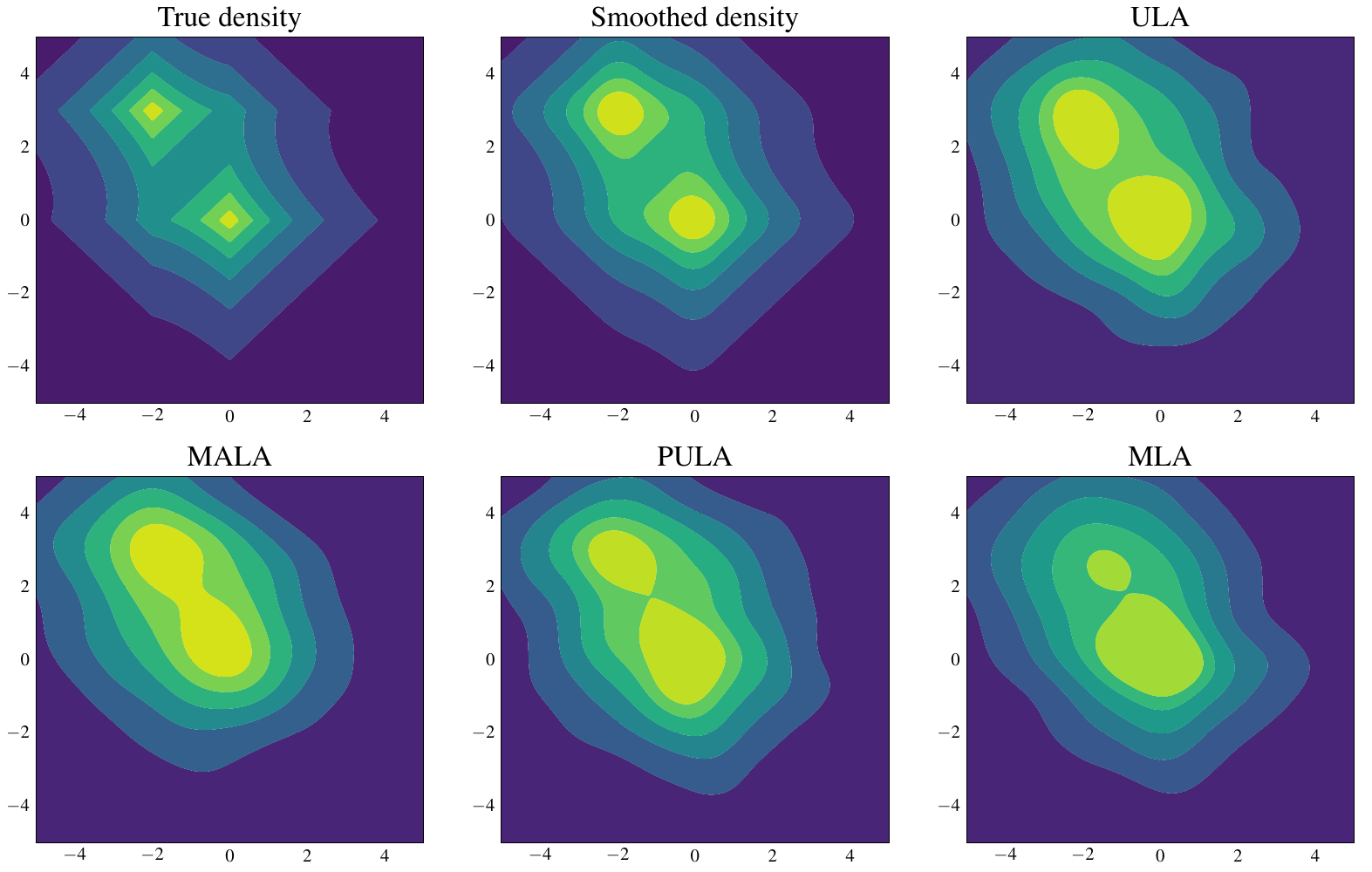}
                \caption{$K=2$}
                \vspace*{2mm}
            \end{subfigure}
            \par\vspace{2mm}
            \begin{subfigure}[h]{\textwidth}
                \centering
                \includegraphics[height=.18\textheight]{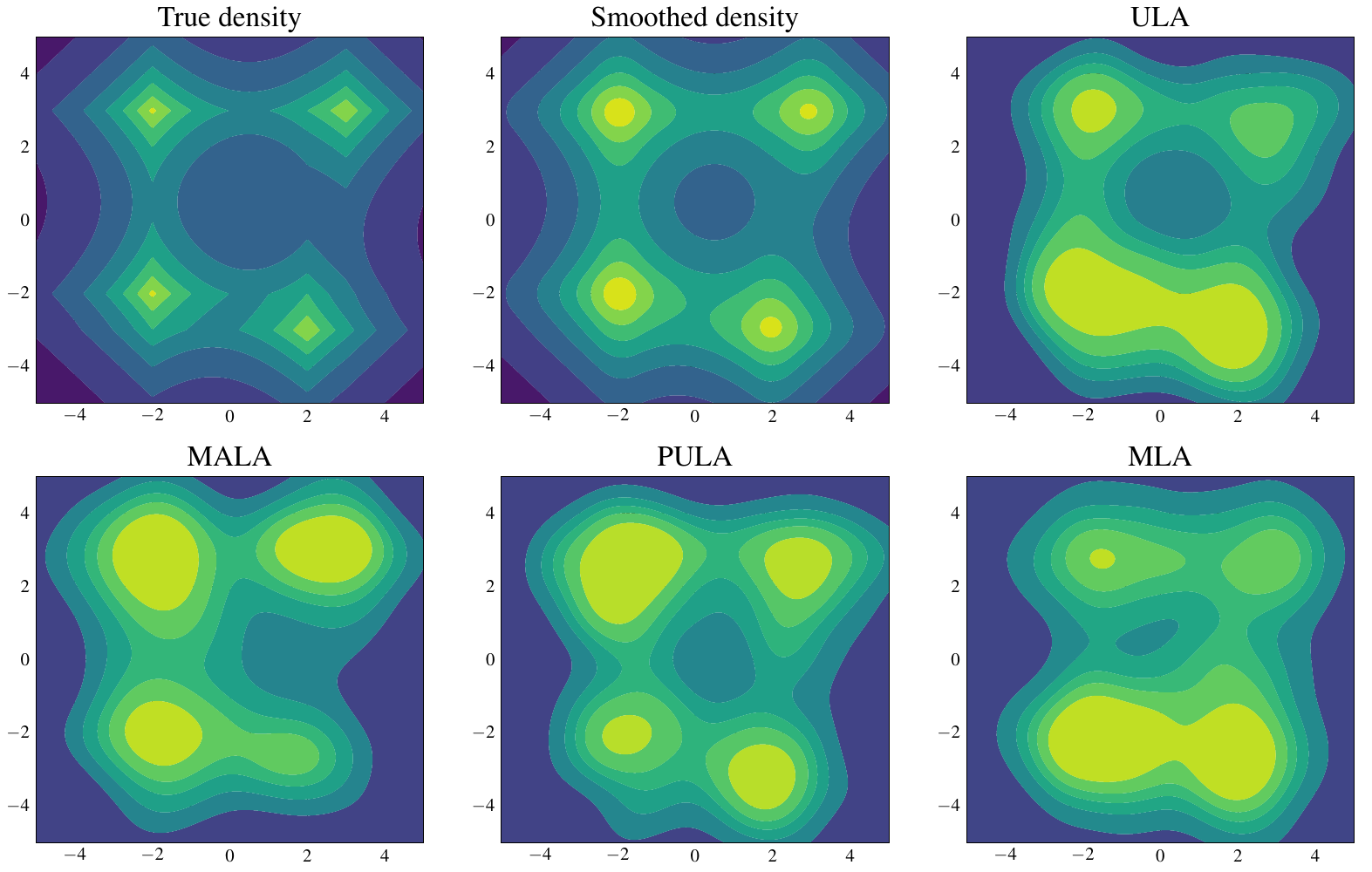}
                \caption{$K=4$}
            \end{subfigure}    
            \caption{Mixture of $K$ Laplacians with $(\gamma, \lambda)=(0.1, 1)$}
        \end{figure} 
        
        \begin{figure}[htbp]
            \begin{subfigure}[t]{.48\textwidth}
                \centering
                \includegraphics[width=\textwidth]{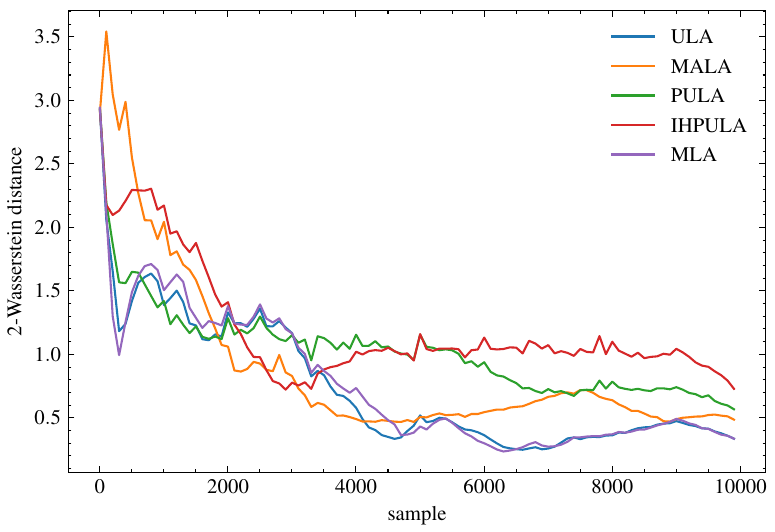}
                \caption{$K=2$}
                \vspace*{1mm}
            \end{subfigure}    
            \hfill
            \begin{subfigure}[t]{.48\textwidth}
                \centering
                \includegraphics[width=\textwidth]{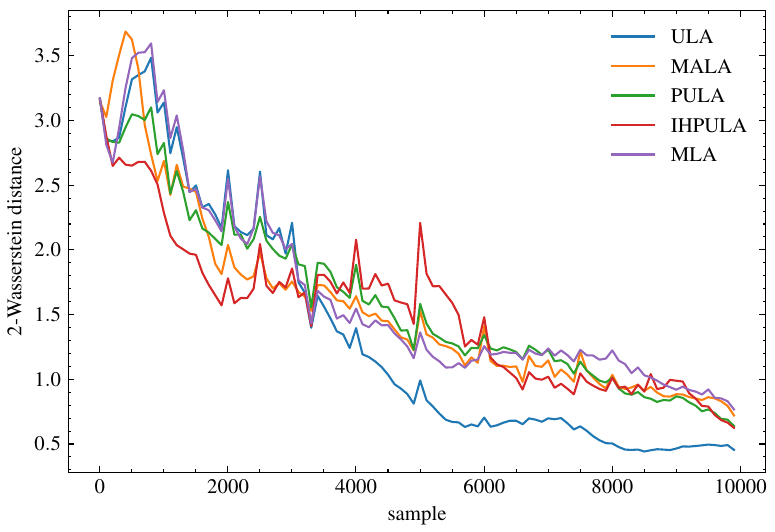}
                \caption{$K=3$}
                \vspace*{1mm}
            \end{subfigure}     
            \par\vspace{2mm}
            \begin{subfigure}[t]{.48\textwidth}
                \centering
                \includegraphics[width=\textwidth]{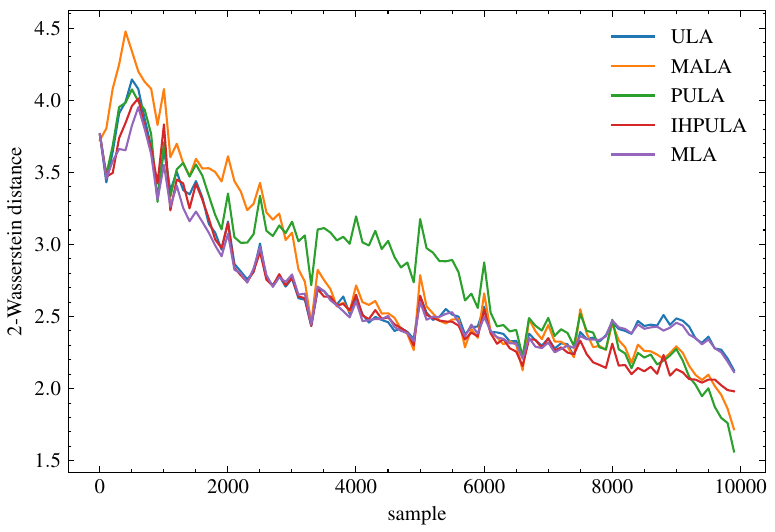}
                \caption{$K=4$}
            \end{subfigure} 
            \hfill
            \begin{subfigure}[t]{.48\textwidth}
                \centering
                \includegraphics[width=\textwidth]{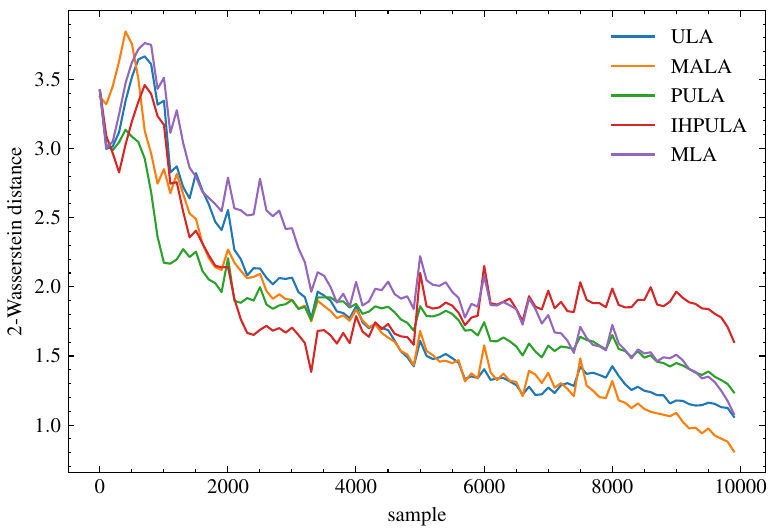}
                \caption{$K=5$}
            \end{subfigure}
            \caption{$2$-Wasserstein distances between generated samples by LMC algorithms and true samples of mixture of $K$ Gaussians with step size $\gamma = 0.01$}
            \label{fig:gaussians_wass_3}
        \end{figure} 
        
        \begin{figure}[htbp]
            \begin{subfigure}[t]{.48\textwidth}
                \centering
                \includegraphics[width=\textwidth]{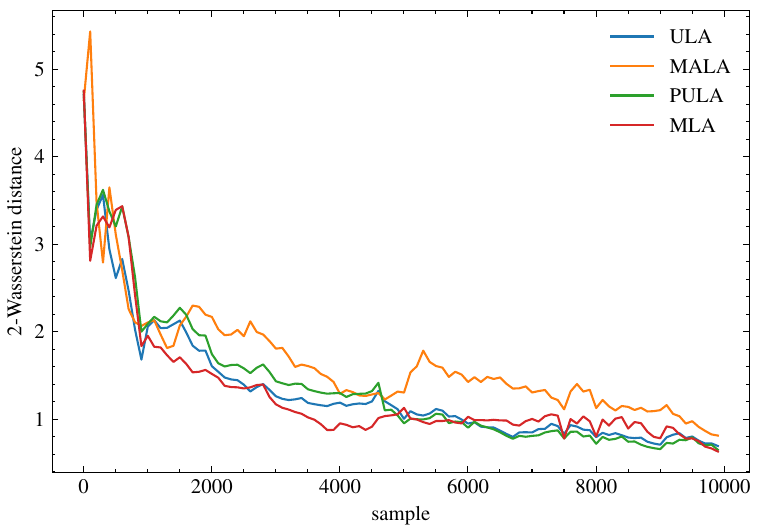}
                \caption{$K=2$}
                \vspace*{1mm}
            \end{subfigure}    
            \hfill
            \begin{subfigure}[t]{.48\textwidth}
                \centering
                \includegraphics[width=\textwidth]{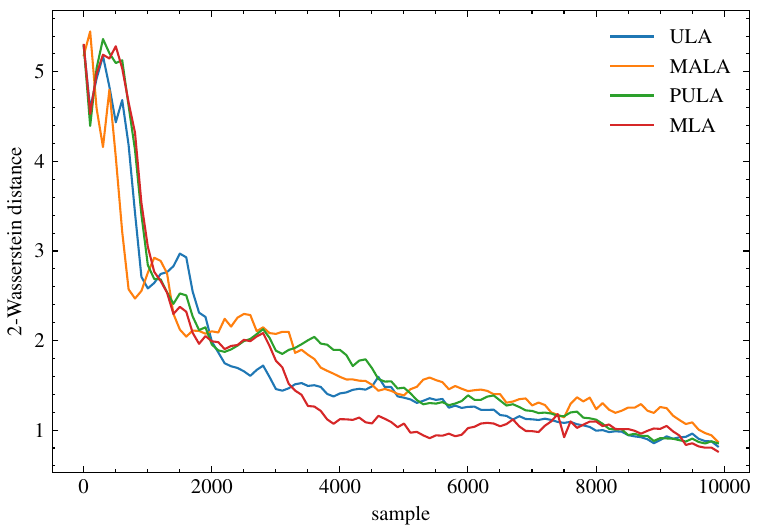}
                \caption{$K=4$}
                \vspace*{1mm}
            \end{subfigure}     
            \caption{$2$-Wasserstein distances between generated samples by LMC algorithms and true samples of mixture of $K$ Laplacians with $(\gamma, \lambda)=(0.1, 1)$ }
            \label{fig:laplacians_wass_1}
        \end{figure}
        
        \begin{figure}[htbp]
            \centering
            \begin{subfigure}[h]{.48\textwidth}
                \centering
                \includegraphics[height=.18\textheight]{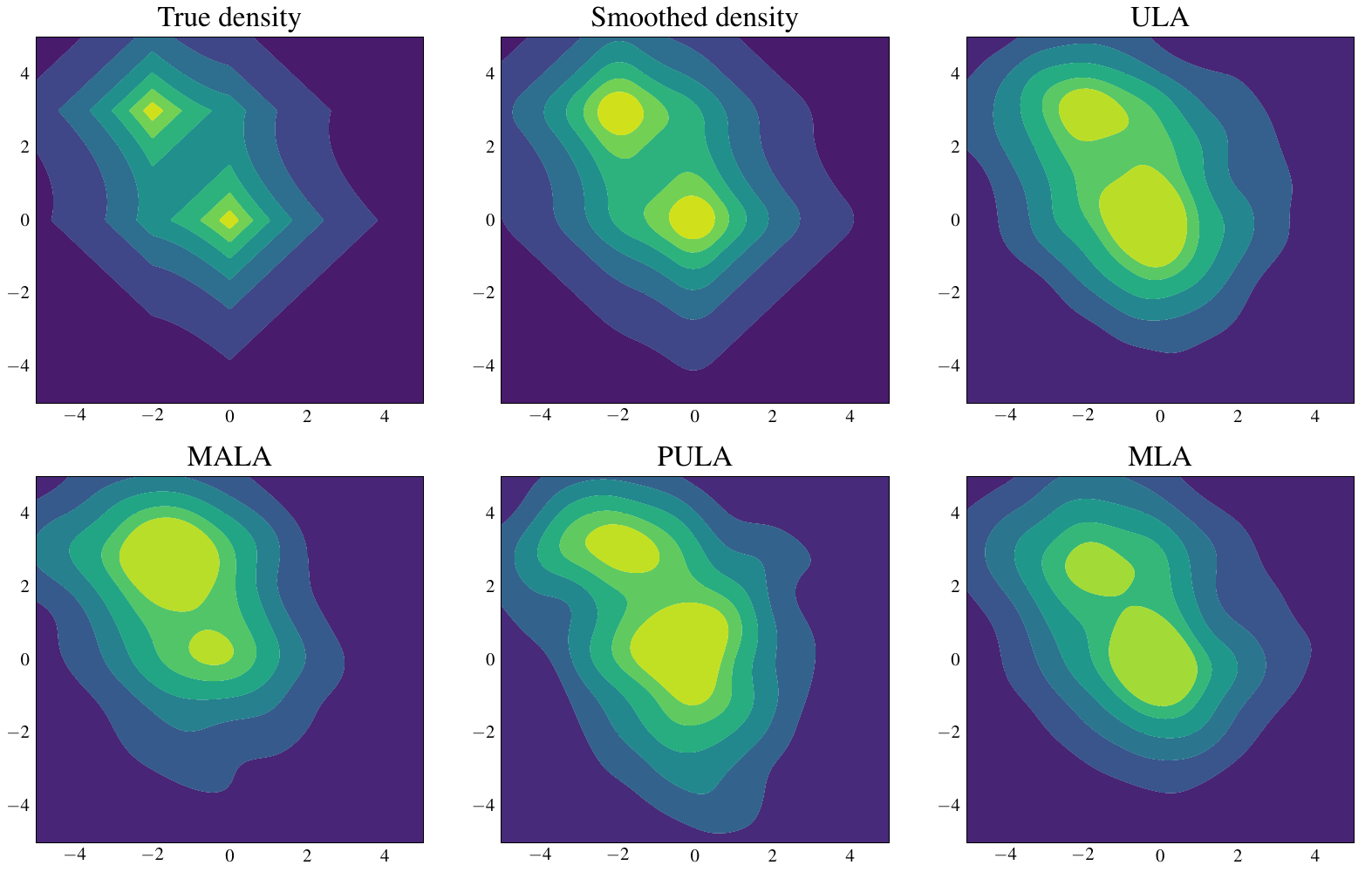}
                \caption{$K=2$}
                \vspace*{1mm}
            \end{subfigure}    
            \hfill
            \begin{subfigure}[h]{.48\textwidth}
                \centering
                \includegraphics[height=.18\textheight]{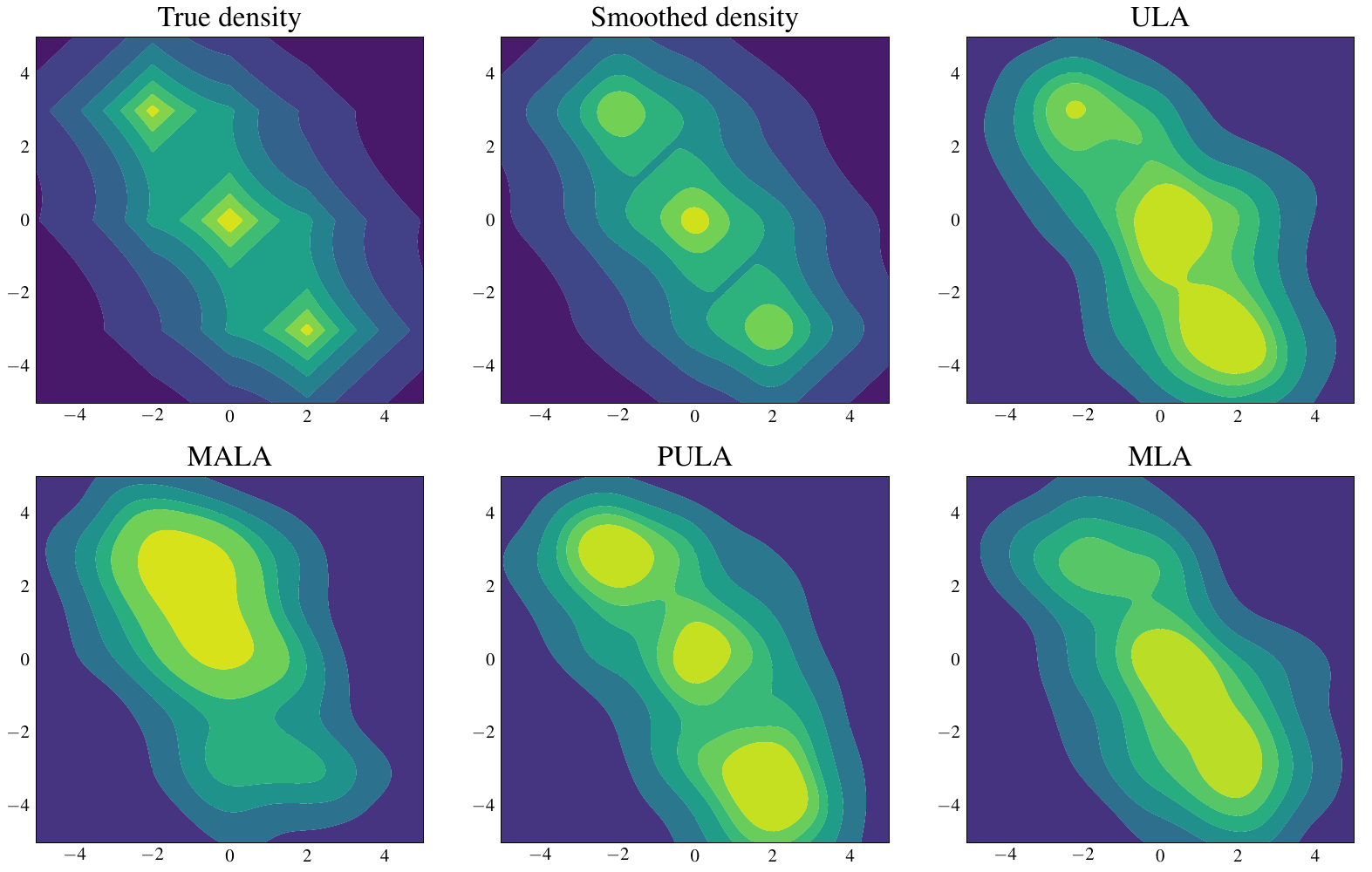}
                \caption{$K=3$}
                \vspace*{1mm}
            \end{subfigure}      
            \par\vspace{2mm}
            \begin{subfigure}[h]{.48\textwidth}
                \centering
                \includegraphics[height=.18\textheight]{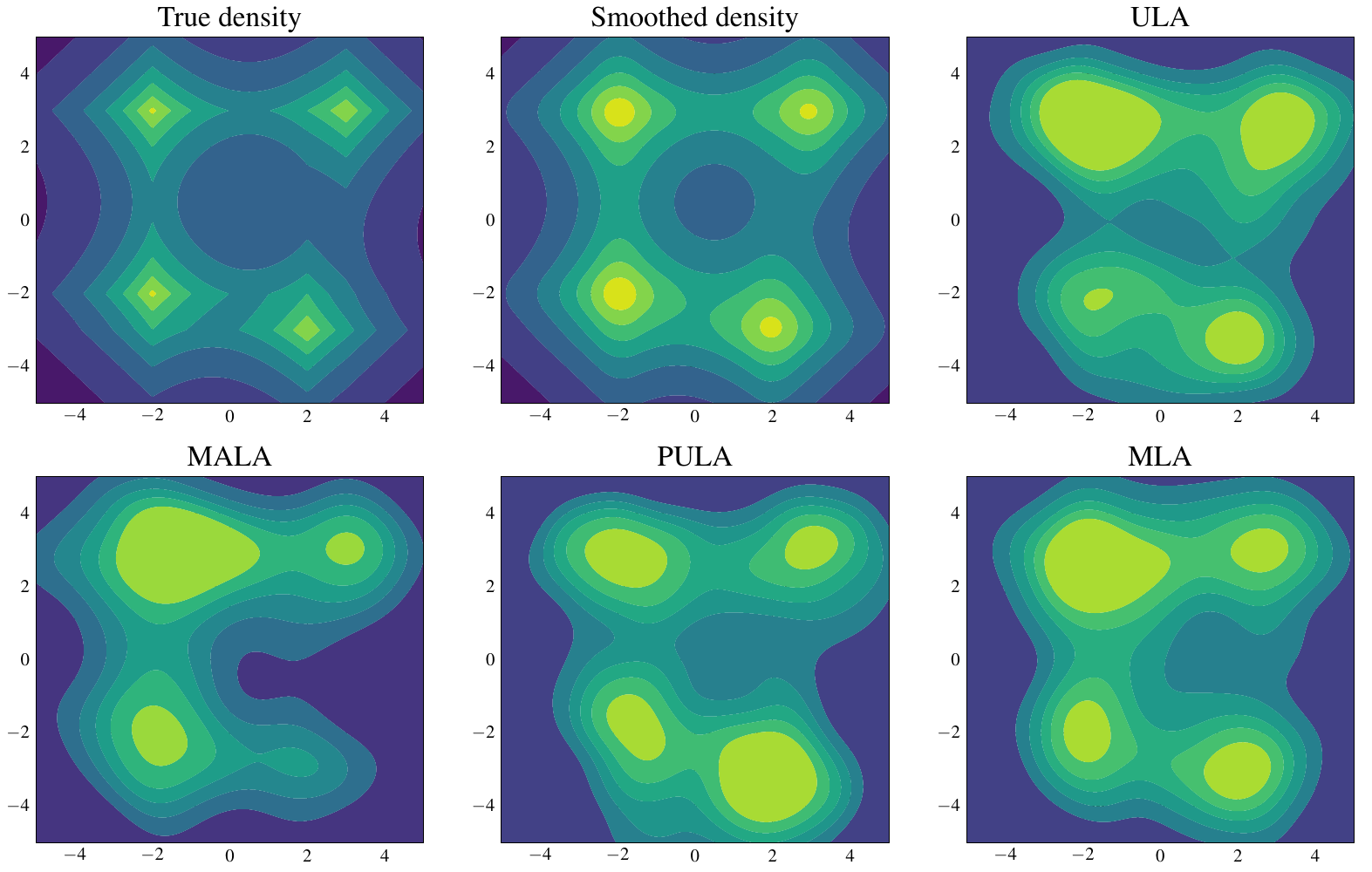}
                \caption{$K=4$}
            \end{subfigure}  
            \hfill
            \begin{subfigure}[h]{.48\textwidth}
                \centering
                \includegraphics[height=.18\textheight]{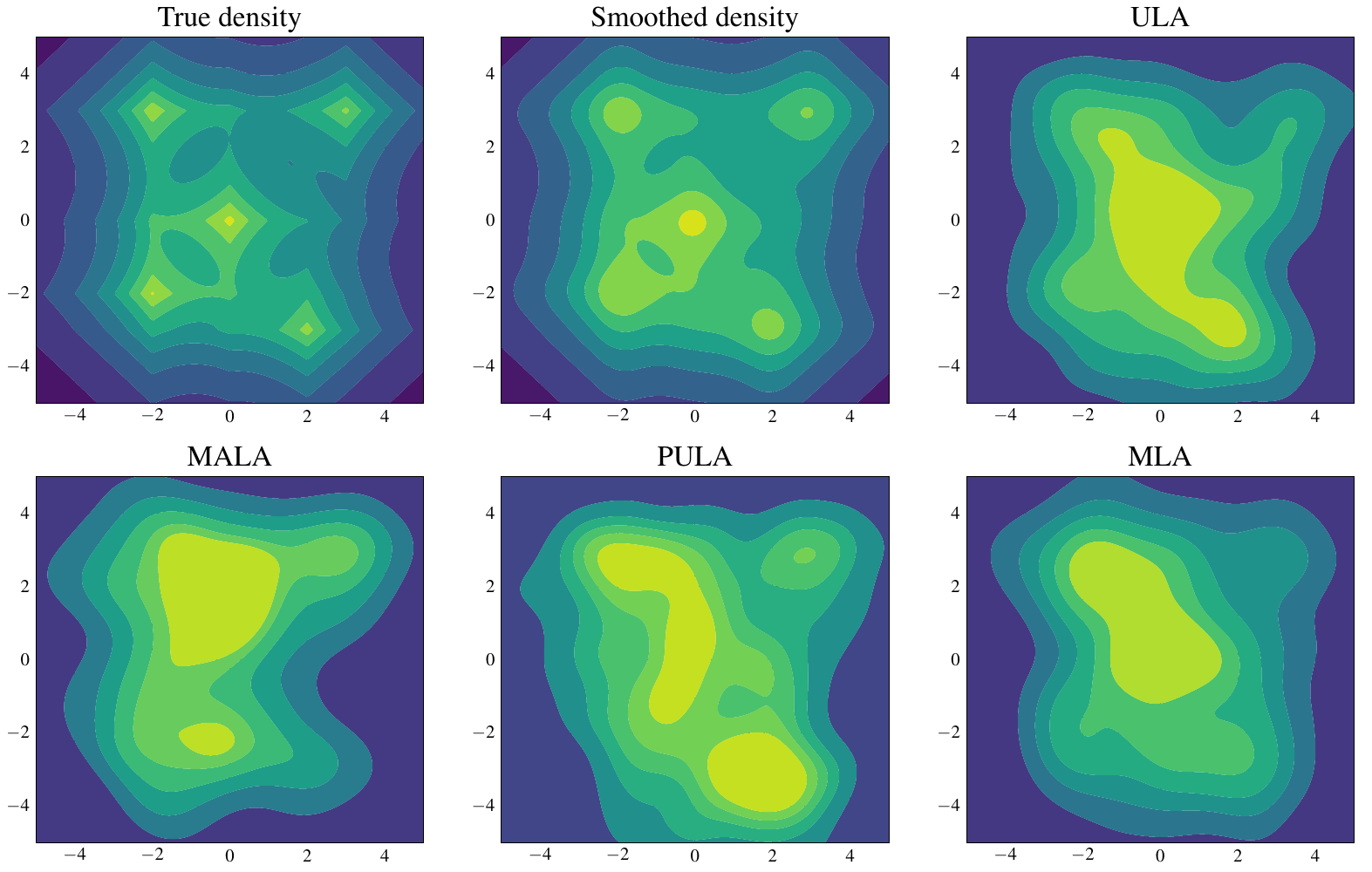}
                \caption{$K=5$}
            \end{subfigure}
            \caption{Mixture of $K$ Laplacians with $(\gamma, \lambda)=(0.05, 1)$}
        \end{figure}

        \begin{figure}[htbp]
            \centering
            \begin{subfigure}[h]{.48\textwidth}
                \centering
                \includegraphics[height=.18\textheight]{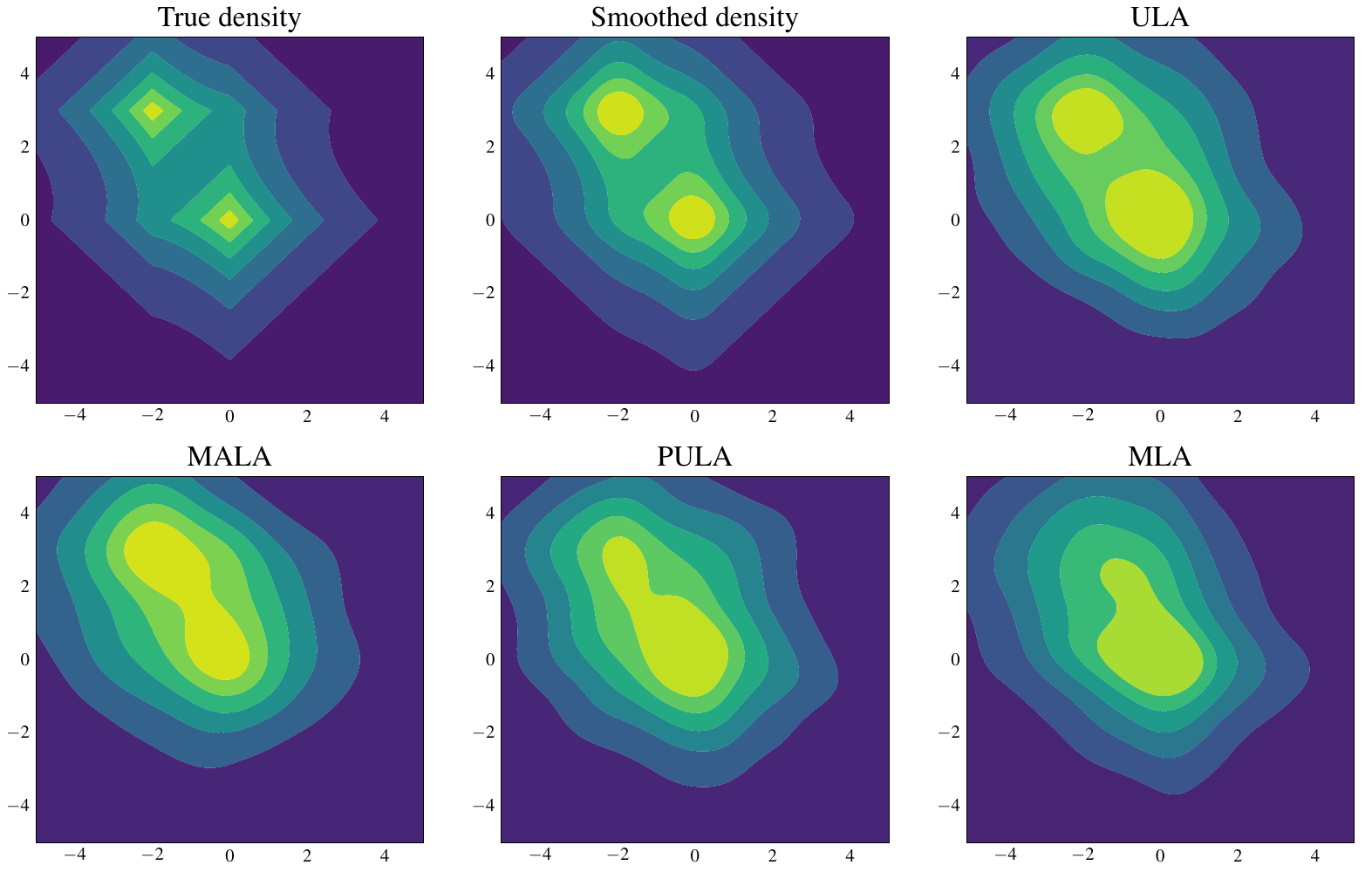}
                \caption{$K=2$}
                \vspace*{1mm}
            \end{subfigure}    
            \hfill
            \begin{subfigure}[h]{.48\textwidth}
                \centering
                \includegraphics[height=.18\textheight]{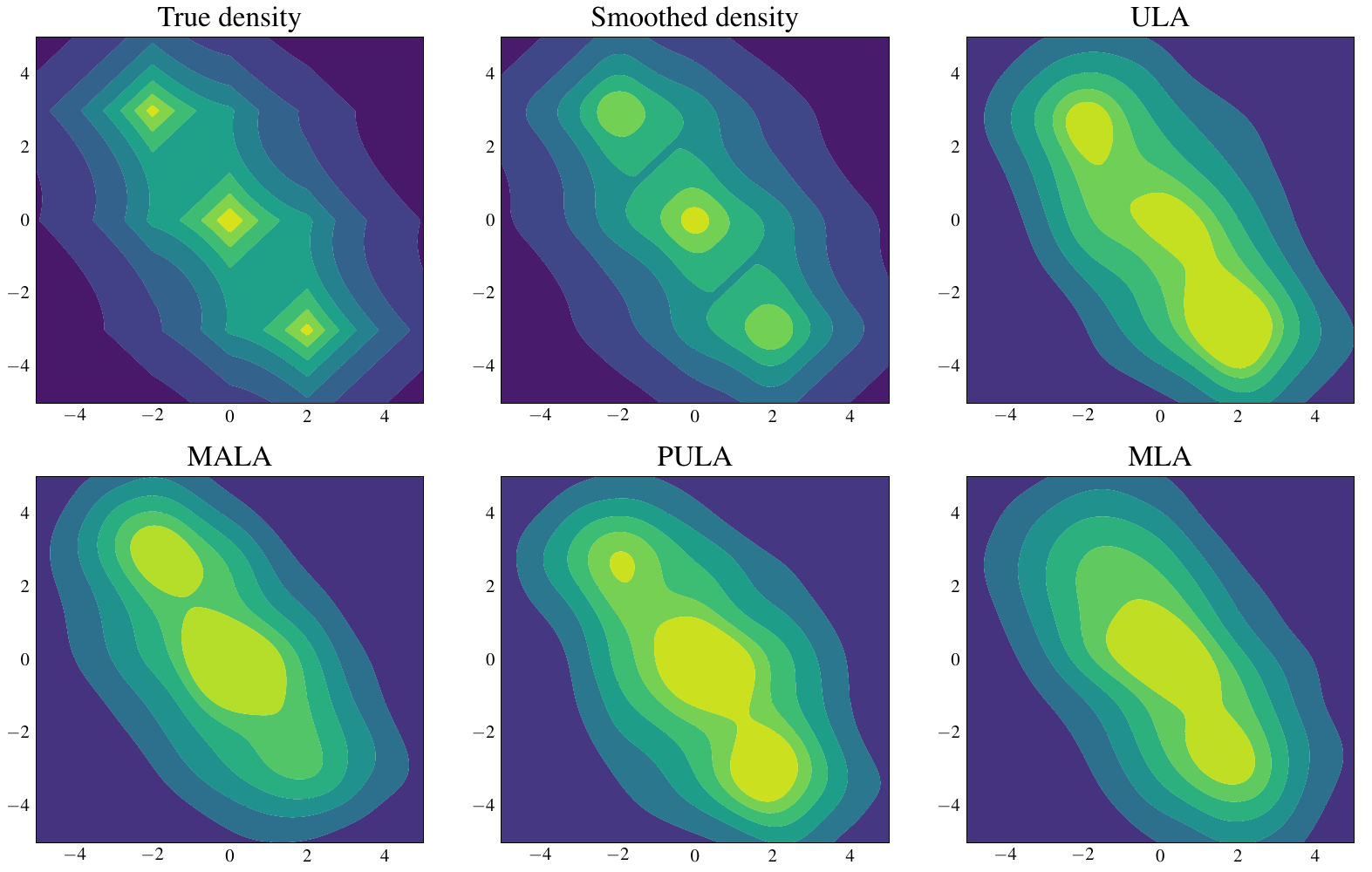}
                \caption{$K=3$}
                \vspace*{1mm}
            \end{subfigure}      
            \par\vspace{2mm}
            \begin{subfigure}[h]{.48\textwidth}
                \centering
                \includegraphics[height=.18\textheight]{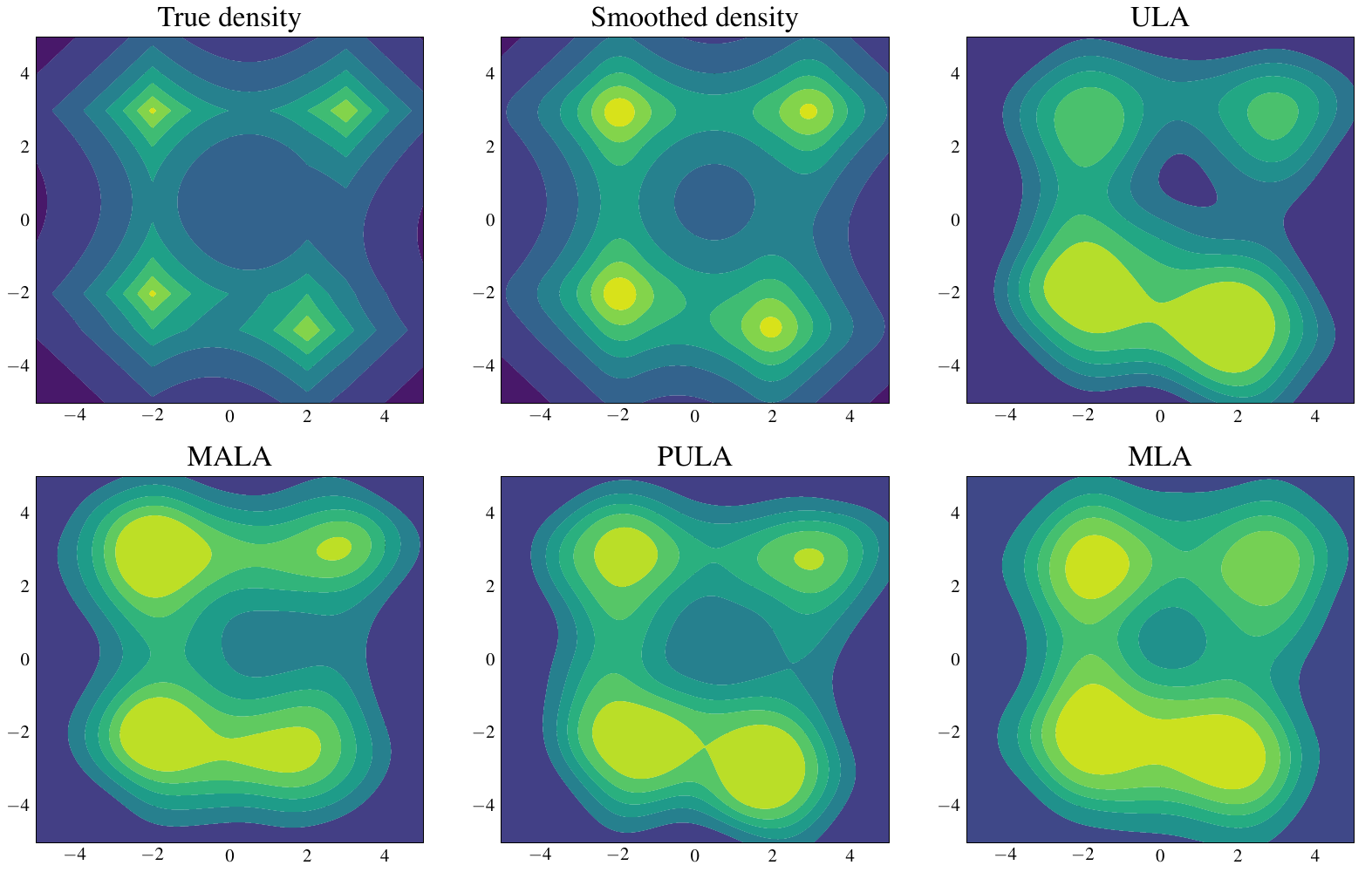}
                \caption{$K=4$}
            \end{subfigure}  
            \hfill
            \begin{subfigure}[h]{.48\textwidth}
                \centering
                \includegraphics[height=.18\textheight]{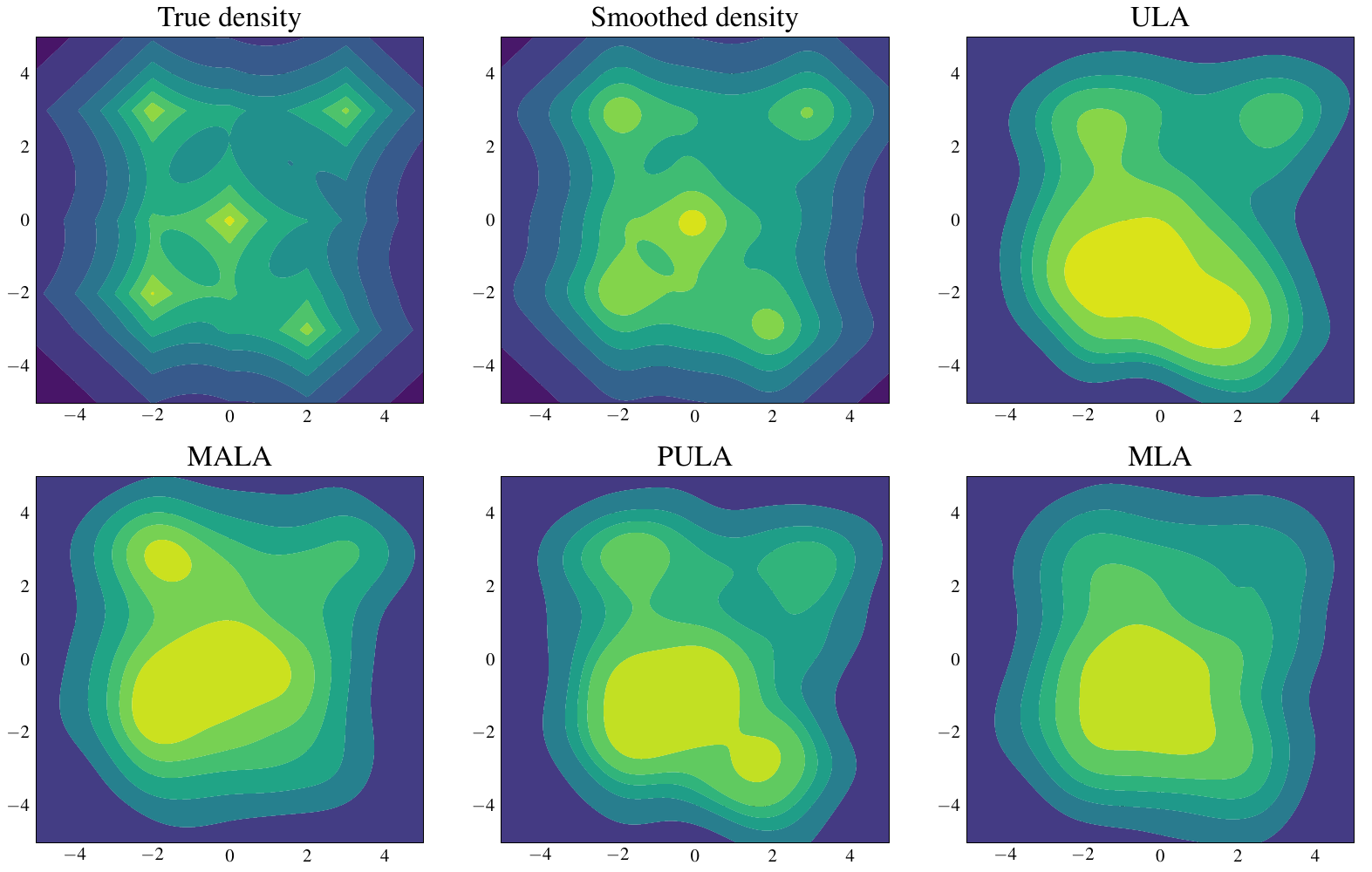}
                \caption{$K=5$}
            \end{subfigure}
            \caption{Mixture of $K$ Laplacians with $(\gamma, \lambda)=(0.15, 1)$}
        \end{figure}       
        
        \begin{figure}[htbp]
            \begin{subfigure}[t]{.48\textwidth}
                \centering
                \includegraphics[height=.18\textheight]{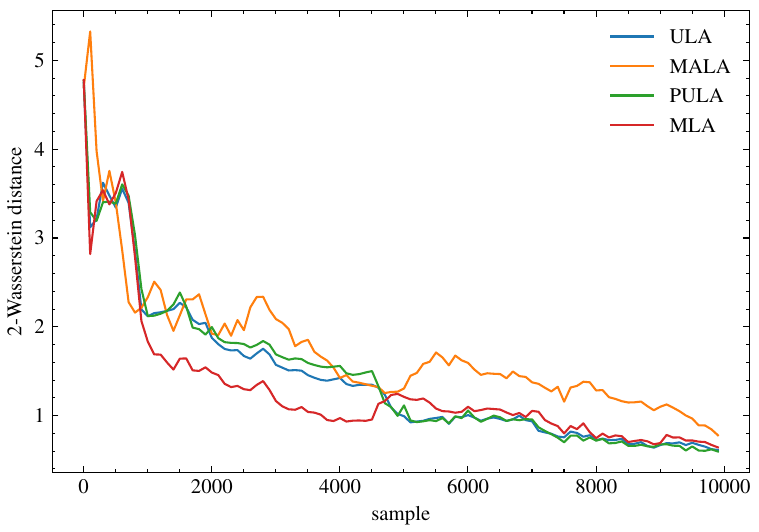}
                \caption{$K=2$}
                \vspace*{1mm}
            \end{subfigure}    
            \hfill
            \begin{subfigure}[t]{.48\textwidth}
                \centering
                \includegraphics[height=.18\textheight]{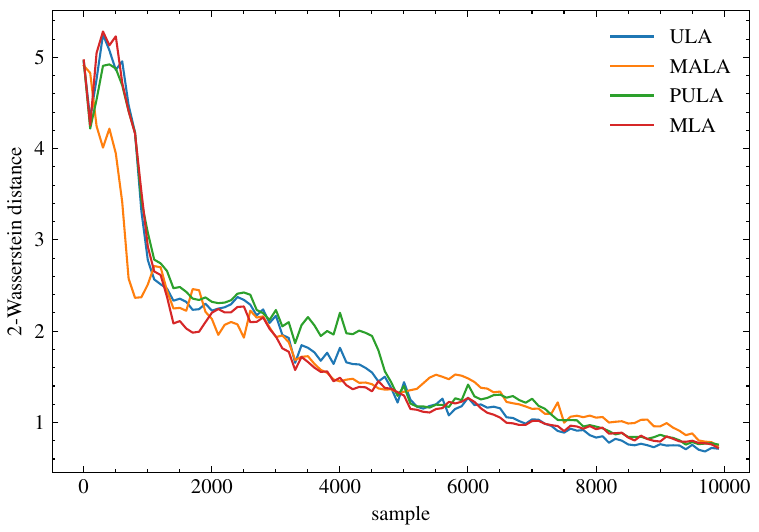}
                \caption{$K=3$}
                \vspace*{1mm}
            \end{subfigure}     
            \par\vspace{2mm}
            \begin{subfigure}[t]{.48\textwidth}
                \centering
                \includegraphics[height=.18\textheight]{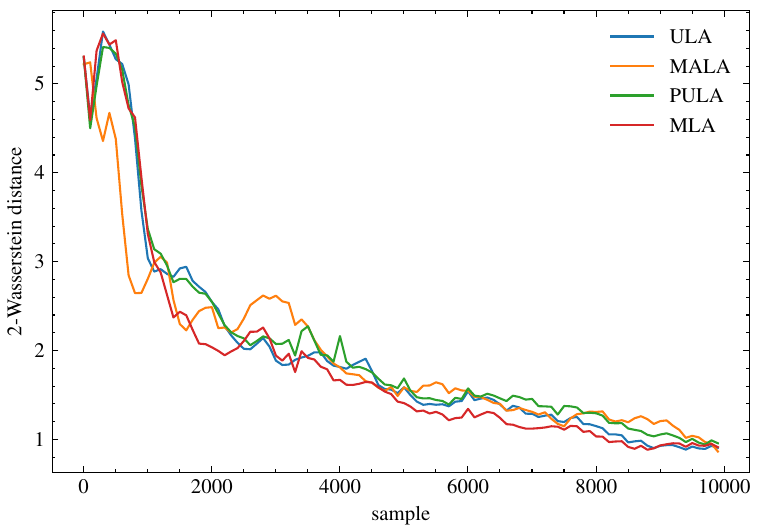}
                \caption{$K=4$}
            \end{subfigure} 
            \hfill
            \begin{subfigure}[t]{.48\textwidth}
                \centering
                \includegraphics[height=.18\textheight]{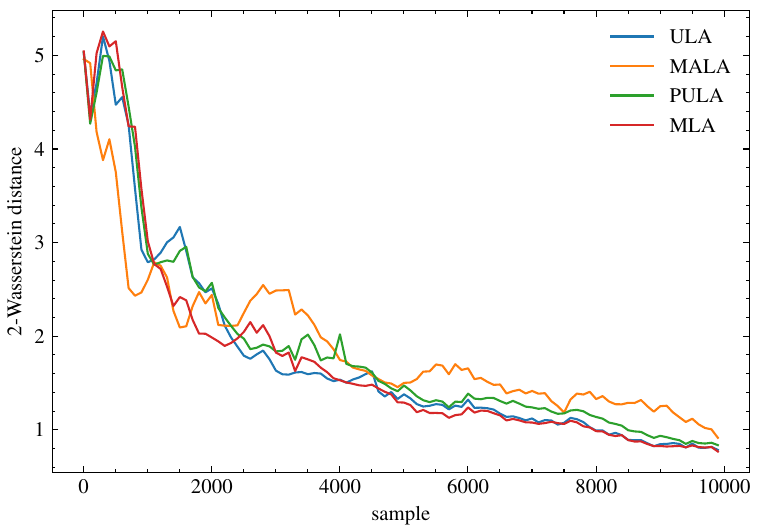}
                \caption{$K=5$}
            \end{subfigure}
            \caption{$2$-Wasserstein distances between generated samples by LMC algorithms and true samples of mixture of $K$ Laplacians with $(\gamma, \lambda)=(0.05, 1)$}
            \label{fig:laplacians_wass_2}
        \end{figure} 
            
        \begin{figure}[htbp]
            \begin{subfigure}[t]{.48\textwidth}
                \centering
                \includegraphics[height=.18\textheight]{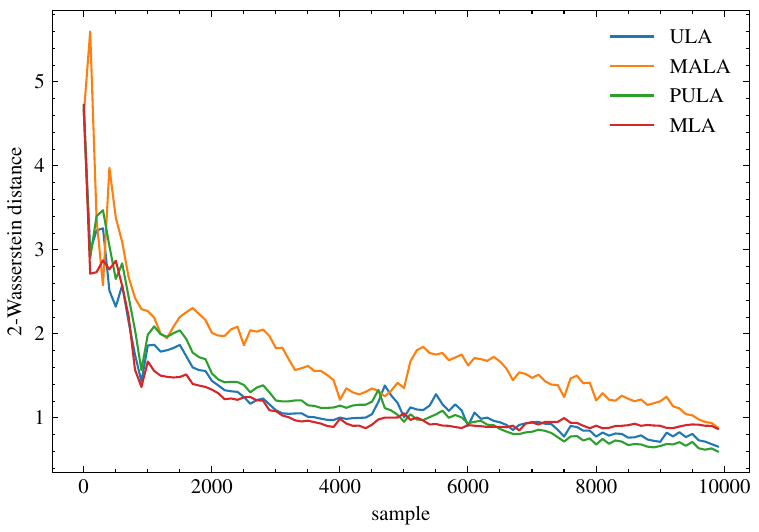}
                \caption{$K=2$}
                \vspace*{1mm}
            \end{subfigure}    
            \hfill
            \begin{subfigure}[t]{.48\textwidth}
                \centering
                \includegraphics[height=.18\textheight]{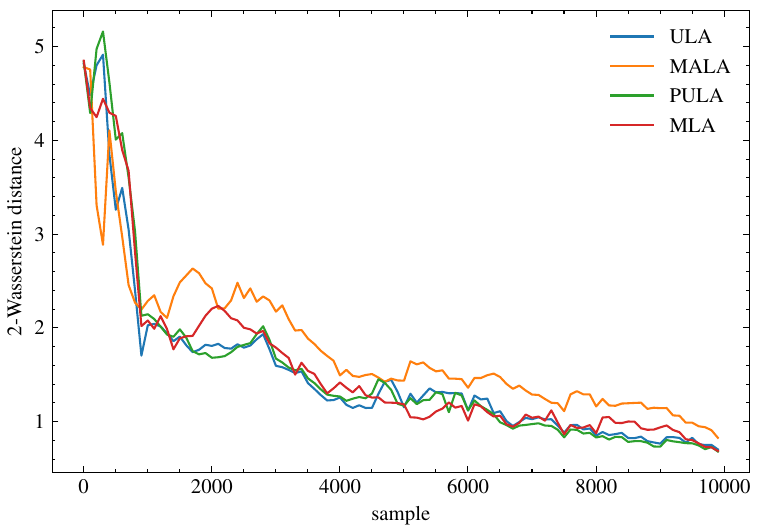}
                \caption{$K=3$}
                \vspace*{1mm}
            \end{subfigure}     
            \par\vspace{2mm}
            \begin{subfigure}[t]{.48\textwidth}
                \centering
                \includegraphics[height=.18\textheight]{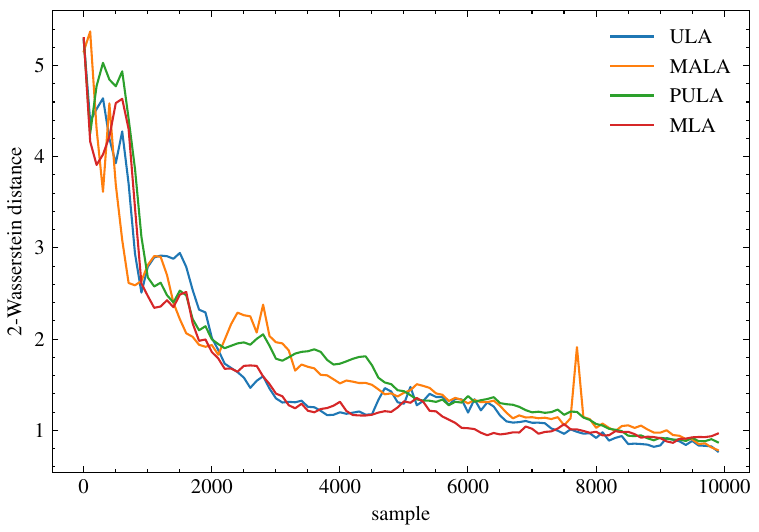}
                \caption{$K=4$}
            \end{subfigure} 
            \hfill
            \begin{subfigure}[t]{.48\textwidth}
                \centering
                \includegraphics[height=.18\textheight]{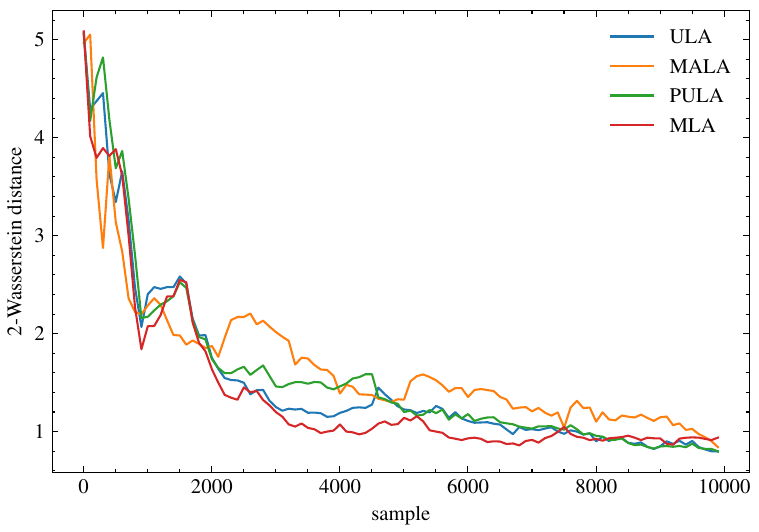}
                \caption{$K=5$}
            \end{subfigure}
            \caption{$2$-Wasserstein distances between generated samples by LMC algorithms and true samples of mixture of $K$ Laplacians with $(\gamma, \lambda)=(0.15, 1)$}
            \label{fig:laplacians_wass_3}
        \end{figure} 
        
        \begin{figure}[htbp]
            \centering
            \begin{subfigure}[h]{.48\textwidth}
                \centering
                \includegraphics[height=.18\textheight]{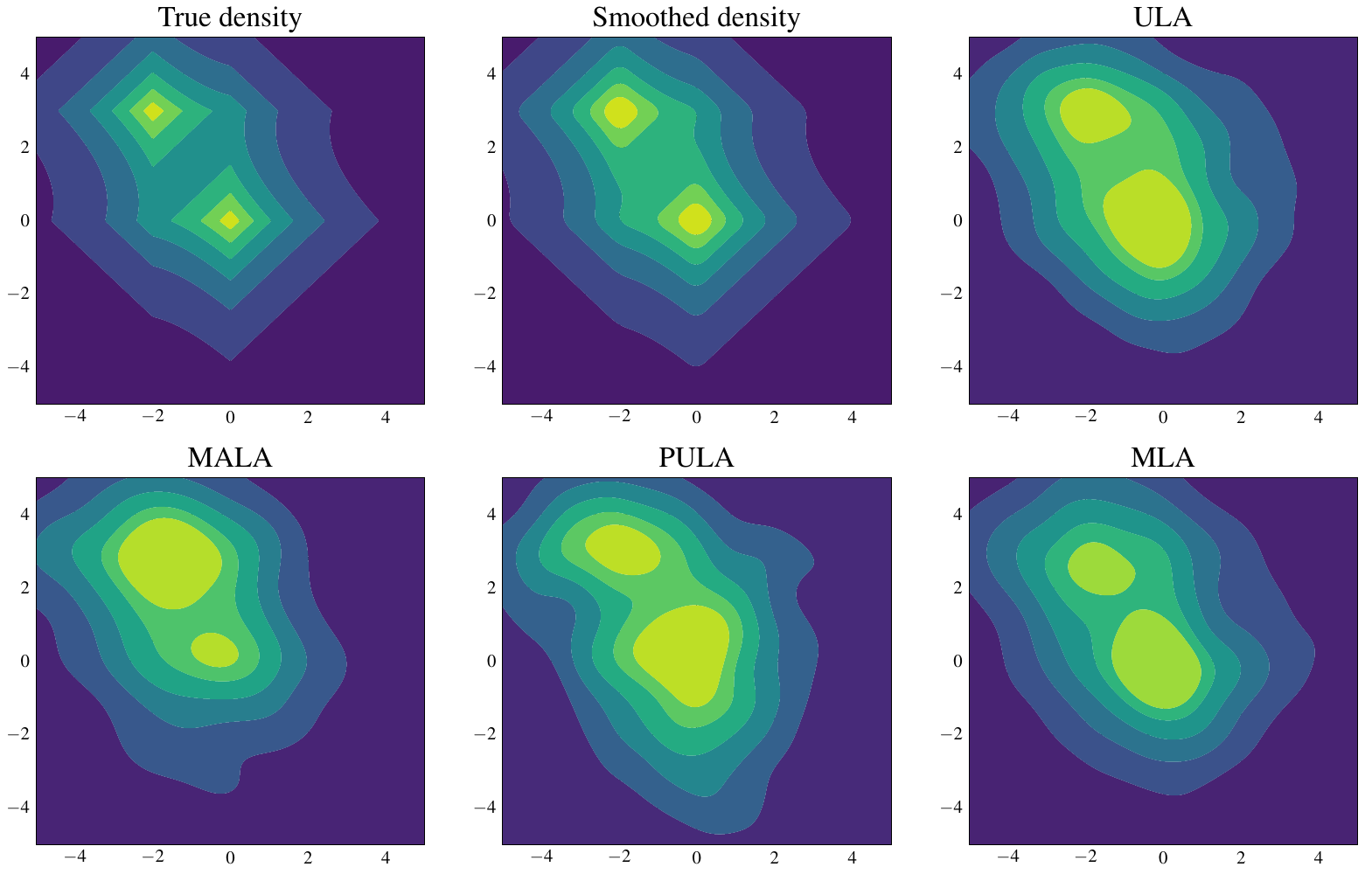}
                \caption{$K=2$}
                \vspace*{1mm}
            \end{subfigure}    
            \hfill
            \begin{subfigure}[h]{.48\textwidth}
                \centering
                \includegraphics[height=.18\textheight]{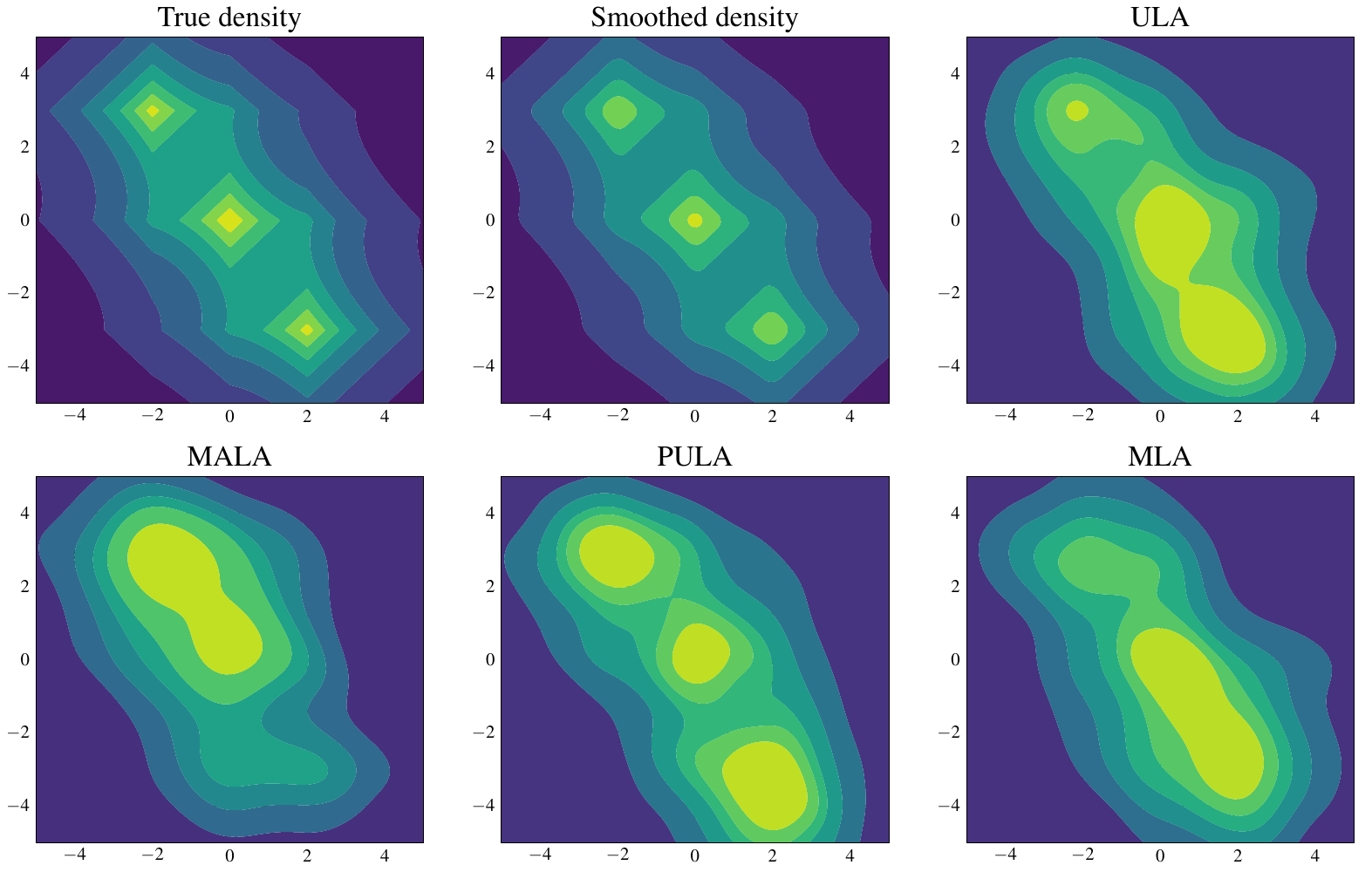}
                \caption{$K=3$}
                \vspace*{1mm}
            \end{subfigure}      
            \par\vspace{2mm}
            \begin{subfigure}[h]{.48\textwidth}
                \centering
                \includegraphics[height=.18\textheight]{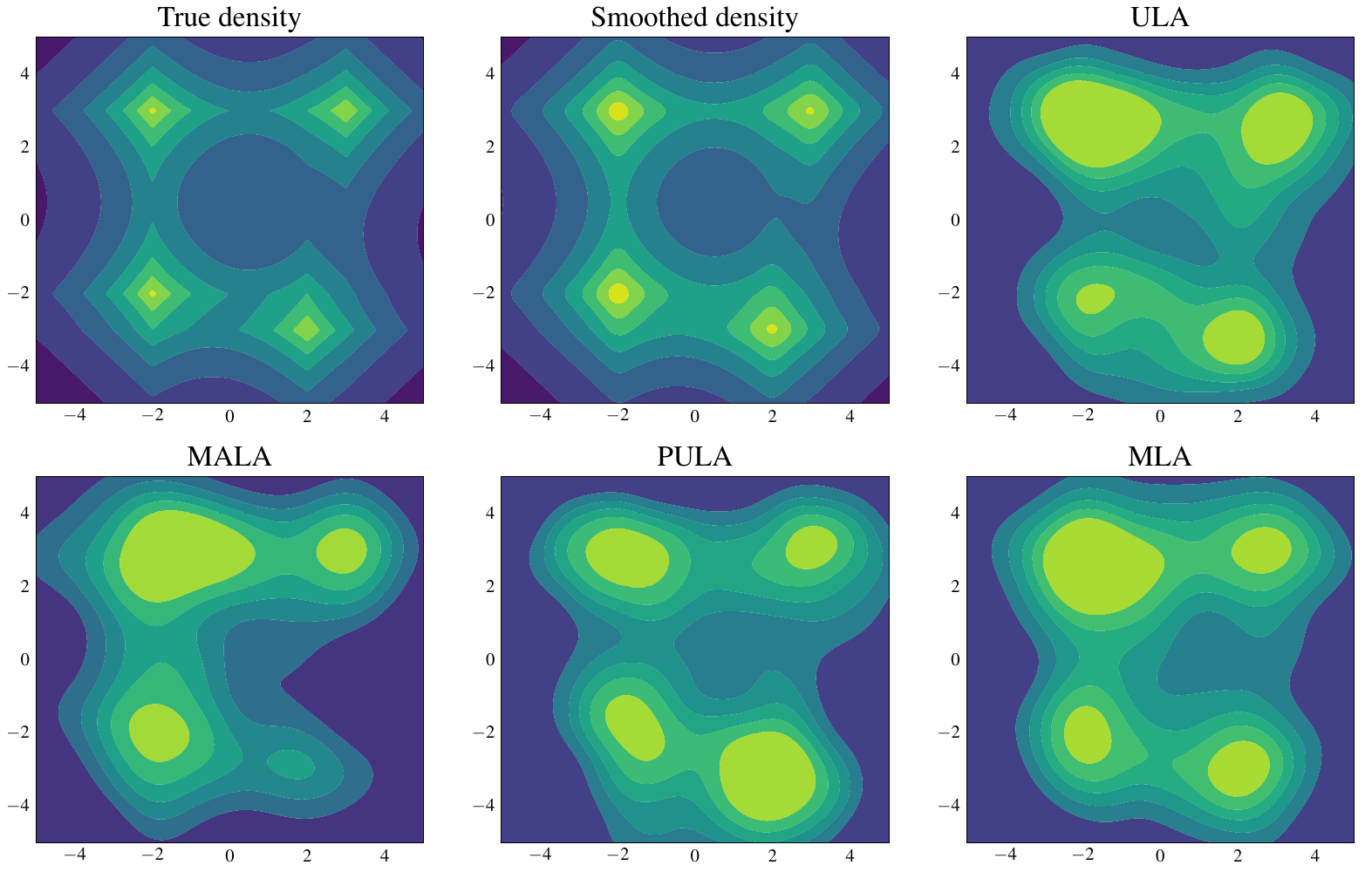}
                \caption{$K=4$}
            \end{subfigure}  
            \hfill
            \begin{subfigure}[h]{.48\textwidth}
                \centering
                \includegraphics[height=.18\textheight]{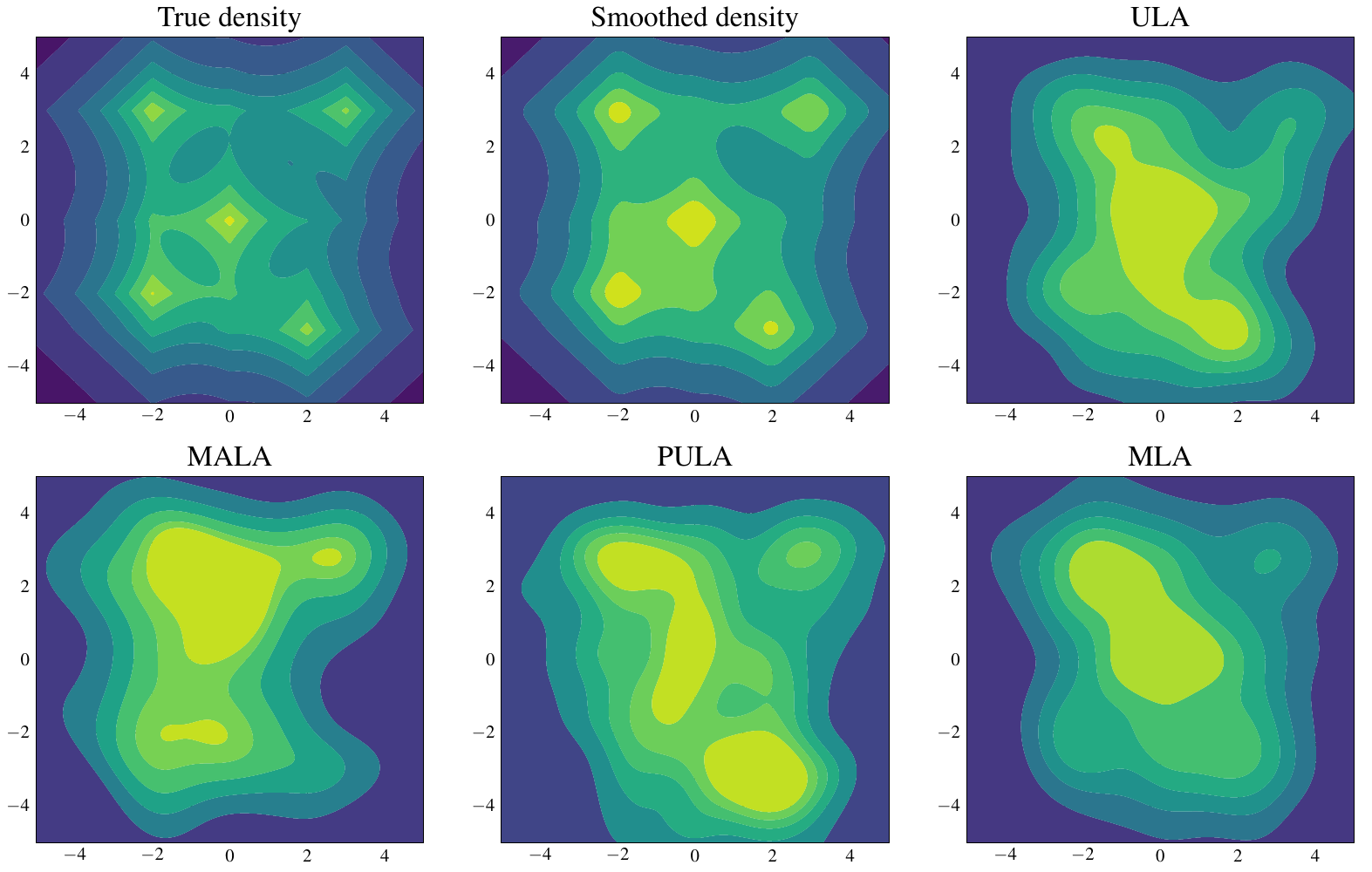}
                \caption{$K=5$}
            \end{subfigure}             
            \caption{Mixture of $K$ Laplacians with $(\gamma, \lambda)=(0.05, 0.5)$}
        \end{figure}

        \begin{figure}[htbp]
            \centering
            \begin{subfigure}[h]{.48\textwidth}
                \centering
                \includegraphics[height=.18\textheight]{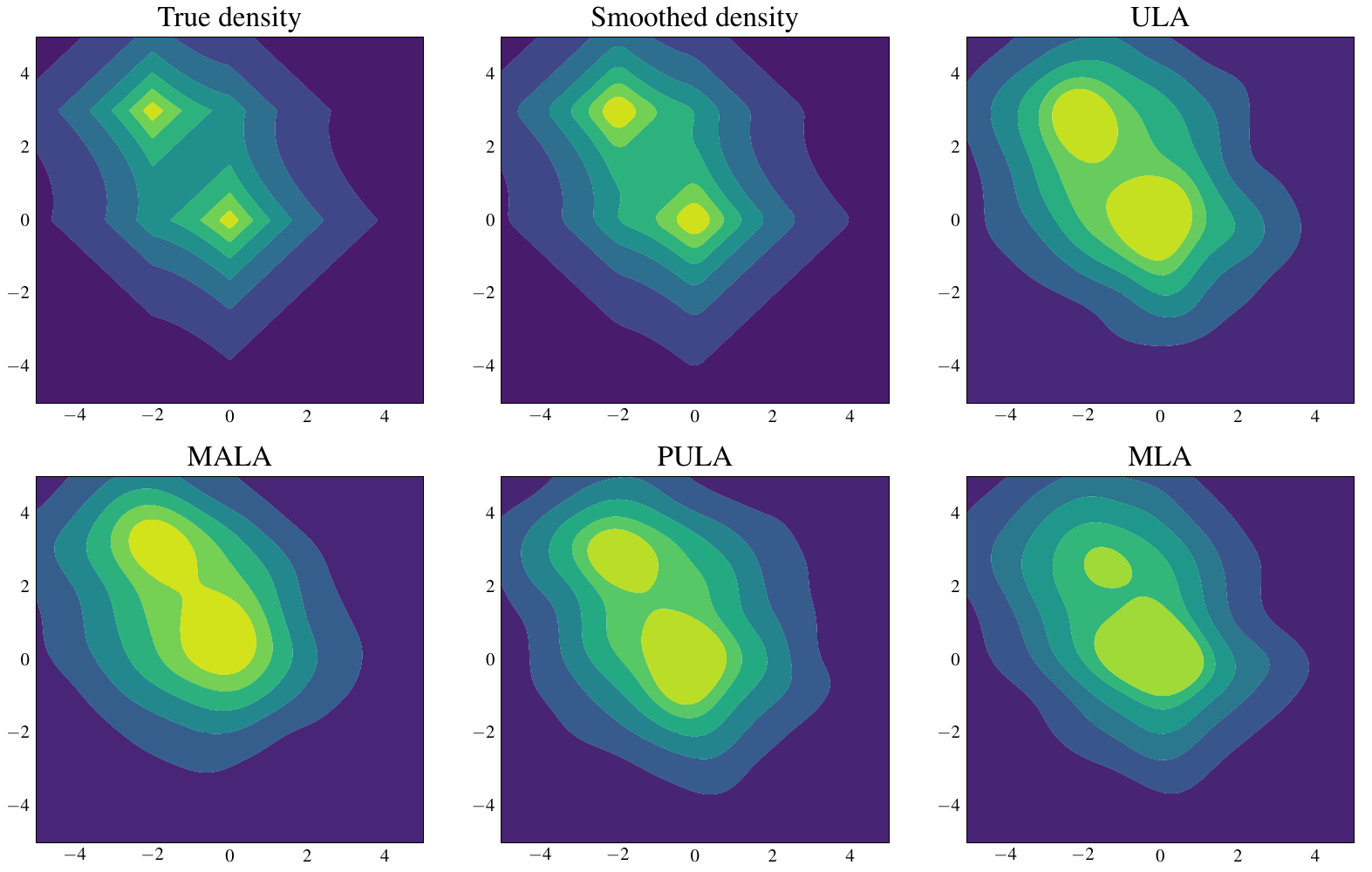}
                \caption{$K=2$}
                \vspace*{1mm}
            \end{subfigure}    
            \hfill
            \begin{subfigure}[h]{.48\textwidth}
                \centering
                \includegraphics[height=.18\textheight]{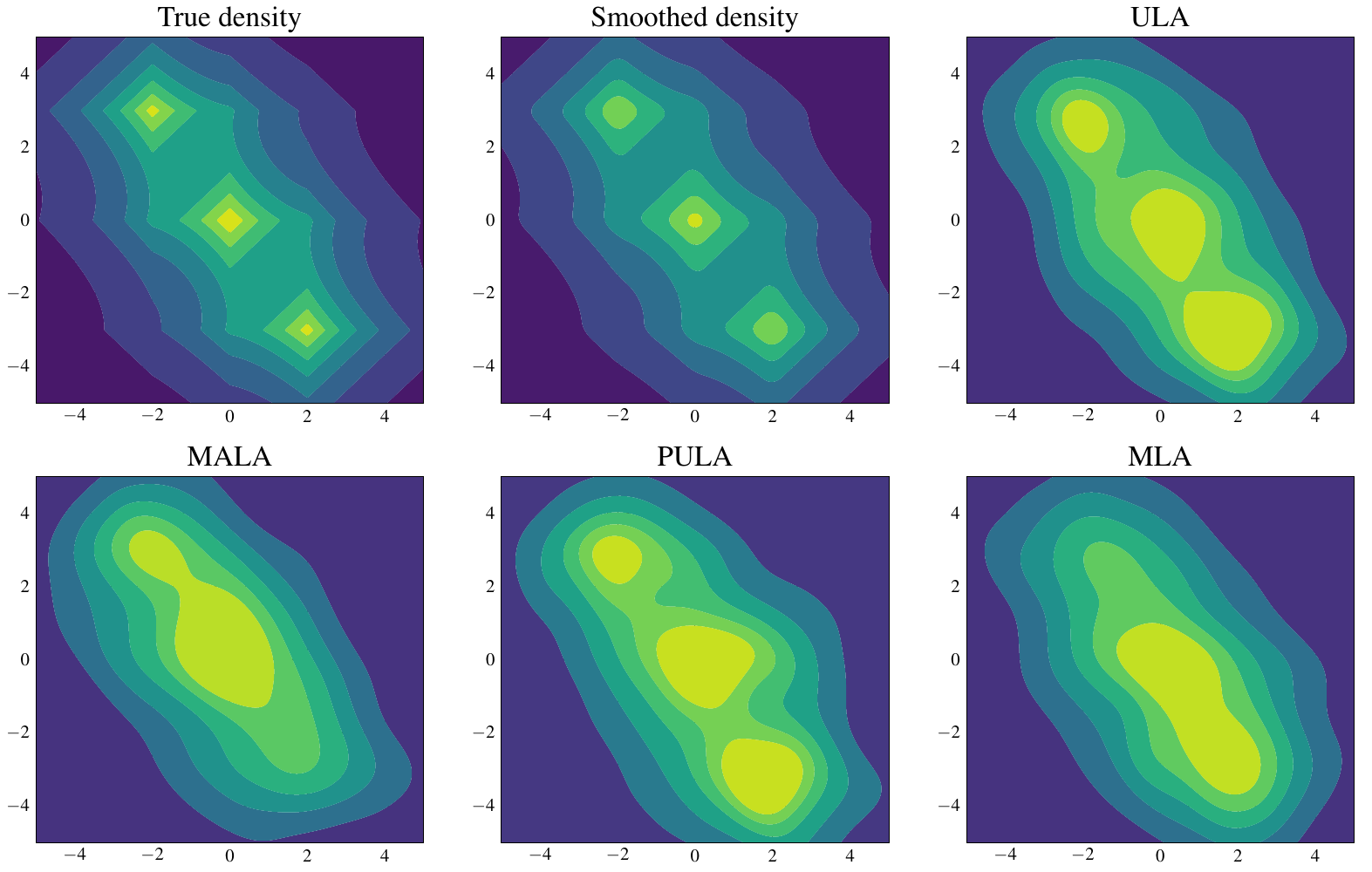}
                \caption{$K=3$}
                \vspace*{1mm}
            \end{subfigure}      
            \par\vspace{2mm}
            \begin{subfigure}[h]{.48\textwidth}
                \centering
                \includegraphics[height=.18\textheight]{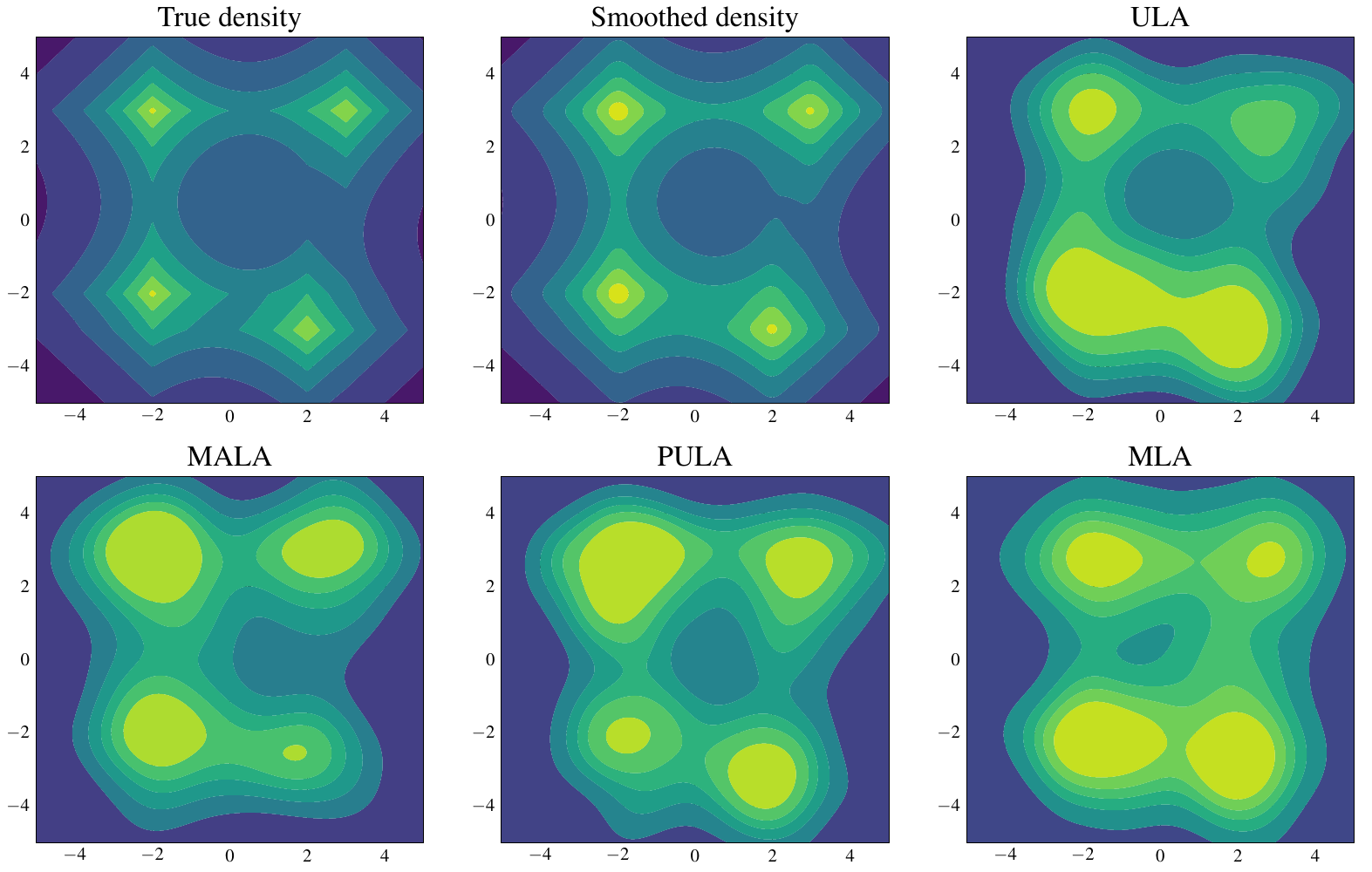}
                \caption{$K=4$}
            \end{subfigure}  
            \hfill
            \begin{subfigure}[h]{.48\textwidth}
                \centering
                \includegraphics[height=.18\textheight]{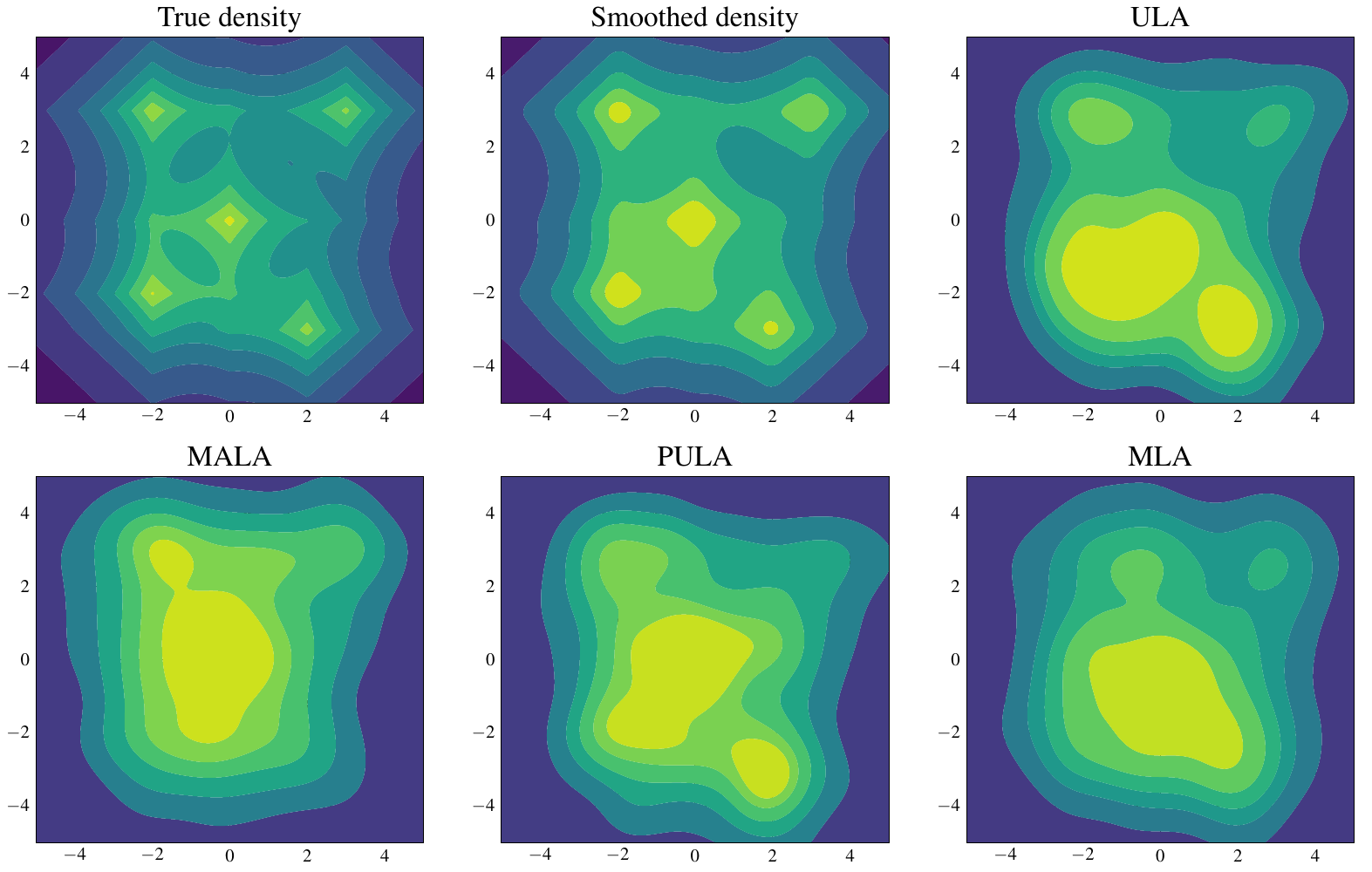}
                \caption{$K=5$}
            \end{subfigure}   
            \caption{Mixture of $K$ Laplacians with $(\gamma, \lambda)=(0.1, 0.5)$}
        \end{figure} 
        
        \begin{figure}[htbp]
            \begin{subfigure}[t]{.48\textwidth}
                \centering
                \includegraphics[height=.18\textheight]{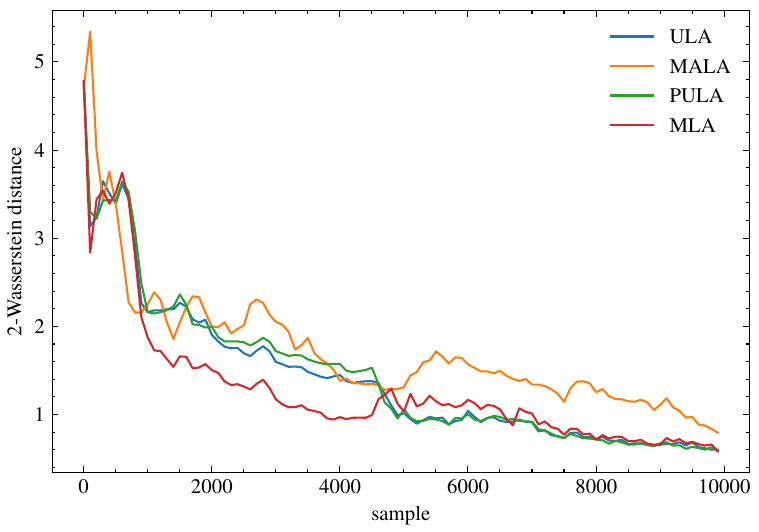}
                \caption{$K=2$}
                \vspace*{1mm}
            \end{subfigure}    
            \hfill
            \begin{subfigure}[t]{.48\textwidth}
                \centering
                \includegraphics[height=.18\textheight]{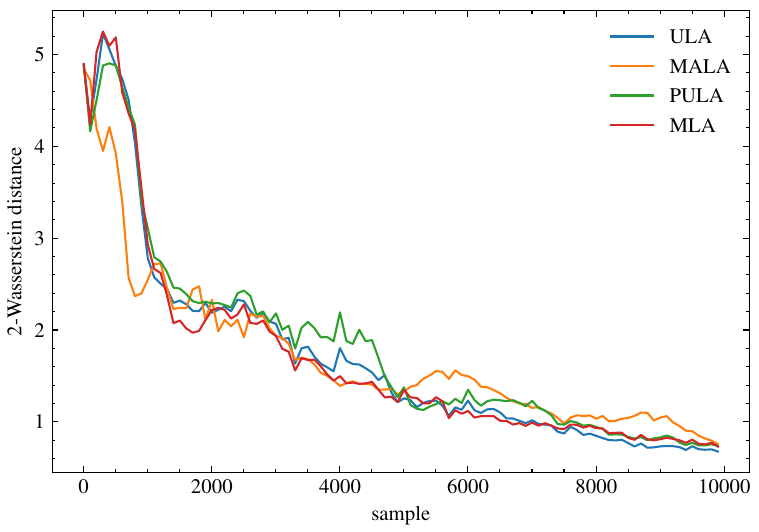}
                \caption{$K=3$}
                \vspace*{1mm}
            \end{subfigure}     
            \par\vspace{2mm}
            \begin{subfigure}[t]{.48\textwidth}
                \centering
                \includegraphics[height=.18\textheight]{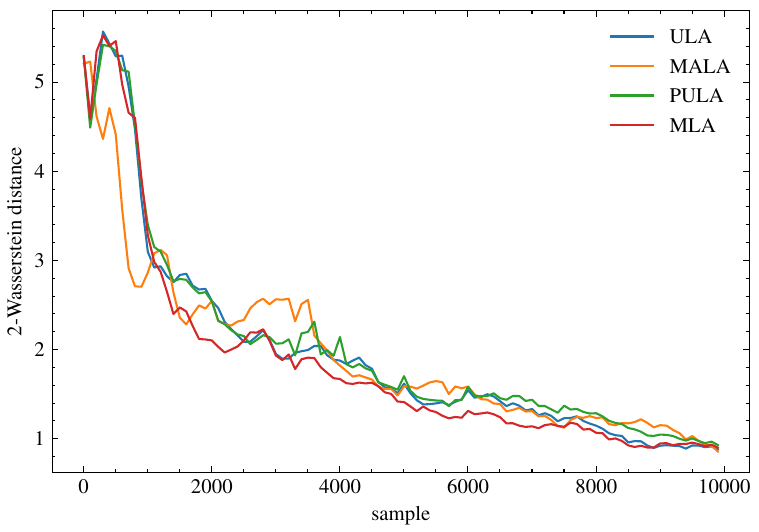}
                \caption{$K=4$}
            \end{subfigure} 
            \hfill
            \begin{subfigure}[t]{.48\textwidth}
                \centering
                \includegraphics[height=.18\textheight]{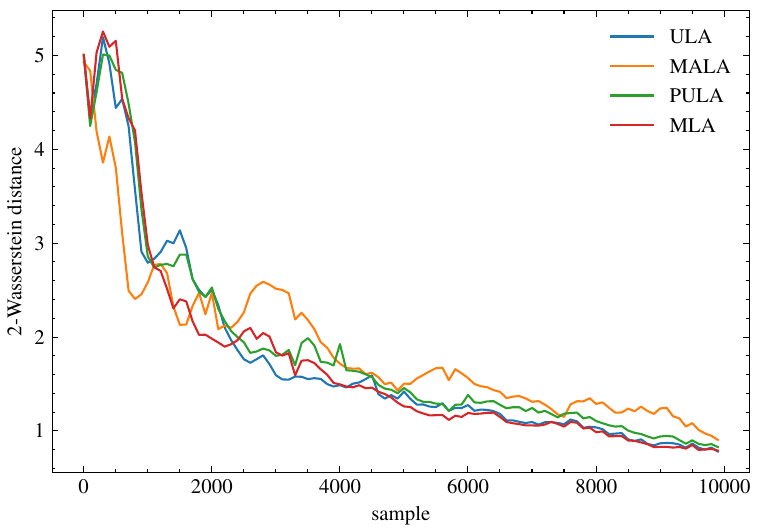}
                \caption{$K=5$}
            \end{subfigure}
            \caption{$2$-Wasserstein distances between generated samples by LMC algorithms and true samples of mixture of $K$ Laplacians with $(\gamma, \lambda)=(0.05, 0.5)$}
            \label{fig:laplacians_wass_4}
        \end{figure} 
            
        \begin{figure}[htbp]
            \begin{subfigure}[t]{.48\textwidth}
                \centering
                \includegraphics[height=.18\textheight]{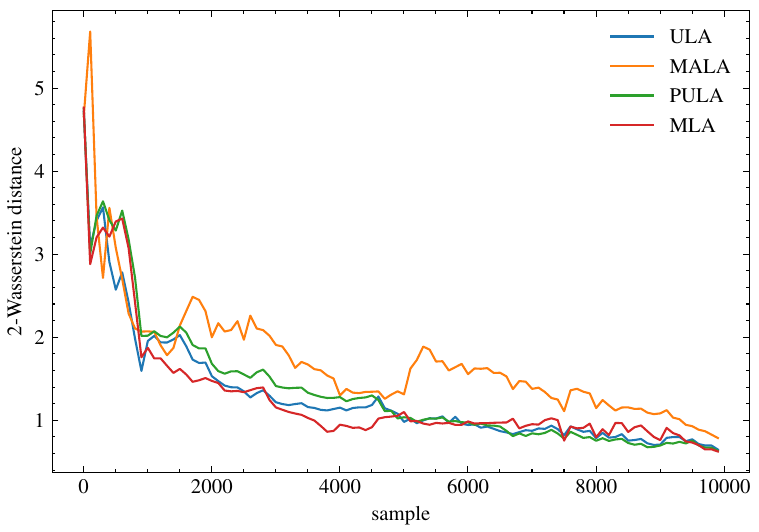}
                \caption{$K=2$}
                \vspace*{1mm}
            \end{subfigure}    
            \hfill
            \begin{subfigure}[t]{.48\textwidth}
                \centering
                \includegraphics[height=.18\textheight]{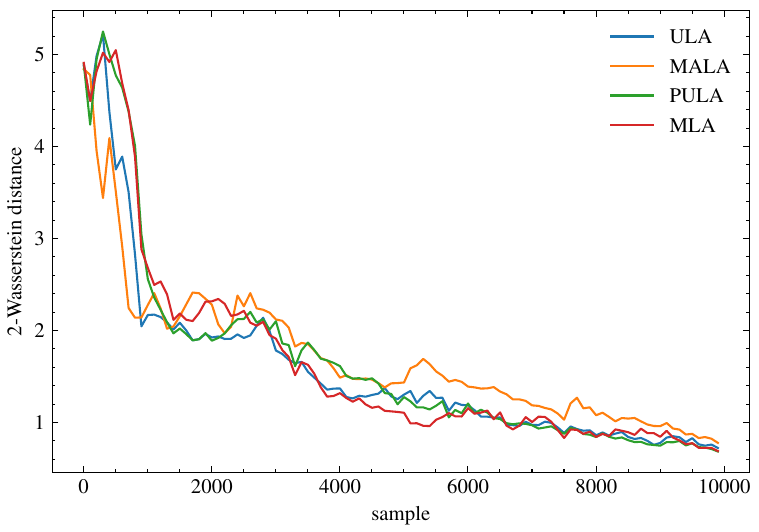}
                \caption{$K=3$}
                \vspace*{1mm}
            \end{subfigure}     
            \par\vspace{2mm}
            \begin{subfigure}[t]{.48\textwidth}
                \centering
                \includegraphics[height=.18\textheight]{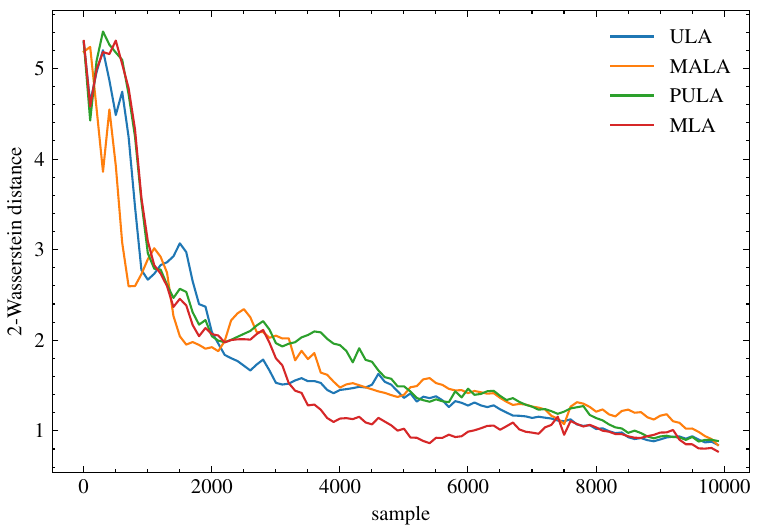}
                \caption{$K=4$}
            \end{subfigure} 
            \hfill
            \begin{subfigure}[t]{.48\textwidth}
                \centering
                \includegraphics[height=.18\textheight]{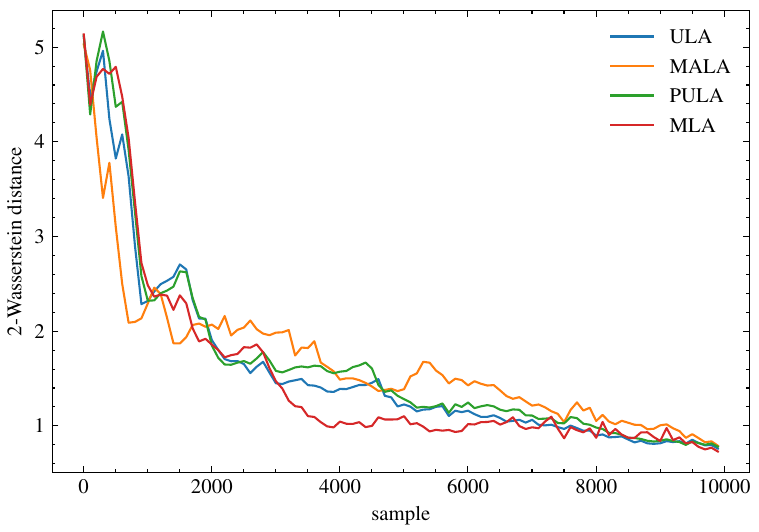}
                \caption{$K=5$}
            \end{subfigure}
            \caption{$2$-Wasserstein distances between generated samples by LMC algorithms and true samples of mixture of $K$ Laplacians with $(\gamma, \lambda)=(0.1, 0.5)$}
            \label{fig:laplacians_wass_5}
        \end{figure} 
        
        \begin{figure}[htbp]
            \centering
            \begin{subfigure}[h]{.48\textwidth}
                \centering
                \includegraphics[height=.18\textheight]{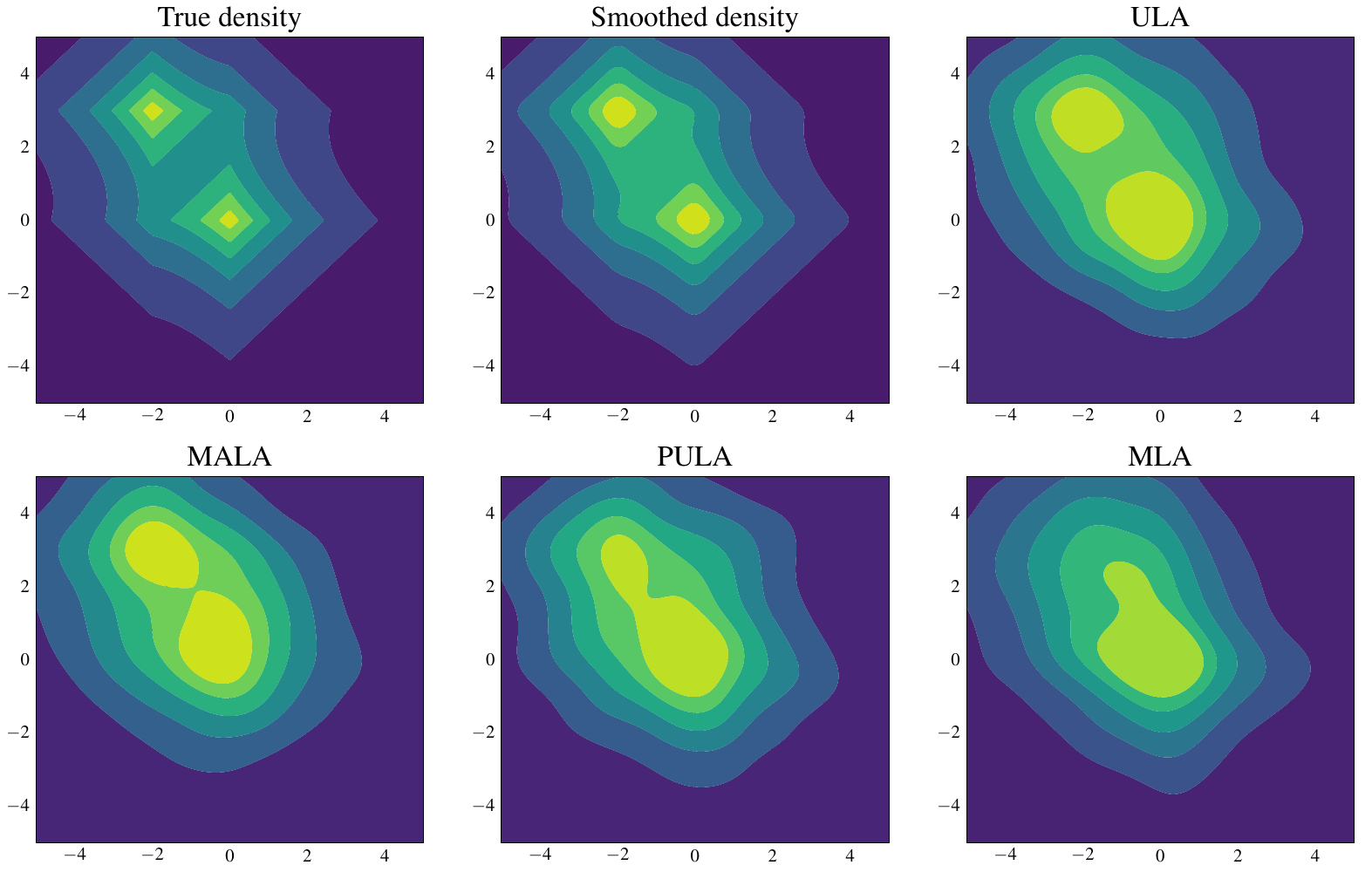}
                \caption{$K=2$}
                \vspace*{1mm}
            \end{subfigure}    
            \hfill
            \begin{subfigure}[h]{.48\textwidth}
                \centering
                \includegraphics[height=.18\textheight]{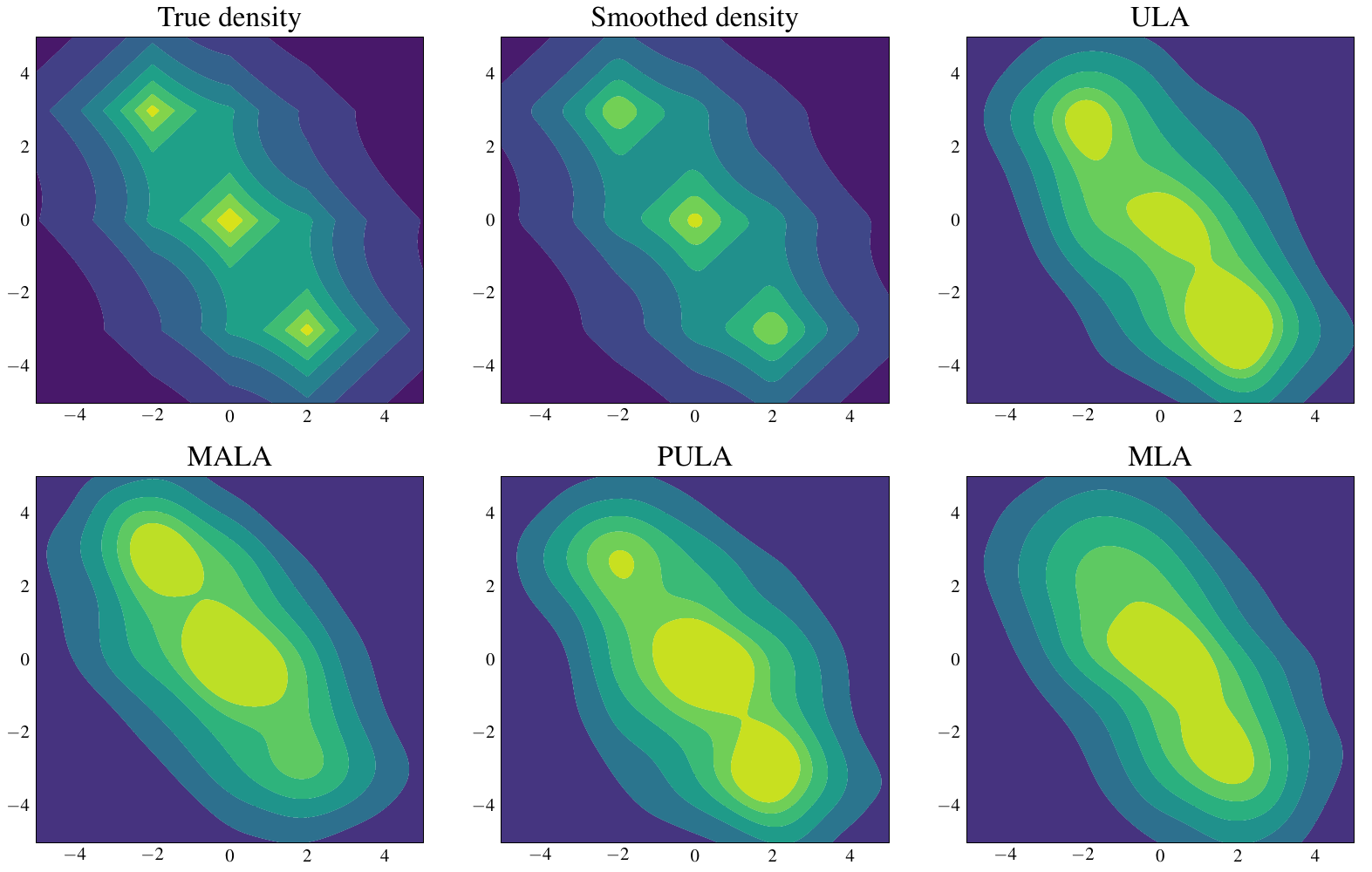}
                \caption{$K=3$}
                \vspace*{1mm}
            \end{subfigure}      
            \par\vspace{2mm}
            \begin{subfigure}[h]{.48\textwidth}
                \centering
                \includegraphics[height=.18\textheight]{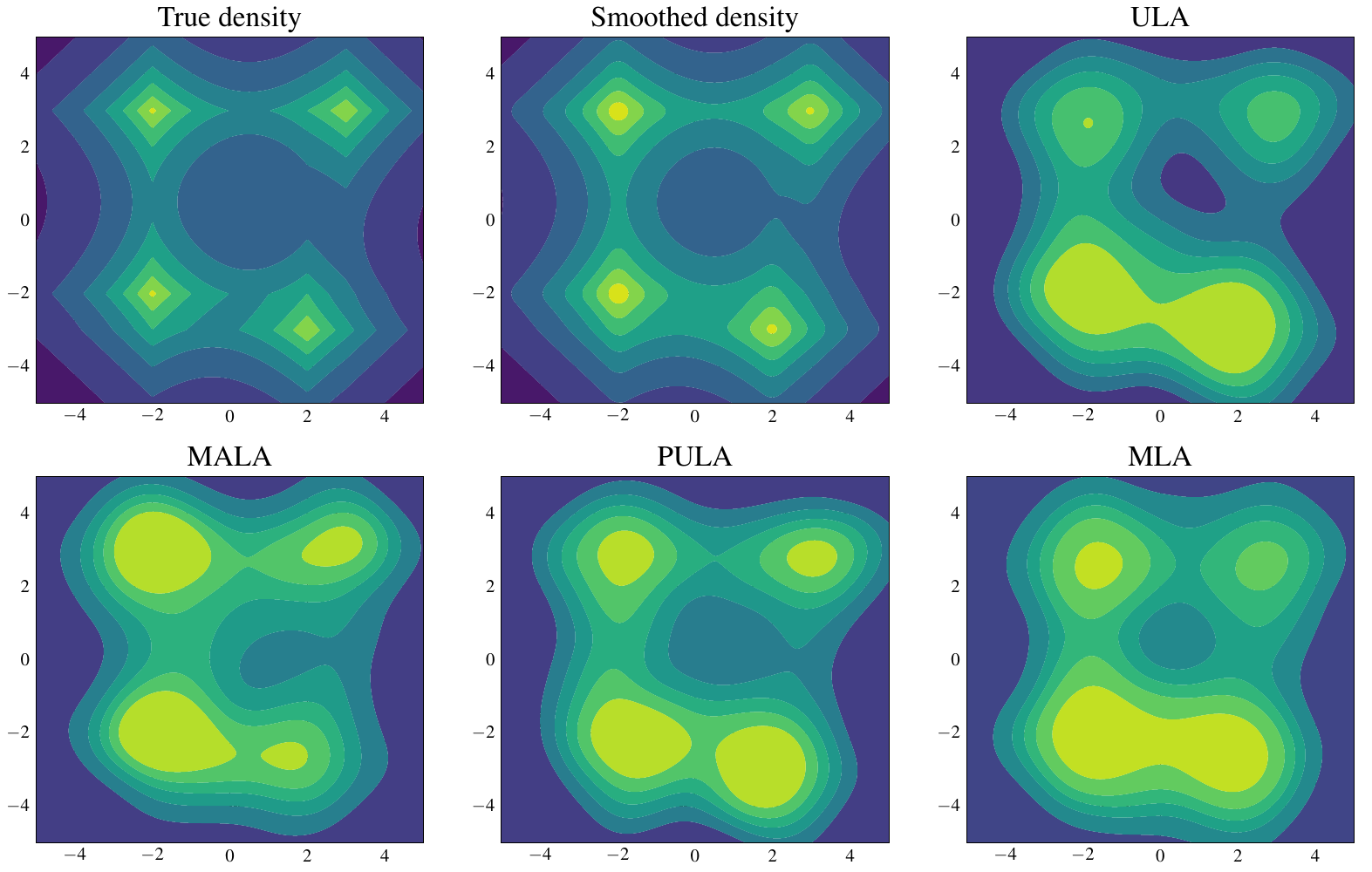}
                \caption{$K=4$}
            \end{subfigure}  
            \hfill
            \begin{subfigure}[h]{.48\textwidth}
                \centering
                \includegraphics[height=.18\textheight]{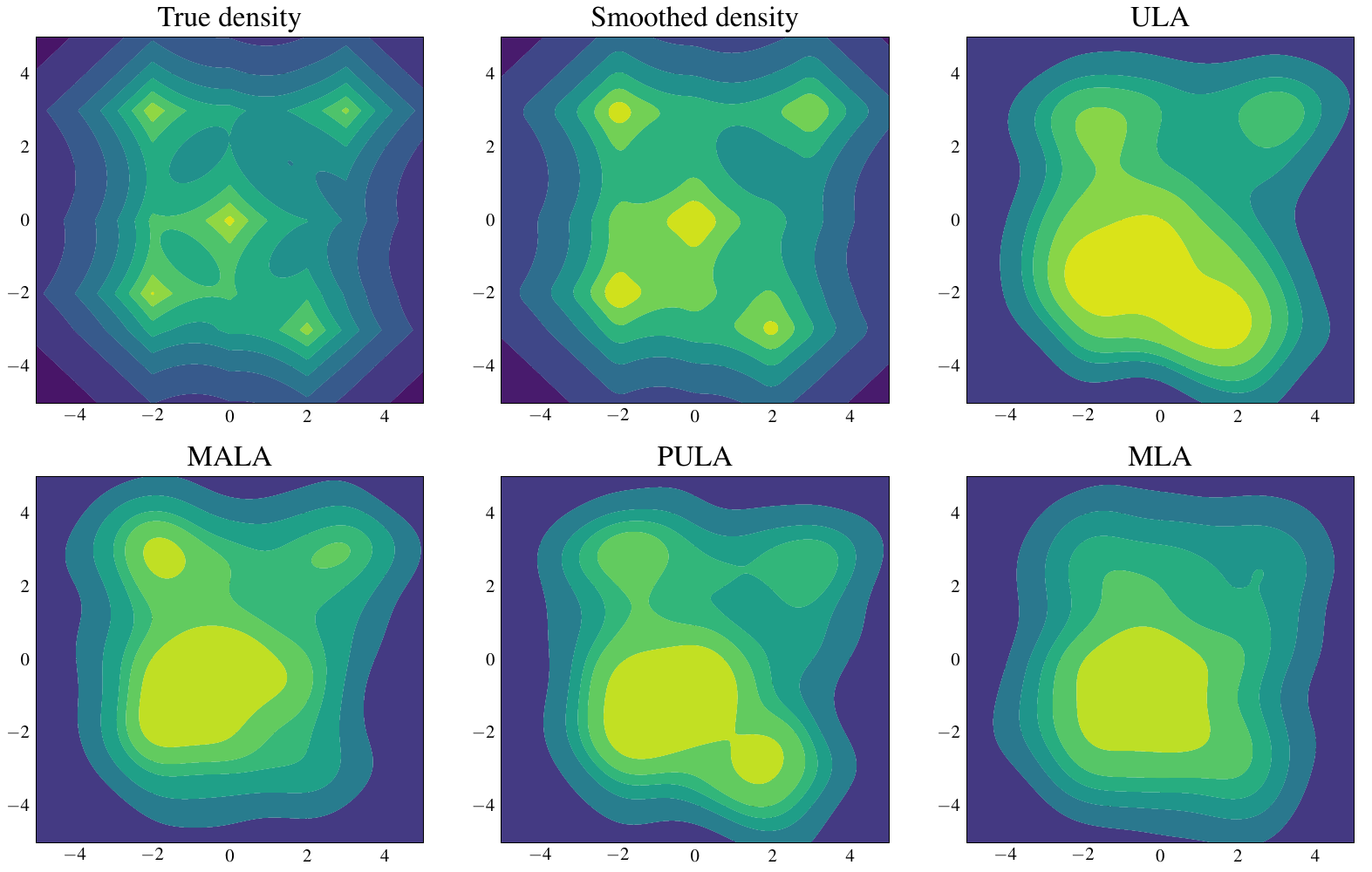}
                \caption{$K=5$}
            \end{subfigure}      
            \caption{Mixture of $K$ Laplacians with $(\gamma, \lambda)=(0.15, 0.5)$}
        \end{figure}

        \begin{figure}[htbp]
            \centering
            \begin{subfigure}[h]{.48\textwidth}
                \centering
                \includegraphics[height=.18\textheight]{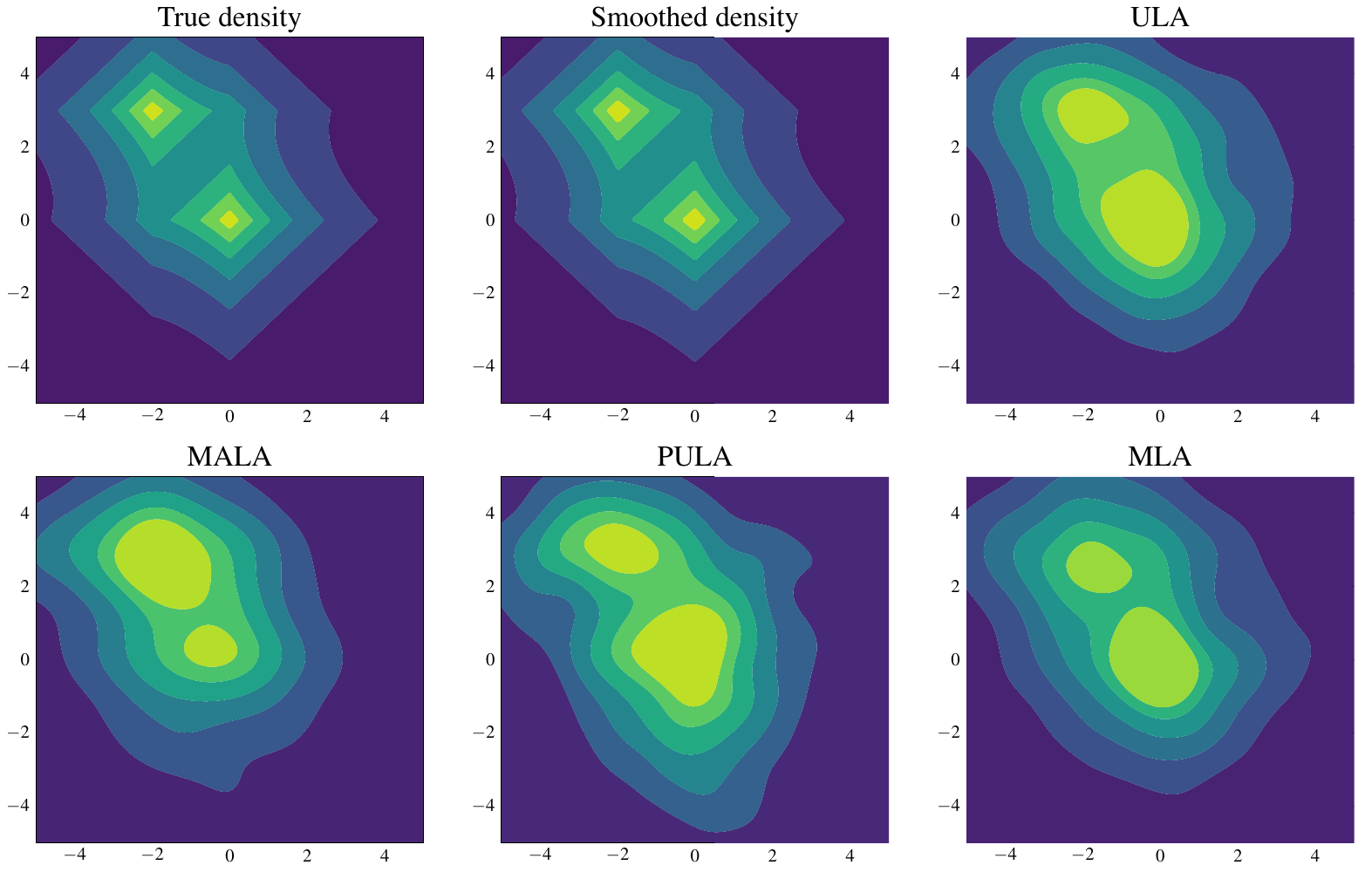}
                \caption{$K=2$}
                \vspace*{1mm}
            \end{subfigure}    
            \hfill
            \begin{subfigure}[h]{.48\textwidth}
                \centering
                \includegraphics[height=.18\textheight]{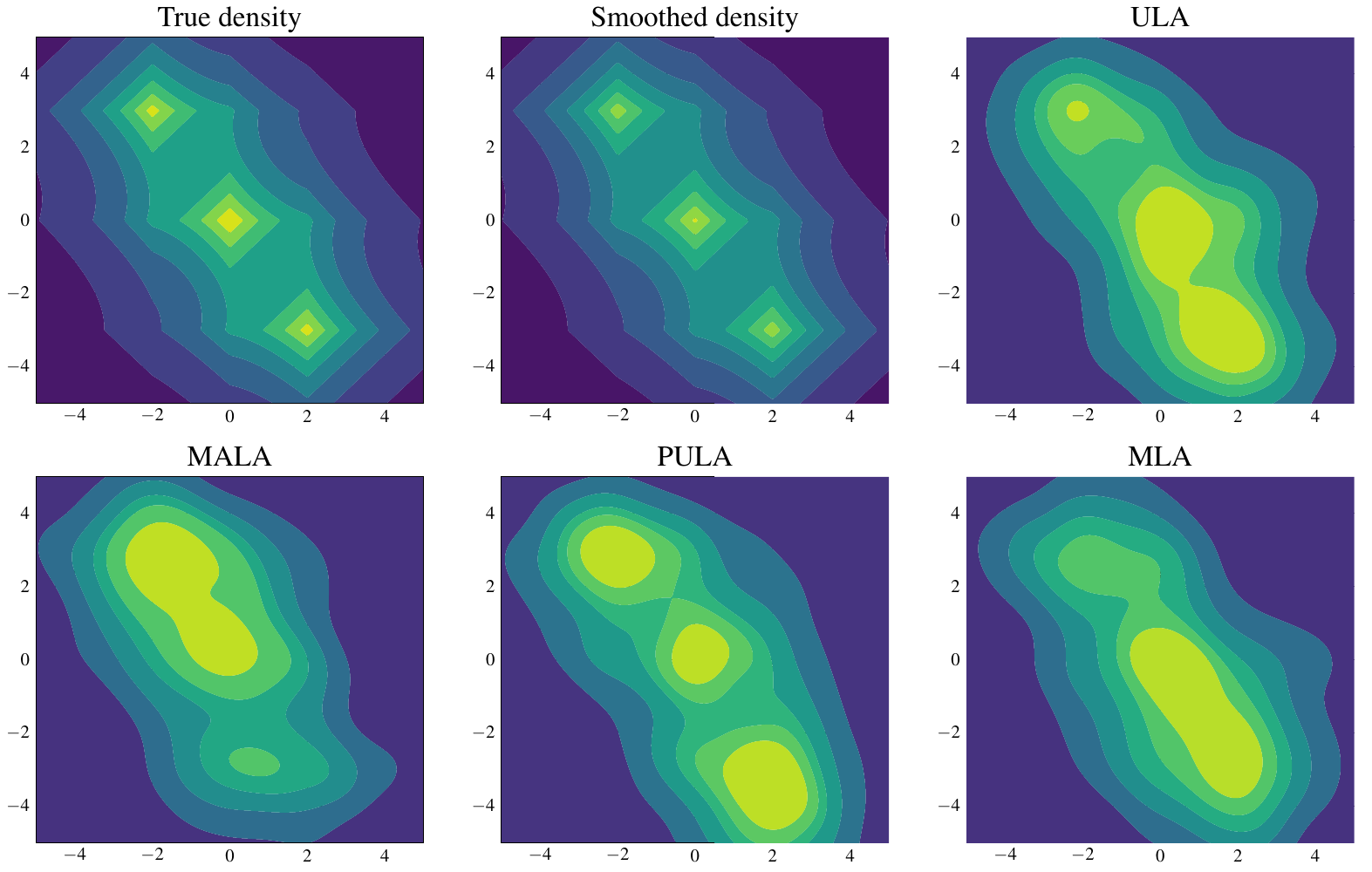}
                \caption{$K=3$}
                \vspace*{1mm}
            \end{subfigure}      
            \par\vspace{2mm}
            \begin{subfigure}[h]{.48\textwidth}
                \centering
                \includegraphics[height=.18\textheight]{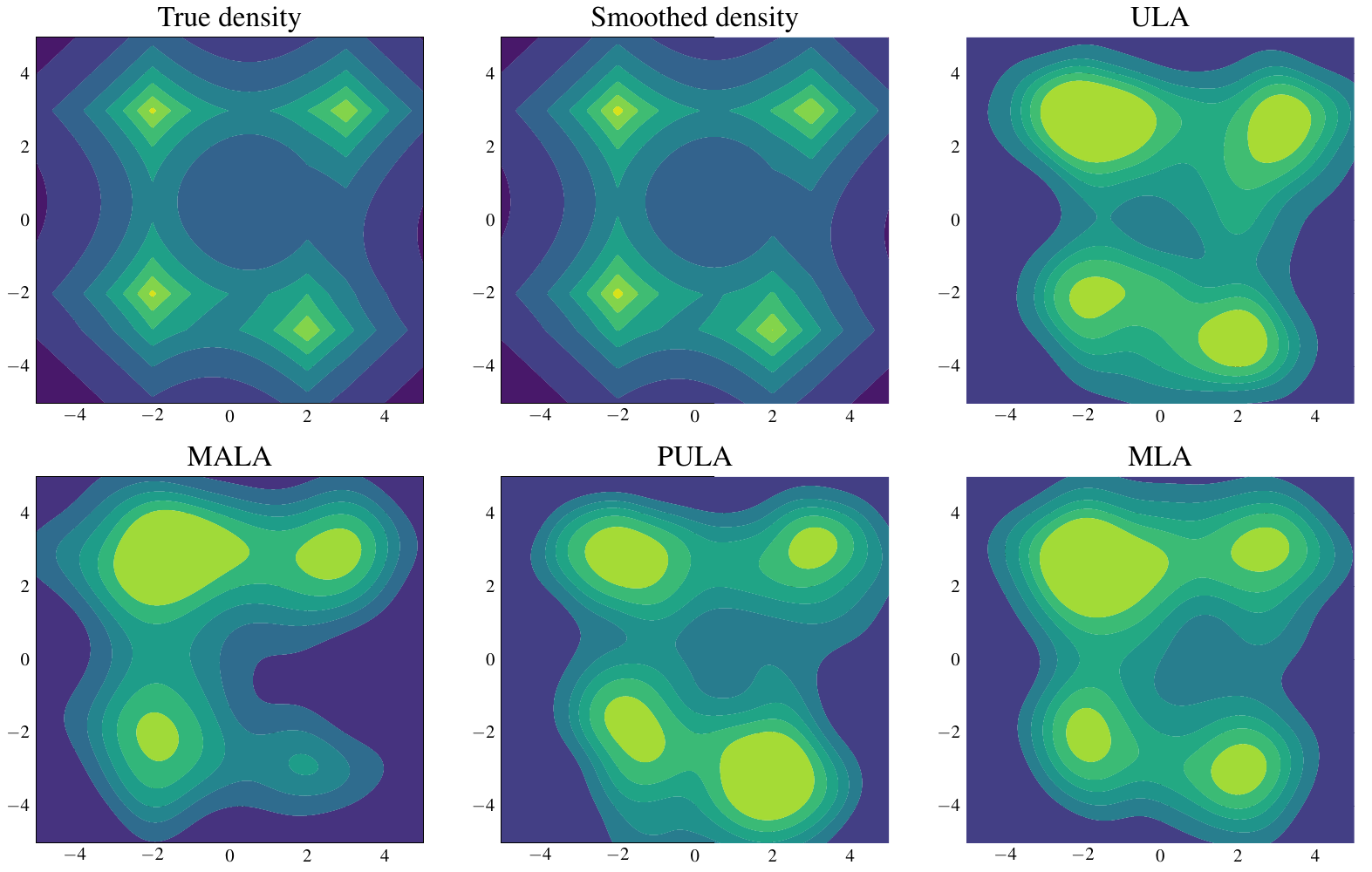}
                \caption{$K=4$}
            \end{subfigure}  
            \hfill
            \begin{subfigure}[h]{.48\textwidth}
                \centering
                \includegraphics[height=.18\textheight]{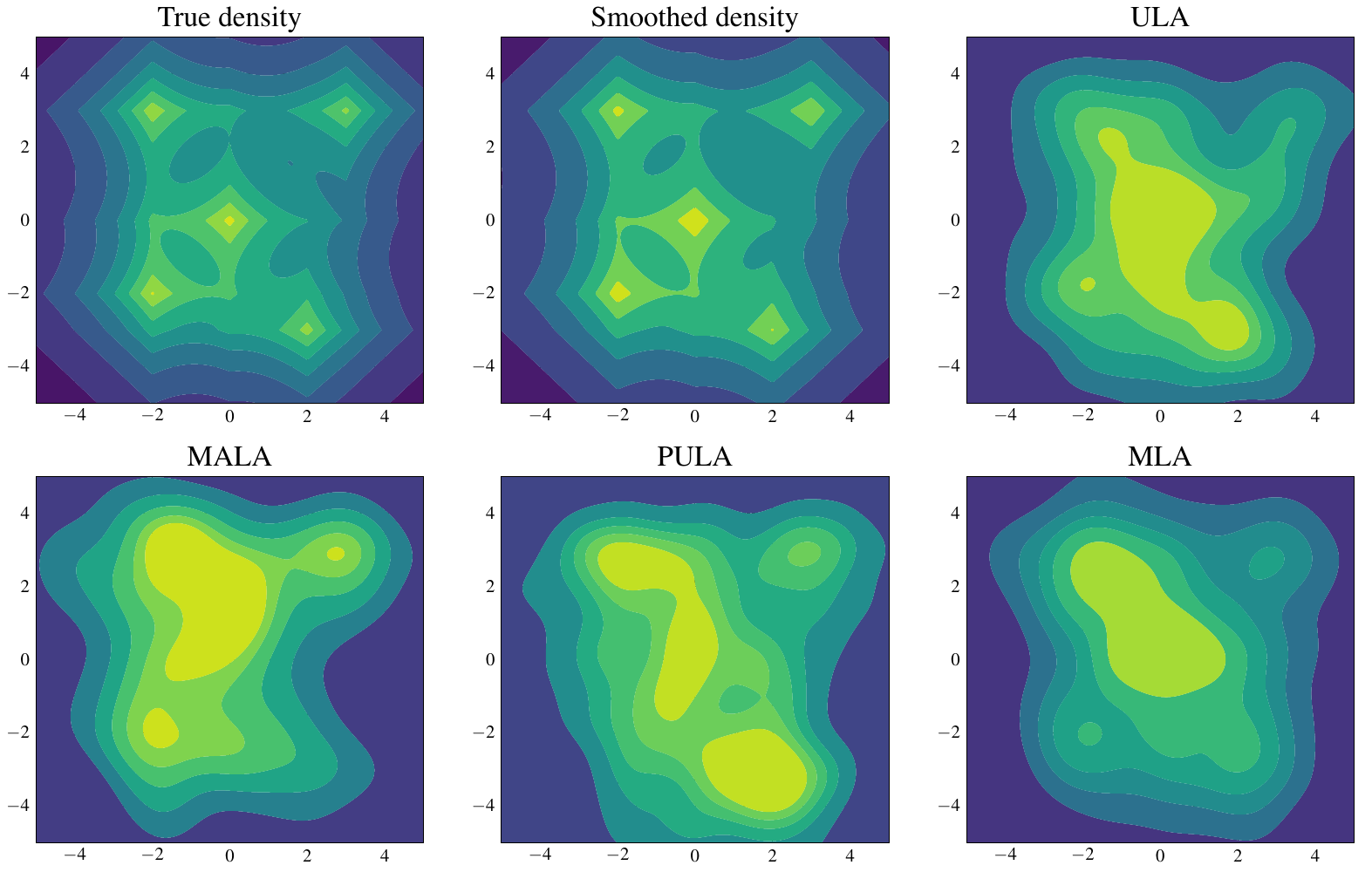}
                \caption{$K=5$}
            \end{subfigure}      
            \caption{Mixture of $K$ Laplacians with $(\gamma, \lambda)=(0.05, 0.1)$}
        \end{figure}    
         
         \begin{figure}[htbp]
             \begin{subfigure}[t]{.48\textwidth}
                 \centering
                 \includegraphics[height=.18\textheight]{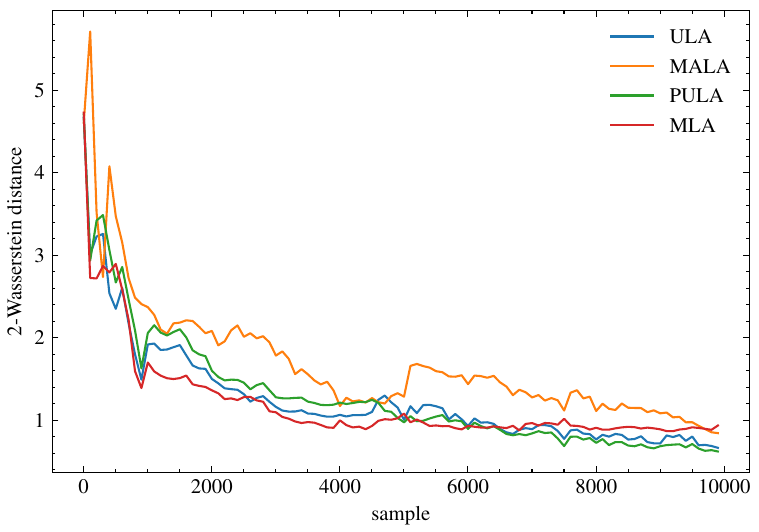}
                 \caption{$K=2$}
                 \vspace*{1mm}
             \end{subfigure}    
             \hfill
             \begin{subfigure}[t]{.48\textwidth}
                 \centering
                 \includegraphics[height=.18\textheight]{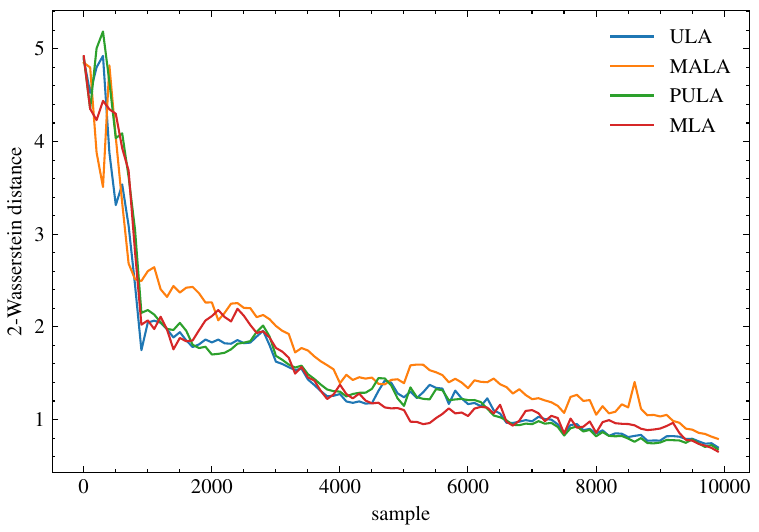}
                 \caption{$K=3$}
                 \vspace*{1mm}
             \end{subfigure}     
             \par\vspace{2mm}
             \begin{subfigure}[t]{.48\textwidth}
                 \centering
                 \includegraphics[height=.18\textheight]{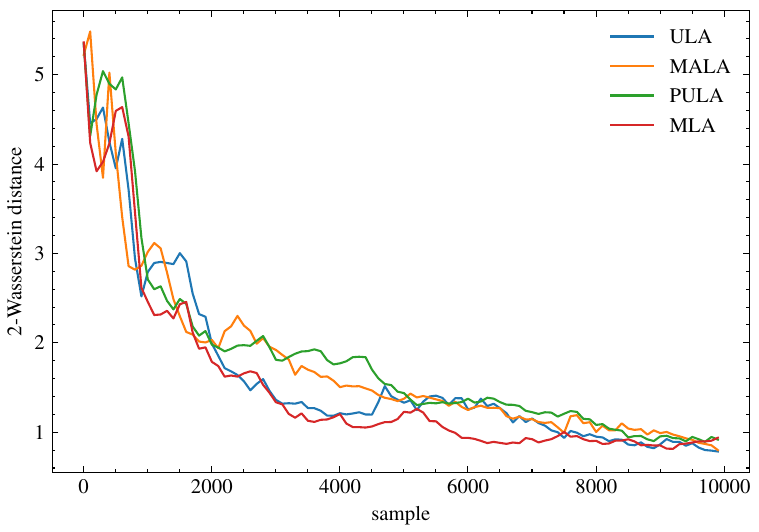}
                 \caption{$K=4$}
             \end{subfigure} 
             \hfill
             \begin{subfigure}[t]{.48\textwidth}
                 \centering
                 \includegraphics[height=.18\textheight]{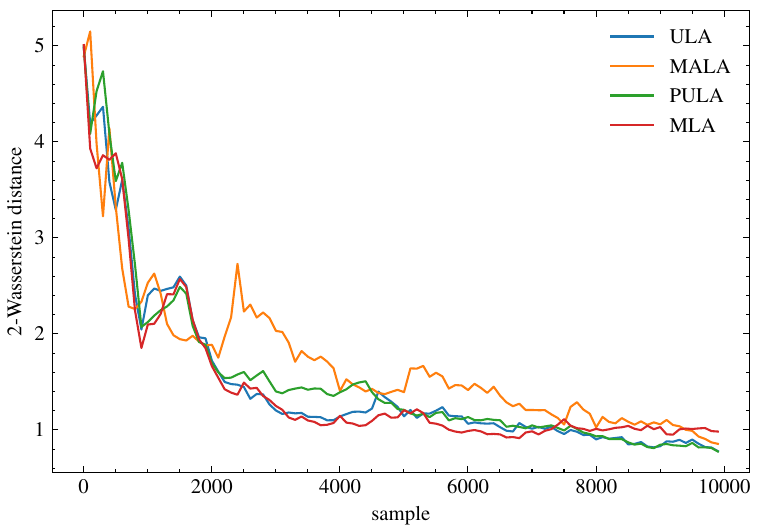}
                 \caption{$K=5$}
             \end{subfigure}
             \caption{$2$-Wasserstein distances between generated samples by LMC algorithms and true samples of mixture of $K$ Laplacians with $(\gamma, \lambda)=(0.15, 0.5)$}
             \label{fig:laplacians_wass_6}
         \end{figure} 
             
         \begin{figure}[htbp]
             \begin{subfigure}[t]{.48\textwidth}
                 \centering
                 \includegraphics[height=.18\textheight]{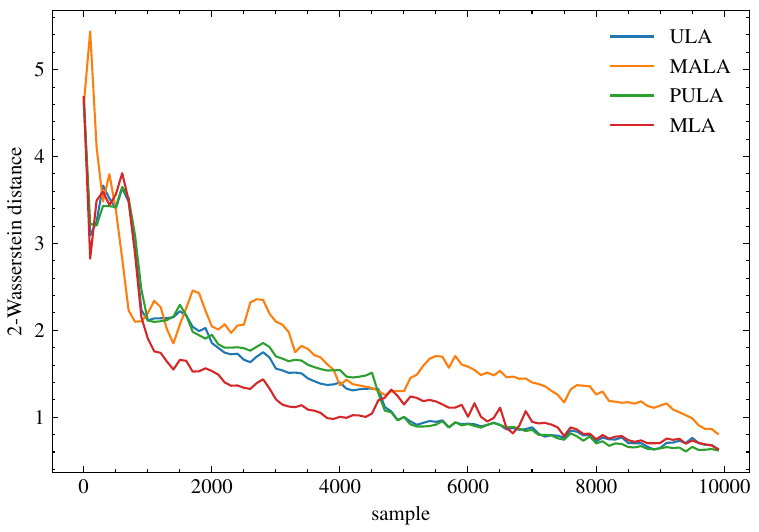}
                 \caption{$K=2$}
                 \vspace*{1mm}
             \end{subfigure}    
             \hfill
             \begin{subfigure}[t]{.48\textwidth}
                 \centering
                 \includegraphics[height=.18\textheight]{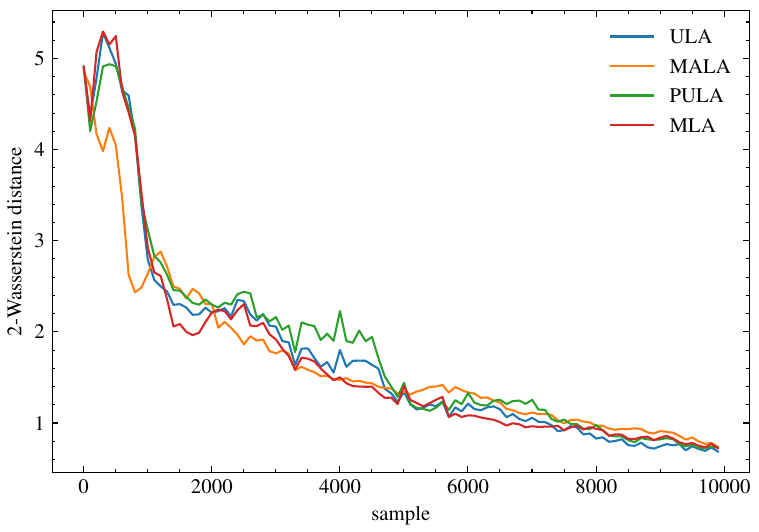}
                 \caption{$K=3$}
                 \vspace*{1mm}
             \end{subfigure}     
             \par\vspace{2mm}
             \begin{subfigure}[t]{.48\textwidth}
                 \centering
                 \includegraphics[height=.18\textheight]{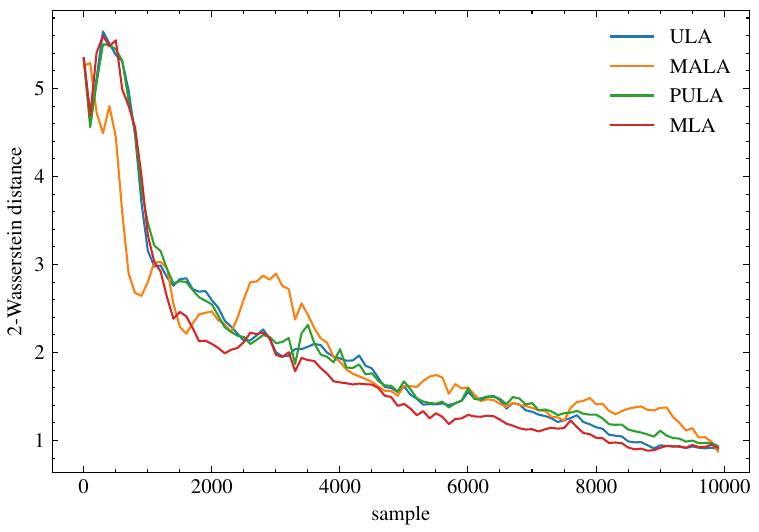}
                 \caption{$K=4$}
             \end{subfigure} 
             \hfill
             \begin{subfigure}[t]{.48\textwidth}
                 \centering
                 \includegraphics[height=.18\textheight]{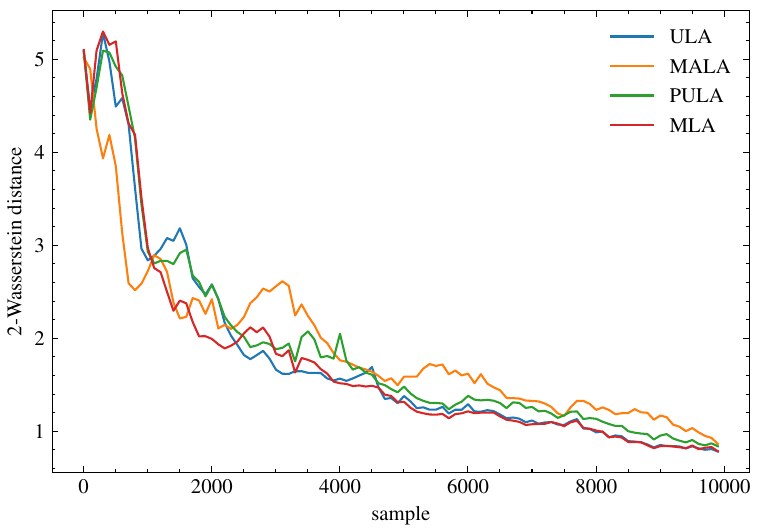}
                 \caption{$K=5$}
             \end{subfigure}
             \caption{$2$-Wasserstein distances between generated samples by LMC algorithms and true samples of mixture of $K$ Laplacians with $(\gamma, \lambda)=(0.05, 0.1)$}
             \label{fig:laplacians_wass_7}
         \end{figure} 
    
        \begin{figure}[htbp]
            \centering
            \begin{subfigure}[h]{.48\textwidth}
                \centering
                \includegraphics[height=.18\textheight]{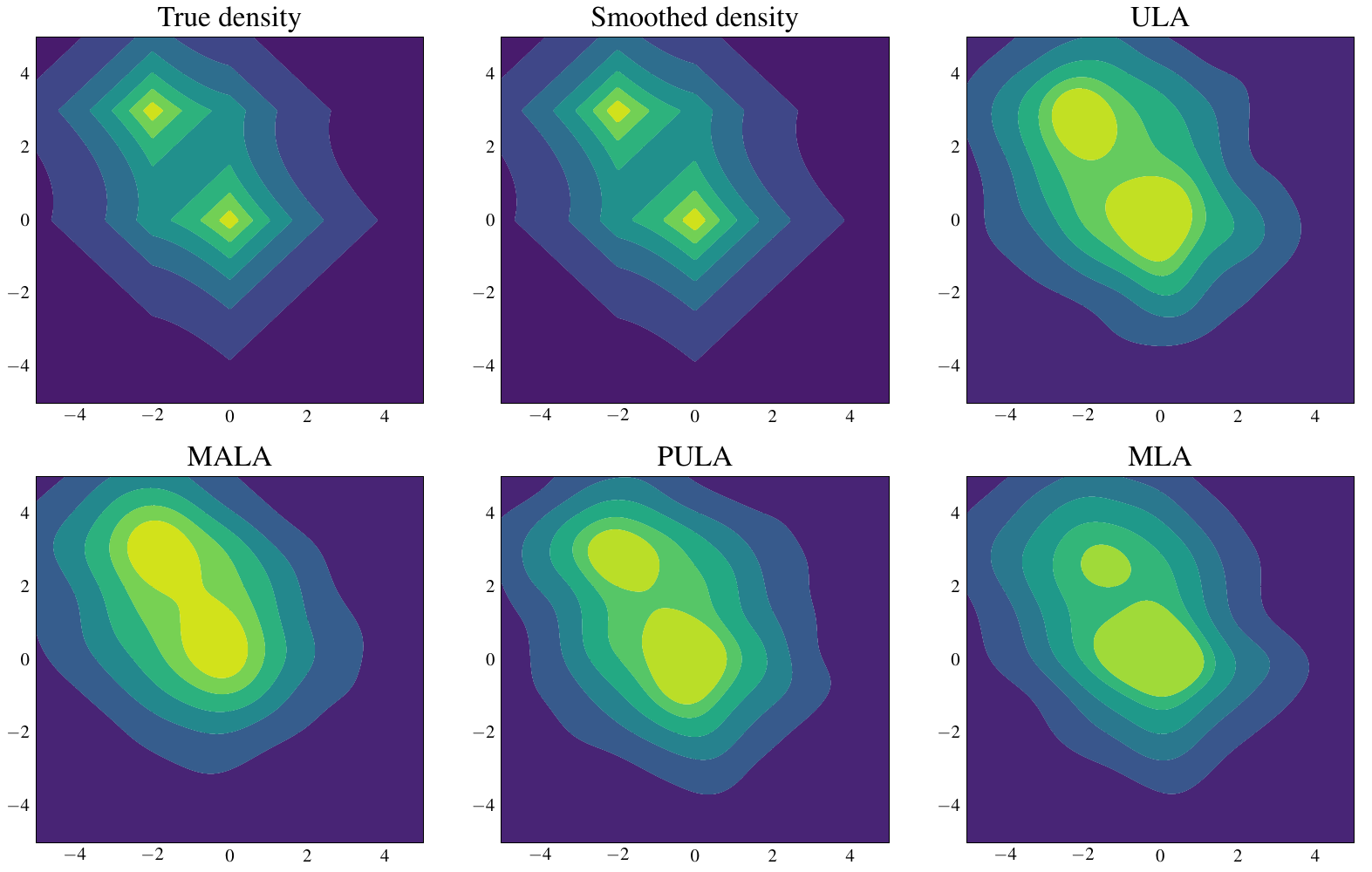}
                \caption{$K=2$}
            \end{subfigure}    
            \hfill
            \begin{subfigure}[h]{.48\textwidth}
                \centering
                \includegraphics[height=.18\textheight]{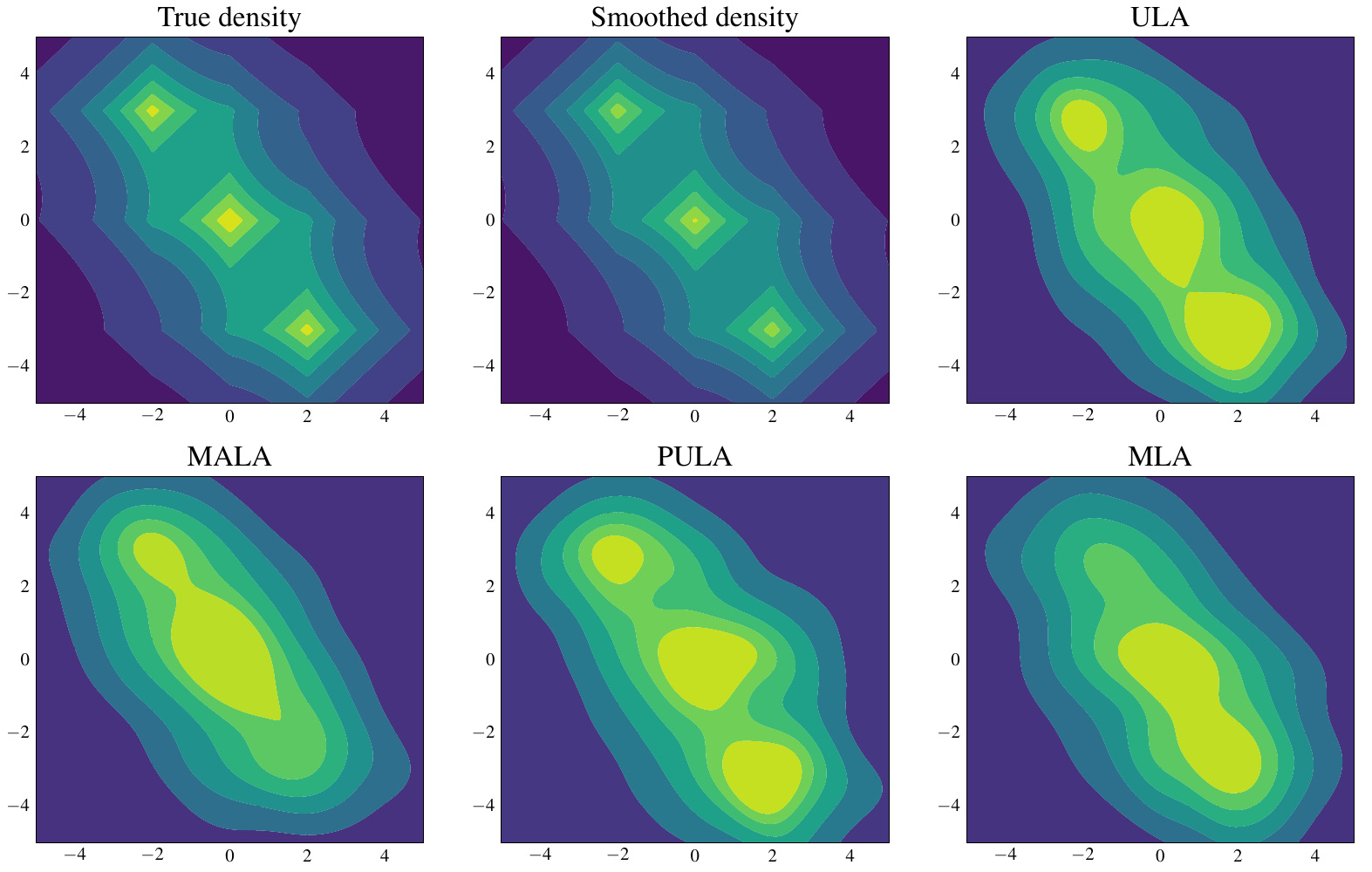}
                \caption{$K=3$}
            \end{subfigure}      
            \par\vspace{2mm}
            \begin{subfigure}[h]{.48\textwidth}
                \centering
                \includegraphics[height=.18\textheight]{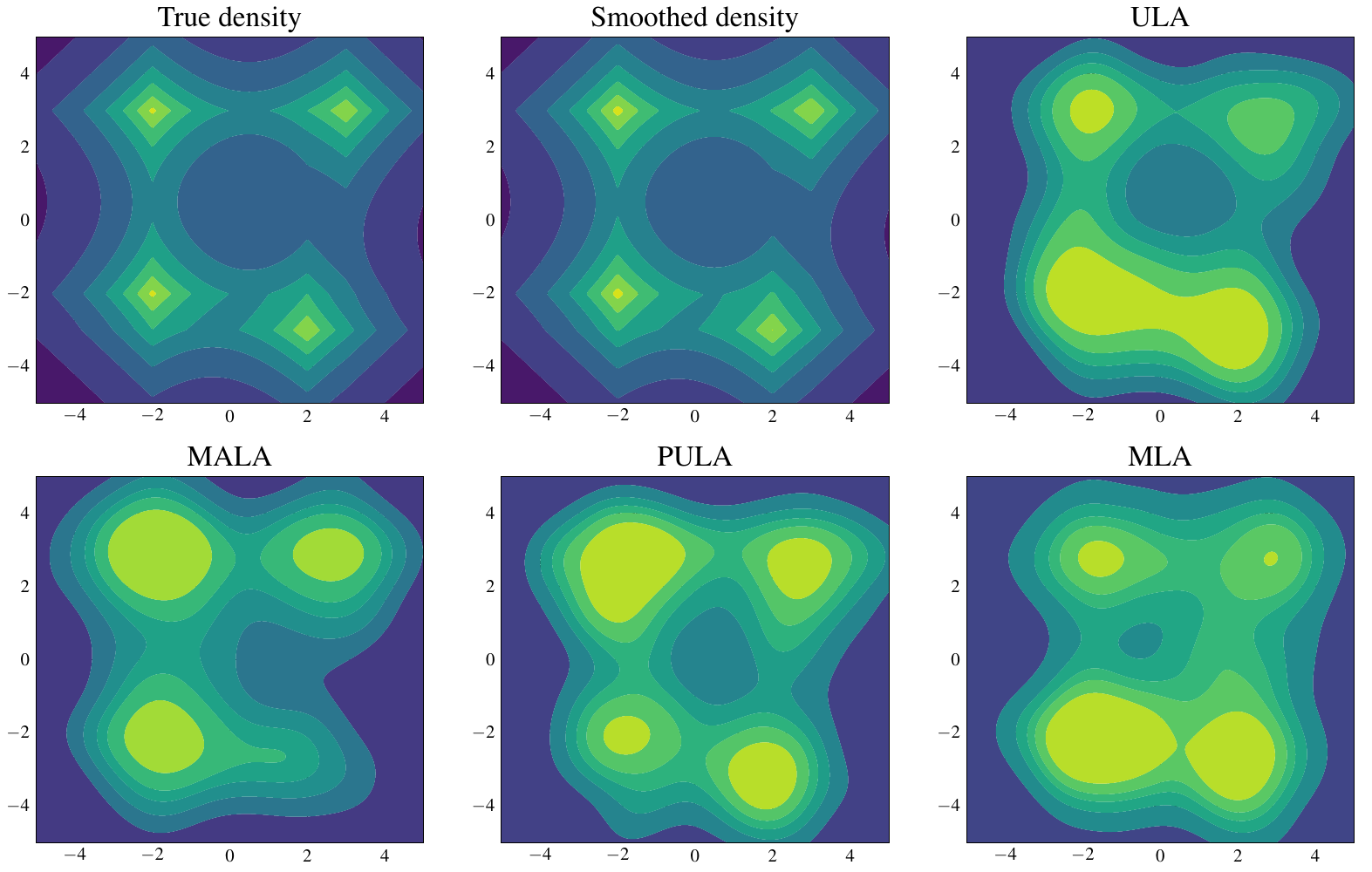}
                \caption{$K=4$}
            \end{subfigure}  
            \hfill
            \begin{subfigure}[h]{.48\textwidth}
                \centering
                \includegraphics[height=.18\textheight]{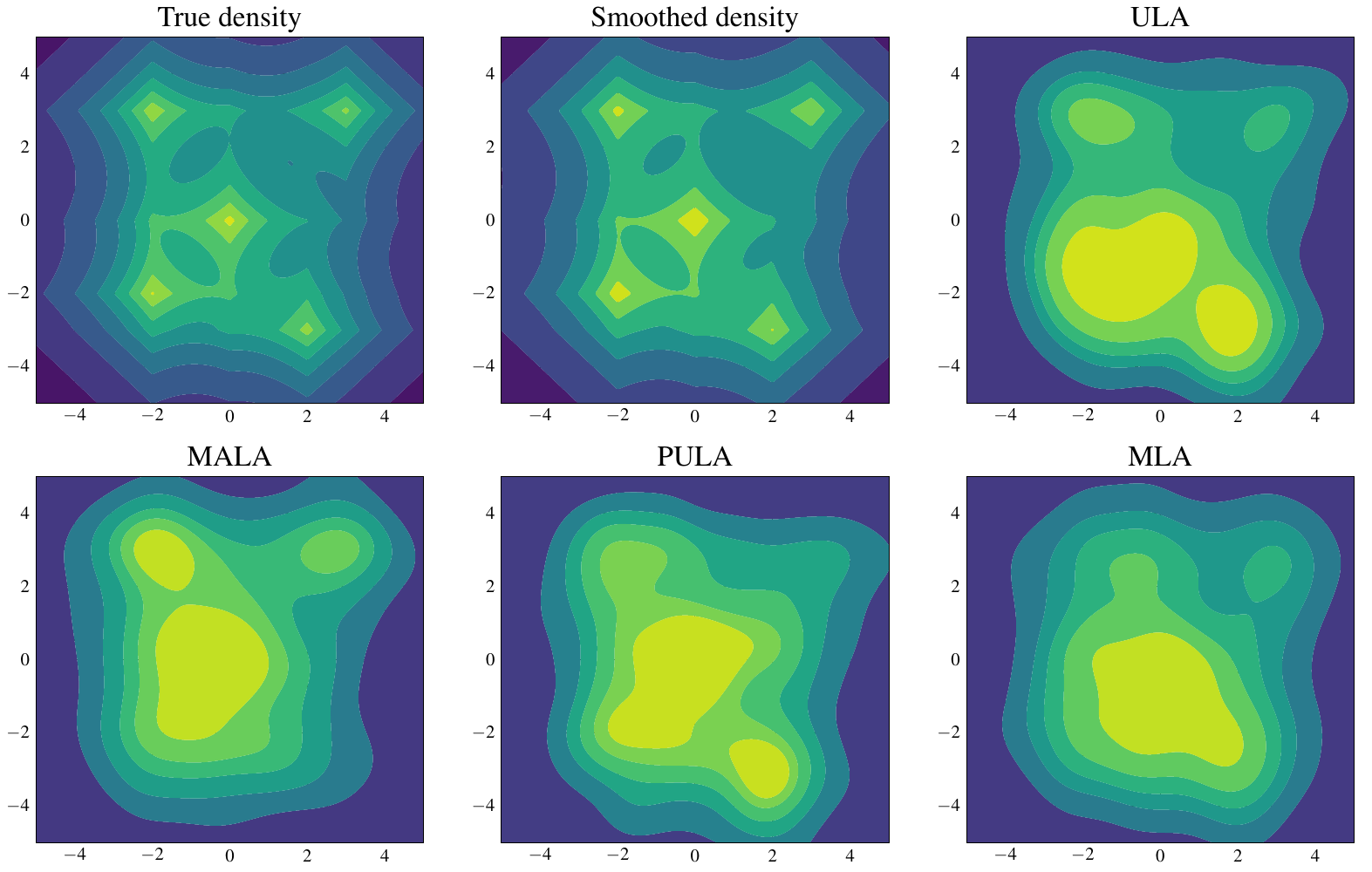}
                \caption{$K=5$}
            \end{subfigure}      
            \caption{Mixture of $K$ Laplacians with $(\gamma, \lambda)=(0.1, 0.1)$}
        \end{figure}

        \begin{figure}[htbp]
            \centering
            \begin{subfigure}[h]{.48\textwidth}
                \centering
                \includegraphics[height=.18\textheight]{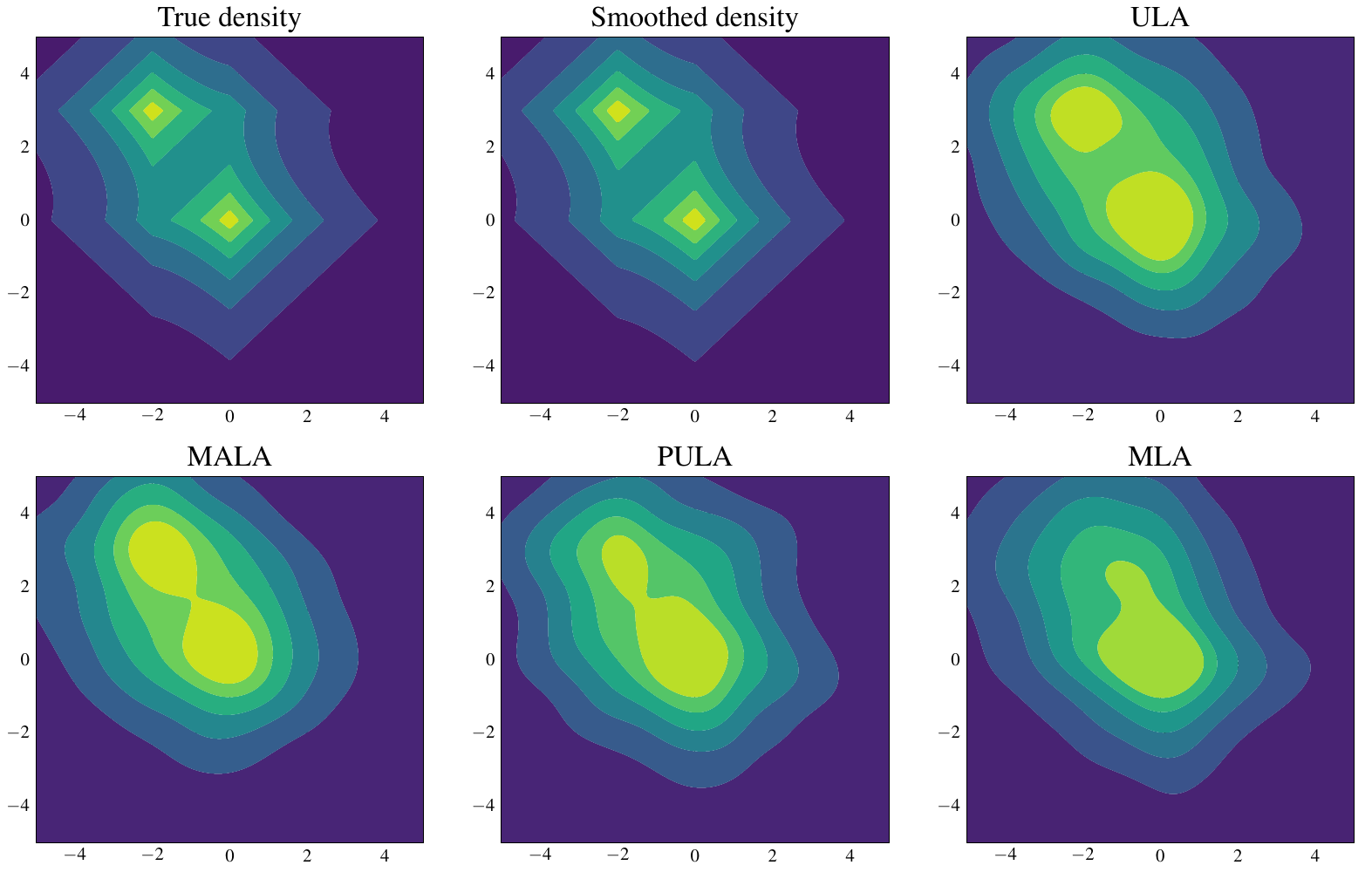}
                \caption{$K=2$}
                \vspace*{1mm}
            \end{subfigure}    
            \hfill
            \begin{subfigure}[h]{.48\textwidth}
                \centering
                \includegraphics[height=.18\textheight]{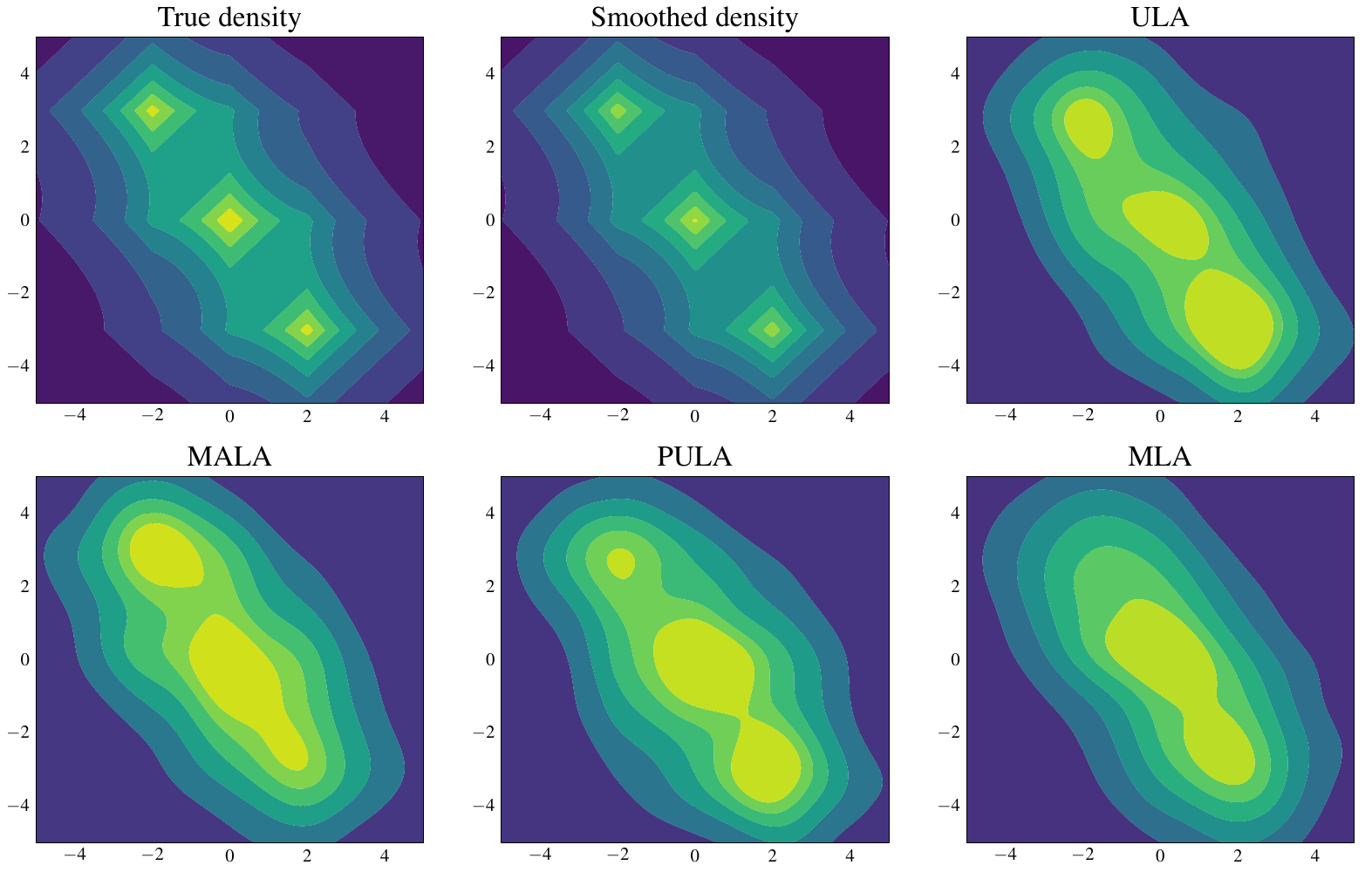}
                \caption{$K=3$}
                \vspace*{1mm}
            \end{subfigure}      
            \par\vspace{2mm}
            \begin{subfigure}[h]{.48\textwidth}
                \centering
                \includegraphics[height=.18\textheight]{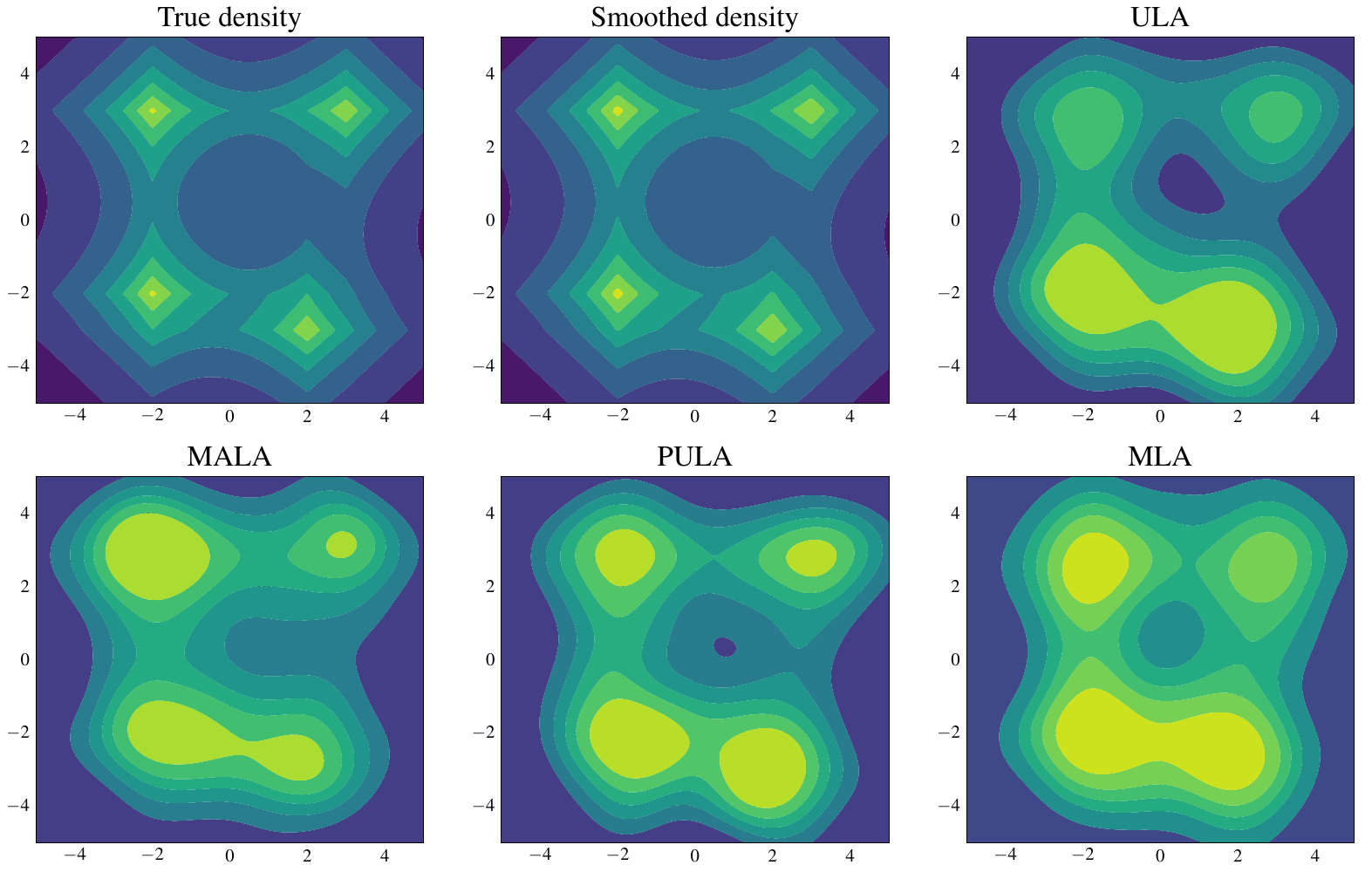}
                \caption{$K=4$}
            \end{subfigure}  
            \hfill
            \begin{subfigure}[h]{.48\textwidth}
                \centering
                \includegraphics[height=.18\textheight]{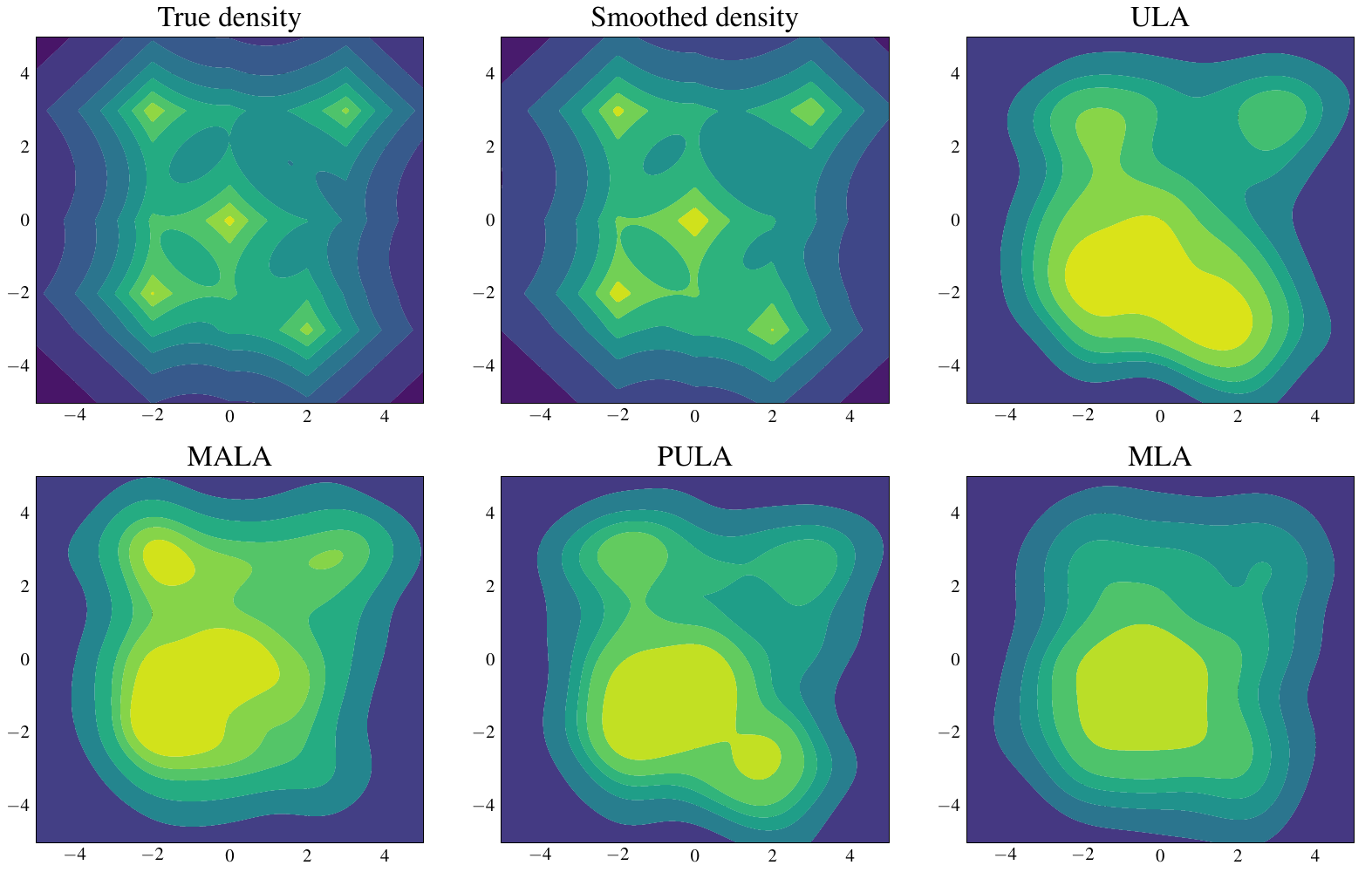}
                \caption{$K=5$}
            \end{subfigure}      
            \caption{Mixture of $K$ Laplacians with $(\gamma, \lambda)=(0.15, 0.1)$}
        \end{figure} 
        
        \begin{figure}[htbp]
             \begin{subfigure}[t]{.48\textwidth}
                 \centering
                 \includegraphics[height=.18\textheight]{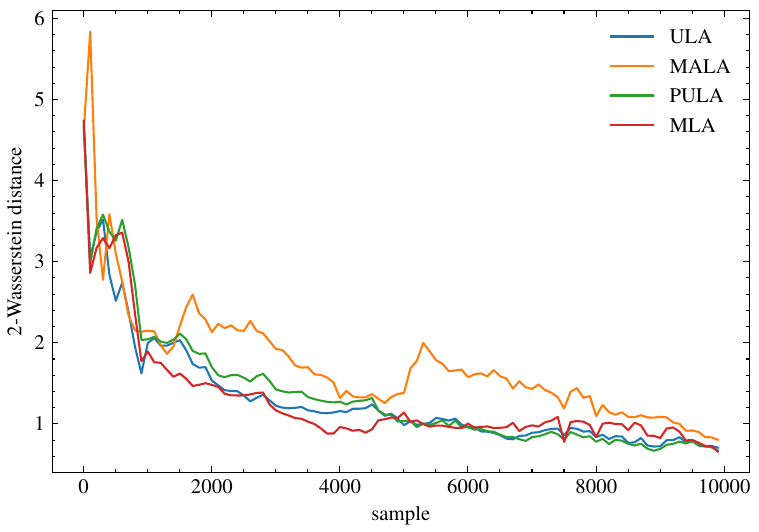}
                 \caption{$K=2$}
                 \vspace*{1mm}
             \end{subfigure}    
             \hfill
             \begin{subfigure}[t]{.48\textwidth}
                 \centering
                 \includegraphics[height=.18\textheight]{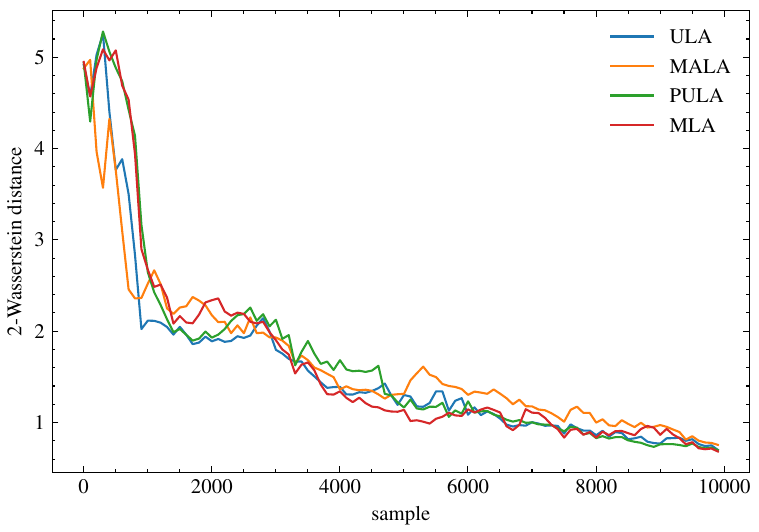}
                 \caption{$K=3$}
                 \vspace*{1mm}
             \end{subfigure}     
             \par\vspace{2mm}
             \begin{subfigure}[t]{.48\textwidth}
                 \centering
                 \includegraphics[height=.18\textheight]{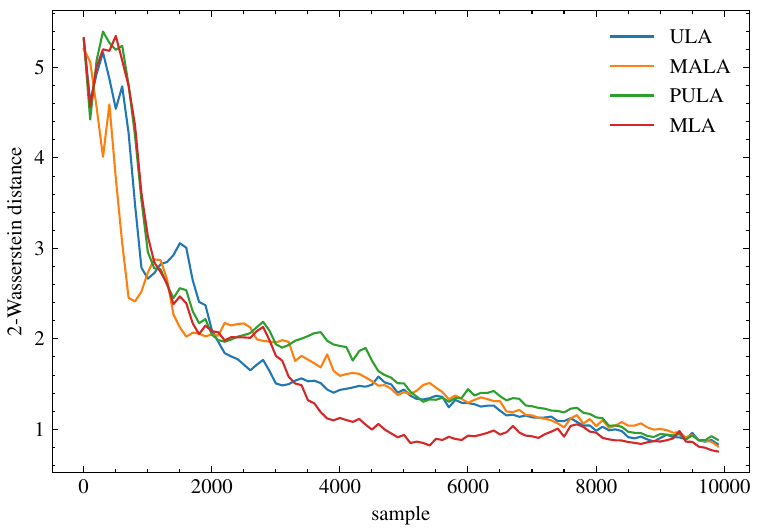}
                 \caption{$K=4$}
             \end{subfigure} 
             \hfill
             \begin{subfigure}[t]{.48\textwidth}
                 \centering
                 \includegraphics[height=.18\textheight]{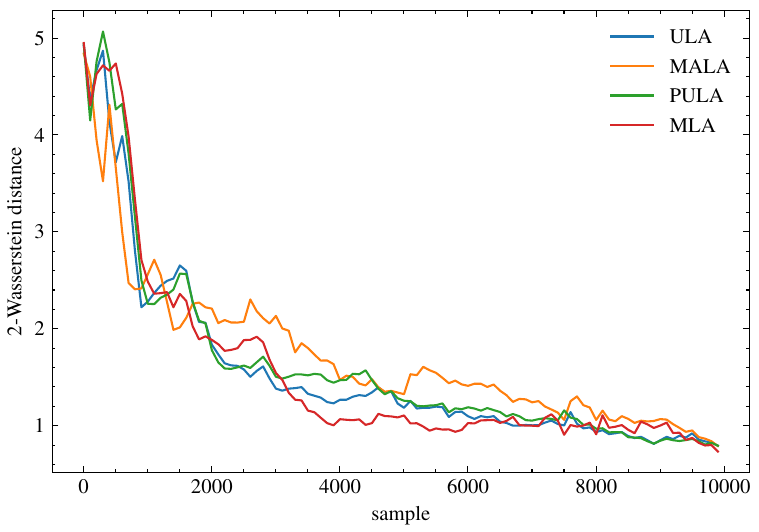}
                 \caption{$K=5$}
             \end{subfigure}
             \caption{$2$-Wasserstein distances between generated samples by LMC algorithms and true samples of mixture of $K$ Laplacians with $(\gamma, \lambda)=(0.1, 0.1)$}
             \label{fig:laplacians_wass_8}
         \end{figure} 
             
         \begin{figure}[htbp]
             \begin{subfigure}[t]{.48\textwidth}
                 \centering
                 \includegraphics[height=.18\textheight]{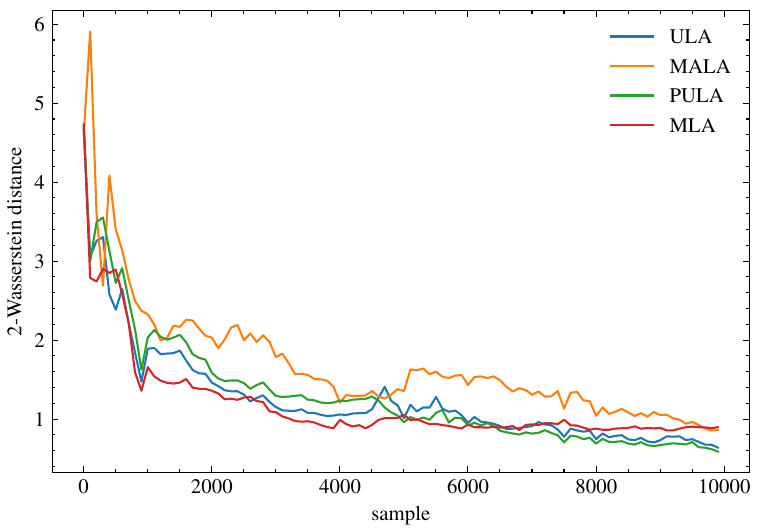}
                 \caption{$K=2$}
                 \vspace*{1mm}
             \end{subfigure}    
             \hfill
             \begin{subfigure}[t]{.48\textwidth}
                 \centering
                 \includegraphics[height=.18\textheight]{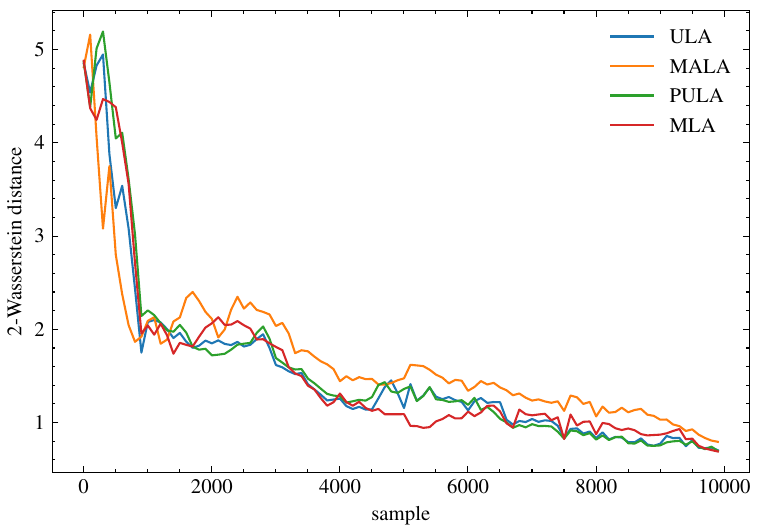}
                 \caption{$K=3$}
                 \vspace*{1mm}
             \end{subfigure}     
             \par\vspace{2mm}
             \begin{subfigure}[t]{.48\textwidth}
                 \centering
                 \includegraphics[height=.18\textheight]{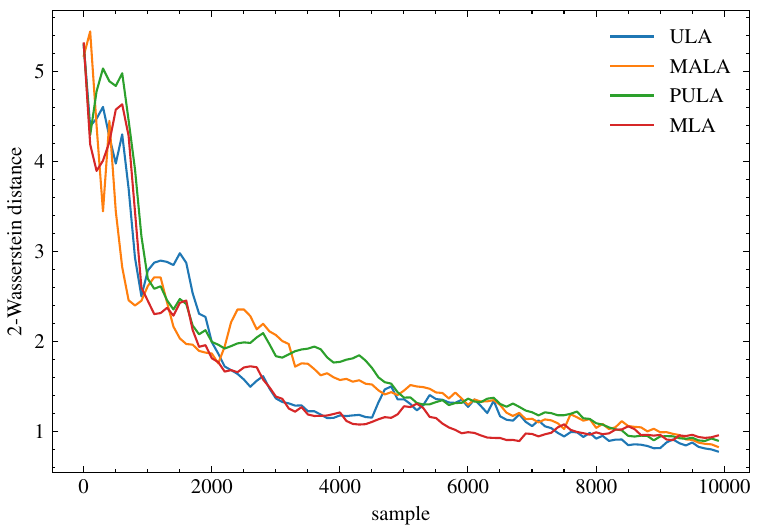}
                 \caption{$K=4$}
             \end{subfigure} 
             \hfill
             \begin{subfigure}[t]{.48\textwidth}
                 \centering
                 \includegraphics[height=.18\textheight]{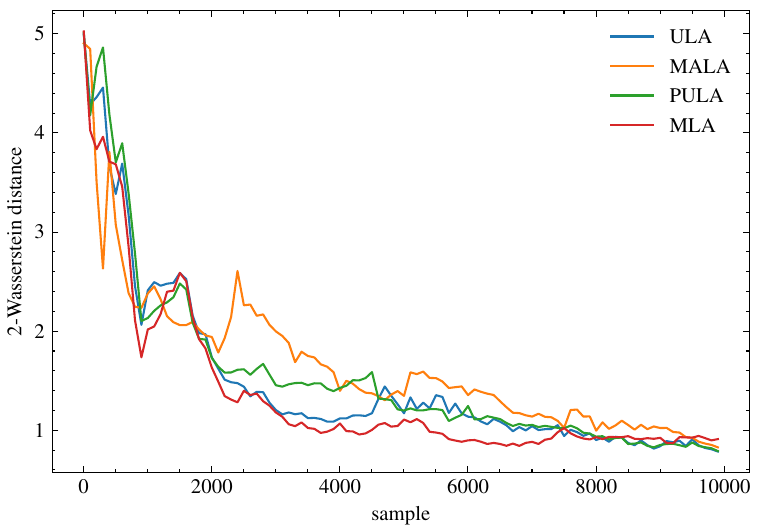}
                 \caption{$K=5$}
             \end{subfigure}
             \caption{$2$-Wasserstein distances between generated samples by LMC algorithms and true samples of mixture of $K$ Laplacians with $(\gamma, \lambda)=(0.15, 0.1)$}
             \label{fig:laplacians_wass_9}
         \end{figure}

        \begin{figure}[htbp]
            \centering
            \begin{subfigure}[h]{\textwidth}
                \centering
                \includegraphics[height=.18\textheight]{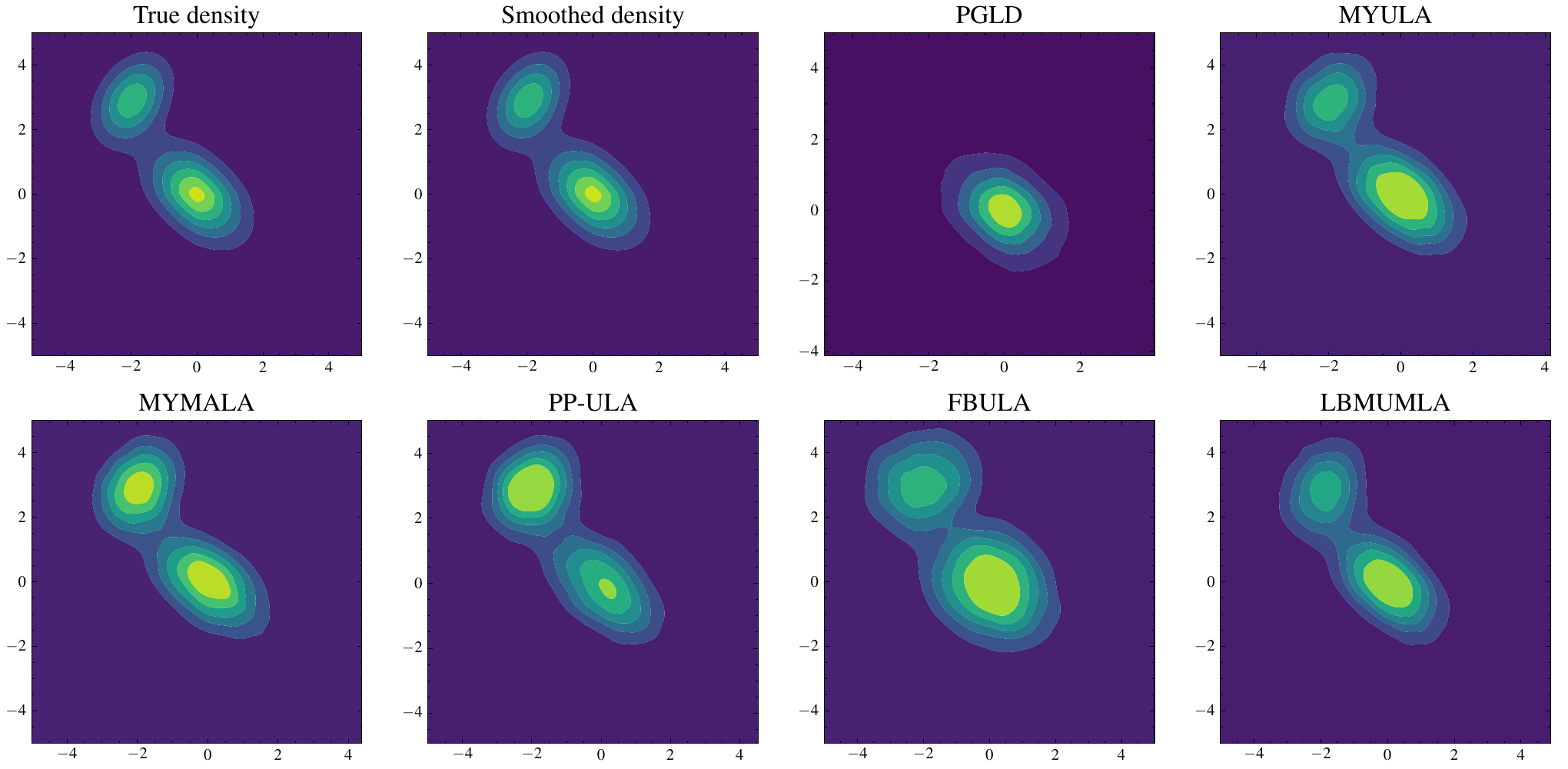}
                \caption{$K=2$}
                \vspace*{2mm}
            \end{subfigure}
            \par\vspace{2mm}
            \begin{subfigure}[h]{\textwidth}
                \centering
                \includegraphics[height=.18\textheight]{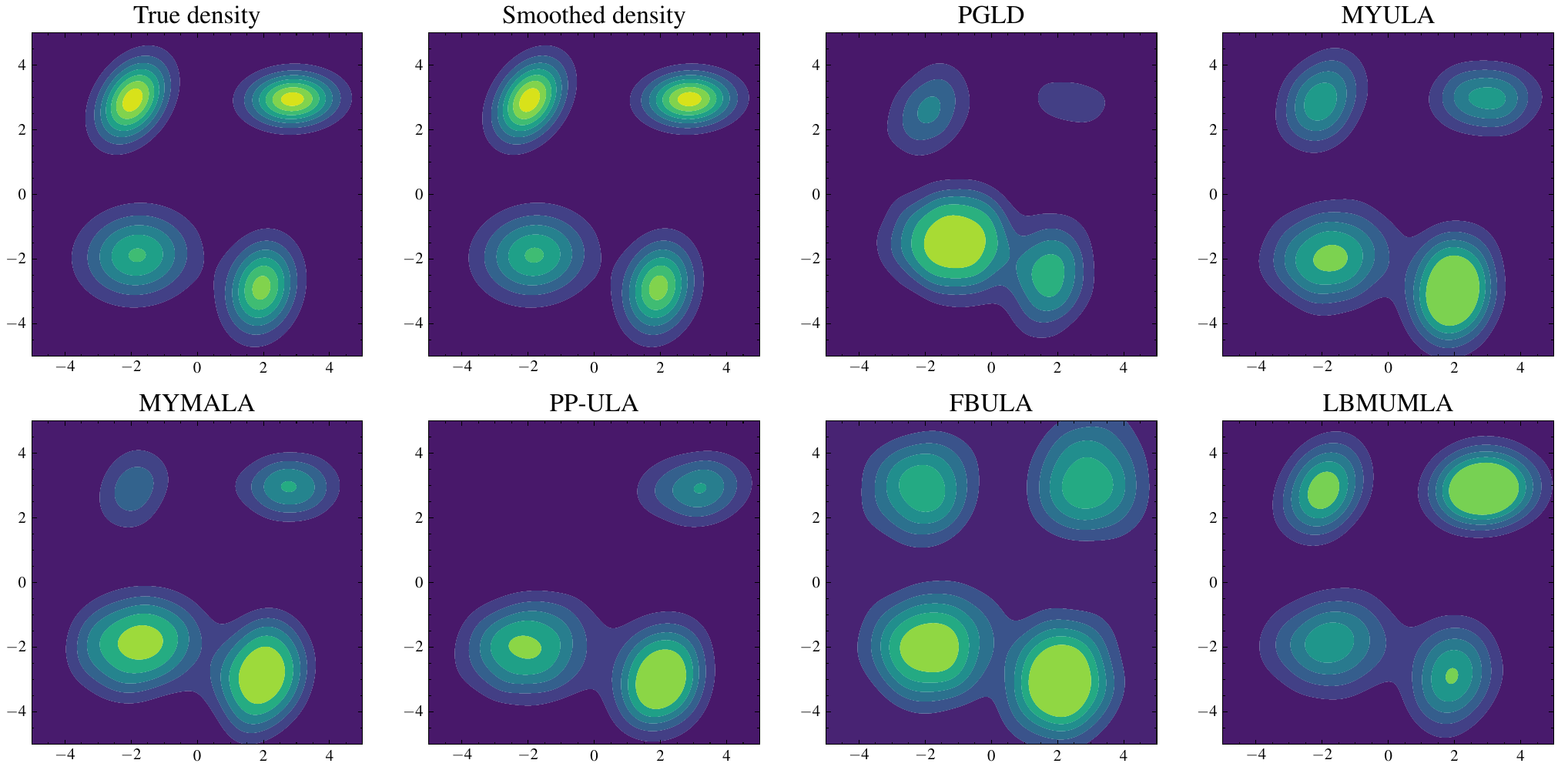}
                \caption{$K=4$}
            \end{subfigure}    
            \caption{Mixture of $K$ Gaussians and a Laplacian prior with step size and smoothing parameter pair $(\gamma, \lambda)=(0.05, 0.25)$}
        \end{figure}

        \begin{figure}[htbp]
            \centering
            \begin{subfigure}[h]{.48\textwidth}
                \centering
                \includegraphics[width=\textwidth]{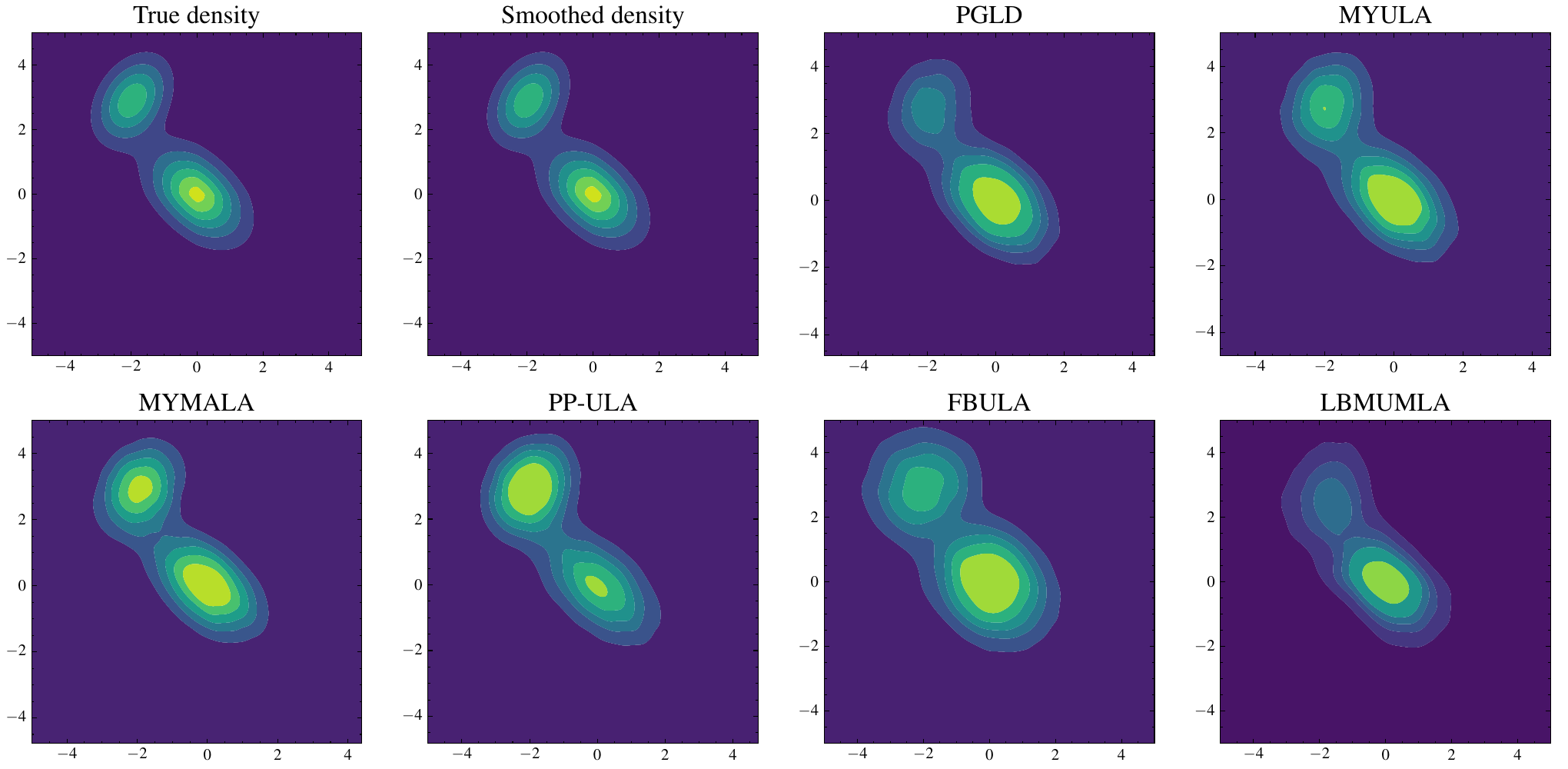}
                \caption{$K=2$}
                \vspace*{1mm}
            \end{subfigure}    
            \hfill
            \begin{subfigure}[h]{.48\textwidth}
                \centering
                \includegraphics[width=\textwidth]{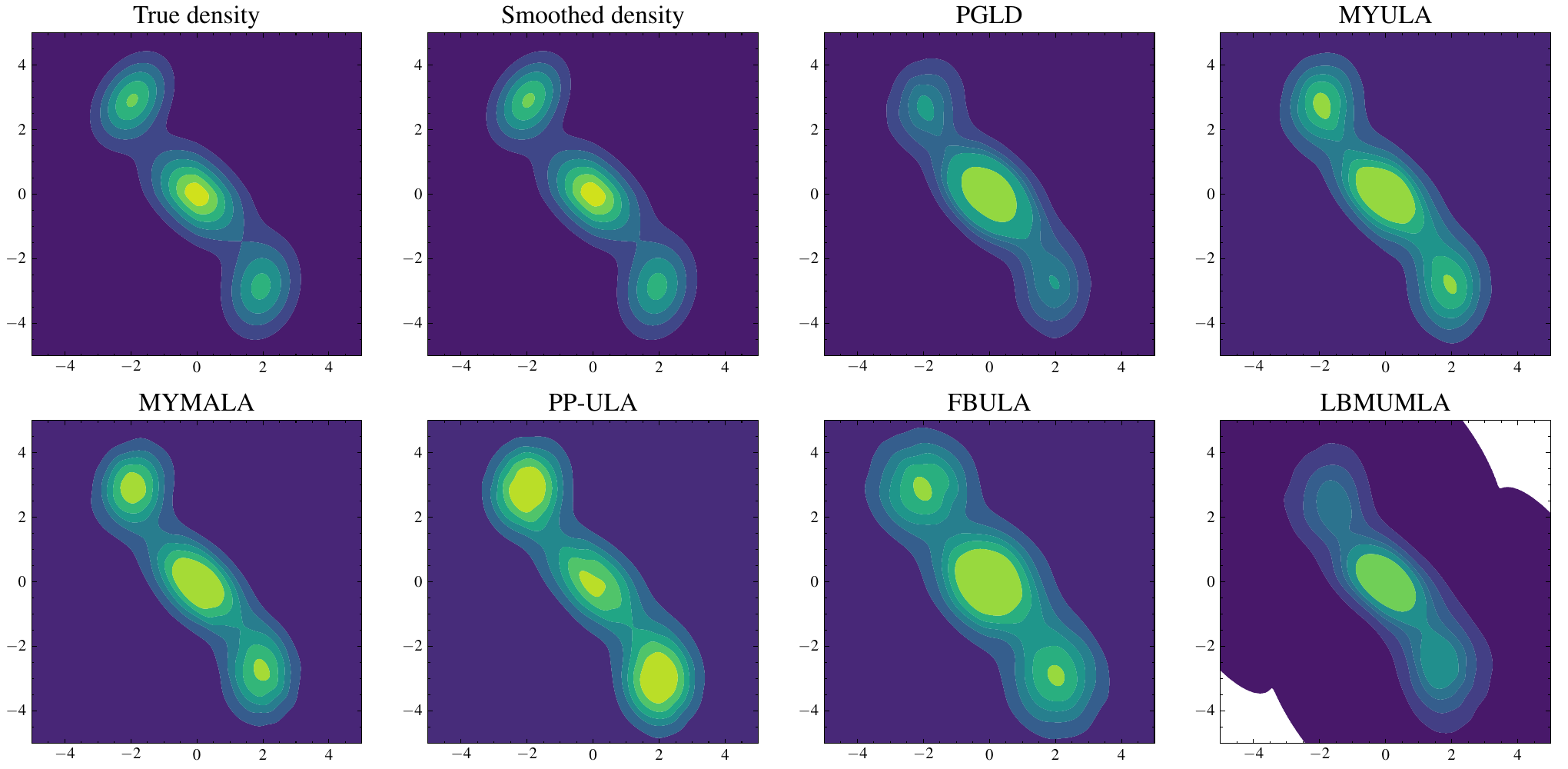}
                \caption{$K=3$}
                \vspace*{1mm}
            \end{subfigure}      
            \par\vspace{2mm}
            \begin{subfigure}[h]{.48\textwidth}
                \centering
                \includegraphics[width=\textwidth]{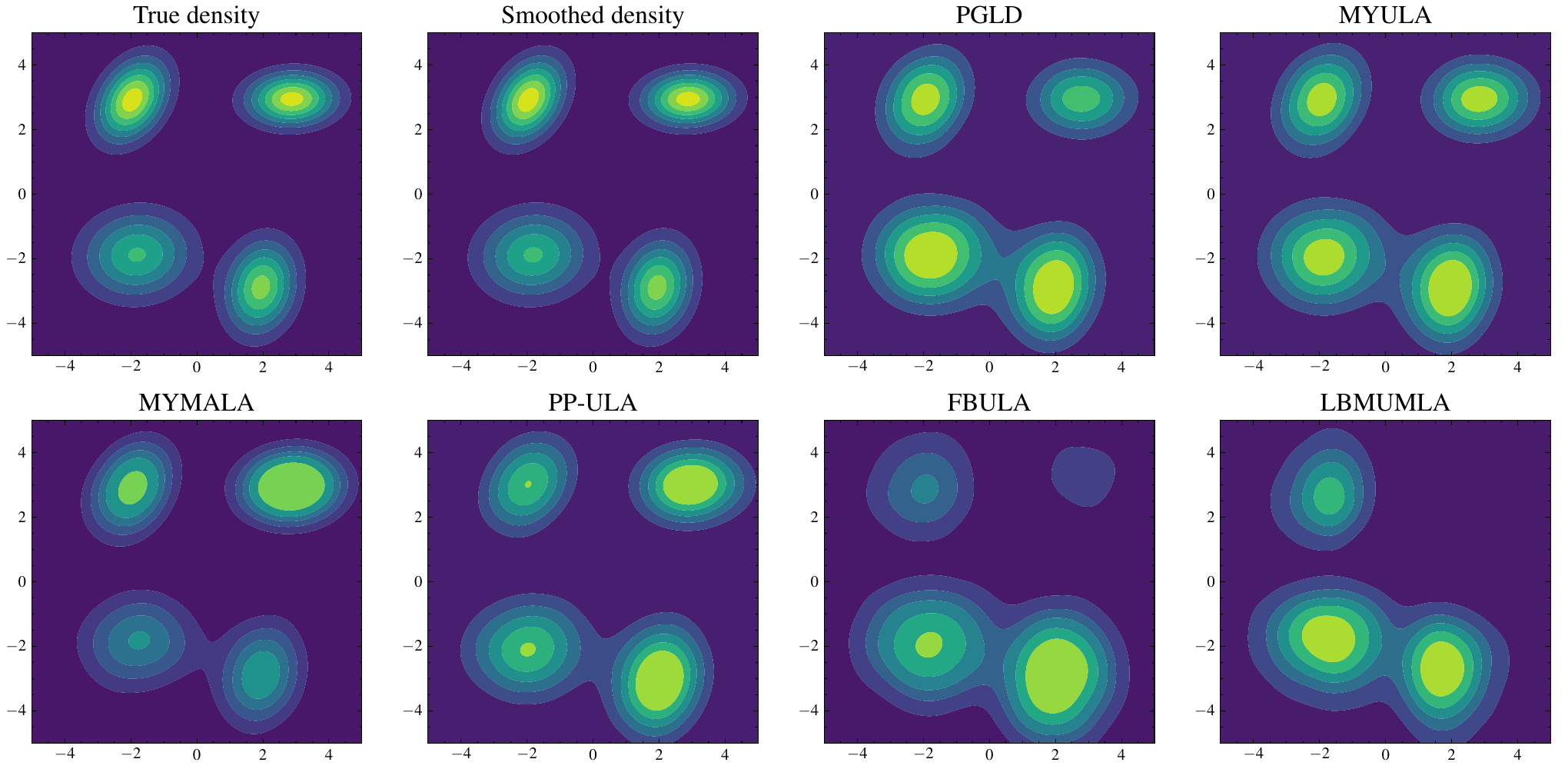}
                \caption{$K=4$}
            \end{subfigure}  
            \hfill
            \begin{subfigure}[h]{.48\textwidth}
                \centering
                \includegraphics[width=\textwidth]{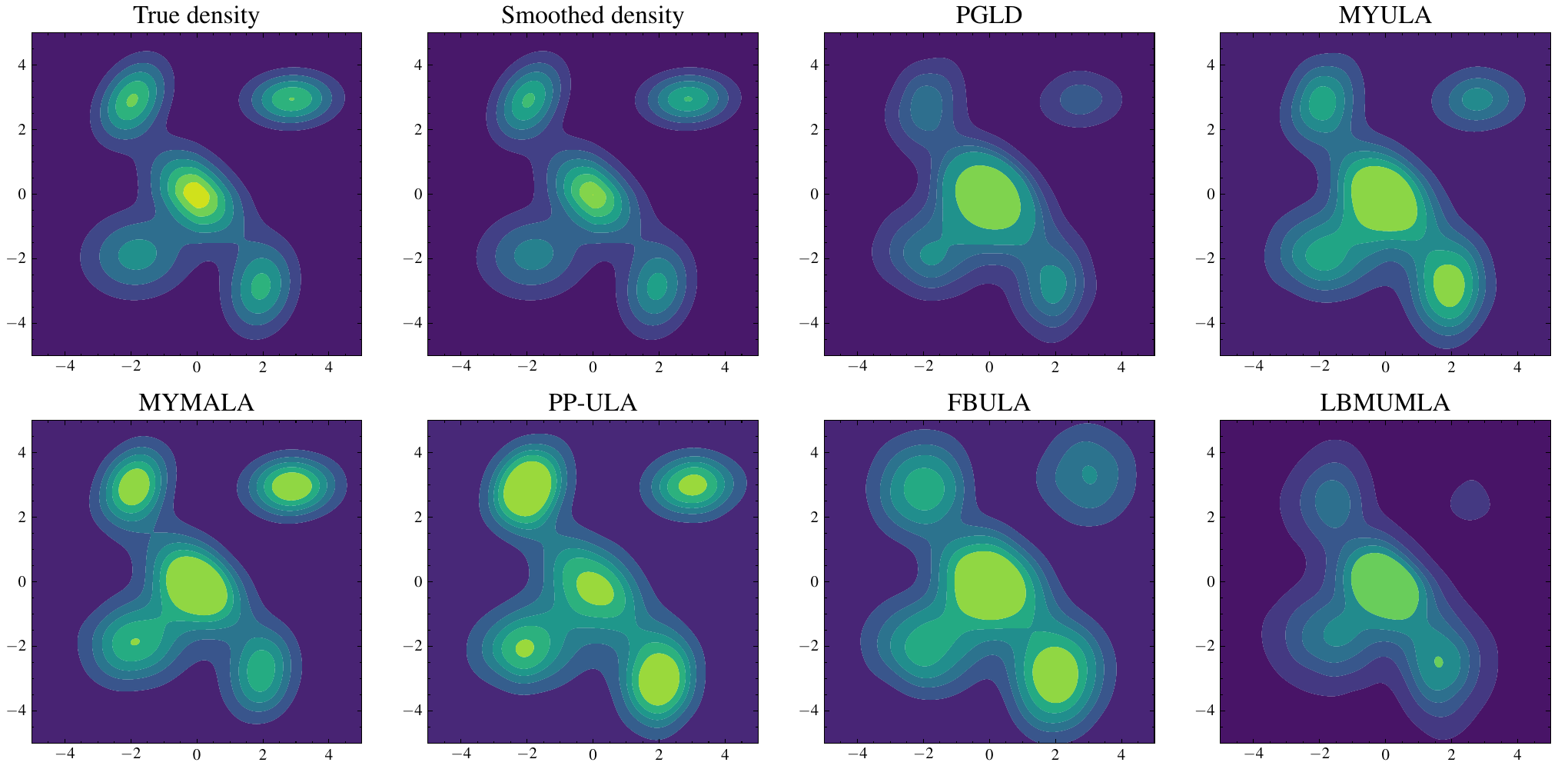}
                \caption{$K=5$}
            \end{subfigure}      
            \caption{Mixture of $K$ Gaussians and a Laplacian prior with step size and smoothing parameter pair $(\gamma, \lambda)=(0.15, 0.25)$}
        \end{figure} 
    
        \begin{figure}[htbp]
            \centering
            \begin{subfigure}[h]{.48\textwidth}
                \centering
                \includegraphics[width=\textwidth]{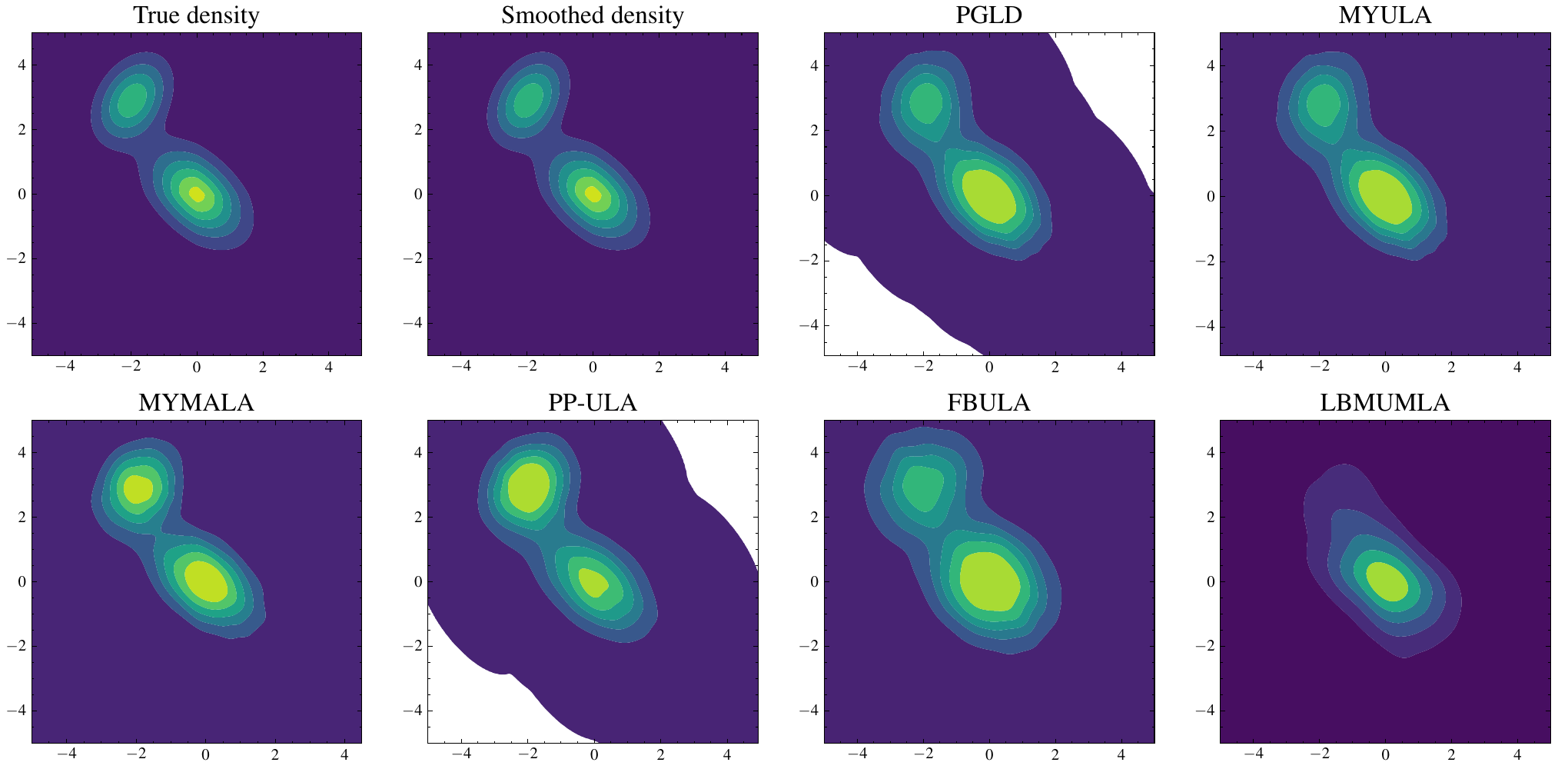}
                \caption{$K=2$}
                \vspace*{1mm}
            \end{subfigure}    
            \hfill
            \begin{subfigure}[h]{.48\textwidth}
                \centering
                \includegraphics[width=\textwidth]{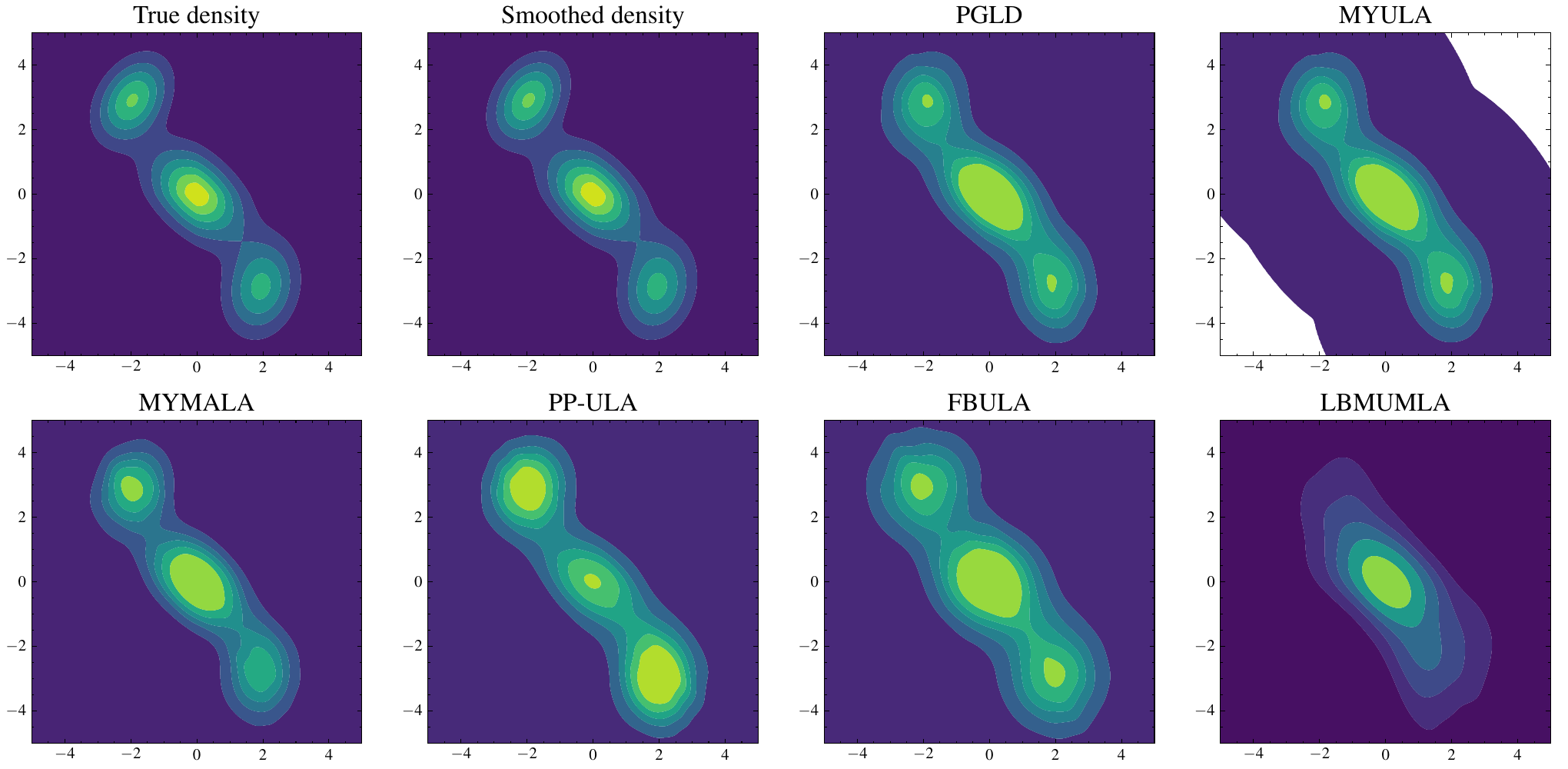}
                \caption{$K=3$}
                \vspace*{1mm}
            \end{subfigure}      
            \par\vspace{2mm}
            \begin{subfigure}[h]{.48\textwidth}
                \centering
                \includegraphics[width=\textwidth]{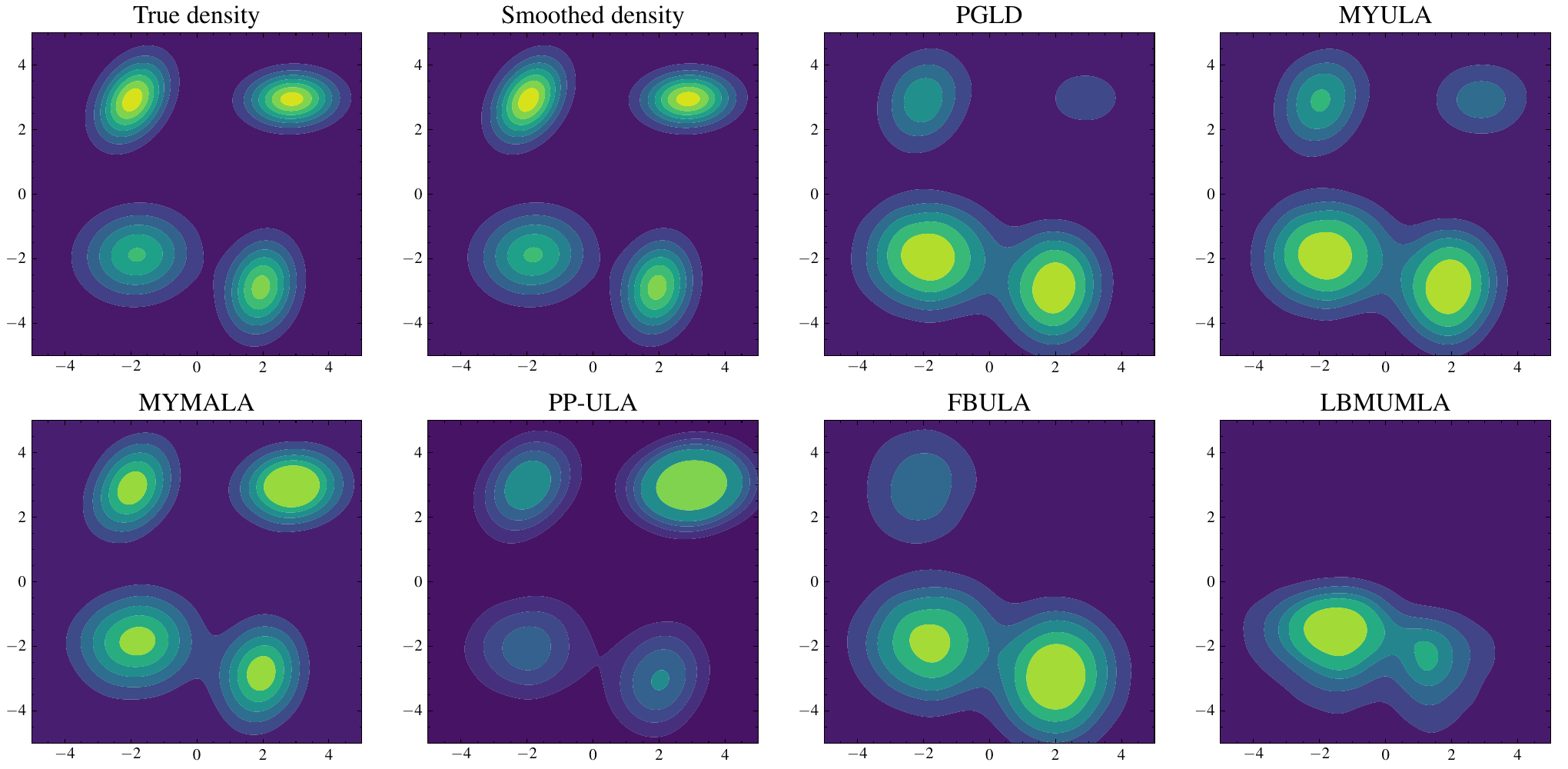}
                \caption{$K=4$}
            \end{subfigure}  
            \hfill
            \begin{subfigure}[h]{.48\textwidth}
                \centering
                \includegraphics[width=\textwidth]{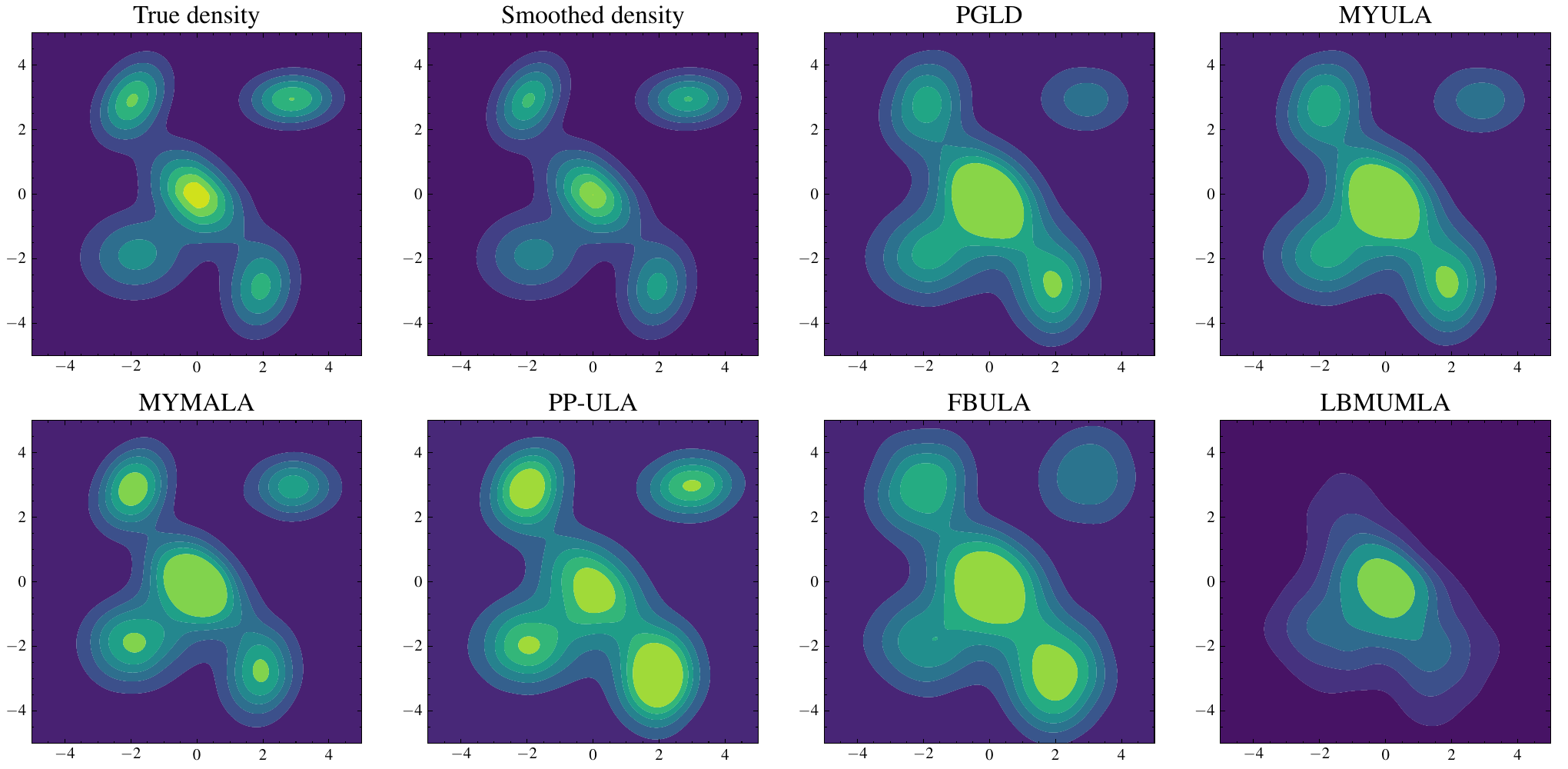}
                \caption{$K=5$}
            \end{subfigure}      
            \caption{Mixture of $K$ Gaussians and a Laplacian prior with step size and smoothing parameter pair $(\gamma, \lambda)=(0.25, 0.25)$}
        \end{figure}

        \begin{figure}[htbp]
            \centering
            \begin{subfigure}[h]{.48\textwidth}
                \centering
                \includegraphics[width=\textwidth]{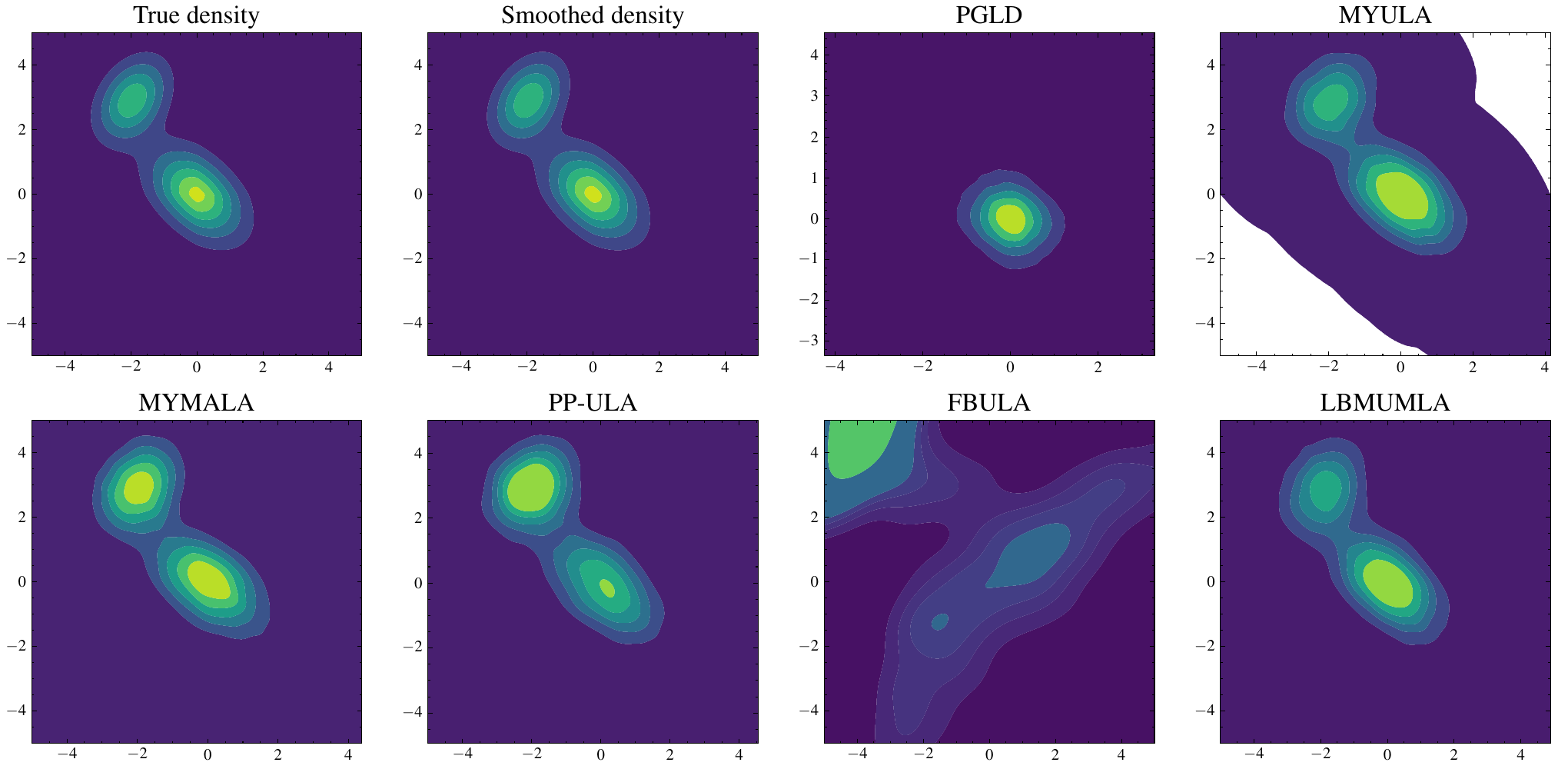}
                \caption{$K=2$}
                \vspace*{1mm}
            \end{subfigure}    
            \hfill
            \begin{subfigure}[h]{.48\textwidth}
                \centering
                \includegraphics[width=\textwidth]{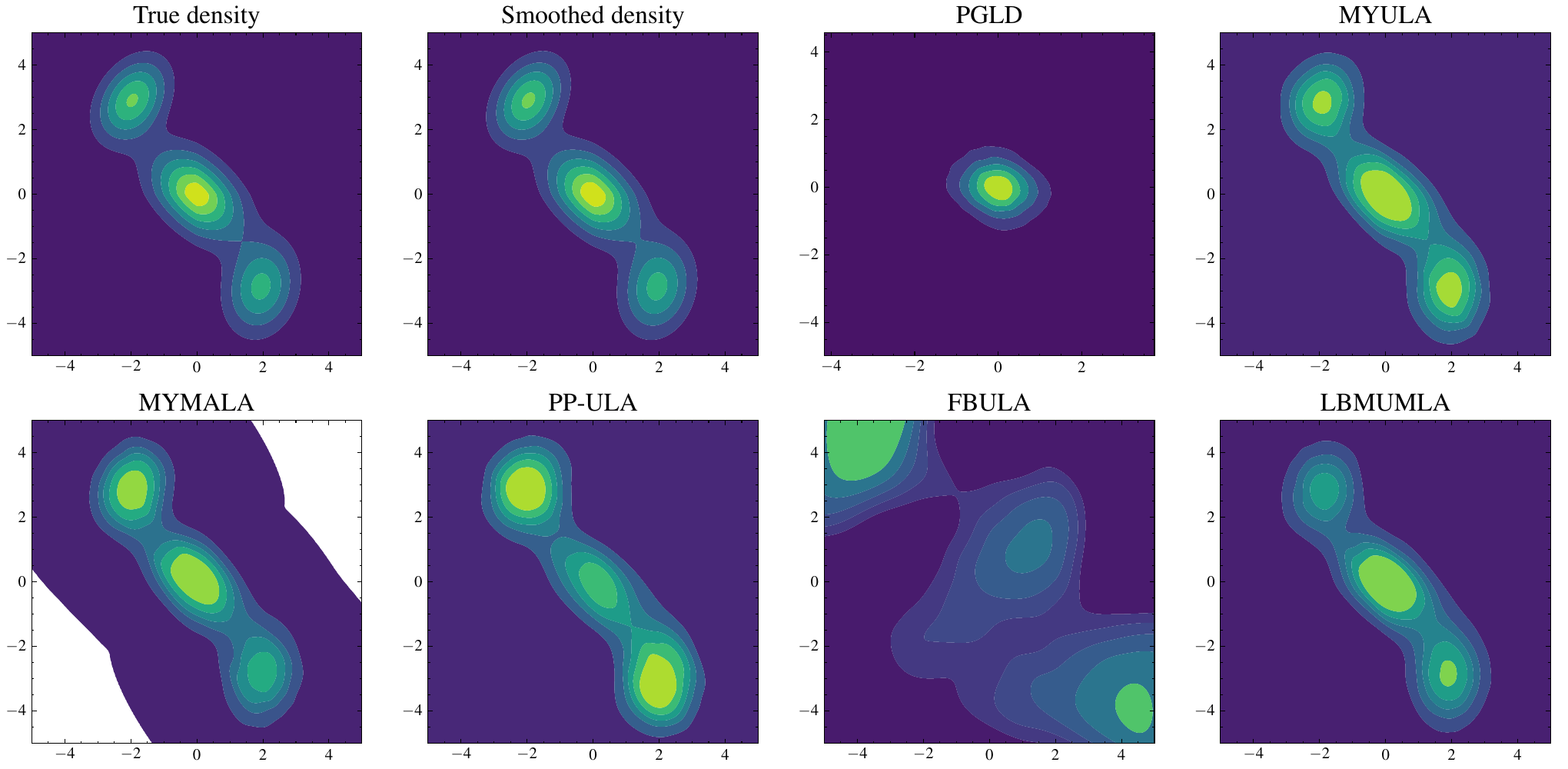}
                \caption{$K=3$}
                \vspace*{1mm}
            \end{subfigure}      
            \par\vspace{2mm}
            \begin{subfigure}[h]{.48\textwidth}
                \centering
                \includegraphics[width=\textwidth]{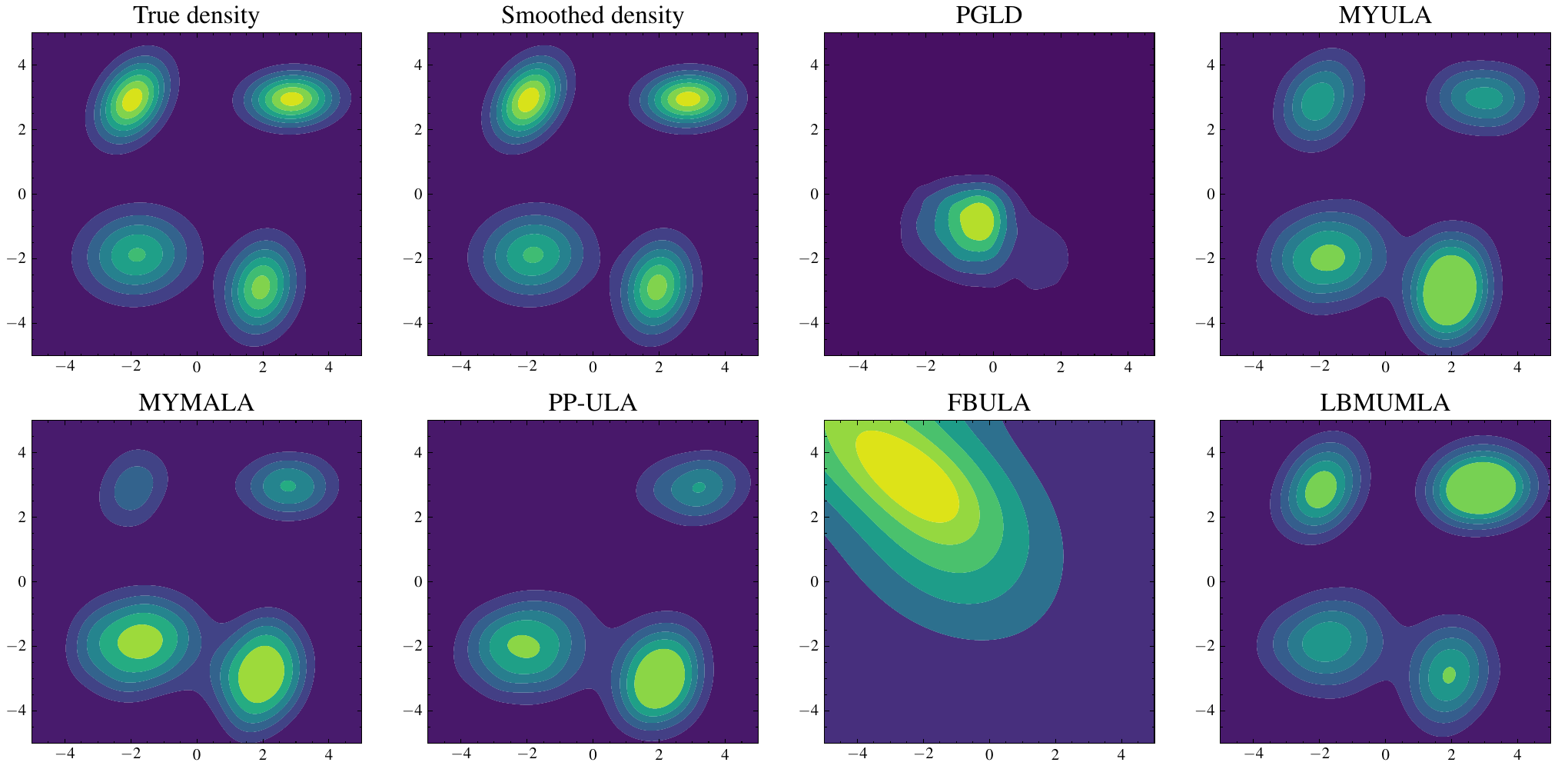}
                \caption{$K=4$}
            \end{subfigure}  
            \hfill
            \begin{subfigure}[h]{.48\textwidth}
                \centering
                \includegraphics[width=\textwidth]{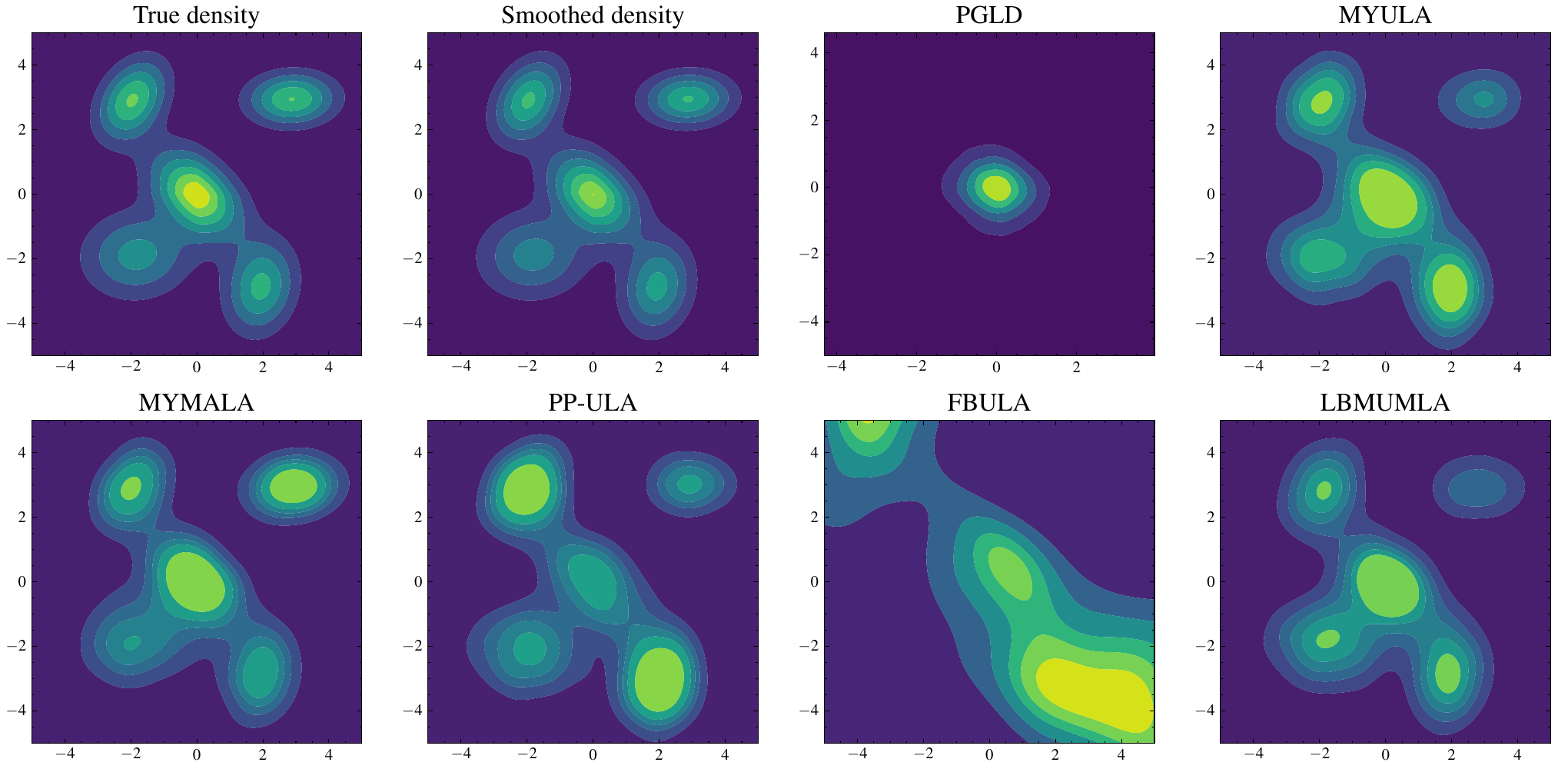}
                \caption{$K=5$}
            \end{subfigure}      
            \caption{Mixture of $K$ Gaussians and a Laplacian prior with step size and smoothing parameter pair $(\gamma, \lambda)=(0.05, 0.5)$}
        \end{figure}

        \begin{figure}[htbp]
            \centering
            \begin{subfigure}[h]{.48\textwidth}
                \centering
                \includegraphics[width=\textwidth]{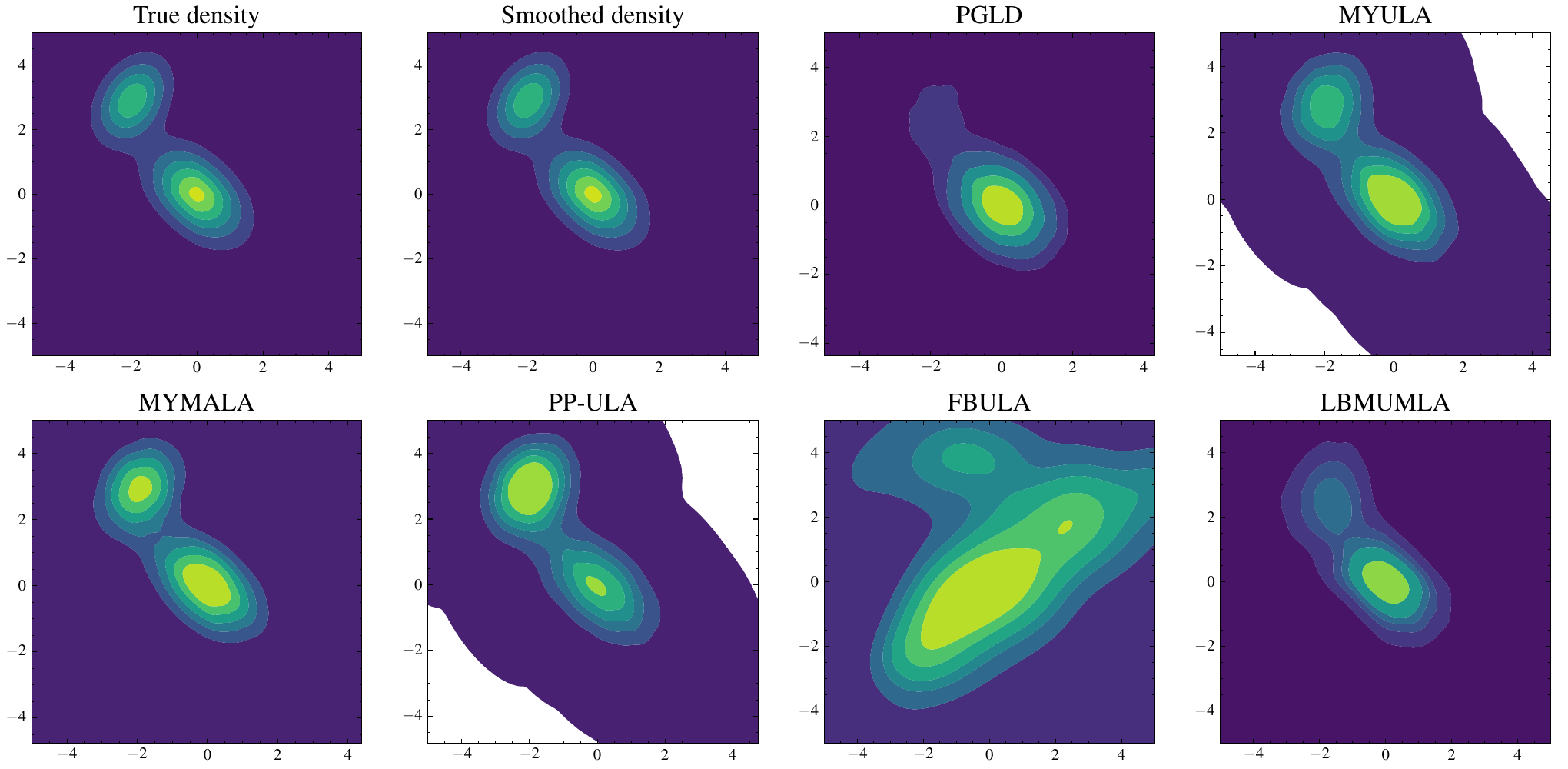}
                \caption{$K=2$}
                \vspace*{1mm}
            \end{subfigure}    
            \hfill
            \begin{subfigure}[h]{.48\textwidth}
                \centering
                \includegraphics[width=\textwidth]{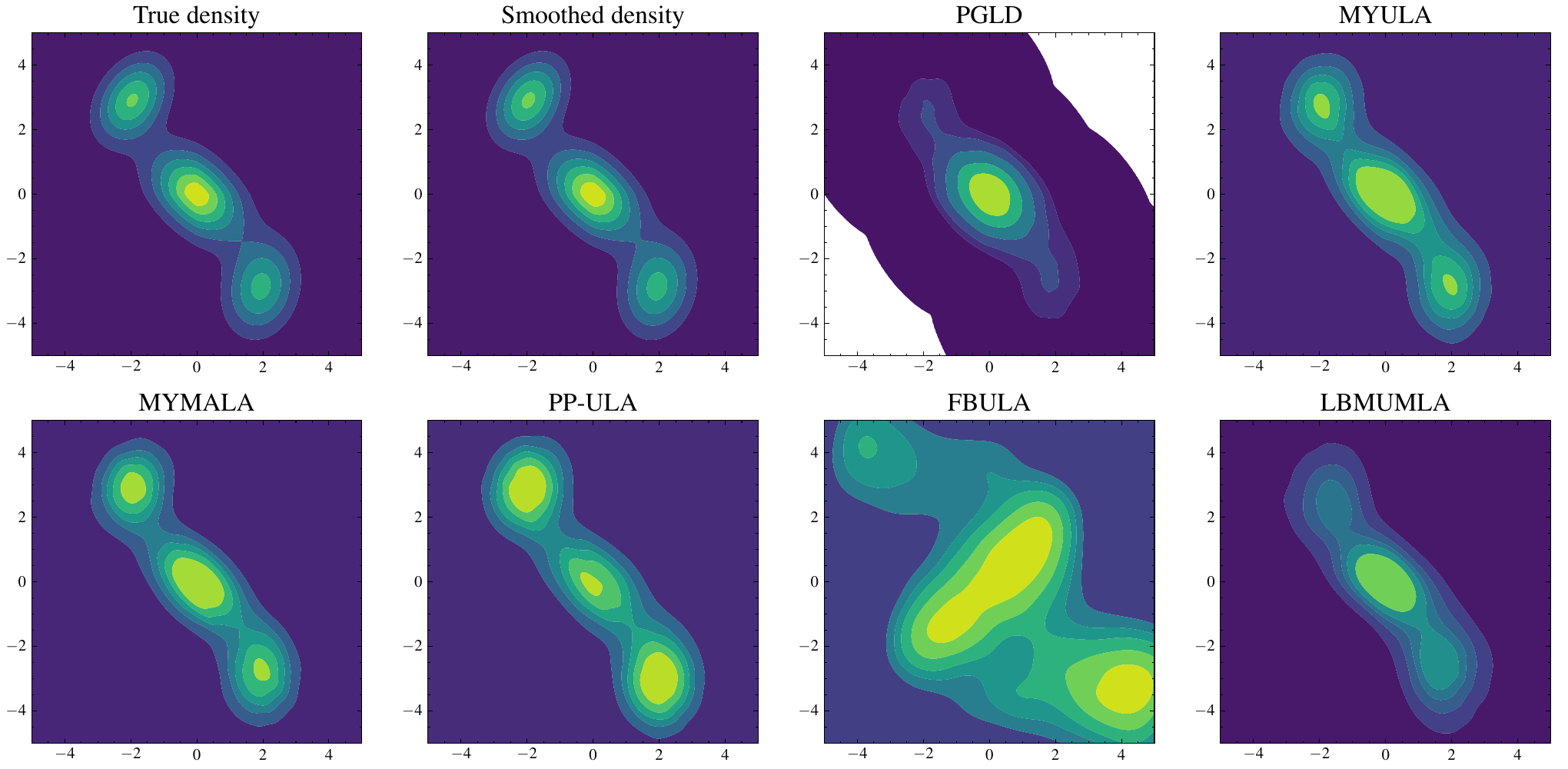}
                \caption{$K=3$}
                \vspace*{1mm}
            \end{subfigure}      
            \par\vspace{2mm}
            \begin{subfigure}[h]{.48\textwidth}
                \centering
                \includegraphics[width=\textwidth]{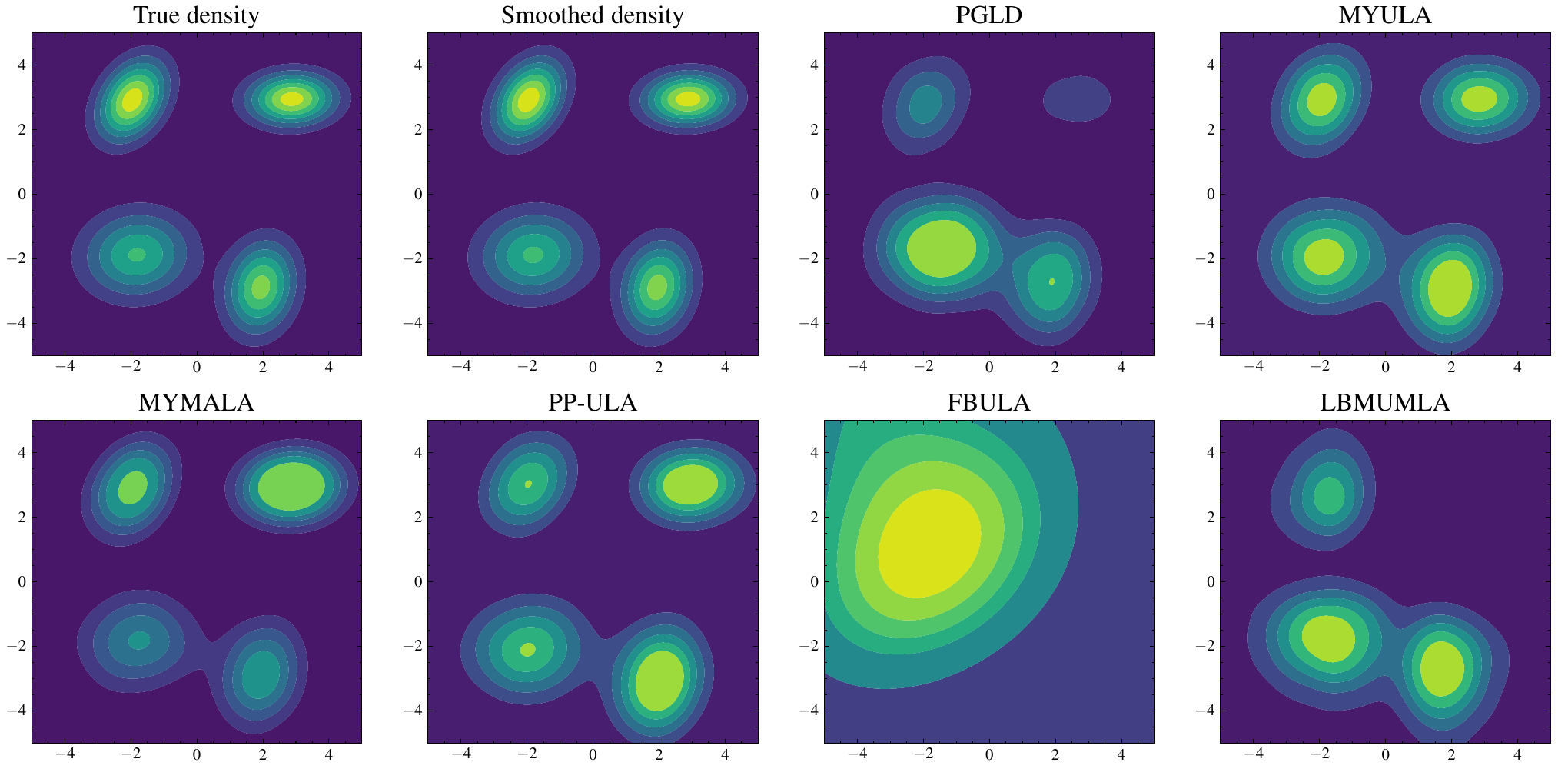}
                \caption{$K=4$}
            \end{subfigure}  
            \hfill
            \begin{subfigure}[h]{.48\textwidth}
                \centering
                \includegraphics[width=\textwidth]{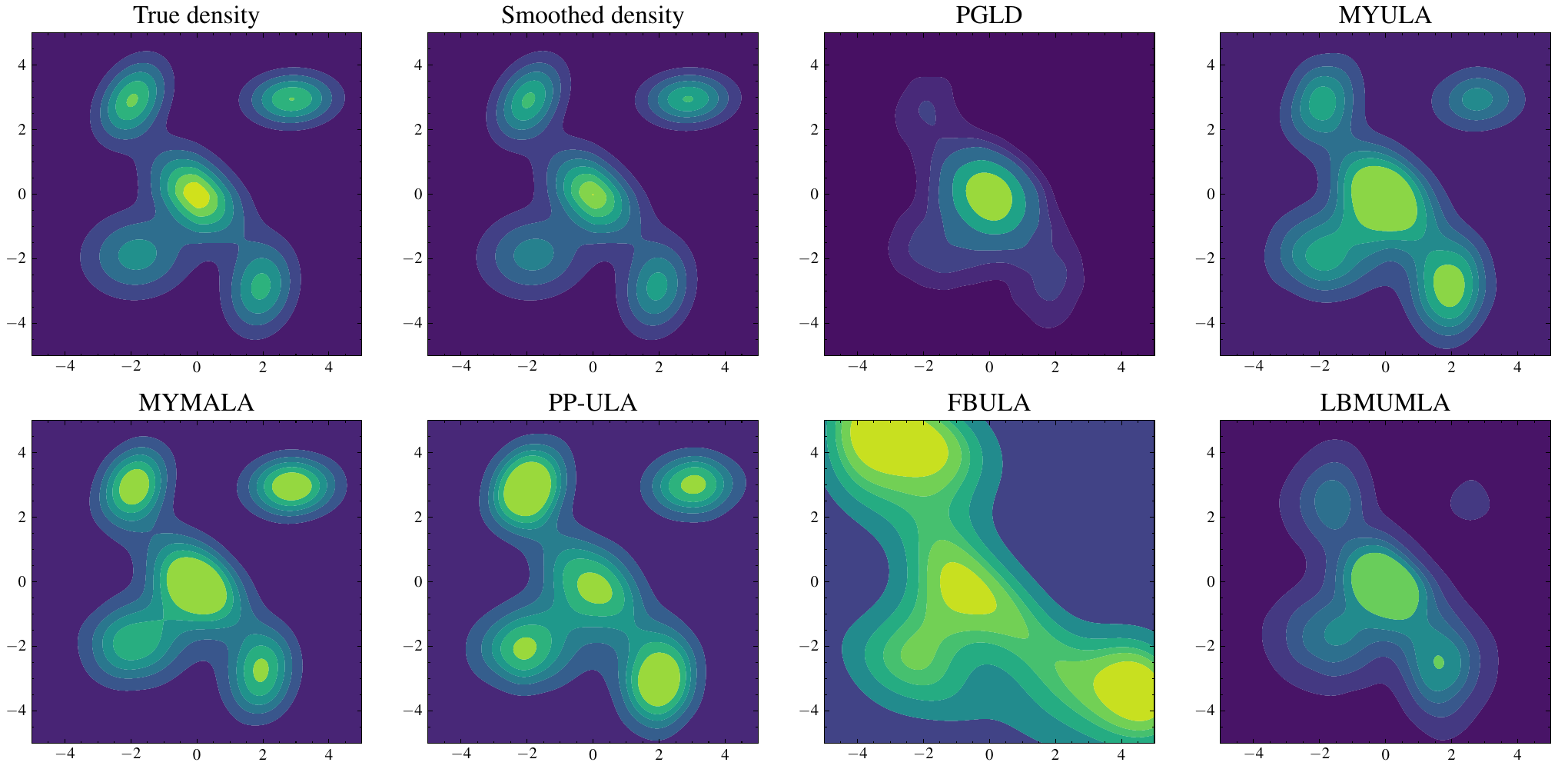}
                \caption{$K=5$}
            \end{subfigure}      
            \caption{Mixture of $K$ Gaussians and a Laplacian prior with step size and smoothing parameter pair $(\gamma, \lambda)=(0.15, 0.5)$}
        \end{figure}

        \begin{figure}[htbp]
            \centering
            \begin{subfigure}[h]{.48\textwidth}
                \centering
                \includegraphics[width=\textwidth]{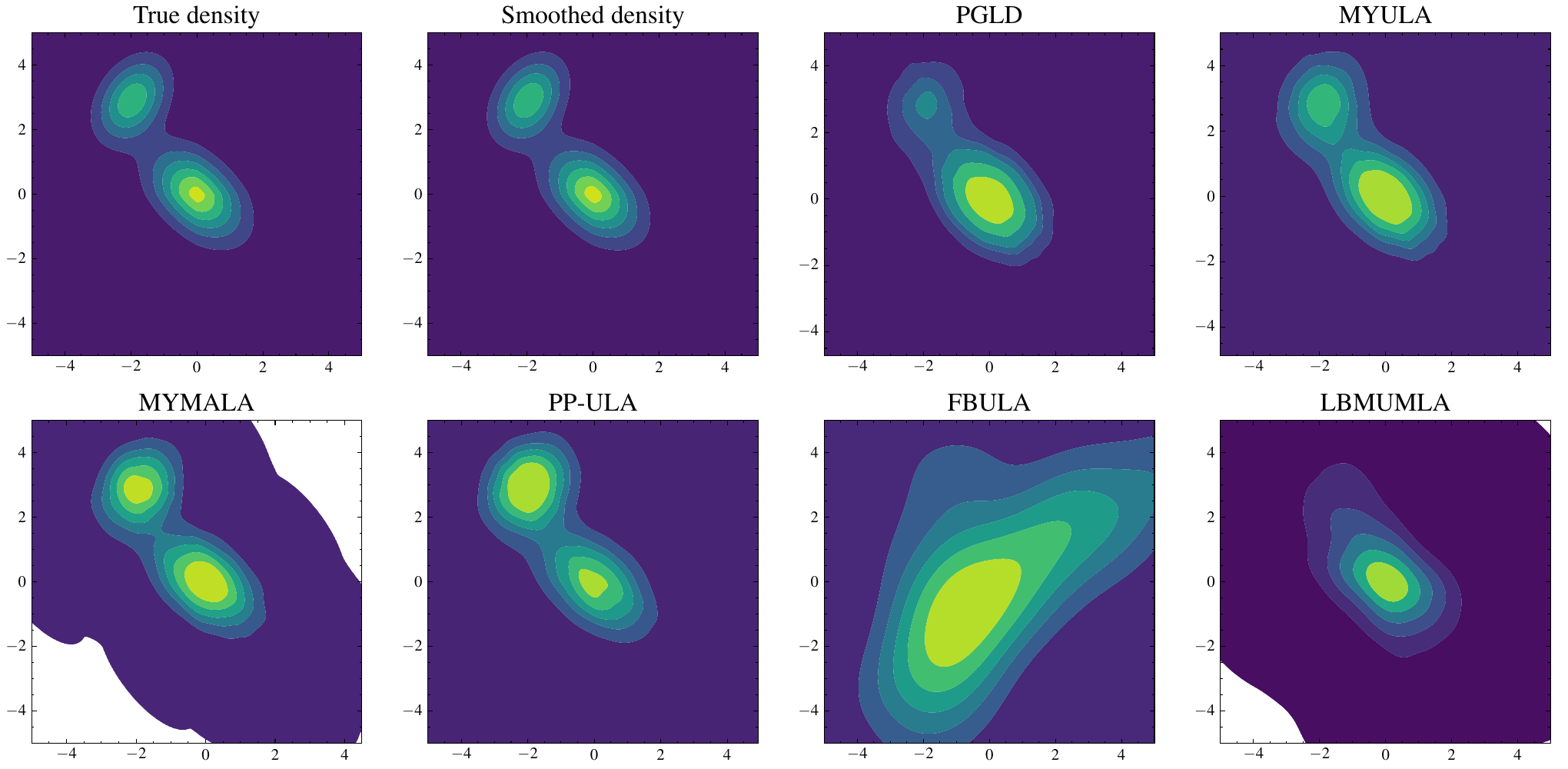}
                \caption{$K=2$}
                \vspace*{1mm}
            \end{subfigure}    
            \hfill
            \begin{subfigure}[h]{.48\textwidth}
                \centering
                \includegraphics[width=\textwidth]{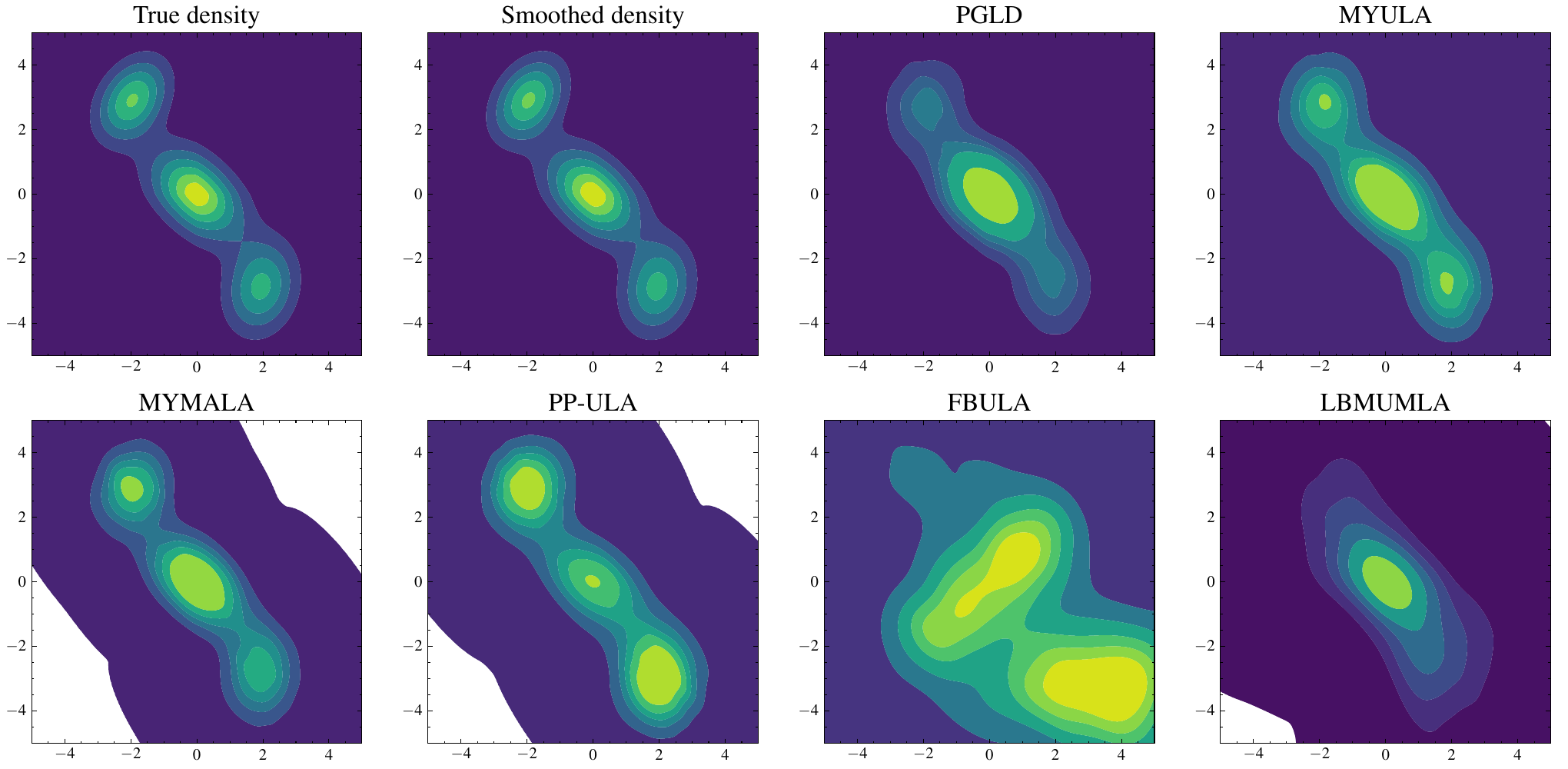}
                \caption{$K=3$}
                \vspace*{1mm}
            \end{subfigure}      
            \par\vspace{2mm}
            \begin{subfigure}[h]{.48\textwidth}
                \centering
                \includegraphics[width=\textwidth]{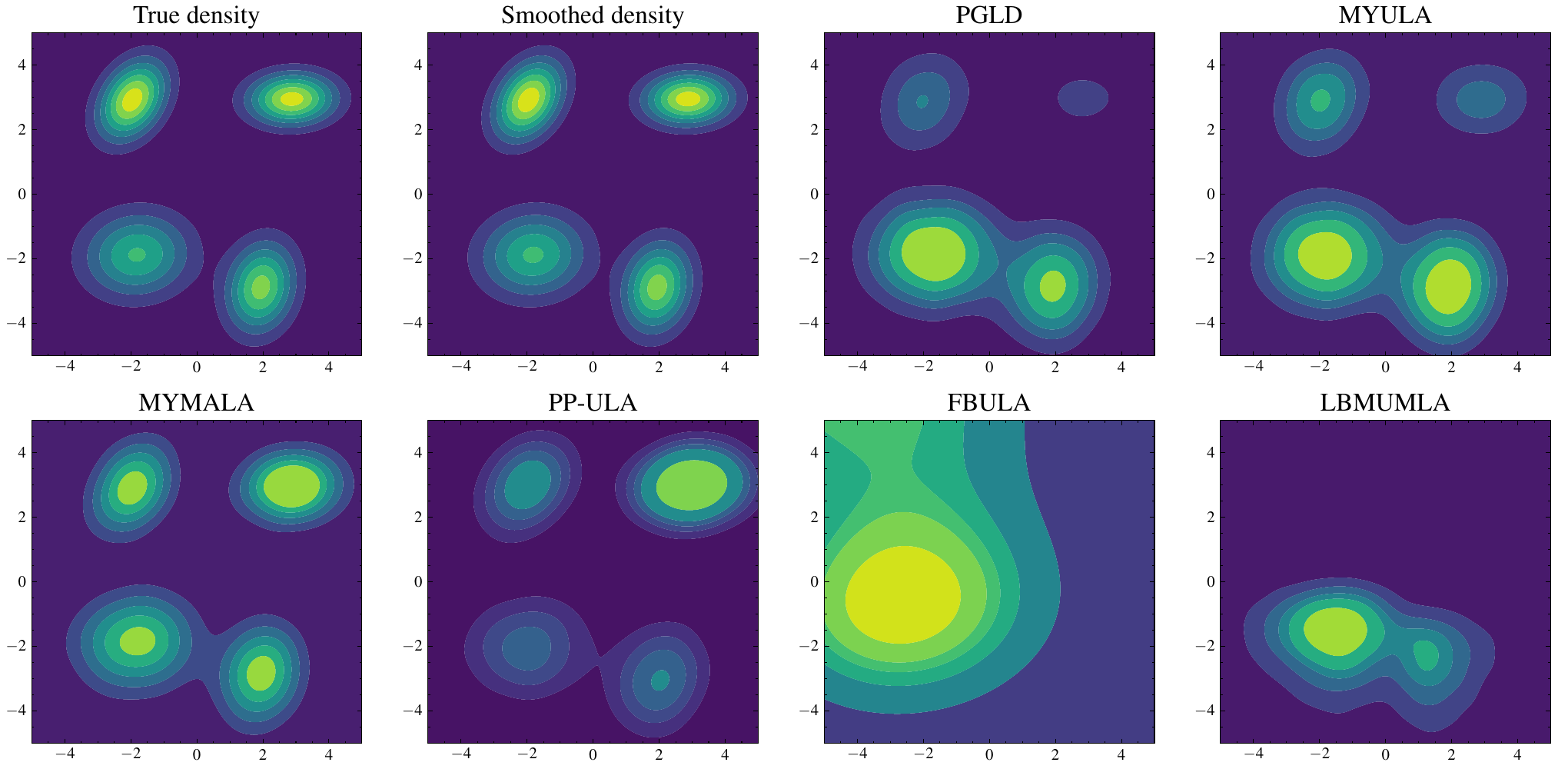}
                \caption{$K=4$}
            \end{subfigure}  
            \hfill
            \begin{subfigure}[h]{.48\textwidth}
                \centering
                \includegraphics[width=\textwidth]{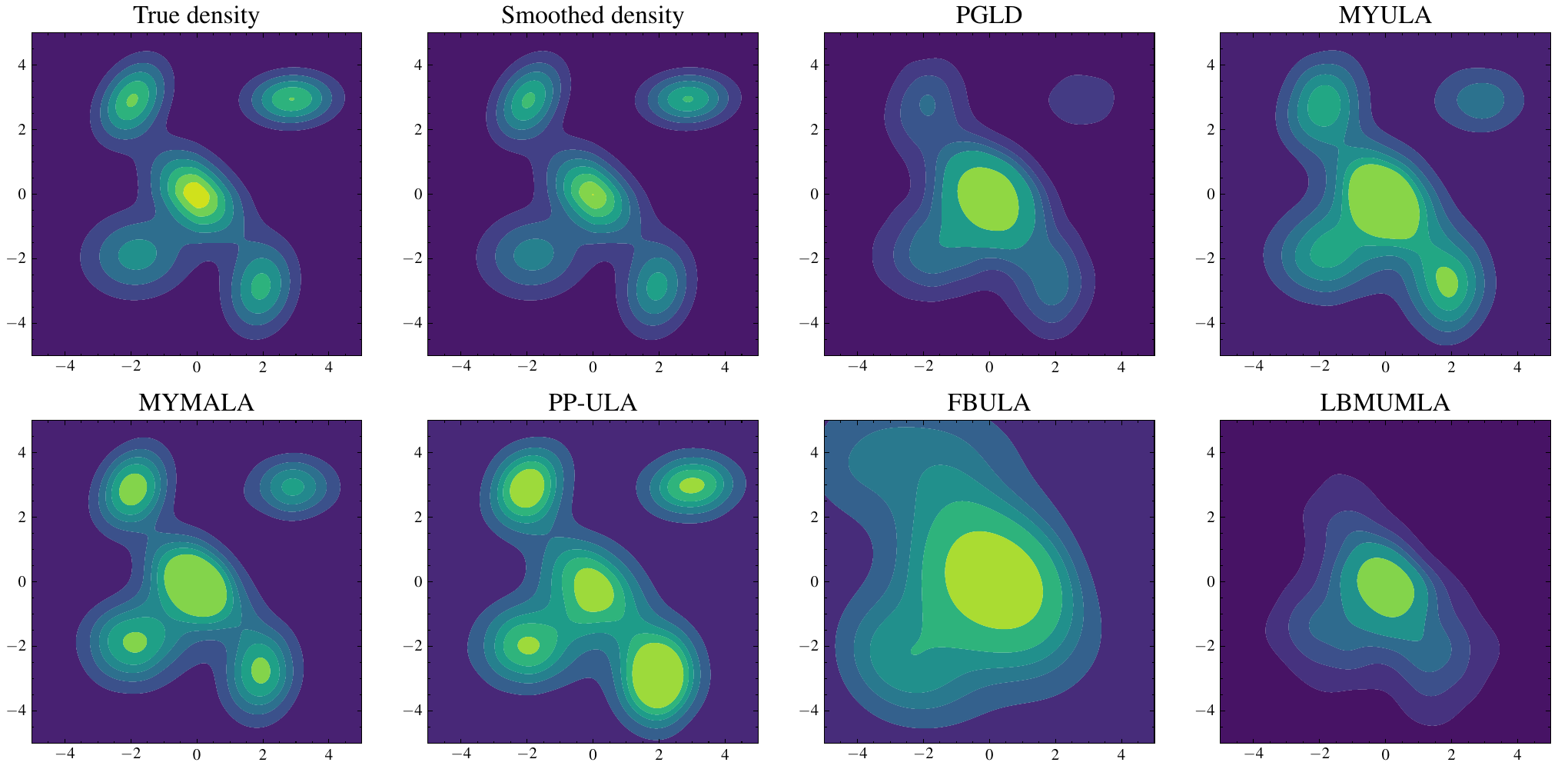}
                \caption{$K=5$}
            \end{subfigure}      
            \caption{Mixture of $K$ Gaussians and a Laplacian prior with step size and smoothing parameter pair $(\gamma, \lambda)=(0.25, 0.5)$}
        \end{figure}

         \begin{figure}[htbp]
             \centering
             \begin{subfigure}[h]{.48\textwidth}
                 \centering
                 \includegraphics[width=\textwidth]{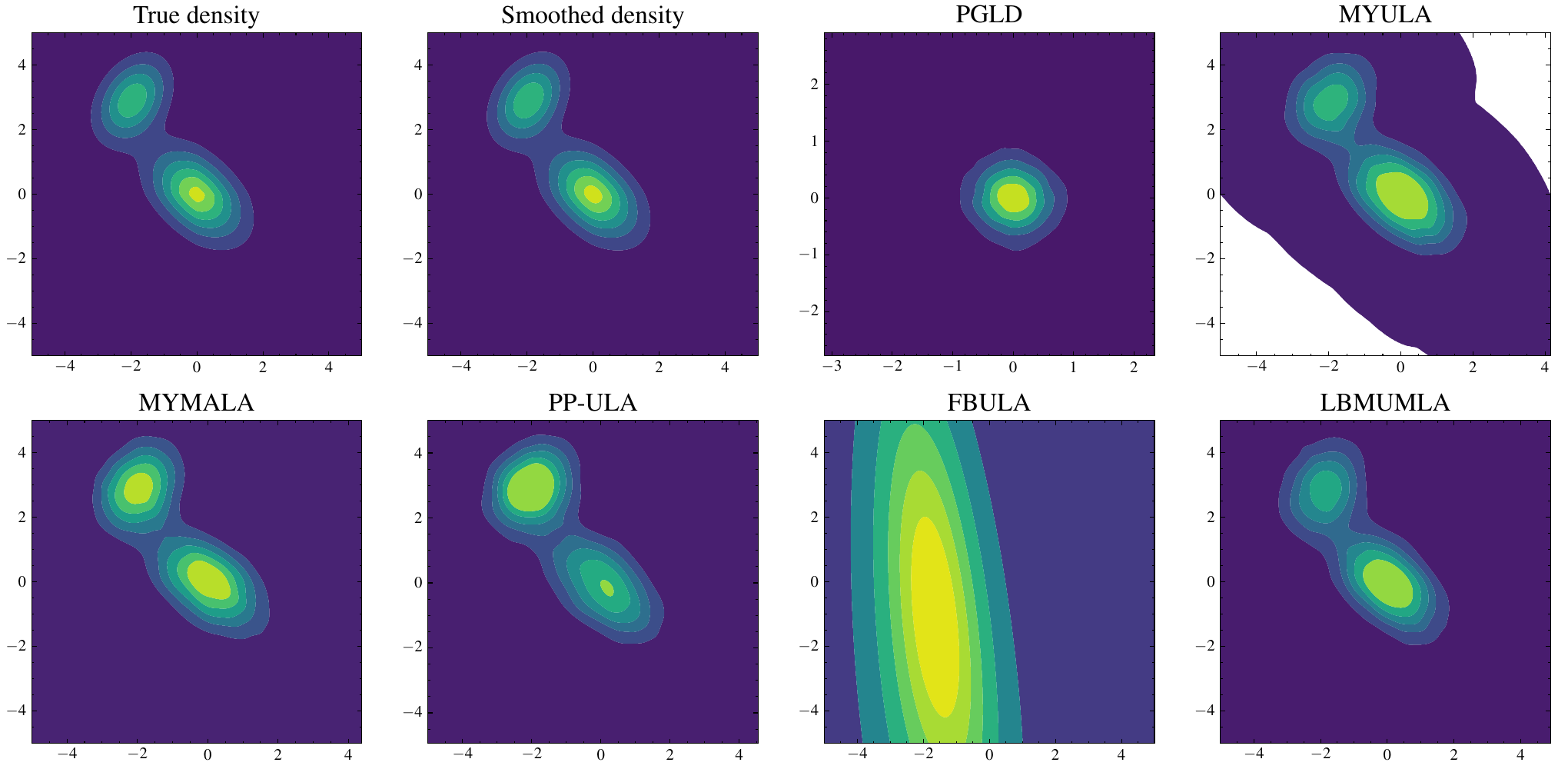}
                 \caption{$K=2$}
                 \vspace*{1mm}
             \end{subfigure}    
             \hfill
             \begin{subfigure}[h]{.48\textwidth}
                 \centering
                 \includegraphics[width=\textwidth]{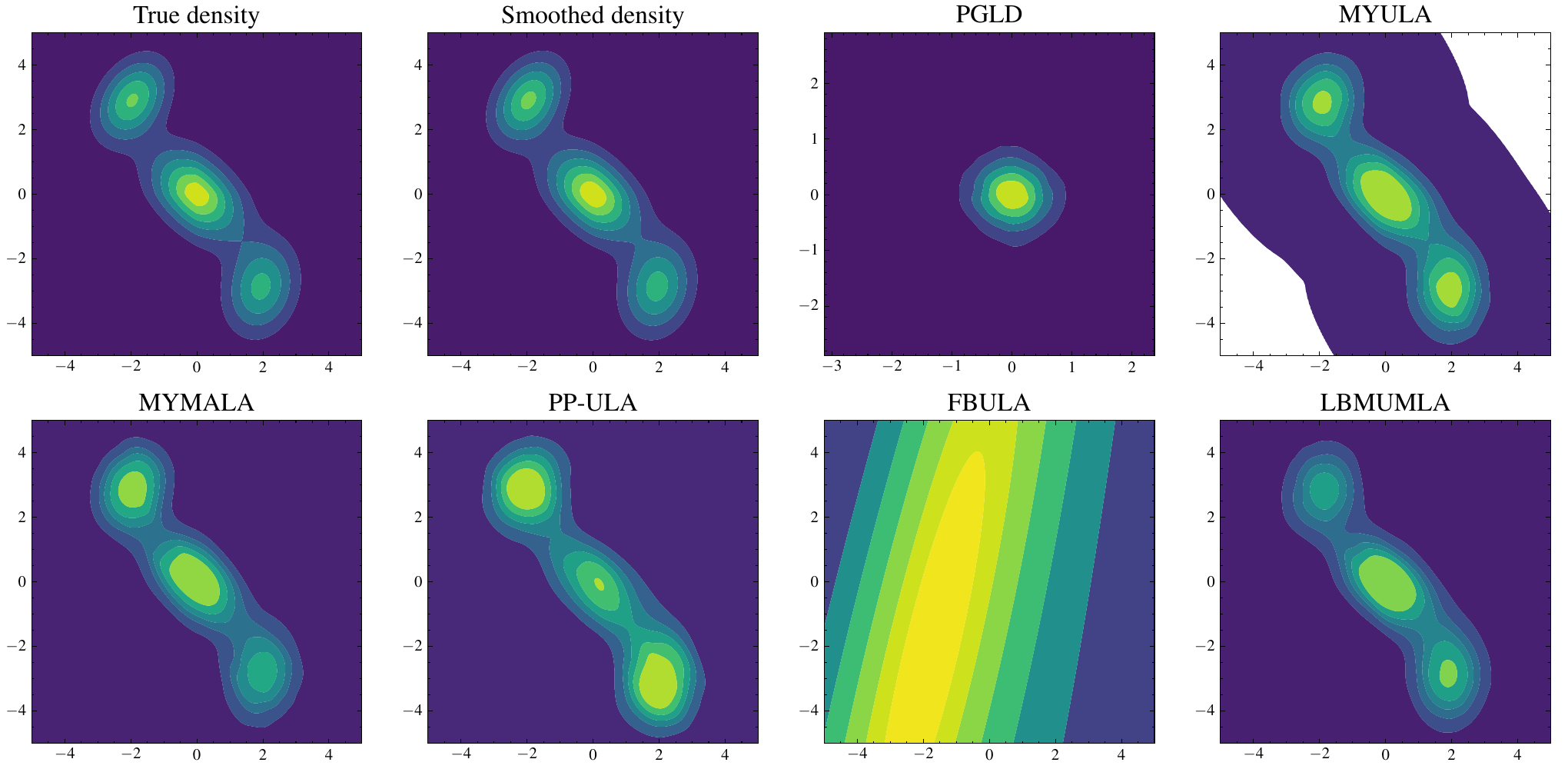}
                 \caption{$K=3$}
                 \vspace*{1mm}
             \end{subfigure}      
             \par\vspace{2mm}
             \begin{subfigure}[h]{.48\textwidth}
                 \centering
                 \includegraphics[width=\textwidth]{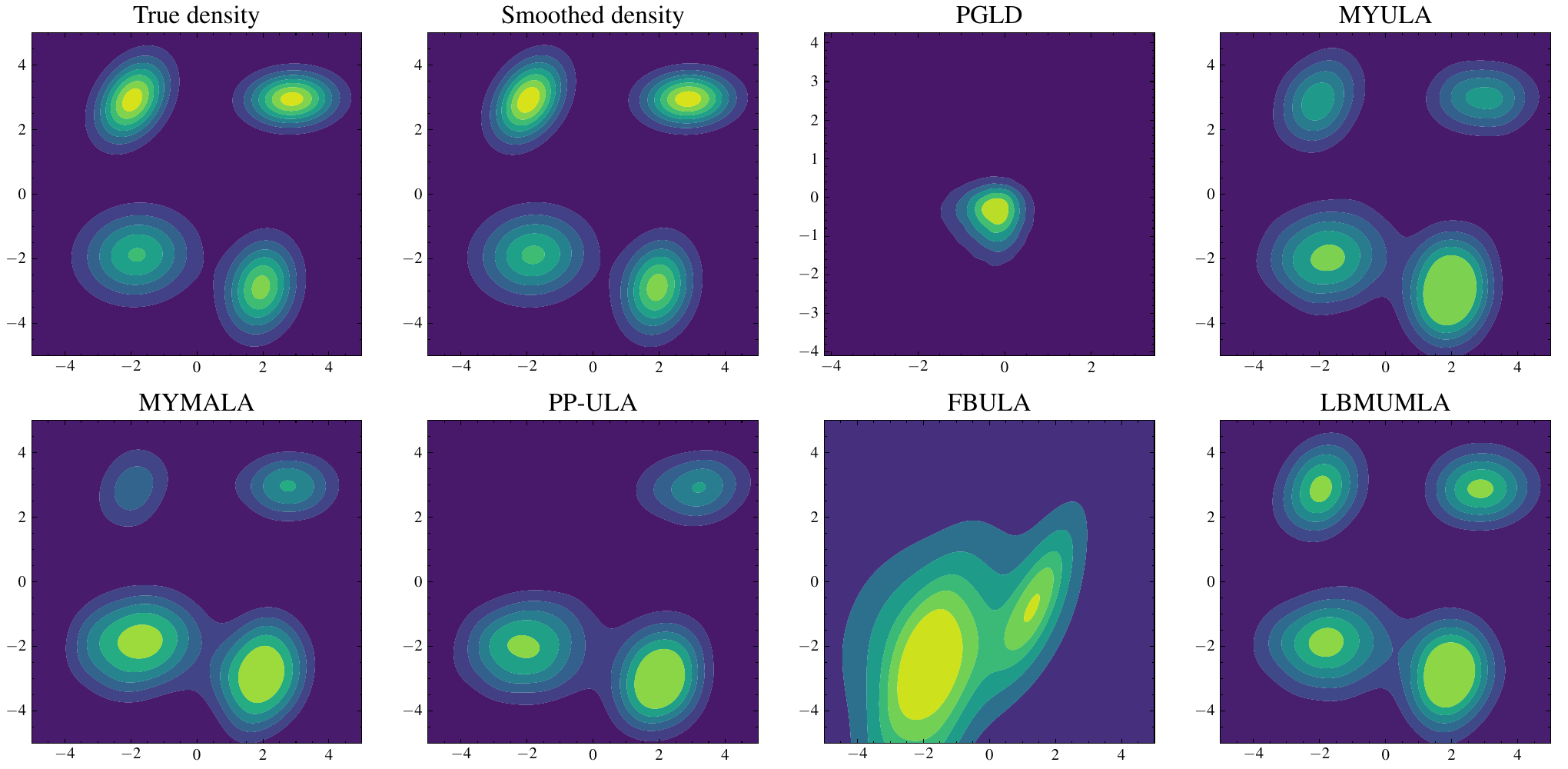}
                 \caption{$K=4$}
             \end{subfigure}  
             \hfill
             \begin{subfigure}[h]{.48\textwidth}
                 \centering
                 \includegraphics[width=\textwidth]{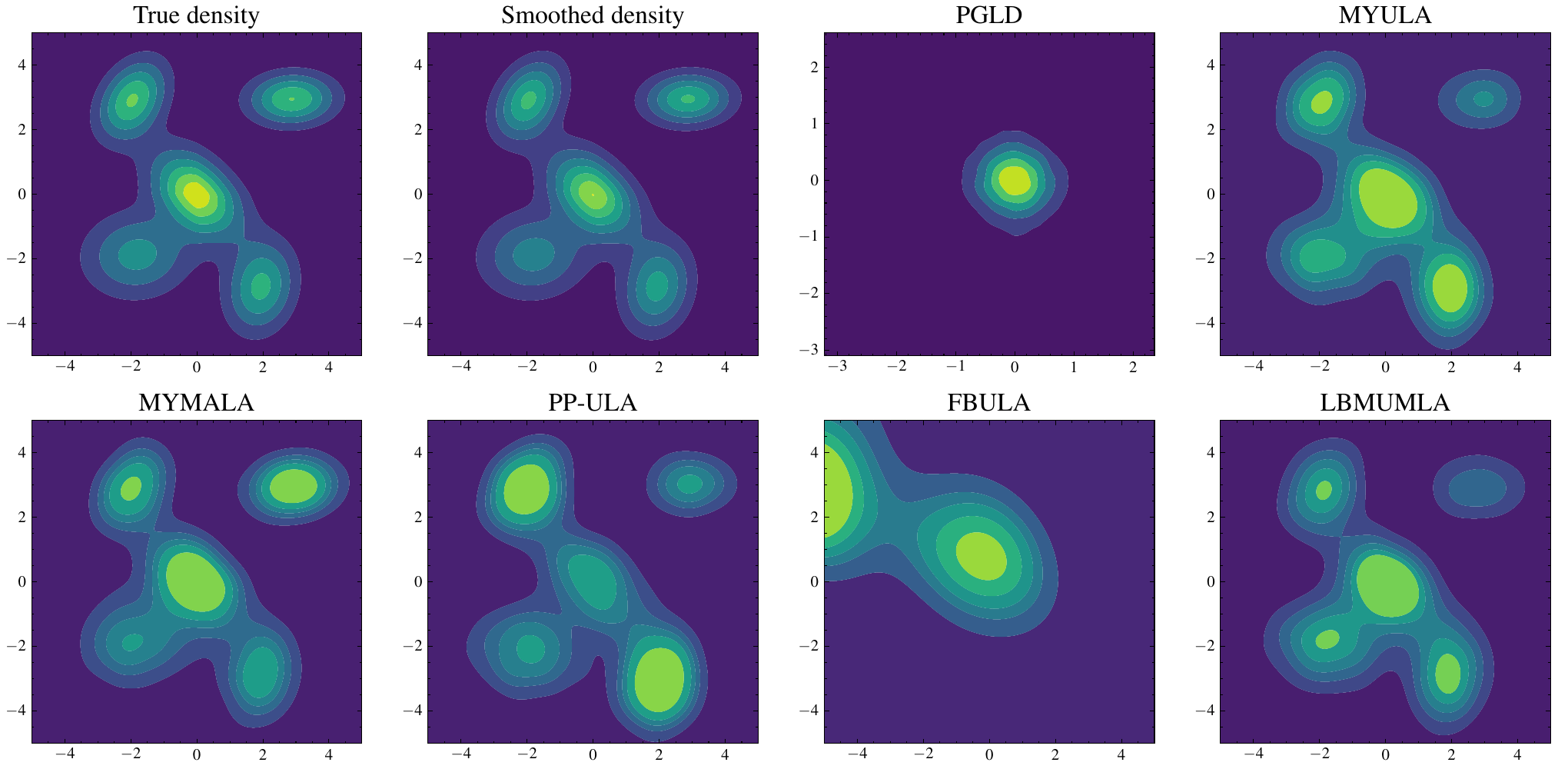}
                 \caption{$K=5$}
             \end{subfigure}      
             \caption{Mixture of $K$ Gaussians and a Laplacian prior with step size and smoothing parameter pair $(\gamma, \lambda)=(0.05, 1)$}
         \end{figure}

         \begin{figure}[htbp]
             \centering
             \begin{subfigure}[h]{.48\textwidth}
                 \centering
                 \includegraphics[width=\textwidth]{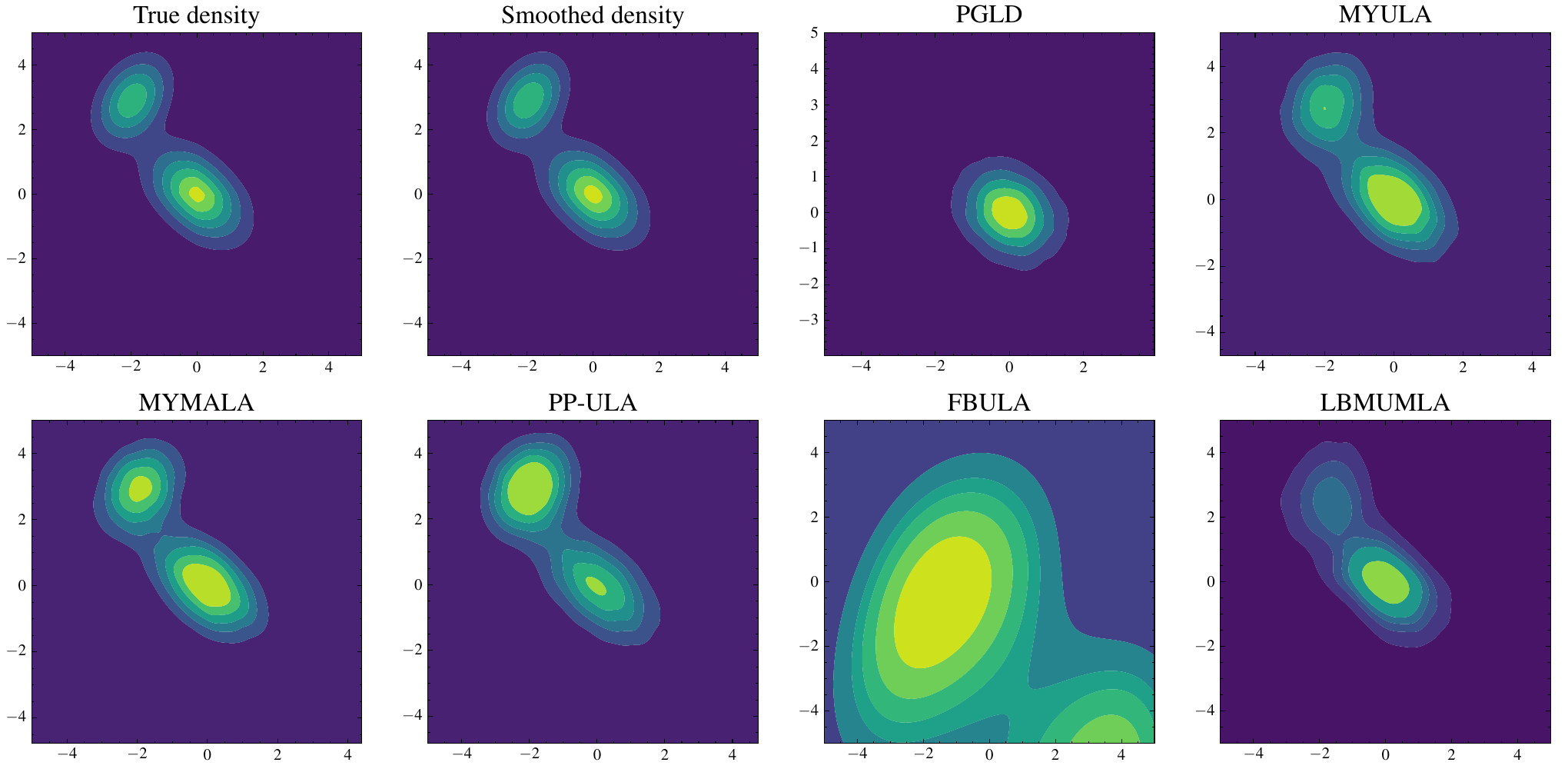}
                 \caption{$K=2$}
                 \vspace*{1mm}
             \end{subfigure}    
             \hfill
             \begin{subfigure}[h]{.48\textwidth}
                 \centering
                 \includegraphics[width=\textwidth]{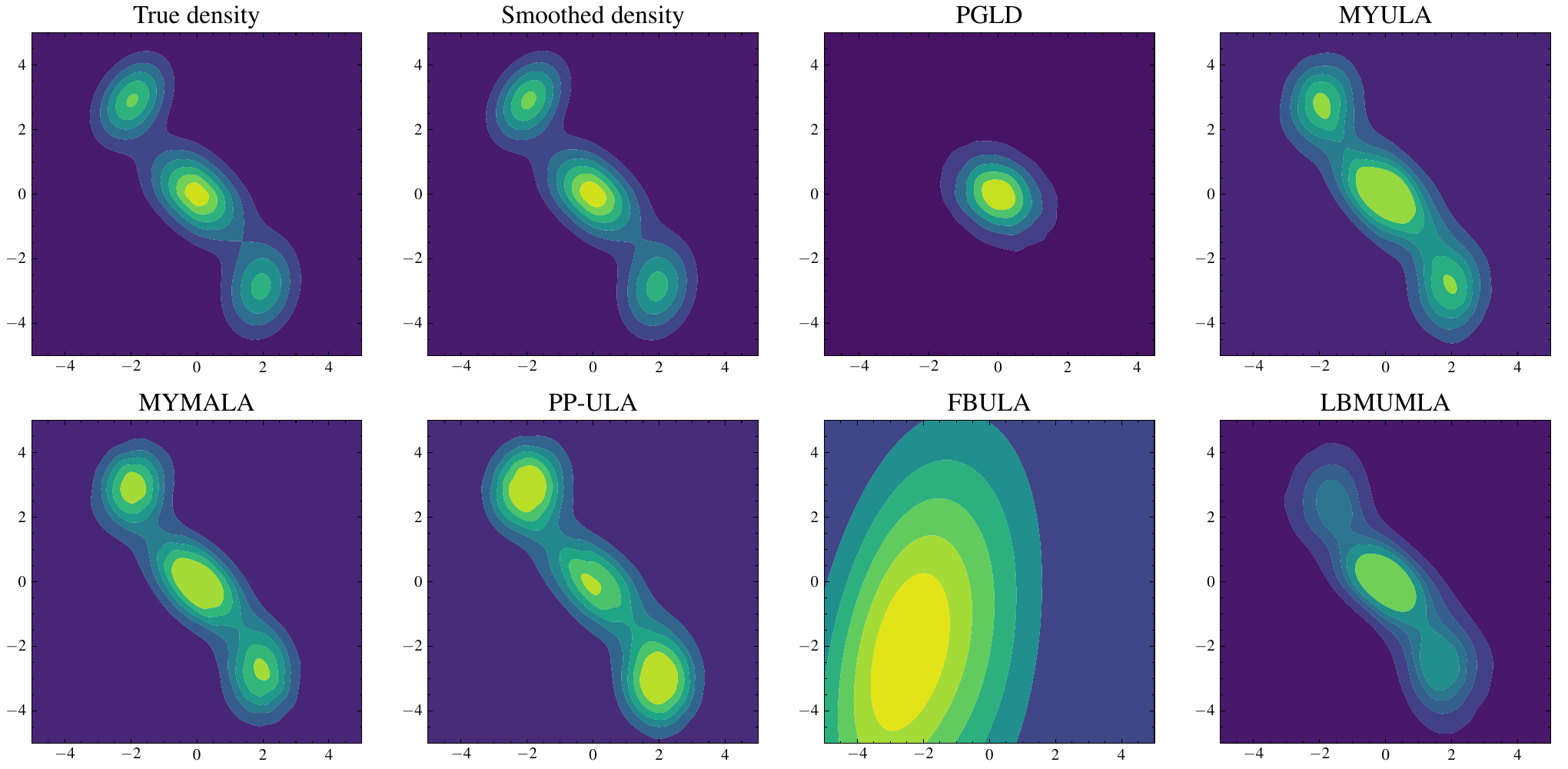}
                 \caption{$K=3$}
                 \vspace*{1mm}
             \end{subfigure}      
             \par\vspace{2mm}
             \begin{subfigure}[h]{.48\textwidth}
                 \centering
                 \includegraphics[width=\textwidth]{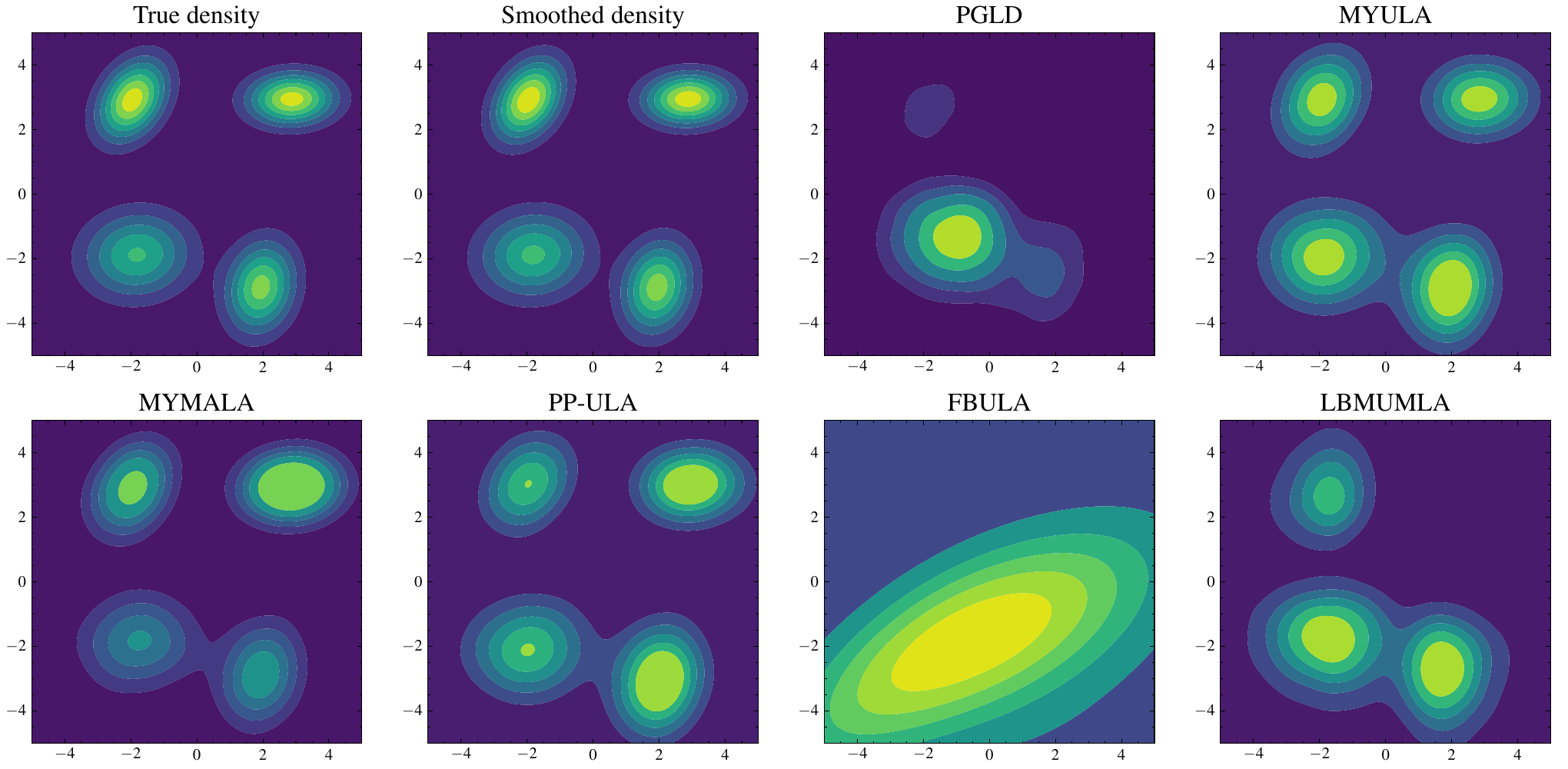}
                 \caption{$K=4$}
             \end{subfigure}  
             \hfill
             \begin{subfigure}[h]{.48\textwidth}
                 \centering
                 \includegraphics[width=\textwidth]{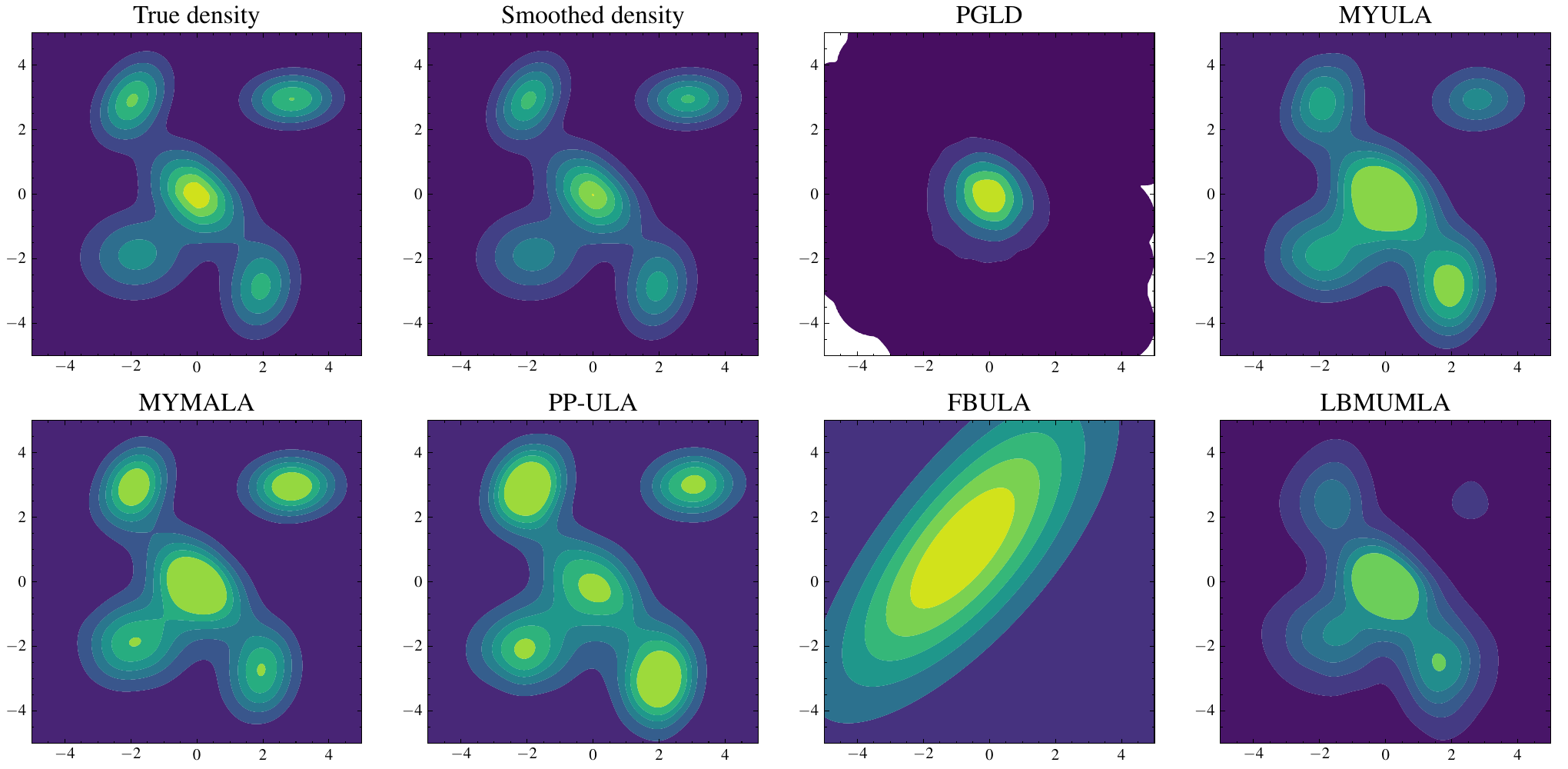}
                 \caption{$K=5$}
             \end{subfigure}      
             \caption{Mixture of $K$ Gaussians and a Laplacian prior with step size and smoothing parameter pair $(\gamma, \lambda)=(0.15, 1)$}
         \end{figure}

         \begin{figure}[htbp]
             \centering
             \begin{subfigure}[h]{.48\textwidth}
                 \centering
                 \includegraphics[width=\textwidth]{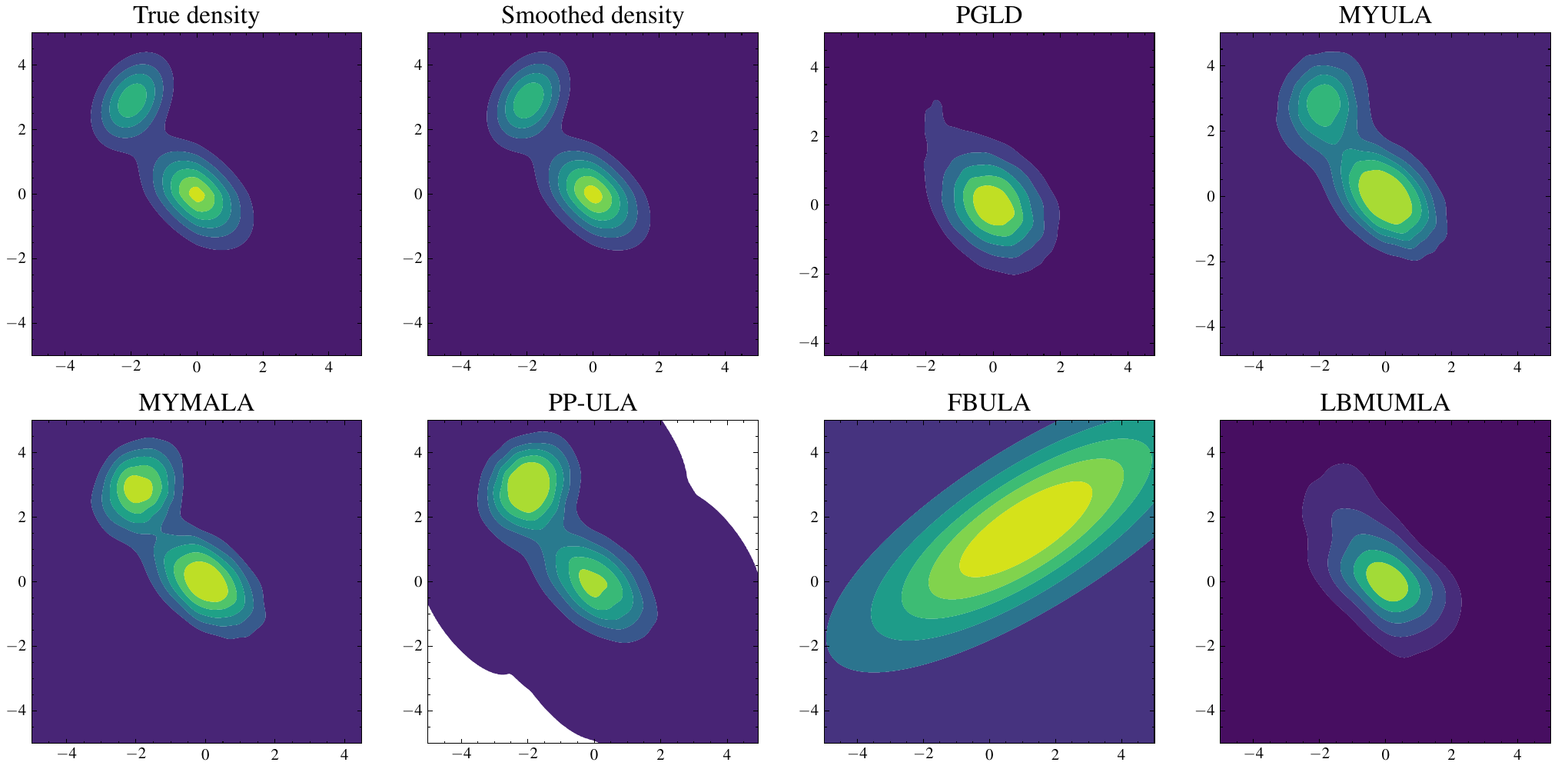}
                 \caption{$K=2$}
                 \vspace*{1mm}
             \end{subfigure}    
             \hfill
             \begin{subfigure}[h]{.48\textwidth}
                 \centering
                 \includegraphics[width=\textwidth]{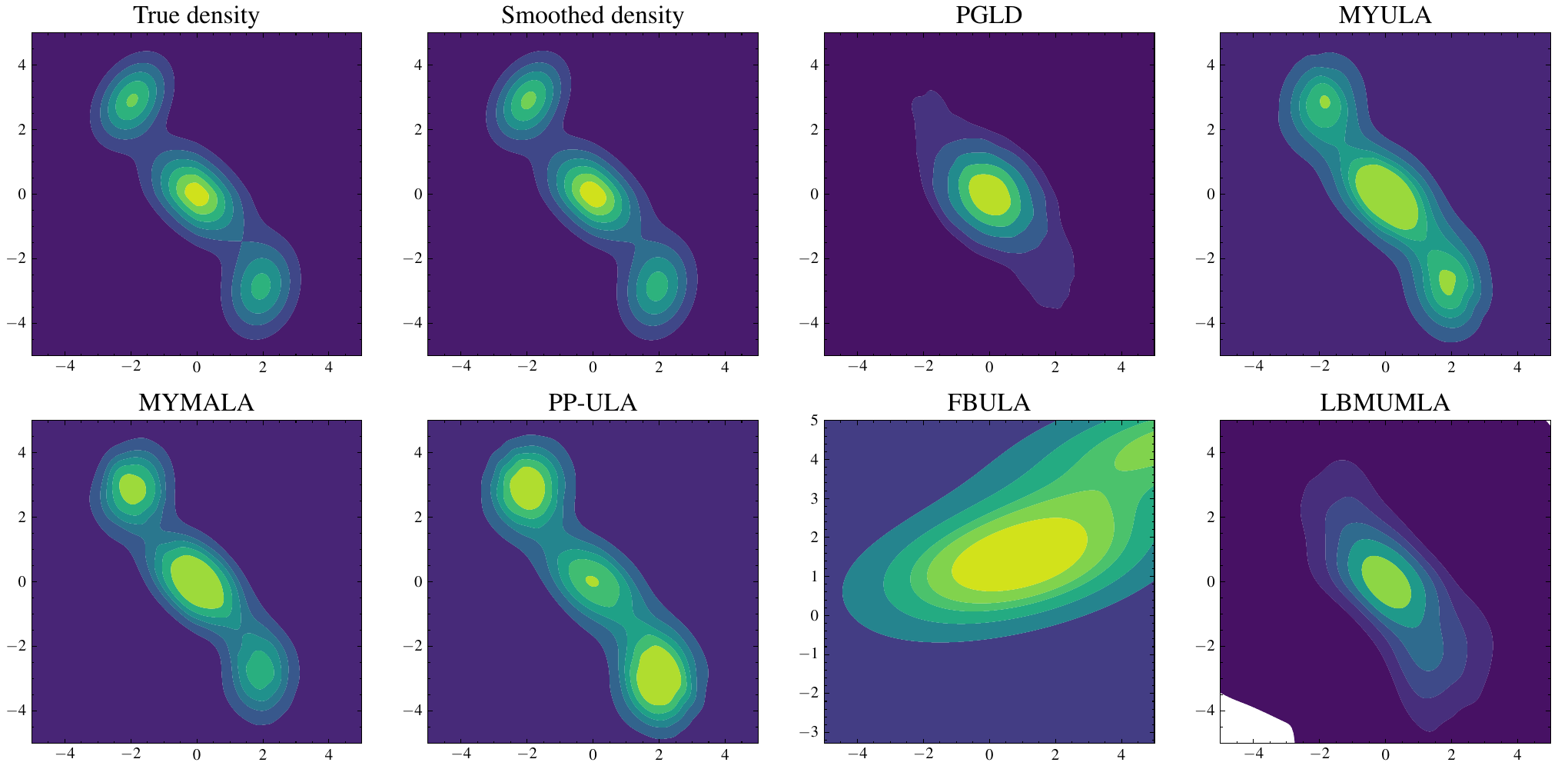}
                 \caption{$K=3$}
                 \vspace*{1mm}
             \end{subfigure}      
             \par\vspace{2mm}
             \begin{subfigure}[h]{.48\textwidth}
                 \centering
                 \includegraphics[width=\textwidth]{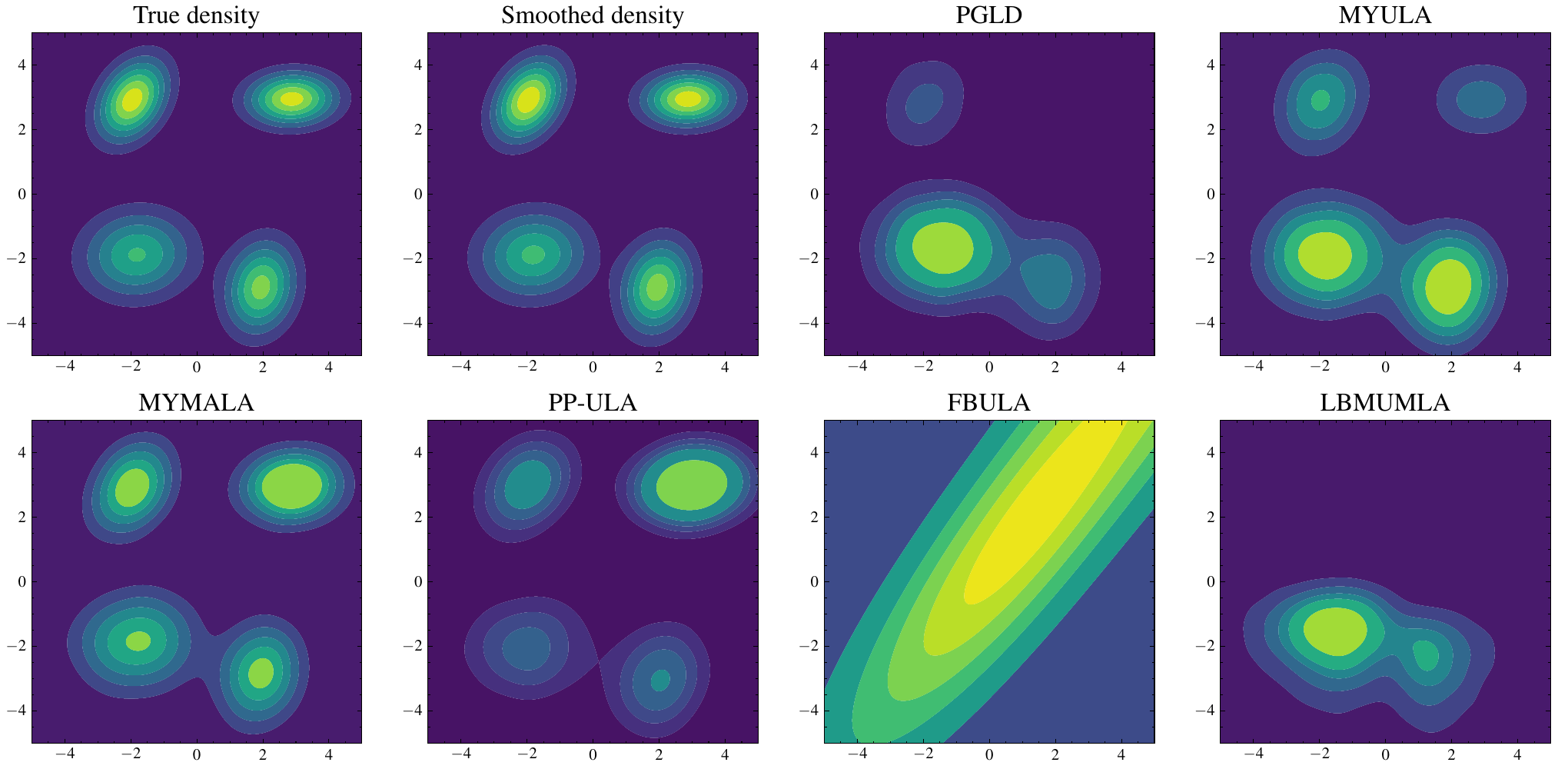}
                 \caption{$K=4$}
             \end{subfigure}  
             \hfill
             \begin{subfigure}[h]{.48\textwidth}
                 \centering
                 \includegraphics[width=\textwidth]{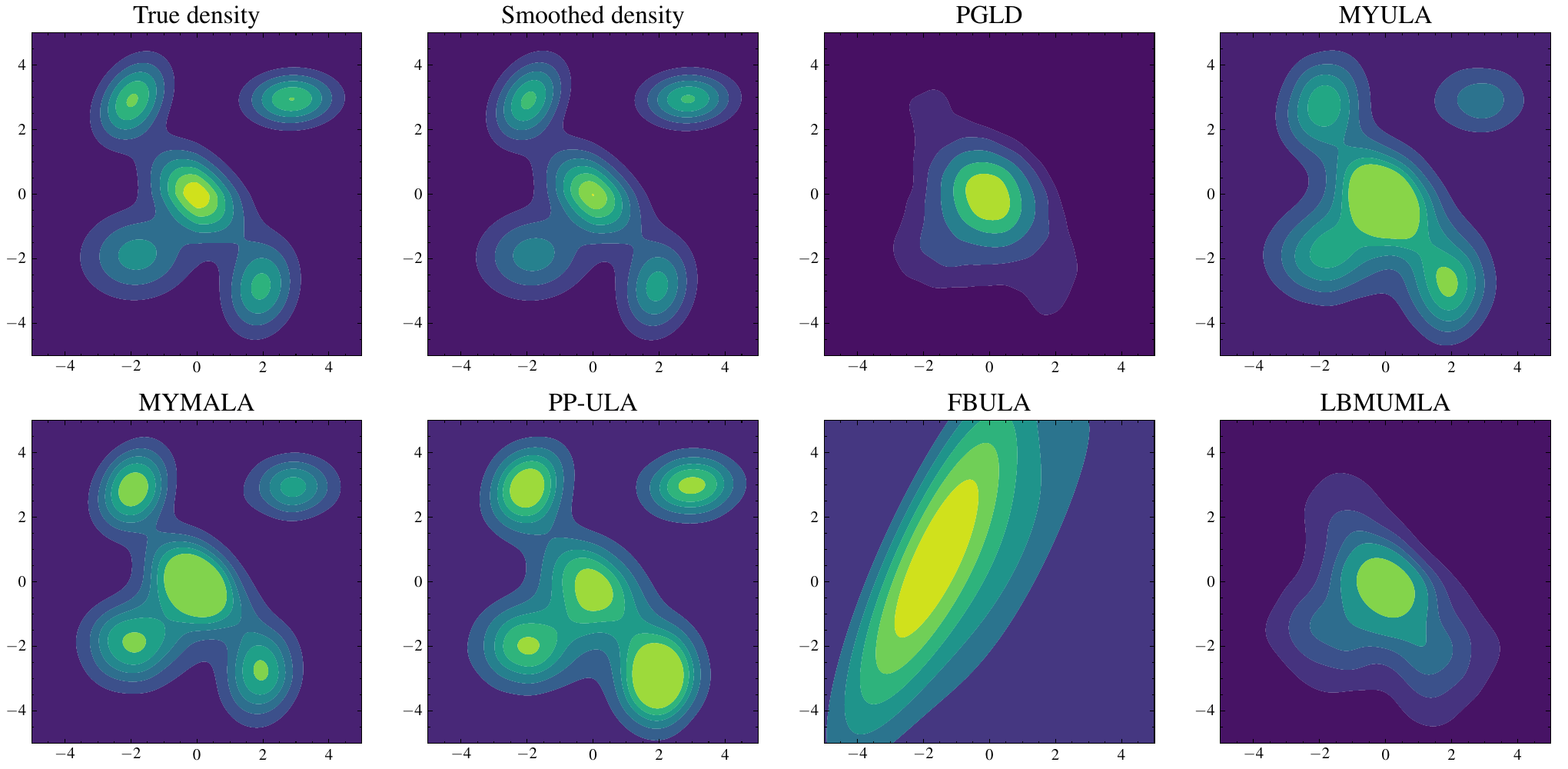}
                 \caption{$K=5$}
             \end{subfigure}      
             \caption{Mixture of $K$ Gaussians and a Laplacian prior with step size and smoothing parameter pair $(\gamma, \lambda)=(0.25, 1)$}
         \end{figure}
         
         \clearpage
        
        \pagebreak
        \section*{Appendix 3}
\begin{figure}[htbp]
            \centering
            \includegraphics[height=0.98\textheight]{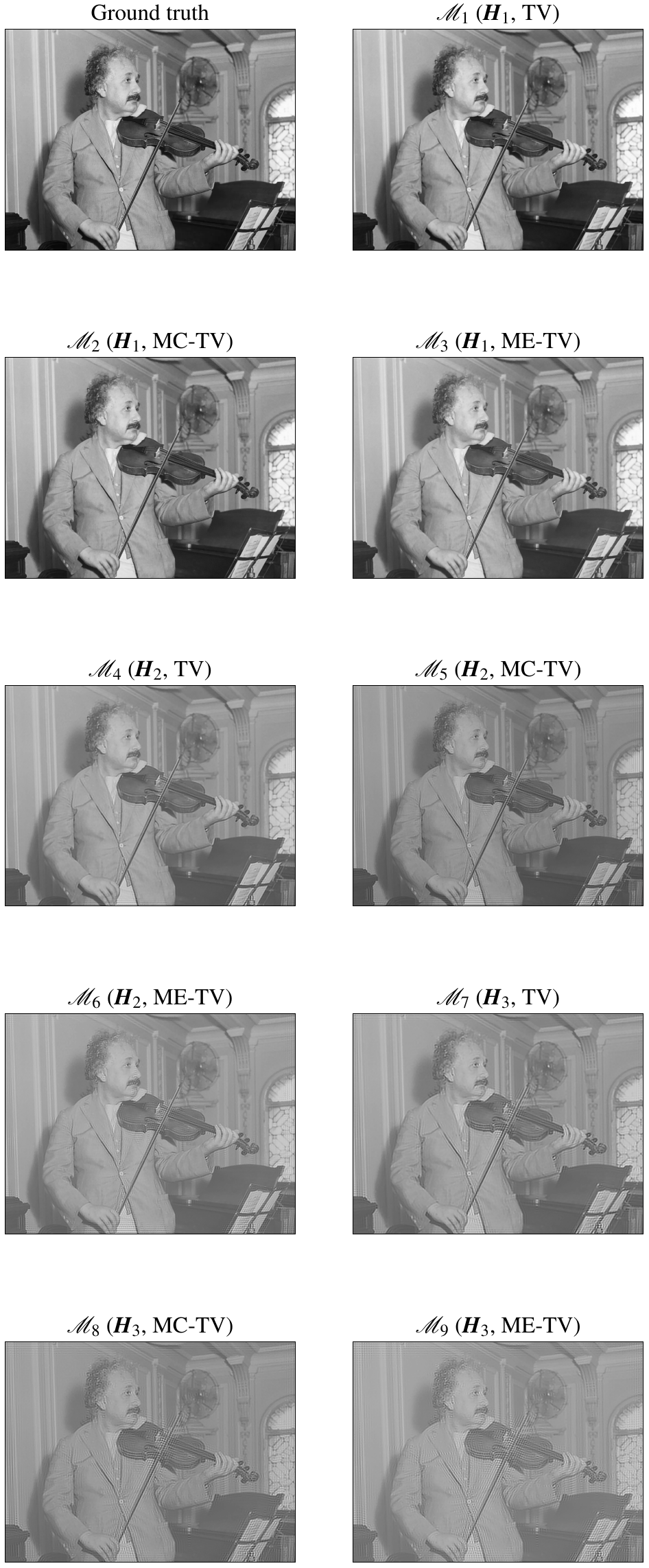}
            \caption{The \texttt{einstein} test image and the MAP estimators of $\scrM_j$, $j\in\set{9}$}    
            \label{fig:einstein}
        \end{figure}

\begin{figure}[htbp]
             \centering
             \includegraphics[height=0.98\textheight]{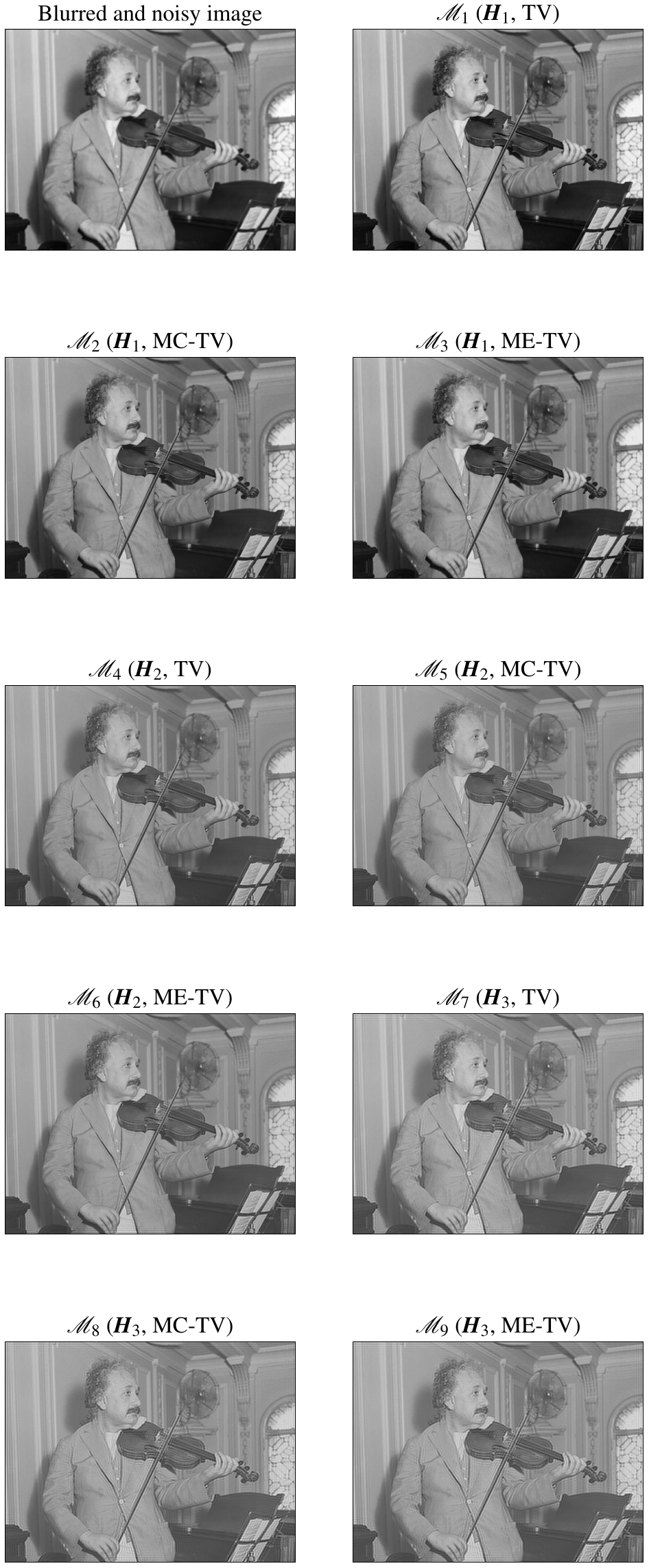}
             \caption{The blurred and noisy version of the \texttt{einstein} image and the posterior means of samples of $\scrM_j$, $j\in\set{9}$, generated using MYULA}    
             \label{fig:einstein_posterior_means_MYULA}
         \end{figure}

\begin{figure}[htbp]
            \centering
            \includegraphics[height=0.98\textheight]{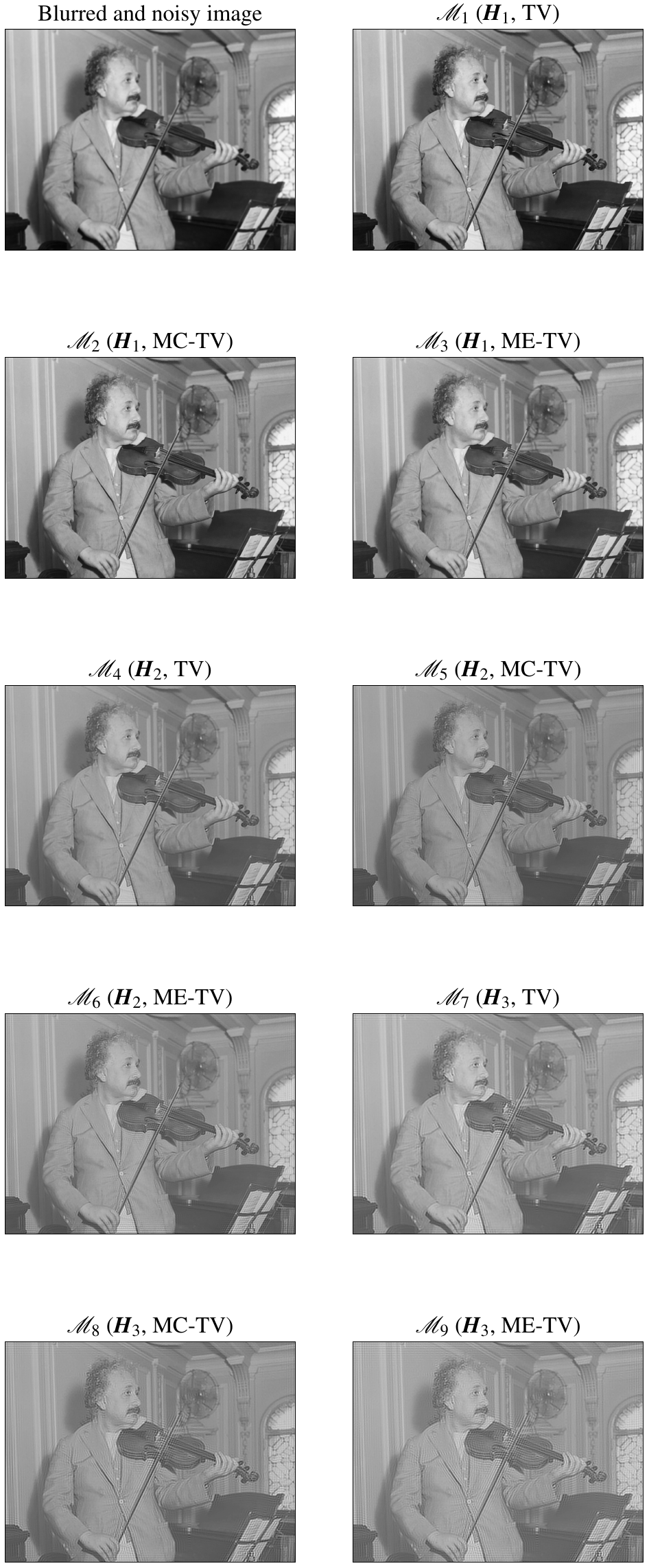}
            \caption{The blurred and noisy version of the \texttt{einstein} image and the posterior means of samples of $\scrM_j$, $j\in\set{9}$, generated using ULPDA}    
            \label{fig:einstein_posterior_means_ULPDA}
        \end{figure}

        \addcontentsline{toc}{section}{Appendix 3}
        We give additional simulation results for the Bayesian image deconvolution task using another test image \texttt{einstein} in this section. Here we use the same settings and hyperparameter configurations as in \Cref{subsec:bayesian_image_deconv}. 
        \Cref{fig:einstein} presents the \texttt{einstein} test image and its MAP estimators for all 9 models computed using the adaptive PDHG algorithm (AdaPDHG) with 1000 iterations. \Cref{fig:einstein_posterior_means_ULPDA,fig:einstein_posterior_means_MYULA} present the blurred and noisy version of the test image, and the posterior means of the samples generated by ULPDA and MYULA. 
        We note that from \Cref{tab:einstein}, observations similar to those for the \texttt{camera} test image can be made for this \texttt{einstein} test image.

         \begin{table}[h]
                 \begin{center}
                 \caption{Signal-to-noise ratios (SNR), peak signal-to-noise ratios (PSNR) and mean-squared errors (MSE) of MAPs and posterior means of the samples based on $\scrM_j$, $j\in\set{9}$ for the \texttt{einstein} image}
                 \label{tab:einstein}
                 \begin{tabular}{@{\extracolsep{\fill}}lrrrrrrrrrrrr@{\extracolsep{\fill}}}
                 \toprule%
                 && \multicolumn{3}{@{}c@{}}{MAP} && \multicolumn{3}{@{}c@{}}{MYULA} && \multicolumn{3}{@{}c@{}}{ULPDA} \\
            \cmidrule{3-5}\cmidrule{7-9}\cmidrule{11-13}%
             && SNR & PSNR & MSE & & SNR & PSNR & MSE & & SNR & PSNR & MSE\\
                 \midrule
                 $\scrM_1$ ($\bH_1$, TV) && 28.10 & 33.32 & 30.22 & & 26.92 & 32.15 & 39.67 & & 28.06 & 33.28 & 30.55 \\
                 $\scrM_2$ ($\bH_1$, MC-TV) && 27.41 & 32.67 & 35.44 & & 25.48 & 30.71 & 55.28 & & 28.19 & 33.42 & 29.62 \\
                 $\scrM_3$ ($\bH_1$, ME-TV) && 28.88 & 34.10 & 25.31 & & 27.03 & 32.25 & 38.73 & & \textbf{28.93} & \textbf{34.15} & \textbf{25.00} \\
                 $\scrM_4$ ($\bH_2$, TV) && 20.89 & 26.12 & 159.06 & & 21.96 & 27.18 & 124.37 & & 21.29 & 26.51 & 145.25 \\
                 $\scrM_5$ ($\bH_2$, MC-TV) && 18.59 & 23.81 & 270.15 & & 20.56 & 25.78 & 171.71 & & 19.24 & 24.47 & 232.55 \\
                 $\scrM_6$ ($\bH_2$, ME-TV) && 17.88 & 23.10 & 318.27 & & 21.30 & 26.56 & 144.72 & & 19.17 & 24.39 & 236.77 \\
                 $\scrM_7$ ($\bH_3$, TV) && 17.95 & 23.17 & 313.42 & & 20.33 & 25.55 & 181.11 & & 18.67 & 23.89 & 265.28 \\
                 $\scrM_8$ ($\bH_3$, MC-TV) && 14.18 & 19.40 & 746.06 & & 18.71 & 23.94 & 262.75 & & 15.90 & 21.12 & 502.43 \\
                 $\scrM_9$ ($\bH_3$, ME-TV) && 13.25 & 18.47 & 925.34 & & 19.26 & 24.48 & 231.58 & & 15.44 & 20.66 & 558.30 \\
                 \bottomrule
                 \end{tabular}
                 \end{center}
             \end{table}
         
         \newpage

        \begin{figure}[htbp]
            \centering
             \begin{subfigure}[h]{0.48\textwidth}
                \centering
                \includegraphics[height=0.46\textheight]{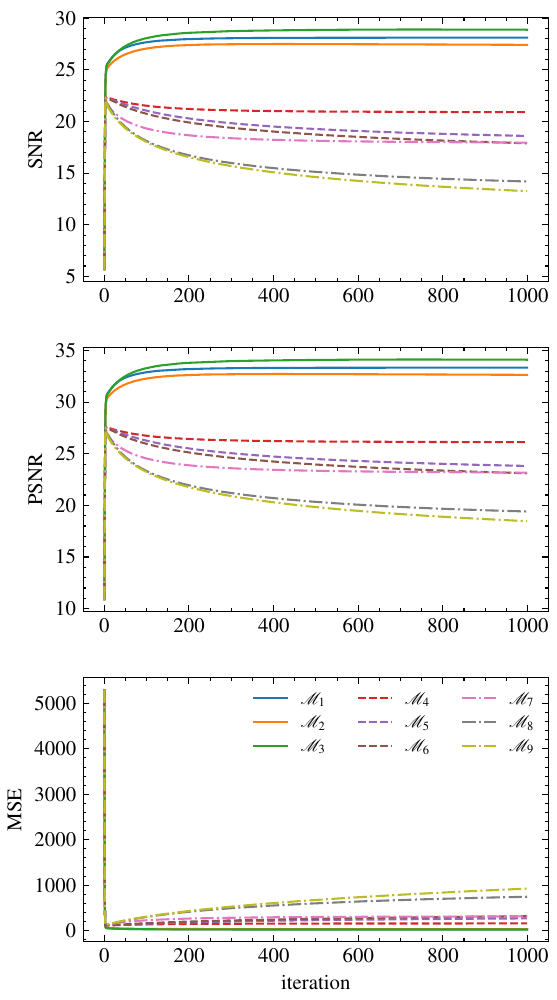}     
                \caption{AdaPDHG}
             \end{subfigure}
             \begin{subfigure}[h]{0.48\textwidth}
                 \centering
                 \includegraphics[height=0.46\textheight]{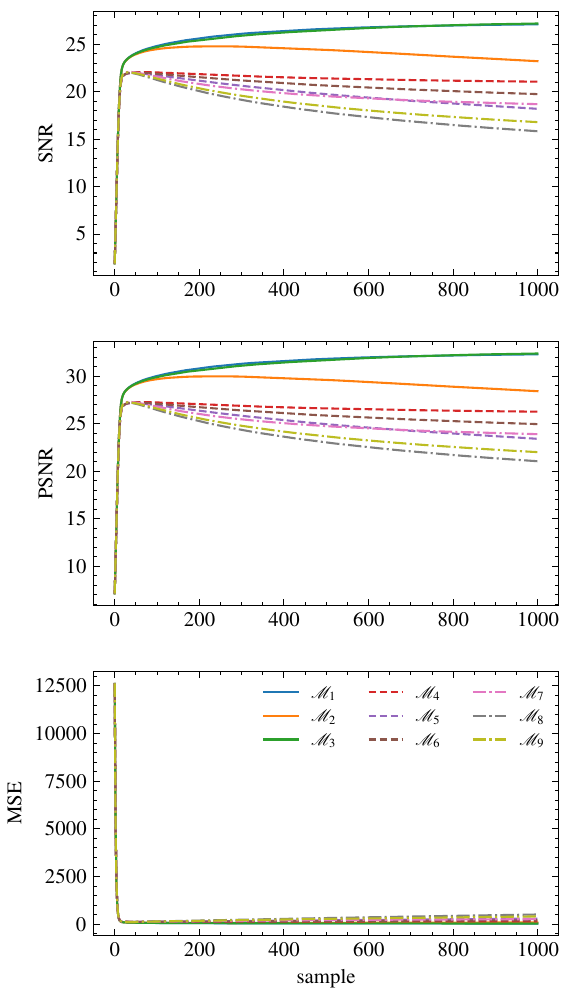}
                 \caption{MYULA}
             \end{subfigure}
            \begin{subfigure}[h]{0.48\textwidth}
                 \centering
                 \includegraphics[height=0.46\textheight]{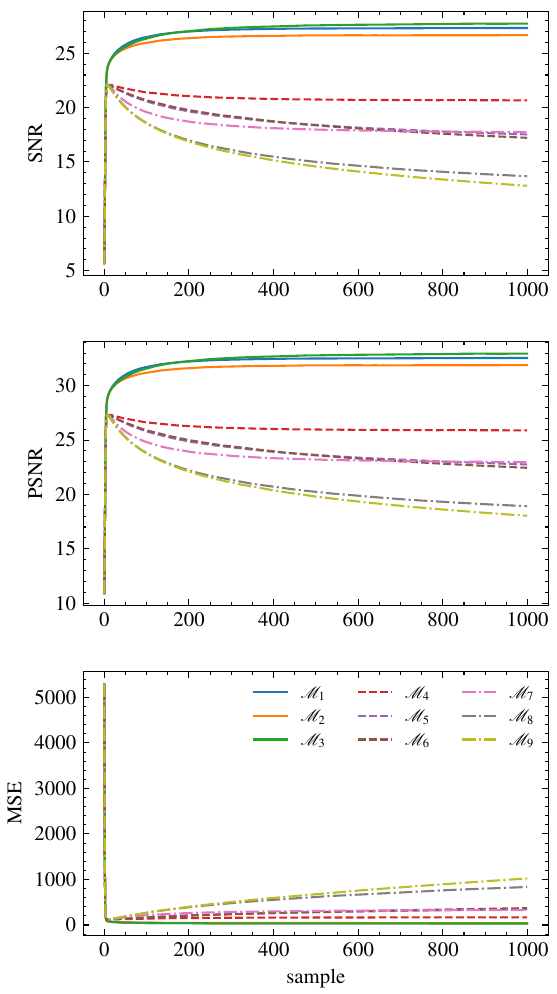}
                 \caption{ULPDA}
             \end{subfigure}
             \caption{SNRs, PSNRs and MSEs of iterates or samples of the \texttt{einstein} image generated by different algorithms based on $\scrM_j$, $j\in\set{9}$}    
             \label{fig:einstein_snr_psnr_mse}
      \end{figure}

\end{document}